\documentclass[english]{article}
\usepackage[T1]{fontenc}
\usepackage[latin9]{inputenc}
\usepackage{babel}
\usepackage{array}
\usepackage{float}
\usepackage{multirow}
\usepackage{amsmath}
\usepackage{amsthm}
\usepackage{graphicx}
\usepackage[unicode=true]
 {hyperref}

\makeatletter

\providecommand{\tabularnewline}{\\}

\theoremstyle{plain}
\newtheorem{prop}{\protect\propositionname}

\PassOptionsToPackage{numbers, compress}{natbib}

\usepackage[preprint]{neurips_2024}

\PassOptionsToPackage{numbers, compress}{natbib}

\usepackage[unicode=true]{hyperref}       
\usepackage{url}            
\usepackage{booktabs}       
\usepackage{amsfonts}       
\usepackage{nicefrac}       
\usepackage{microtype}      
\usepackage{xcolor} 

\usepackage{microtype}
\usepackage{graphicx}
\usepackage{subfigure}
\usepackage{mathtools}

\usepackage{xcolor}
\usepackage{soul}

\definecolor{red}{HTML}{FF0000}
\definecolor{lawngreen}{HTML}{7CFC00}
\definecolor{orange}{HTML}{FFA500}
\definecolor{peru}{HTML}{CD853F}
\definecolor{gray}{HTML}{808080}
\definecolor{lightgray}{RGB}{240,240,240}

\makeatother

\providecommand{\propositionname}{Proposition}

\begin{document}
\renewcommand{\contentsname}{Table of Content for Appendix}
\title{Variational Flow Models: Flowing in Your Style}
\author{\textbf{Kien Do, Duc Kieu, Toan Nguyen, Dang Nguyen, Hung Le, Dung
Nguyen, Thin Nguyen}\\
Applied Artificial Intelligence Institute (A2I2), Deakin University,
Australia\\
\emph{\{k.do, v.kieu, k.nguyen, d.nguyen, thai.le, dung.nguyen, thin.nguyen\}@deakin.edu.au}}

\maketitle
\global\long\def\Expect{\mathbb{E}}%
\global\long\def\Real{\mathbb{R}}%
\global\long\def\Var{\mathbb{V}}%
\global\long\def\Data{\mathcal{D}}%
\global\long\def\Loss{\mathcal{L}}%
\global\long\def\Normal{\mathcal{N}}%
\global\long\def\Id{\mathrm{I}}%
\global\long\def\Rectify{\mathrm{Rectify}}%
\global\long\def\ELBO{\text{ELBO}}%
\global\long\def\diag{\text{diag}}%
\global\long\def\softmax{\text{softmax}}%
\global\long\def\argmin#1{\underset{#1}{\text{argmin}}}%
\global\long\def\argmax#1{\underset{#1}{\text{argmax}}}%
\global\long\def\betamax{\overline{\beta}_{\text{max}}}%
\global\long\def\betamin{\overline{\beta}_{\text{min}}}%

\begin{abstract}
We propose a systematic \emph{training-free} method to transform
the probability flow of a \emph{``linear''} stochastic process characterized
by the equation $X_{t}=a_{t}X_{0}+\sigma_{t}X_{1}$ into a \emph{straight}
\emph{constant-speed} (SC) flow, reminiscent of Rectified Flow. This
transformation facilitates fast sampling along the original probability
flow via Euler method without training a new model of the SC flow.
The flexibility of our approach allows us to extend our transformation
to inter-convert two posterior flows of two distinct ``linear''
stochastic processes. Moreover, we can easily integrate high-order
numerical solvers into the transformed SC flow, further enhancing
the sampling accuracy and efficiency. Rigorous theoretical analysis
and extensive experimental results substantiate the advantages of
our framework. Our code is available at this \href{https://github.com/clarken92/VFM}{link}.

\end{abstract}
\addtocontents{toc}{\protect\setcounter{tocdepth}{-1}}

\section{Introduction}

Diffusion models \cite{ho2020denoising,sohl2015deep,song2019generative,song2020improved}
have recently achieved remarkable results across diverse machine learning
tasks, spanning image generation \cite{dhariwal2021diffusion,rombach2022high,saharia2022photorealistic},
audio synthesis \cite{kong2020diffwave}, image editing \cite{meng2021sdedit,brooks2023instructpix2pix},
and video generation \cite{ho2022video}.

The strength of diffusion models mainly stems from their iterative
sampling process which gradually removes noise from a noisy latent
sample to recover the clean sample. This iterative process can be
viewed as an advanced version of the generation process in hierarchical
VAEs \cite{maaloe2019biva,vahdat2020nvae}, allowing diffusion models
to approximate the data distribution more accurately. Intriguingly,
in the context of continuous time, this process becomes the diffusion
process - a fundamental mechanism underpinning various natural phenomena
in our physical world \cite{sohl2015deep,song2020score}. In this
continuous-time setting, diffusion models can be formulated using
stochastic differential equations (SDEs) \cite{song2020score}. However,
this sampling process tends to be slow, typically requiring hundreds
to thousands of function evaluations to generate high-quality images
\cite{ho2020denoising,song2020score}. This problem is mainly caused
by the random noise present in each SDE-based update step. It can
be mitigated by leveraging the ``probability flow'' ODE, which has
the same marginal distribution of samples as the diffusion SDE, for
sampling \cite{maoutsa2020interacting,song2020score}. This class
of \emph{statistically equivalent} ODEs provides an interesting connection
between diffusion models and neural ODEs \cite{chen2018neural}.

Recent studies have extended diffusion SDEs to a general class of
\emph{``linear''} stochastic processes defined by the equation $X_{t}=a_{t}X_{0}+\sigma_{t}X_{1}$
\cite{albergo2023stochastic,liu2022flow}, which bridge two arbitrary
distributions $p_{0}$ and $p_{1}$, corresponding to the random variables
$X_{0}$ and $X_{1}$, respectively. The statistically equivalent
ODEs of these processes are modeled by minimizing the ``flow matching''
loss \cite{lipman2022flow} - a variant of the ``score matching''
loss in diffusion models \cite{hyvarinen2005estimation,vincent2011connection}.
Since the non-parametric formulas of these ODEs involve the posterior
distribution $p\left(x_{0},x_{1}|x_{t}\right)$, we term these ODEs
\emph{``posterior flow''} ODEs. We interpret their parameterized
models through the lens of variational inference when $p_{1}$ is
a Gaussian distribution, coining these models as \emph{``Variational
Flow Models''} (VFMs).

One notable VFM is Rectified Flow which approximates the posterior
flow of the stochastic process $X_{t}=\left(1-t\right)X_{0}+tX_{1}$
\cite{liu2022flow,liu2022rectified}. Demonstrated to exhibit straightness
and constant speed, this flow significantly reduces the number of
sampling steps even with the use of simple numerical methods like
the Euler method. Currently, obtaining a Rectified Flow involves either
training it from scratch \cite{liu2022flow} or distilling it from
an existing posterior flow \cite{liu2023instaflow}. These approaches
are resource-intensive, especially for large models like Stable Diffusion
\cite{rombach2022high}. Moreover, they lack flexibility as new training
are required when converting \emph{other} posterior flows into Rectified
Flow and vice versa.

\begin{figure}
\begin{centering}
\includegraphics[width=0.6\textwidth]{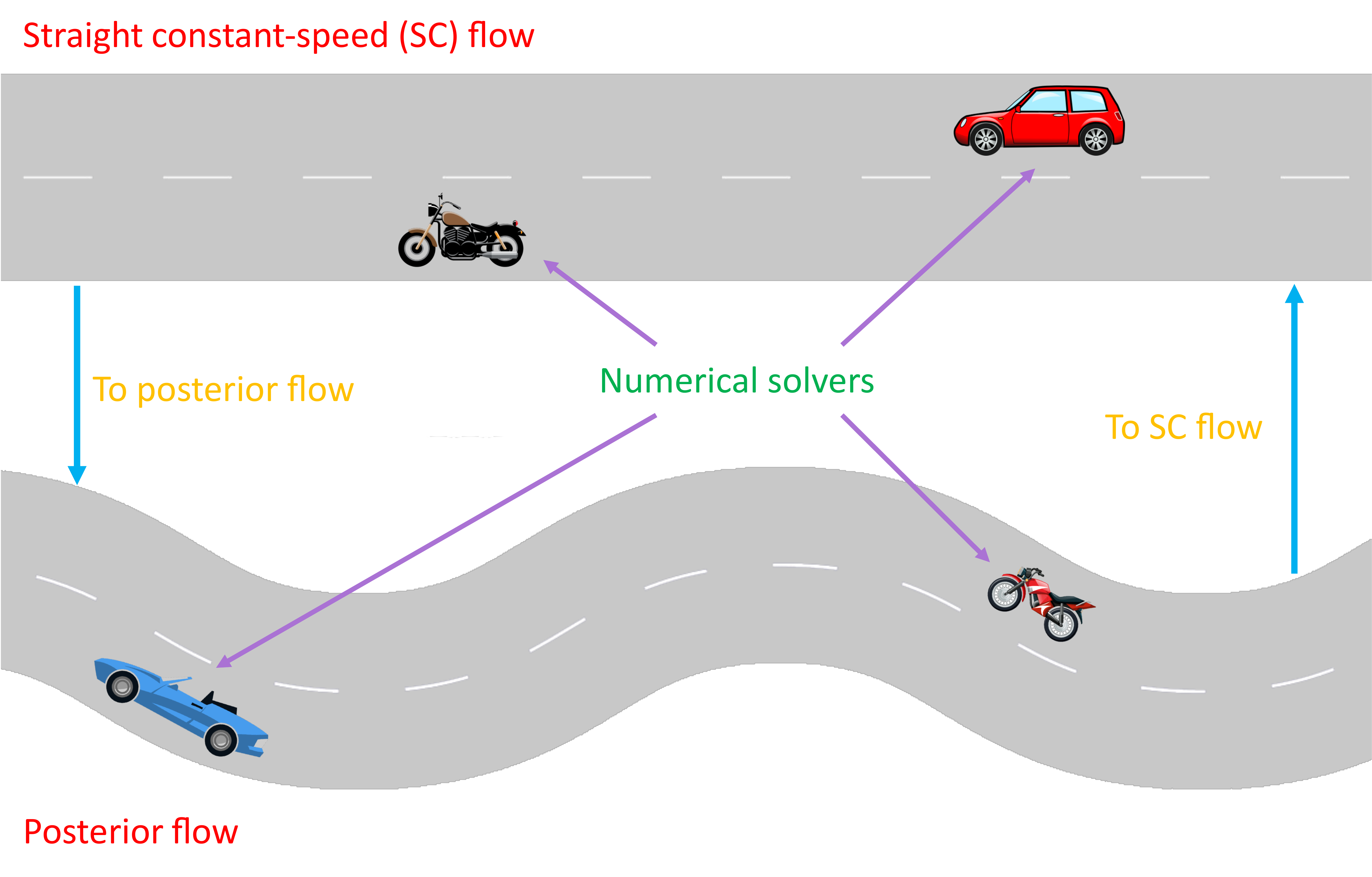}
\par\end{centering}
\caption{An illustration of our method that transforms a non-straight posterior
flow into a straight constant-speed one and performs updates on it.\label{fig:Illustration-of-our-method}}
\end{figure}

To overcome this challenge, in this work we propose a systematic \emph{training-free}
method to transform the posterior flow (or its VFM) of \emph{any}
``linear'' stochastic process into a \emph{straight constant-speed
(SC) flow} akin to Rectified Flow, and vice versa. Our key idea is
to convert the original process $X_{t}=a_{t}X_{0}+\sigma_{t}X_{1}$
into the process $\bar{X}_{t}=(1-t)X_{0}+tX_{1}$ associated with
Rectified Flow (or the processes associated with other SC flows) via
some clever change-of-variables operations which allow us to apply
rules of calculus to easily compute the velocity of the SC flow from
the velocity of the original (posterior) flow. With our proposed transformation,
instead of directly moving from a sample $x_{t}$ at time $t$ to
a sample $x_{t'}$ at time $t'$ along the original flow, we first
map $x_{t}$ to the sample $\bar{x}_{t}$ in the transformed SC flow.
Then, we transition from $\bar{x}_{t}$ to $\bar{x}_{t'}$ along the
SC flow and finally recover $x_{t'}$ from $\bar{x}_{t'}$. Consequently,
we can make use of straight constant-speed flows to speed up sampling
with the Euler method even when the original flow is strongly curved.
The essence of this approach lies in opting for a faster parallel
path, akin to a straight constant-speed highway, instead of navigating
a designated curved road when traveling from one place to another
as illustrated in Fig.~\ref{fig:Illustration-of-our-method}.

\begin{figure*}
\begin{centering}
{\resizebox{\textwidth}{!}{%
\par\end{centering}
\begin{centering}
\begin{tabular}{cccccc}
Real Image & DDIM (NFE=999) & DDIM & DEIS-1 & DPM-Solver++ (1M) & \textbf{SC-Euler}\tabularnewline
\includegraphics[width=0.2\textwidth]{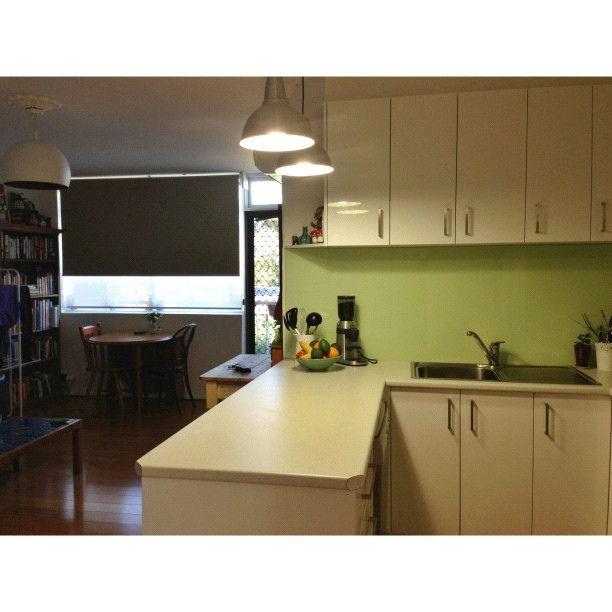} & \includegraphics[width=0.2\textwidth]{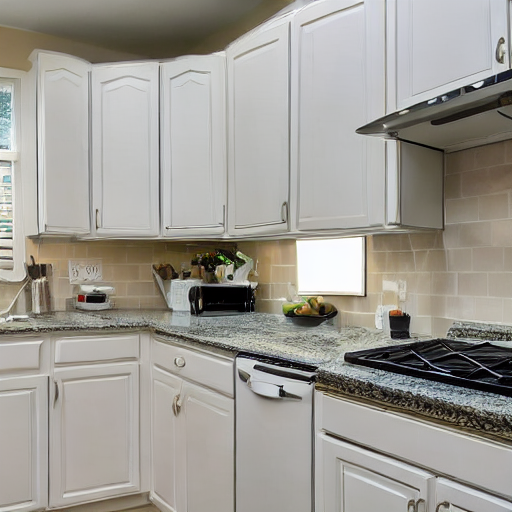} & \includegraphics[width=0.2\textwidth]{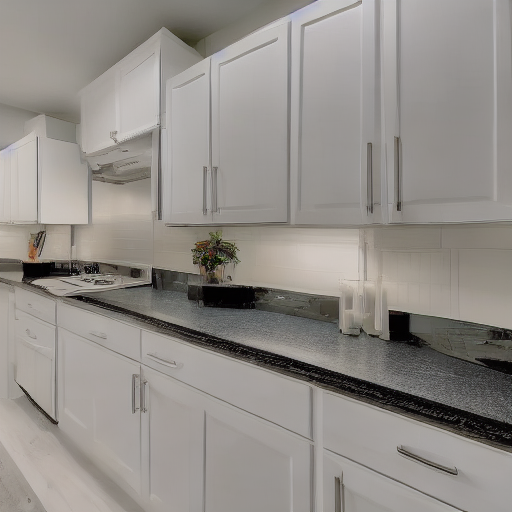} & \includegraphics[width=0.2\textwidth]{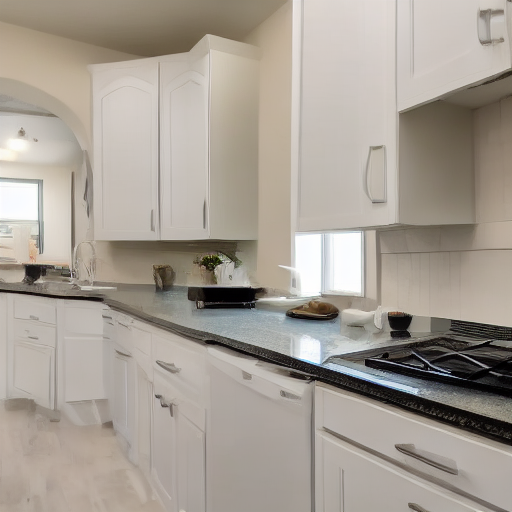} & \includegraphics[width=0.2\textwidth]{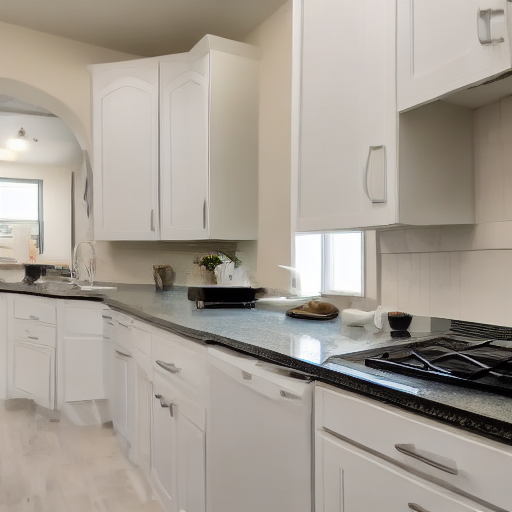} & \includegraphics[width=0.2\textwidth]{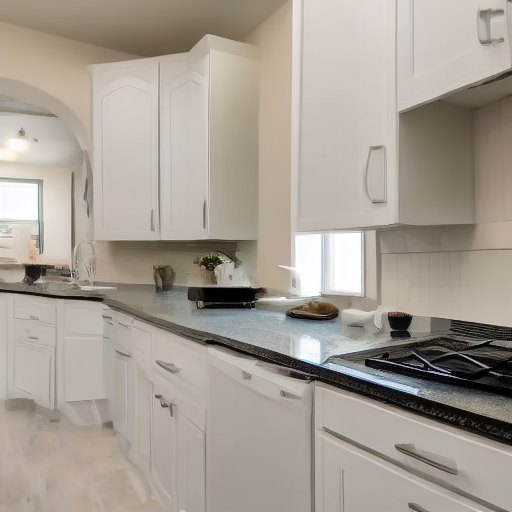}\tabularnewline
 &  &  &  &  & \tabularnewline
DEIS-2 & DPM-Solver++ (2M) & UniPC-1 & \textbf{SC-AB2} & \textbf{SC-AB1AM2} & \textbf{SC-AB2AM2}\tabularnewline
\includegraphics[width=0.2\textwidth]{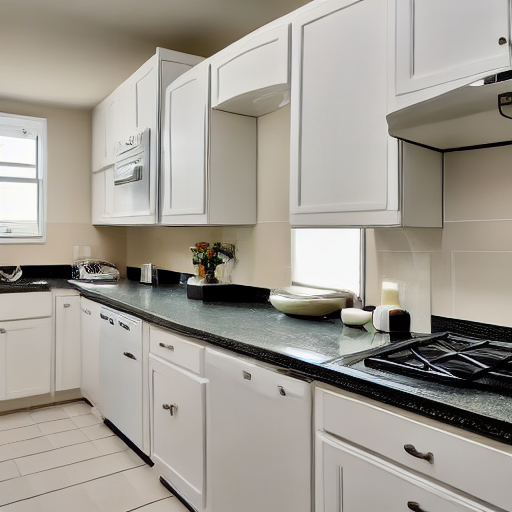} & \includegraphics[width=0.2\textwidth]{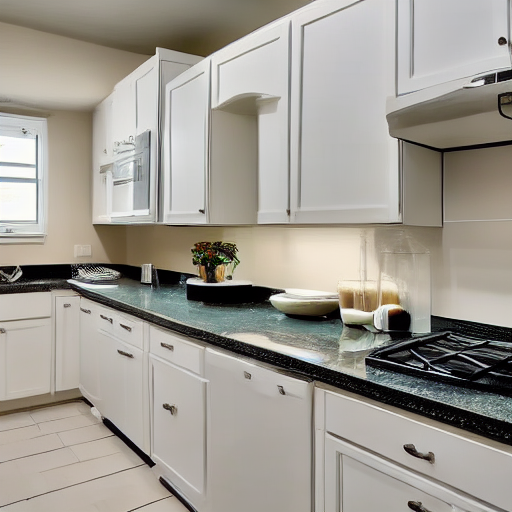} & \includegraphics[width=0.2\textwidth]{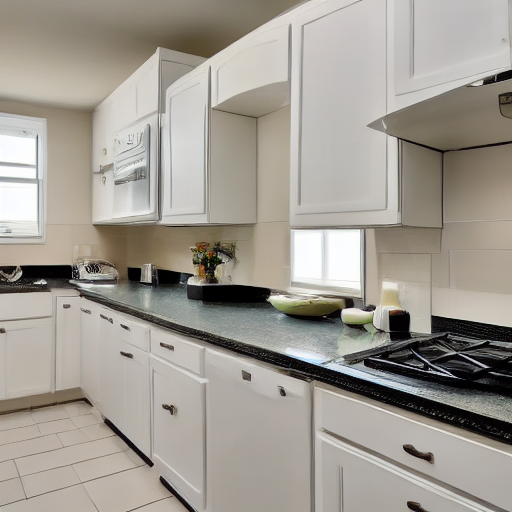} & \includegraphics[width=0.2\textwidth]{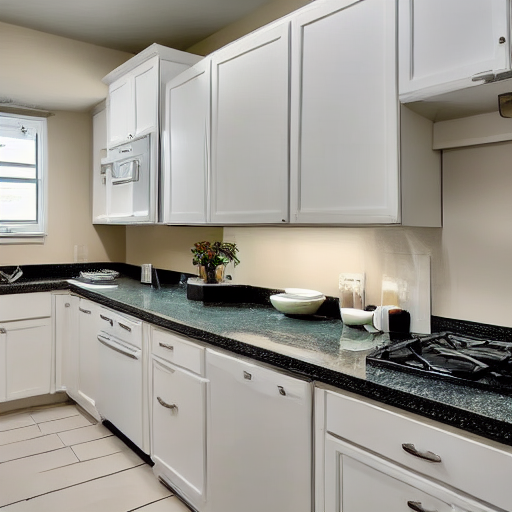} & \includegraphics[width=0.2\textwidth]{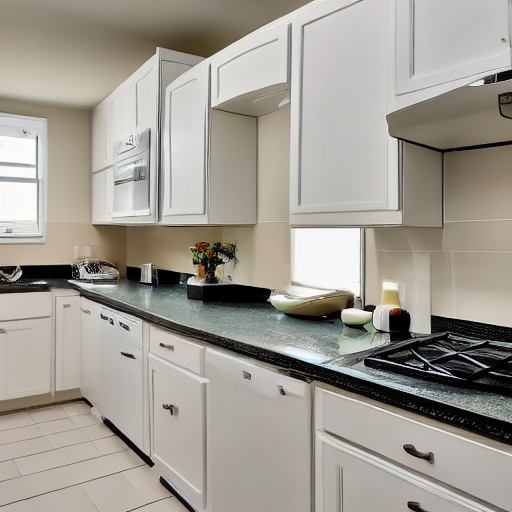} & \includegraphics[width=0.2\textwidth]{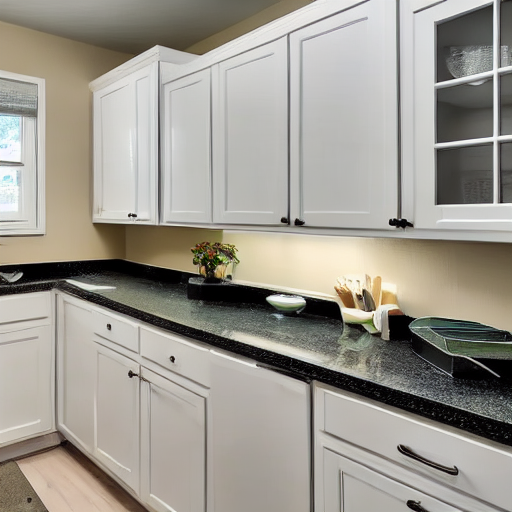}\tabularnewline
 &  &  &  &  & \tabularnewline
DEIS-3 & DPM-Solver++ (3M) & UniPC-2 & \textbf{SC-AB3} & \textbf{SC-AB2AM3} & \textbf{SC-AB3AM3}\tabularnewline
\includegraphics[width=0.2\textwidth]{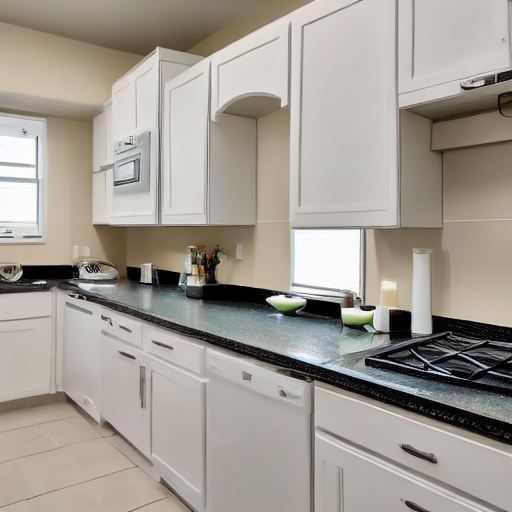} & \includegraphics[width=0.2\textwidth]{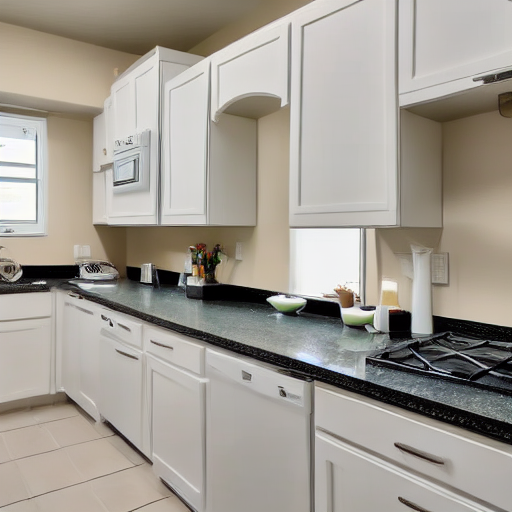} & \includegraphics[width=0.2\textwidth]{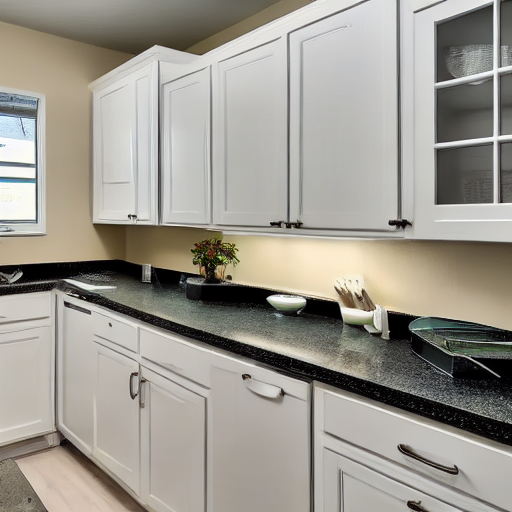} & \includegraphics[width=0.2\textwidth]{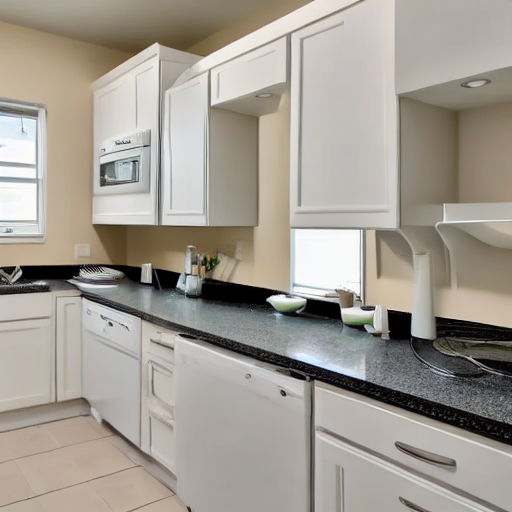} & \includegraphics[width=0.2\textwidth]{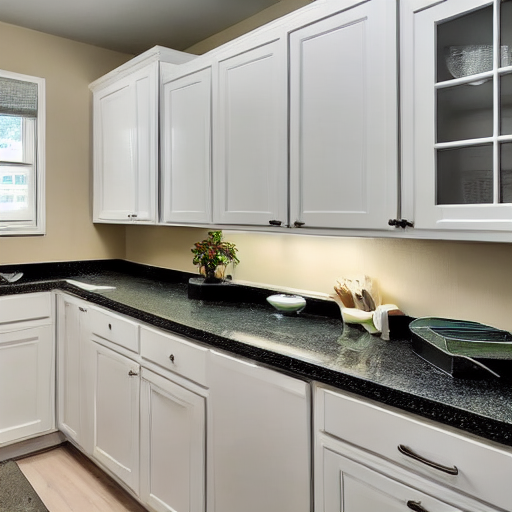} & \includegraphics[width=0.2\textwidth]{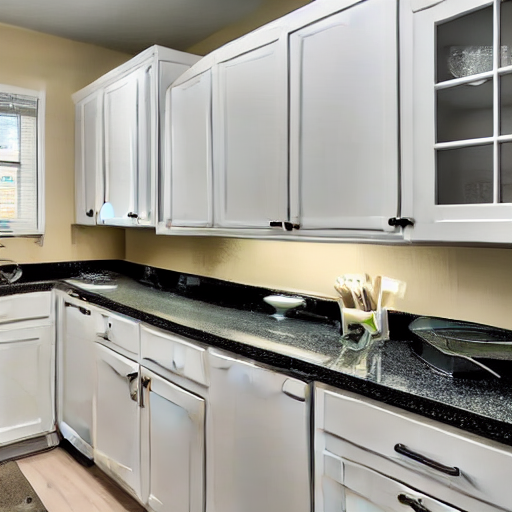}\tabularnewline
\end{tabular}}}
\par\end{centering}
\caption{Real images and images generated by our proposed numerical solvers
(\textbf{SC-{*}}) and baseline solvers with NFE=10. The model is Stable
Diffusion v1.4 and the text prompt is \emph{\textquotedblleft a kitchen
with white counter tops and cabinets.\textquotedblright}.\label{fig:GenImagesByDiffMethods}}
\end{figure*}

Our method is versatile, applicable to all existing diffusion models,
including DDPM / VP SDE \cite{ho2020denoising,song2020score}, SMLD
/ VE SDE \cite{song2019generative,song2020score} as the equation
$X_{t}=a_{t}X_{0}+\sigma_{t}X_{1}$ encompasses the SDE forms of these
models \cite{liu2022flow}. Moreover, our framework is adaptable and
can readily accommodate additional functionalities. Firstly, beyond
converting a posterior flow into a SC flow, we can extend the transformation
to two posterior flows of two different ``linear'' stochastic processes.
This is particularly impactful in ODE trajectory simulation, where
learning a single velocity model for one ODE is sufficient to simulate
the trajectories of an infinite number of other ODEs. Secondly, our
framework facilitates the seamless adoption of high-order numerical
ODE solvers for the transformed SC flow with minimal adjustments compared
to working on the original flow. These solvers significantly enhance
sample quality and reduce the number of sampling steps, positioning
them on par with current state-of-the-art training-free samplers for
diffusion models like DPM-Solver++ \cite{lu2022dpmpp} and UniPC \cite{zhao2023unipc}
(Fig.~\ref{fig:GenImagesByDiffMethods}) which \emph{implicitly}
employ a special variant of our proposed SC transformations.

Comprehensive mathematical derivations, coupled with extensive experiments
on a 2D toy dataset and large-scale image generation tasks, validate
the correctness of our proposed transformation method and showcase
the effectiveness of our high-order numerical solvers.

\section{Background}

Below we provide a brief overview of diffusion models \cite{ho2020denoising,song2019generative}
and rectified flows \cite{liu2022flow,liu2022rectified}. To simplify
our presentation, we will focus on continuous-time models, as the
discrete-time counterparts can be easily obtained by selecting an
appropriate discretization scheme.

\subsection{Diffusion Models\label{subsec:Diffusion-Models}}

Diffusion models (DMs) \cite{ho2020denoising,sohl2015deep,song2019generative,song2020improved}
are a class of generative models that iteratively add Gaussian noise
to input data to transform the data distribution $p_{0}$ to a Gaussian
distribution $p_{1}$. In DDPM \cite{ho2020denoising}, given an input
image $x_{0}\sim p_{0}$, a noisy image $x_{t}$ at time $t$ ($0<t\leq1$)
can be sampled from $p\left(x_{t}|x_{0}\right)$. This distribution
is the Gaussian distribution of the form $\Normal\left(x_{t};\sqrt{\overline{\alpha}_{t}}x_{0},\left(1-\overline{\alpha}_{t}\right)\mathrm{I}\right)$
where $\overline{\alpha}_{t}\equiv\overline{\alpha}\left(t\right)$
is a positive, decreasing function of $t$ with $\overline{\alpha}_{0}\approx1$
and $\overline{\alpha}_{1}\approx0$. This means $x_{t}$ can be computed
as: 
\begin{equation}
x_{t}=\sqrt{\overline{\alpha}_{t}}x_{0}+\sqrt{1-\overline{\alpha}_{t}}\epsilon\label{eq:DDPM_forward_xt_x0}
\end{equation}
where $\epsilon\sim\Normal\left(0,\mathrm{I}\right)$. Song et al.
\cite{song2020score} have shown that the diffusion process in Eq.~\ref{eq:DDPM_forward_xt_x0}
can be expressed as the variance-preserving stochastic differential
equation (VP SDE):
\begin{equation}
dx_{t}=-\frac{\beta_{t}}{2}x_{t}dt+\sqrt{\beta_{t}}dw_{t}\label{eq:DDPM_SDE}
\end{equation}
where $w_{t}$ denotes the Wiener process, and $\beta_{t}\equiv\beta\left(t\right)$
is a function of $t$ derived from $\overline{\alpha}_{t}$. The associated
\emph{probability flow} ordinary differential equation (PF ODE) of
this SDE, which has the same marginal distribution of samples $p_{t}\left(x_{t}\right)$,
is expressed as:
\begin{align}
dx_{t} & =-\frac{\beta_{t}}{2}\left(x_{t}+\nabla\log p\left(x_{t}\right)\right)dt
\end{align}
where $\nabla\log p\left(x_{t}\right)$ is the \emph{score function},
approximated as $\frac{-\epsilon_{\theta}\left(x_{t},t\right)}{\sqrt{1-\overline{\alpha}_{t}}}$.
$\epsilon_{\theta}$ is a noise network trained by minimizing the
\emph{noise matching} (NM) loss:
\begin{equation}
\Loss_{\text{NM}}\left(\theta\right)=\Expect_{t}\left[\lambda_{t}\Expect_{x_{0},\epsilon}\left[\left\Vert \epsilon_{\theta}\left(x_{t}\left(x_{0},\epsilon\right),t\right)-\epsilon\right\Vert _{2}^{2}\right]\right]\label{eq:noise_matching}
\end{equation}
where $\lambda_{t}$ denotes the weight at time $t$ and is set to
1 by default. $\Loss_{\text{NM}}$ is a variant of the \emph{score
matching} (SM) loss \cite{hyvarinen2005estimation,vincent2011connection}.
In Eq.~\ref{eq:noise_matching}, $t\sim\mathcal{U}\left(\left[0,1\right]\right)$,
$x_{0}\sim p_{0}$, $\epsilon\sim\Normal\left(0,\mathrm{I}\right)$,
and $x_{t}$ is computed from $x_{0}$ and $\epsilon$ via Eq.~\ref{eq:DDPM_forward_xt_x0}.

Another well-known diffusion model is SMLD \cite{song2019generative,song2020improved},
characterized by the diffusion process:
\begin{equation}
x_{t}=x_{0}+\sigma_{t}\epsilon\label{eq:DDIM_forward_xt_x0}
\end{equation}
Eq.~\ref{eq:DDIM_forward_xt_x0} is just a reparameterization of
Eq.~\ref{eq:DDPM_forward_xt_x0} by regarding $x_{t}$ and $\sigma_{t}$
in Eq.~\ref{eq:DDIM_forward_xt_x0} as $\frac{x_{t}}{\sqrt{\overline{\alpha}_{t}}}$
and $\sqrt{\frac{1-\overline{\alpha}_{t}}{\overline{\alpha}_{t}}}$
in Eq.~\ref{eq:DDPM_forward_xt_x0}, respectively. The diffusion
process has been shown to be equivalent to the variance exploding
SDE (VE SDE) \cite{song2020score,karras2022elucidating}:
\begin{equation}
dx_{t}=\sqrt{\frac{d\left(\sigma_{t}^{2}\right)}{dt}}dw_{t}
\end{equation}
The corresponding PF ODE of this SDE is:
\begin{align}
dx_{t} & =-\sigma_{t}\nabla_{x}\log p_{t}\left(x_{t}\right)d\sigma_{t}\label{eq:DDIM_ODE_1}\\
 & \approx\epsilon_{\theta}\left(x_{t},t\right)d\sigma_{t}\label{eq:DDIM_ODE_2}
\end{align}
where $\epsilon_{\theta}$ is the noise network identical to the one
in DDPM.

\subsection{Generalized Rectified Flows\label{subsec:Generalized-Rectified-Flows}}

Liu et al. \cite{liu2022flow} have explored a broader class of stochastic
processes between two boundary distributions $p_{0}$ and $p_{1}$,
characterized by the general formula:
\begin{equation}
X_{t}=\phi\left(X_{0},X_{1},t\right)\label{eq:general_stochastic_process}
\end{equation}
In this equation, $X_{0}$, $X_{1}$, $X_{t}$ represent random variables
with marginal distributions $p_{0}$, $p_{1}$, $p_{t}$, respectively.
The function $\phi:\Real^{d}\times\Real^{d}\times\left[0,1\right]\rightarrow\Real^{d}$
maps a coupling $\left(X_{0},X_{1}\right)$ (w.r.t. the joint distribution
$p_{0,1}$) to $X_{t}$ at time $t$ $\in$ $\left[0,1\right]$. Their
goal is to find a \emph{deterministic mapping} from $X_{0}$ to $X_{1}$
governed by the ODE: $dx_{t}=v_{\theta}\left(x_{t},t\right)dt$, such
that the marginal distribution of $X_{t}$ obtained from the ODE matches
that of $X_{t}$ derived from the stochastic process in Eq.~\ref{eq:general_stochastic_process}.
This \emph{statistically equivalent} ODE is referred to as a \emph{Generalized
Rectified Flow }(GRF) and $v_{\theta}$ is learned by minimizing the
\emph{velocity matching} (or \emph{flow matching}) loss $\Loss_{\text{VM}}$
\cite{lipman2022flow,liu2022flow} below:
\begin{equation}
\Loss_{\text{VM}}\left(\theta\right)=\Expect_{t}\Expect_{x_{0},x_{1}}\left[\left\Vert v_{\theta}\left(x_{t},t\right)-v_{\phi}\left(x_{0},x_{1},t\right)\right\Vert _{2}^{2}\right]\label{eq:velocity_matching_loss_reframed}
\end{equation}
where $t\sim\mathcal{U}\left(\left[0,1\right]\right)$, $x_{0},x_{1}\sim p_{0,1}$,
$x_{t}=\phi\left(x_{0},x_{1},t\right)$, and $v_{\phi}\left(x_{0},x_{1},t\right):=\dot{x}_{t}=\frac{d\phi\left(x_{0},x_{1},t\right)}{dt}$
denotes the target velocity at time $t$ for the trajectory between
$x_{0}$, $x_{1}$. Sometimes, we express the target velocity as $v_{\phi}\left(x_{t},t\right):=\dot{x}_{t}$
instead of $v_{\phi}\left(x_{0},x_{1},t\right):=\frac{d\phi\left(x_{0},x_{1},t\right)}{dt}$
to emphasize that it is a function of $x_{t}=\phi\left(x_{0},x_{1},t\right)$
rather than $x_{0}$, $x_{1}$.

A remarkable variant of Eq.~\ref{eq:general_stochastic_process}
is:
\begin{equation}
X_{t}=a_{t}X_{0}+\sigma_{t}X_{1}\label{eq:GRF_DiffusionProcess_v2}
\end{equation}
where $a_{t}\equiv a\left(t\right)$, $\sigma_{t}\equiv\sigma\left(t\right)$
are continuously differentiable functions of $t$ satisfying that
$a_{0}=1$, $a_{1}=0$, and $\sigma_{0}=0$. When $X_{1}$ follows
the standard normal distribution (i.e., $p_{1}=\Normal\left(0,\mathrm{I}\right)$),
Eq.~\ref{eq:general_stochastic_process} describes various diffusion
processes including VP SDE and VE SDE. This implies that probability
flow ODEs can be regarded as specific instances of GRFs. Eq.~\ref{eq:general_stochastic_process}
also has an important special case:
\begin{equation}
X_{t}=\left(1-t\right)X_{0}+tX_{1}\label{eq:RectifiedFlow}
\end{equation}
which constitutes a linear interpolation between $X_{0}$ and $X_{1}$.
The statistically equivalent ODE of this stochastic process is called
\emph{Rectified Flow} (RF).

\section{Variational Flow Models}

By introducing the Generalized Rectified Flow (GRF) framework (Eqs.~\ref{eq:general_stochastic_process},
\ref{eq:velocity_matching_loss_reframed}), deriving a special variant
of the framework (Eq.~\ref{eq:GRF_DiffusionProcess_v2}) that encompasses
various diffusion processes and their corresponding PF ODEs, and providing
theoretical analysis for this special variant, Liu et al. \cite{liu2022flow}
have done an inspiring work. In this section, we aim to (i) \emph{convey
their findings using simpler notations}, (ii) \emph{interpret the
GRF framework from the perspective of variational inference}, and
(iii) \emph{derive explicit formulas for the velocities of well-known
diffusion models}. Detailed proofs are available in the appendix,
with some not covered in \cite{liu2022flow}. The content in this
section will greatly assist in presenting our method in the next section.

If we regard $\Loss_{\text{VM}}$ in Eq.~\ref{eq:velocity_matching_loss_reframed}
as a function of a non-parameteric velocity $v$ rather than the parameters
$\theta$, we can prove that this loss attains its global minimum
at $v^{*}\left(x_{t},t\right):=\Expect_{x_{0},x_{1}|x_{t}}\left[v_{\phi}\left(x_{0},x_{1},t\right)\right]$
(Appdx.~\ref{subsec:Derivation-of-posterior-velocity}). Furthermore,
the ODE characterized by $v^{*}$ exhibits the same marginal distribution
of samples (i.e., $p\left(x_{t}\right)$) as the stochastic process
in Eq.~\ref{eq:general_stochastic_process} at every time $t\in\left[0,1\right]$
(proof in Appdx.~\ref{subsec:Statistical-equivalence-between-process-and-flow}).
The intuition for this is that given either of the boundary distributions
$p_{0}$, $p_{1}$ is fixed, the match in velocity results in a corresponding
match in probability. This alignment is substantiated by the \emph{continuity
equation} $\frac{\partial p_{t}\left(x_{t}\right)}{\partial t}=\nabla\cdot\left(p_{t}\left(x_{t}\right)v\left(x_{t},t\right)\right)$,
where the temporal evolution of $p_{t}\left(x_{t}\right)$ primarily
relies on the velocity $v\left(x_{t},t\right)$. We refer to $v^{*}$
as the\emph{ posterior velocity} based on its formula and the ODE
characterized by $v^{*}$ as the \emph{posterior flow ODE} or simply
\emph{posterior flow} of the given stochastic process. Since $v_{\theta}$
can be regarded as a \emph{variational approximation} of $v^{*}$
with $v_{\theta}=v^{*}$ leading to the maximum likelihood when $p_{1}$
is a Gaussian distribution (details in Appdx.~\ref{subsec:Variational-interpretation-of}),
we refer to $v_{\theta}$ as a \emph{variational velocity} and its
associated ODE as a \emph{variational flow ODE} or a \emph{Variational
Flow Model} (VFM).

For the class of \emph{``linear''} stochastic processes defined
in Eq.~\ref{eq:GRF_DiffusionProcess_v2}, the corresponding target
velocities $v_{\phi}\left(x_{0},x_{1},t\right)$ and posterior velocities
$v^{*}\left(x_{t},t\right)$ of these processes are represented as
follows:
\begin{align}
v_{\phi}\left(x_{0},x_{1},t\right) & :=\frac{d\phi\left(x_{0},x_{1},t\right)}{dt}=\dot{a}_{t}x_{0}+\dot{\sigma}_{t}x_{1}\label{eq:target_velo_1}\\
v^{*}\left(x_{t},t\right) & :=\Expect_{x_{0},x_{1}|x_{t}}\left[v_{\phi}\left(x_{0},x_{1},t\right)\right]\\
 & =\Expect_{x_{0},x_{1}|x_{t}}\left[\dot{a}_{t}x_{0}+\dot{\sigma}_{t}x_{1}\right]
\end{align}
Interestingly, $v^{*}\left(x_{t},t\right)$ can also be expressed
in the following forms (details in Appdx.~\ref{subsec:Derivation-of-posterior-velocity}):
\begin{align}
v^{*}\left(x_{t},t\right) & =\frac{\dot{\sigma}_{t}}{\sigma_{t}}x_{t}+a_{t}\left(\frac{\dot{a}_{t}}{a_{t}}-\frac{\dot{\sigma}_{t}}{\sigma_{t}}\right)x_{0|t}\label{eq:nonparam_velo_xt_x0}\\
 & =\frac{\dot{a}_{t}}{a_{t}}x_{t}+\sigma_{t}\left(\frac{\dot{\sigma}_{t}}{\sigma_{t}}-\frac{\dot{a}_{t}}{a_{t}}\right)x_{1|t}\label{eq:nonparam_velo_xt_x1}
\end{align}
where $x_{0|t}:=\Expect_{x_{0}|x_{t}}\left[x_{0}\right]$ and $x_{1|t}:=\Expect_{x_{1}|x_{t}}\left[x_{1}\right]$.
The reason behind Eqs.~\ref{eq:nonparam_velo_xt_x0}, \ref{eq:nonparam_velo_xt_x1}
is that given $x_{t}$ and $x_{0}$ ($x_{1}$), $x_{1}$ ($x_{\ensuremath{0}}$)
can be computed with certainty via Eq.~\ref{eq:general_stochastic_process}.Thus,
the stochasticity in $p\left(x_{0},x_{1}|x_{t}\right)$ can be fully
transformed into the stochasticity in either $p\left(x_{0}|x_{t}\right)$
or $p\left(x_{1}|x_{t}\right)$. Eqs.~\ref{eq:nonparam_velo_xt_x0},
\ref{eq:nonparam_velo_xt_x1} suggest that $v^{*}\left(x_{t},t\right)$
can be modeled indirectly via modeling either $x_{0|t}$ or $x_{1|t}$.
In the context of diffusion models, $x_{0|t}$ and $x_{1|t}$ are
modeled using a denoising autoencoder $x_{0,\theta}\left(x_{t},t\right)$
\cite{karras2022elucidating,vincent2010stacked,vincent2011connection}
and a noise network $x_{1,\theta}\left(x_{t},t\right)$ (i.e., $\epsilon_{\theta}\left(x_{t},t\right)$)
\cite{song2019generative,ho2020denoising}, respectively.

In Table~\ref{tab:velocity-formulas-summary}, we provide explicit
formulas for the target and posterior velocities of well-known diffusion
models. The detailed derivations can be found in Appdx.~\ref{subsec:Explicit-formulas-of}.

\begin{table*}
\begin{centering}
{\resizebox{\textwidth}{!}{
\begin{tabular}{cccccc}
\hline 
\multirow{2}{*}{Model} & \multirow{2}{*}{$a_{\ensuremath{t}}$} & \multirow{2}{*}{$\sigma_{t}$} & \multirow{2}{*}{$v_{\phi}\left(x_{0},x_{1},t\right)$} & \multicolumn{2}{c}{$v^{*}\left(x_{t},t\right)$}\tabularnewline
\cline{5-6} \cline{6-6} 
 &  &  &  & $x_{t}$, $x_{0|t}$ & $x_{\ensuremath{t}}$, $x_{1|t}$\tabularnewline
\hline 
\hline 
VP SDE & $\sqrt{\overline{\alpha}_{t}}$ & $\sqrt{1-\overline{\alpha}_{t}}$ & $\frac{\beta_{t}\overline{\alpha}_{t}}{2}\left(\frac{x_{1}}{\sqrt{1-\overline{\alpha}_{t}}}-\frac{x_{0}}{\sqrt{\overline{\alpha}_{t}}}\right)$ & $\frac{\beta_{t}\overline{\alpha}_{t}}{2\left(1-\overline{\alpha}_{t}\right)}\left(x_{t}-\frac{x_{0|t}}{\sqrt{\overline{\alpha}_{t}}}\right)$ & $-\frac{\beta_{t}}{2}\left(x_{t}-\frac{x_{1|t}}{\sqrt{1-\overline{\alpha}_{t}}}\right)$\tabularnewline
sub-VP SDE & $\sqrt{\overline{\alpha}_{t}}$ & $1-\overline{\alpha}_{t}$ & $\frac{\beta_{t}\sqrt{\overline{\alpha}_{t}}}{2}\left(2\sqrt{\overline{\alpha}_{t}}x_{1}-x_{0}\right)$ & $\frac{\beta_{t}\sqrt{\overline{\alpha}_{t}}}{1-\overline{\alpha}_{t}}\left(\sqrt{\overline{\alpha}_{t}}x_{t}-\frac{1+\overline{\alpha}_{t}}{2}x_{0|t}\right)$ & $-\frac{\beta_{t}}{2}\left(x_{t}-\left(1+\overline{\alpha}_{t}\right)x_{1|t}\right)$\tabularnewline
VE SDE & $1$ & $\sigma_{t}$ & $\dot{\sigma}_{t}x_{1}$ & $\frac{\dot{\sigma}_{t}}{\sigma_{t}}\left(x_{t}-x_{0|t}\right)$ & $\dot{\sigma}_{t}x_{1|t}$\tabularnewline
Rect. Flow & $1-t$ & $t$ & $x_{1}-x_{0}$ & $\frac{1}{t}\left(x_{t}-x_{0|t}\right)$ & $\frac{1}{1-t}\left(x_{1|t}-x_{t}\right)$\tabularnewline
\hline 
\end{tabular}}}
\par\end{centering}
\caption{The coefficients $a_{t}$, $\sigma_{t}$, target velocities $v_{\phi}\left(x_{0},x_{1},t\right)$,
and posterior velocities $v^{*}\left(x_{t},t\right)$ corresponding
to well-known diffusion models \cite{song2020score} and Rectified
Flow (Rect. Flow) \cite{liu2022flow}. In the case of (sub-)VP SDE,
$\beta_{t}=t\left(\protect\betamax-\protect\betamin\right)+\protect\betamin$
and $\overline{\alpha}_{t}=e^{-\frac{1}{2}t\left(\beta_{t}+\protect\betamin\right)}$.\label{tab:velocity-formulas-summary}}
\end{table*}

\section{Method\label{sec:Method}}

\subsection{Transforming posterior flows into straight constant-speed flows\label{subsec:Transforming-a-flow}}

In this section, we establish theoretically that the posterior flow
of any linear stochastic process defined by Eq.~\ref{eq:GRF_DiffusionProcess_v2},
no matter how intricately curved, can be transformed into straight
constant-speed flows reminiscent of Rectified Flow \cite{liu2022flow}.
Our investigation carries substantial implications in practical applications
by offering a systematic \emph{training-free} method to perform efficient
sampling with \emph{minimal} updates for a wide range of stochastic
processes, including the diffusion processes in Table~\ref{tab:velocity-formulas-summary}.

Let us begin by examining the stochastic processes described by the
equations\footnote{We don't include the stochastic process $X_{t}=\varphi_{t}X_{0}+X_{1}$
here because all the results w.r.t. this process can be easily obtained
from the results w.r.t. the process $X_{t}=X_{0}+\varphi_{t}X_{1}$
by swapping the role of $a_{t}$, $X_{0}$ and $\sigma_{t}$, $X_{1}$,
respectively.}:
\begin{align}
X_{t} & =\left(1-\varphi_{t}\right)X_{0}+\varphi_{t}X_{1}\text{ and}\label{eq:straight_stochastic_1}\\
X_{t} & =X_{0}+\varphi_{t}X_{1}\label{eq:straight_stochastic_2}
\end{align}
where $\varphi_{t}$ is a continuously differentiable function of
$t$ satisfying that $\varphi_{0}=0$. This condition ensures that
the distribution at $t=0$ aligns with $p_{0}$. The posterior velocities
corresponding to these processes are computed as follows:
\begin{align}
v^{*}\left(x_{t},t\right) & =\dot{\varphi}_{t}\left(x_{1|t}-x_{0|1,t}\right)\text{ and}\label{eq:straight_posterior_velo_1}\\
v^{*}\left(x_{t},t\right) & =\dot{\varphi}_{t}x_{1|t}\label{eq:straight_posterior_velo_2}
\end{align}
where $x_{0|1,t}$ in Eq.~\ref{eq:straight_posterior_velo_1} is
computed from $x_{t}$ and $x_{1|t}$ as $x_{0|1,t}=\frac{x_{t}-\varphi_{t}x_{1|t}}{1-\varphi_{t}}$.
In Eq.~\ref{eq:straight_posterior_velo_1} and Eq.~\ref{eq:straight_posterior_velo_2},
the two vectors $x_{1|t}-x_{0|1,t}$ and $x_{1|t}$ represent the
\emph{direction} of $v^{*}$, respectively while $\dot{\varphi}_{t}$
represents the \emph{scale in length} of $v^{*}$.

Since these two direction vectors depend on the estimated boundary
points $x_{1|t}$, $x_{0|1,t}$ which are \emph{expected} to be invariant
w.r.t. $t$ (Appdx.~\ref{subsec:Explanation-of-invariant-wrt-t}),
the velocities of samples in the \emph{same trajectory} should have
\emph{almost the same direction}. This implies that the flows characterized
by either Eq.~\ref{eq:straight_posterior_velo_1} or Eq.~\ref{eq:straight_posterior_velo_2}
are \emph{nearly straight}. If $\dot{\varphi}_{t}$ varies across
time, these flows experience varying speeds. By contrast, when $\dot{\varphi}_{t}$
is constant w.r.t. $t$, these flows maintain a constant speed. The
case where $\dot{\varphi}_{t}=1$, or equivalently $\varphi_{t}=t$
\footnote{$\dot{\varphi}_{t}=1$ implies that $\varphi_{t}=t+c$ with $c$ is
a constant. Since $\varphi_{0}=0$, we have $c=0$.}, is of special interest because in this case, the \emph{instantaneous}
velocity $v^{*}\left(x_{t},t\right)$ becomes the \emph{average} velocity
(i.e., $x_{1|t}-x_{0|1,t}$ for Eq.~\ref{eq:straight_posterior_velo_1}
and $x_{1|t}$ for Eq.~\ref{eq:straight_posterior_velo_2}). This
implies that in ideal cases we only need one Euler update to generate
samples of a boundary distribution given samples of the other. When
$\varphi_{t}=t$, Eqs.~\ref{eq:straight_stochastic_1}, \ref{eq:straight_stochastic_2}
become:
\begin{align}
X_{t} & =\left(1-t\right)X_{0}+tX_{1}\text{ and}\label{eq:straight_constant_speed_stochastic_1}\\
X_{t} & =X_{0}+tX_{1}\label{eq:straight_constant_speed_stochastic_2}
\end{align}
Eq.~\ref{eq:straight_constant_speed_stochastic_1} aligns with the
stochastic process corresponding to Rectified Flow \cite{liu2022flow},
while Eq.~\ref{eq:straight_constant_speed_stochastic_2} resembles
VE SDE (Table~\ref{tab:velocity-formulas-summary}), differing only
in the coefficient of $X_{1}$, which is $t\in\left[0,1\right]$ rather
than some function $\sigma_{t}$ of $t$ \cite{karras2022elucidating}.
Given the correspondence between a stochastic process and its posterior
flow, we term the \emph{stochastic process associated with a straight
(constant-speed) posterior flow} as a \emph{straight (constant-speed)
stochastic process}. The transformation of a posterior flow into a
straight constant-speed one is tantamount to converting its associated
stochastic process into a straight constant-speed process. Below,
we present in detail how this can be accomplished in two steps. Initially,
a non-straight flow (process) is transformed into a straight flow
(process). Subsequently, the straight flow (process) obtained in the
first step is converted into a straight constant-speed flow (process).
This procedure turns a sample $x_{t}$ of the original flow into a
sample $\bar{x}_{t}$ of the straight constant-speed flow. Following
the straight constant-speed flow, $\bar{x}_{t}$ seamlessly transitions
to $\bar{x}_{t'}$ and is then mapped back to $x_{t'}$.

\subsubsection{Non-straight flows $\rightarrow$ Straight flows\label{subsec:Non-straight-flows-2-straight-flows}}

We can transform a stochastic process of the form $X_{t}=a_{t}X_{0}+\sigma_{t}X_{1}$
into a straight one of the form $X_{t}=\left(1-\varphi_{t}\right)X_{0}+\varphi_{t}X_{1}$
(Eq.~\ref{eq:straight_stochastic_1}) by reformulating the original
stochastic process as follows:
\begin{align}
 & X_{t}=a_{t}X_{0}+\sigma_{t}X_{1}\label{eq:non_straight_to_straight_1}\\
\Leftrightarrow\  & \frac{X_{t}}{a_{t}+\sigma_{t}}=\frac{a_{t}}{a_{t}+\sigma_{t}}X_{0}+\frac{\sigma_{t}}{a_{t}+\sigma_{t}}X_{1}\label{eq:non_straight_to_straight_2}\\
\Leftrightarrow\  & \tilde{X}_{t}=\left(1-\varphi_{t}\right)X_{0}+\varphi_{t}X_{1}\label{eq:non_straight_to_straight_3}
\end{align}
where $\tilde{X}_{t}:=\frac{X_{t}}{a_{t}+\sigma_{t}}$, $\varphi_{t}:=\frac{\sigma_{t}}{a_{t}+\sigma_{t}}$.
$\varphi_{0}=0$ because $a_{0}=1$ and $\sigma_{0}=0$. Possible
numerical problems related to this transformation are discussed in
Section~\ref{subsec:Numerical-robustness}.

Let $x_{t}$ be a sample in the original posterior flow associated
with the process in Eq.~\ref{eq:non_straight_to_straight_1}, then
$\tilde{x}_{t}=\frac{x_{t}}{a_{t}+\sigma_{t}}$ becomes a sample in
the \emph{straight} posterior flow associated with the process in
Eq.~\ref{eq:non_straight_to_straight_3}. From $\tilde{x}_{t}$,
we perform an Euler update with the velocity $v^{*}\left(\tilde{x}_{t},t\right)$
and step size $t'-t$ to obtain $\tilde{x}_{t'}$:
\begin{equation}
\tilde{x}_{t'}=\tilde{x}_{t}+\left(t'-t\right)v^{*}\left(\tilde{x}_{t},t\right)\label{eq:Euler_update_SN}
\end{equation}
We then recover $x_{t'}$ from $\tilde{x}_{t'}$ as:
\begin{equation}
x_{t'}=\left(a_{t'}+\sigma_{t'}\right)\tilde{x}_{t'}\label{eq:x_t_recovered}
\end{equation}

Since $\tilde{X}_{t}$ is a scaled version of $X_{t}$ (by a factor
of $\frac{1}{a_{t}+\sigma_{t}}$), we refer to the transformation
from $X_{t}$ to $\tilde{X}_{t}$ in Eqs.~\ref{eq:non_straight_to_straight_1}-\ref{eq:non_straight_to_straight_3}
as \emph{``variable scaling''}. The remaining question revolves
around computing $v^{*}\left(\tilde{x}_{t},t\right)$. Intuitively,
$v^{*}\left(\tilde{x}_{t},t\right)$ can be approximated by $v_{\theta}\left(\tilde{x}_{t},t\right)$
if $p_{t}\left(x_{t}\right)$ has support over the entire sample space.
However, in practice, this approximation is unreliable. It is because
at time $t$, $v_{\theta}$ only yields accurate outputs for samples
from the original stochastic process (i.e., $x_{t}\sim p_{t}\left(x_{t}\right)$)
on which $v_{\theta}$ was trained, while $\tilde{x}_{t}$ and $p_{t}\left(\tilde{x}_{t}\right)$
significantly differ from $x_{t}$ and $p_{t}\left(x_{t}\right)$,
respectively. Fortunately, we can overcome this problem by expressing
$v^{*}\left(\tilde{x}_{t},t\right)$ via $v^{*}\left(x_{t},t\right)$
based on the following results:
\begin{prop}
[\textbf{Velocities of a scaled process}] Let $X_{t}=a_{t}X_{0}+\sigma_{t}X_{1}$
be a stochastic process defined in Eq.~\ref{eq:GRF_DiffusionProcess_v2}
and let $\tilde{X_{t}}=k_{t}X_{t}$ be another stochastic process
derived from $X_{t}$ with $k_{t}$ being a differentiable function
of $t$. The target velocity $v_{\phi}\left(\tilde{x}_{t},t\right)$
and posterior velocity $v^{*}\left(\tilde{x}_{t},t\right)$ w.r.t.
the stochastic process $\tilde{X}_{t}$ can be computed via the target
velocity $v_{\phi}\left(x_{t},t\right)$ and posterior velocity $v^{*}\left(x_{t},t\right)$
w.r.t. the stochastic process $X_{t}$, respectively, as follows:\label{Prop_1}
\begin{align}
v_{\phi}\left(\tilde{x}_{t},t\right) & =\dot{k}_{t}x_{t}+k_{t}v_{\phi}\left(x_{t},t\right)\label{eq:prop_1_target_velo}\\
v^{*}\left(\tilde{x}_{t},t\right) & =\dot{k}_{t}x_{t}+k_{t}v^{*}\left(x_{t},t\right)\label{eq:prop_1_posterior_velo}
\end{align}
\end{prop}
\begin{proof}
Please refer to Appdx.~\ref{subsec:Proof-of-Proposition-1} for a
detailed proof.
\end{proof}
Applying Eq.~\ref{eq:prop_1_posterior_velo} to the process in Eq.~\ref{eq:non_straight_to_straight_3}
with a note that $k_{t}=\frac{1}{a_{t}+\sigma_{t}}$, we have:
\begin{equation}
v^{*}\left(\tilde{x}_{t},t\right)=\frac{\left(a_{t}+\sigma_{t}\right)v^{*}\left(x_{t},t\right)-\left(\dot{a}_{t}+\dot{\sigma}_{t}\right)x_{t}}{\left(a_{t}+\sigma_{t}\right)^{2}}\label{eq:straight_velo_itpl}
\end{equation}

Given the relationships between $v^{*}\left(x_{t},t\right)$, $x_{t}$,
$x_{1|t}$ and $x_{0|t}$ in Eqs.~\ref{eq:nonparam_velo_xt_x0},
\ref{eq:nonparam_velo_xt_x1}, we can also express $v^{*}\left(\tilde{x}_{t},t\right)$
in terms of $x_{t}$, $x_{1|t}$ or $x_{t}$, $x_{0|t}$ or $x_{1|t}$,
$x_{0|1,t}$. Details about these expressions of $v^{*}\left(\tilde{x}_{t},t\right)$
are provided in Appdx.~\ref{subsec:Appdx_NonStraight2Straight}.
In the last case, $v^{*}\left(\tilde{x}_{t},t\right)=\dot{\varphi}_{t}\left(x_{1|t}-x_{0|1,t}\right)$
and Eq.~\ref{eq:Euler_update_SN} can be expressed as follows:
\begin{equation}
\tilde{x}_{t'}=\tilde{x}_{t}+\left(t'-t\right)\dot{\varphi}_{t}\left(x_{1|t}-x_{0|1,t}\right)\label{eq:Euler_update_SN_2}
\end{equation}
It is interesting to see that since $\tilde{x}_{t}$ represents a
sample from the straight flow $\tilde{X}_{t}=\left(1-\varphi_{t}\right)X_{0}+\varphi_{t}X_{1}$,
$v^{*}\left(\tilde{x}_{t},t\right)$ equals $\dot{\varphi}_{t}\left(\tilde{x}_{1|t}-\tilde{x}_{0|1,t}\right)$,
as per Eq.~\ref{eq:straight_posterior_velo_1}. It might suggest
that $\tilde{x}_{1|t}=x_{1|t}$ and $\tilde{x}_{0|1,t}=x_{0|1,t}$,
and this impression is correct because $\tilde{x}_{t}$ is a deterministic
transformation of $x_{t}$.

We can also transform the original stochastic process into the straight
process $\tilde{X}_{t}=X_{0}+\frac{\sigma_{t}}{a_{t}}X_{1}$ by setting
$\tilde{X}_{t}$ to $\frac{X_{t}}{a_{t}}$. In this case, the posterior
velocity $v^{*}\left(\tilde{x}_{t},t\right)$ can be computed as follows:
\begin{equation}
v^{*}\left(\tilde{x}_{t},t\right)=\frac{a_{t}v^{*}\left(x_{t},t\right)-\dot{a}_{t}x_{t}}{a_{t}^{2}}\label{eq:straight_velo_scaled_x1}
\end{equation}
After computing $\tilde{x}_{t'}$ via the Euler update in Eq.~\ref{eq:Euler_update_SN},
we retrieve $x_{t'}$ from $\tilde{x}_{t'}$ as $x_{t'}=a_{t}\tilde{x}_{t'}$.

\subsubsection{Straight flows $\rightarrow$ Straight constant-speed flows\label{subsec:Straight-flows-2-straight-constant-speed-flows}}

A straight flow often requires fewer Euler updates than the original
one, especially when the original flow is strongly curved. However,
if a straight flow has varying speeds over time, using large step
sizes will lead to inaccurate updates. Therefore, it is desirable
to make a straight flow constant-speed. Due to space limit, we will
only present the transformation procedure for the straight process
$\tilde{X}_{t}=\left(1-\varphi_{t}\right)X_{0}+\varphi_{t}X_{1}$
here and provide the discussion for $\tilde{X}_{t}=X_{0}+\varphi_{t}X_{1}$
in Appdx.~\ref{subsec:Appdx_Straight2StraightConstant}.

There are two ways to make $\tilde{X}_{t}=\left(1-\varphi_{t}\right)X_{0}+\varphi_{t}X_{1}$
with $\varphi_{t}=\frac{\sigma_{t}}{a_{t}+\sigma_{t}}$ constant-speed.
The first approach is based on the observation that the posterior
flow corresponding to this process has the form $d\tilde{x}_{t}=v^{*}\left(\tilde{x}_{t},t\right)dt$
with $v^{*}\left(\tilde{x}_{t},t\right)=\dot{\varphi}_{t}\left(x_{1|t}-x_{0|1,t}\right)$,
which allows us to rewrite the posterior flow as follows:
\begin{align}
d\tilde{x}_{t} & =\dot{\varphi}_{t}\left(x_{1|t}-x_{0|1,t}\right)dt\label{eq:time_adj_1}\\
 & =\left(x_{1|t}-x_{0|1,t}\right)d\varphi_{t}\label{eq:time_adj_2}
\end{align}
Eqs.~\ref{eq:time_adj_1}, \ref{eq:time_adj_2} suggest that \emph{instead
of moving with non-constant speed when the change of time is constant}
 (i.e., $dt$), we can \emph{move with constant speed when the change
of time is non-constant} (i.e., $d\varphi_{t}$). Therefore, we refer
to this approach as \emph{``time adjustment''}. We can compute $x_{1|t}-x_{0|1,t}$
as $\frac{v^{*}\left(\tilde{x}_{t},t\right)}{\dot{\varphi}_{t}}$
with $v^{*}\left(\tilde{x}_{t},t\right)$ given in Eq.~\ref{eq:straight_velo_itpl}.
We can also compute $x_{1|t}$, $x_{0|1,t}$ directly via $v^{*}\left(x_{t},t\right)$
and $x_{t}$ and then take the difference between the two terms. Both
ways lead to the same result:
\begin{equation}
x_{1|t}-x_{0|1,t}=\frac{\left(a_{t}+\sigma_{t}\right)v^{*}\left(x_{t},t\right)-\left(\dot{a}_{t}+\dot{\sigma}_{t}\right)x_{t}}{a_{t}\dot{\sigma}_{t}-\dot{a}_{t}\sigma_{t}}\label{eq:straight_constant_velo_way_1_velo_net}
\end{equation}
Eq.~\ref{eq:straight_constant_velo_way_1_velo_net} is preferred
if we model $v^{*}\left(x_{t},t\right)$ as $v_{\theta}\left(x_{t},t\right)$.
On the other hand, if we model $x_{1|t}$ as $x_{1,\theta}\left(x_{t},t\right)$
($x_{1,\theta}$ is the noise network $\epsilon_{\theta}$ in diffusion
models), it will be better to compute $x_{1|t}-x_{0|1,t}$ as follows:
\begin{align}
x_{1|t}-x_{0|1,t} & =\frac{\left(a_{t}+\sigma_{t}\right)x_{1|t}-x_{t}}{a_{t}}\label{eq:straight_constant_velo_way_1_noise_net}
\end{align}
Once $x_{1|t}-x_{0|1,t}$ has been computed via either Eq.~\ref{eq:straight_constant_velo_way_1_velo_net}
or Eq.~\ref{eq:straight_constant_velo_way_1_noise_net}, we can perform
an Euler update with the constant-speed velocity $x_{1|t}-x_{0|1,t}$
to transition $\tilde{x}_{t}$ to $\tilde{x}_{t'}$:
\begin{equation}
\tilde{x}_{t'}=\tilde{x}_{t}+\left(\varphi_{t'}-\varphi_{t}\right)\left(x_{1|t}-x_{0|1,t}\right)\label{eq:straight_constant_update_time_adj}
\end{equation}
and then recover $x_{t'}$ from $\tilde{x}_{t'}$ via Eq.~\ref{eq:x_t_recovered}.

It's intriguing to observe the similarity between the Euler updates
in Eqs.~\ref{eq:straight_constant_update_time_adj} and \ref{eq:Euler_update_SN_2},
with the distinction lying in the coefficient of the constant velocity
$x_{1|t}-x_{0|1,t}$, where it is $\left(\varphi_{t'}-\varphi_{t}\right)$
in the former and $\dot{\varphi}_{t}\left(t'-t\right)$ in the latter.
The former discretizes $d\varphi_{t}$, while the latter is merely
the first-order Taylor approximation of the former. To highlight that
Eq.~\ref{eq:straight_constant_update_time_adj} is an update on a
straight constant-speed flow, we present it with new notations as
follows:
\begin{align}
\bar{x}_{t'} & =\bar{x}_{t}+\left(\varphi_{t'}-\varphi_{t}\right)\left(x_{1|t}-x_{0|1,t}\right)\label{eq:Euler_update_time_adjust_1}\\
 & =\bar{x}_{t}+\left(\varphi_{t'}-\varphi_{t}\right)v^{*}\left(\bar{x}_{t},t\right)\label{eq:Euler_update_time_adjust_2}
\end{align}
where $\bar{x}_{t}$ denotes a sample from the straight constant-speed
flow $d\bar{X}_{t}=\left(X_{1}-X_{0}\right)d\varphi_{t}$ with $\bar{X}_{t}=\tilde{X}_{t}$.

The second approach is to transform $\tilde{X}_{t}=\left(1-\varphi_{t}\right)X_{0}+\varphi_{t}X_{1}$
into the process $\bar{X}_{t}=\left(1-t\right)X_{0}+tX_{1}$ by setting
$\bar{X}_{t}=\tilde{X}_{t}+\left(t-\varphi_{t}\right)\left(X_{1}-X_{0}\right)$.
Since $\bar{X}_{t}$ is shifted from $\tilde{X}_{t}$ with the offset
$\left(t-\varphi_{t}\right)\left(X_{1}-X_{0}\right)$, we name this
transformation \emph{``variable shifting''}. In this approach, $v^{*}\left(\bar{x}_{t},t\right)$
also equals $x_{1|t}-x_{0|1,t}$ (details in Appdx.~\ref{subsec:Appdx_Straight2StraightConstant})
and can be approximated via Eq.~\ref{eq:straight_constant_velo_way_1_velo_net}
or Eq.~\ref{eq:straight_constant_velo_way_1_noise_net}.

With $\bar{x}_{t}$ and $v^{*}\left(\bar{x}_{t},t\right)$ in hand,
we perform the following Euler update to obtain $\bar{x}_{t'}$:
\begin{align}
\bar{x}_{t'} & =\bar{x}_{t}+\left(t'-t\right)v^{*}\left(\bar{x}_{t},t\right)\label{eq:Euler_sample_shifting_1}\\
 & =\bar{x}_{t}+\left(t'-t\right)\left(x_{1|t}-x_{0|1,t}\right)\label{eq:Euler_sample_shifting_2}
\end{align}
Subsequently, we retrieve $\tilde{x}_{t'}$ from $\bar{x}_{t'}$ as
follows:
\begin{equation}
\tilde{x}_{t'}=\bar{x}_{t'}-\left(t-\varphi_{t}\right)\left(x_{1|t'}-x_{0|1,t'}\right)\label{eq:x_tilde_t_recovered_sample_shifting}
\end{equation}
Finally, we restore $x_{t'}$ from $\tilde{x}_{t'}$ using Eq.~\ref{eq:x_t_recovered}.

However, the computation of $x_{1|t'}-x_{0|1,t'}$ in Eq.~\ref{eq:x_tilde_t_recovered_sample_shifting}
necessitates knowledge of $x_{t'}$, which is presently unknown. To
circumvent this, we can instead compute $x_{1|t}-x_{0|1,t}$. Given
that $x_{t}$ and $x_{t'}$ belong to the same trajectory, their estimated
boundary pairs $\left(x_{1|t},x_{0|1,t}\right)$ and $\left(x_{1|t'},x_{0|1,t'}\right)$
are expected to be similar. Therefore, $x_{1|t}-x_{0|1,t}$ serves
as an approximation for $x_{1|t'}-x_{0|1,t'}$. Eq.~\ref{eq:x_tilde_t_recovered_sample_shifting}
can be reformulated as follows:
\begin{equation}
\tilde{x}_{t'}\approx\bar{x}_{t'}-\left(t-\varphi_{t}\right)\left(x_{1|t}-x_{0|1,t}\right)\label{eq:x_tilde_t_recovered_approx}
\end{equation}
Interestingly, when the Euler method is used, Eq.~\ref{eq:x_tilde_t_recovered_approx}
renders the ``variable shifting'' approach equivalent to the ``time
adjustment'' approach, which is proven in Appdx.~\ref{subsec:Appdx_Straight2StraightConstant}.

We summarize our proposed transformations in Tables~\ref{tab:straight_transform_summary_x0x1_itpl}
and \ref{tab:straight_transform_summary_x1scale}. To enhance readability
in subsequent sections, we will use the term\emph{ }``\emph{SN }flow/process\emph{''
}to denote a \emph{straight non-constant-speed} flow/process and ``\emph{SC}
flow/process'' to indicate a \emph{straight constant-speed} flow/process.

\begin{table*}
\begin{centering}
{\resizebox{\textwidth}{!}{%
\par\end{centering}
\begin{centering}
\begin{tabular}{|c|c|c|c|}
\hline 
Type & Stochastic process & Sample & Posterior velocity\tabularnewline
\hline 
Ori. & $X_{t}=a_{t}X_{0}+\sigma_{t}X_{1}$ & $x_{t}$ & $v^{*}\left(x_{t},t\right)$\tabularnewline
\hline 
\multirow{2}{*}{SN} & $\tilde{X}_{t}=\left(1-\varphi_{t}\right)X_{0}+\varphi_{t}X_{1}$,
with & \multirow{2}{*}{$\tilde{x}_{t}=\frac{x_{t}}{a_{t}+\sigma_{t}}$} & $v^{*}\left(\tilde{x}_{t},t\right)=\frac{\left(a_{t}+\sigma_{t}\right)v^{*}\left(x_{t},t\right)-\left(\dot{a}_{t}+\dot{\sigma}_{t}\right)x_{t}}{\left(a_{t}+\sigma_{t}\right)^{2}}$,
or\tabularnewline
 & $\varphi_{t}=\frac{\sigma_{t}}{a_{t}+\sigma_{t}}$, $\dot{\varphi}_{t}=\frac{a_{t}\dot{\sigma}_{t}-\dot{a}_{t}\sigma_{t}}{\left(a_{t}+\sigma_{t}\right)^{2}}$ &  & $v^{*}\left(\tilde{x}_{t},t\right)=\frac{a_{t}\dot{\sigma}_{t}-\dot{a}_{t}\sigma_{t}}{\left(a_{t}+\sigma_{t}\right)^{2}}\left(\frac{\left(a_{t}+\sigma_{t}\right)x_{1|t}-x_{t}}{a_{t}}\right)$\tabularnewline
\hline 
\multirow{2}{*}{SC} & $d\bar{X}_{t}=\left(X_{1}-X_{0}\right)d\varphi_{t}$ & $\bar{x}_{t}=\tilde{x}_{t}$ & $v^{*}\left(\bar{x}_{t},t\right)=\frac{\left(a_{t}+\sigma_{t}\right)x_{1|t}-x_{t}}{a_{t}}$,
or\tabularnewline
\cline{2-3} \cline{3-3} 
 & $\bar{X}_{t}=\left(1-t\right)X_{0}+tX_{1}$ & $\bar{x}_{t}=\tilde{x}_{t}+\left(t-\varphi_{t}\right)\left(x_{1}-x_{0}\right)$ & $v^{*}\left(\bar{x}_{t},t\right)=\frac{\left(a_{t}+\sigma_{t}\right)v^{*}\left(x_{t},t\right)-\left(\dot{a}_{t}+\dot{\sigma}_{t}\right)x_{t}}{a_{t}\dot{\sigma}_{t}-\dot{a}_{t}\sigma_{t}}$\tabularnewline
\hline 
\end{tabular}}}
\par\end{centering}
\caption{Summary of the transformations from a (stochastic) process of the
form $X_{t}=a_{t}X_{0}+\sigma_{t}X_{1}$ to the corresponding straight
process of the form $\tilde{X}_{t}=\left(1-\varphi_{t}\right)X_{0}+\varphi_{t}X_{1}$
and straight constant-speed process of either the form $d\bar{X}_{t}=\left(X_{1}-X_{0}\right)d\varphi_{t}$
or the form $\bar{X}_{t}=\left(1-t\right)X_{0}+tX_{1}$.\label{tab:straight_transform_summary_x0x1_itpl}}
\end{table*}

\begin{table*}
\begin{centering}
\begin{tabular}{|c|c|c|c|}
\hline 
Type & Stochastic process & Sample & Posterior velocity\tabularnewline
\hline 
Ori. & $X_{t}=a_{t}X_{0}+\sigma_{t}X_{1}$ & $x_{t}$ & $v^{*}\left(x_{t},t\right)$\tabularnewline
\hline 
\multirow{2}{*}{SN} & $\tilde{X}_{t}=X_{0}+\varphi_{t}X_{1}$, with & \multirow{2}{*}{$\tilde{x}_{t}=\frac{x_{t}}{a_{t}}$} & $v^{*}\left(\tilde{x}_{t},t\right)=\frac{a_{t}v^{*}\left(x_{t},t\right)-\dot{a}_{t}x_{t}}{a_{t}^{2}}$,
or\tabularnewline
 & $\varphi_{t}=\frac{\sigma_{t}}{a_{t}}$, $\dot{\varphi}_{t}=\frac{a_{t}\dot{\sigma}_{t}-\dot{a}_{t}\sigma_{t}}{a_{t}^{2}}$ &  & $v^{*}\left(\tilde{x}_{t},t\right)=\frac{a_{t}\dot{\sigma}_{t}-\dot{a}_{t}\sigma_{t}}{a_{t}^{2}}x_{1|t}$\tabularnewline
\hline 
\multirow{2}{*}{SC} & $d\bar{X}_{t}=X_{1}d\varphi_{t}$ & $\bar{x}_{t}=\tilde{x}_{t}$ & $v^{*}\left(\bar{x}_{t},t\right)=x_{1|t}$, or\tabularnewline
\cline{2-3} \cline{3-3} 
 & $\bar{X}_{t}=X_{0}+tX_{1}$ & $\bar{x}_{t}=\tilde{x}_{t}+\left(t-\varphi_{t}\right)x_{1}$ & $v^{*}\left(\bar{x}_{t},t\right)=\frac{a_{t}v^{*}\left(x_{t},t\right)-\dot{a}_{t}x_{t}}{a_{t}\dot{\sigma}_{t}-\dot{a}_{t}\sigma_{t}}$\tabularnewline
\hline 
\end{tabular}
\par\end{centering}
\caption{Summary of the transformations from a (stochastic) process of the
form $X_{t}=a_{t}X_{0}+\sigma_{t}X_{1}$ to the corresponding straight
process of the form $\tilde{X}_{t}=X_{0}+\varphi_{t}X_{1}$ and straight
constant-speed process of either the form $d\bar{X}_{t}=X_{1}d\varphi_{t}$
or the form $\bar{X}_{t}=X_{0}+tX_{1}$.\label{tab:straight_transform_summary_x1scale}}
\end{table*}

\subsection{Ensuring the numerical robustness of our method\label{subsec:Numerical-robustness}}

One factor affecting the quality of samples generated by our method
is the numerical robustness of some operations in our proposed SN
and SC transformations. As shown in Tables~\ref{tab:straight_transform_summary_x0x1_itpl},
\ref{tab:straight_transform_summary_x1scale}, the computations of
$\tilde{x}_{t}$, $v^{*}\left(\tilde{x}_{t},t\right)$, $\bar{x}_{t}$,
and $v^{*}\left(\bar{x}_{t},t\right)$ from $x_{t}$, $v^{*}\left(x_{t},t\right)$
(or $x_{1|t}$) often involves division by some time-dependent terms,
which can lead to a severe numerical problem if these terms approach
0 at a specific time $t$. We can mitigate this problem by clipping
the denominators in the formulas of $\tilde{x}_{t}$, $v^{*}\left(\tilde{x}_{t},t\right)$,
$\bar{x}_{t}$ and $v^{*}\left(\bar{x}_{t},t\right)$ to a small value.
For example, $v^{*}\left(\bar{x}_{t},t\right)$ computed via Eq.~\ref{eq:straight_constant_velo_way_1_noise_net}
has $a_{t}$ in its denominator, making it undefined at $t=1$ since
$a_{1}=0$ by design. To ensure numerical stability, we adjust $a_{t}$
to $\text{sign}\left(a_{t}\right)*\max\left(\left|a_{t}\right|,\epsilon\right)$
where $\text{sign}\left(a_{t}\right)$ = 1 if $a_{t}\geq0$ and -1
otherwise and $\epsilon$ is a small positive number. Through our
experiments, we have observed that using this simple clipping technique
is sufficient to maintain numerically stable updates along SN and
SC flows in most practical scenarios. It is worth noting that constraining
the time $t$ within the range $\left[0,1-\epsilon\right]$ does \emph{not}
always guarantee robustness, particularly in cases where $a_{1}$,
as a function of $\epsilon$, can become exceedingly small without
reaching precisely 0. For example, if $a_{t}=\left(1-t\right)^{5}$
and $\epsilon=1e^{-3}$ then $a_{1}=\epsilon^{5}=1e^{-15}$, which,
although not exactly 0, is extremely small and can cause numerical
instability.

\subsection{Using higher-order numerical solvers on straight constant-speed flows\label{subsec:Higher-order-numerical-solvers}}

In Section~\ref{subsec:Transforming-a-flow}, we have discussed how
to transform a non-straight flow into a straight constant-speed (SC)
flow and perform Euler updates on the SC flow. While achieving a perfectly
straight constant-speed flow is ideal, it is seldom realized in practice
due to the inherent randomness of the independent coupling typically
used in training the velocity/noise model (Eqs.~\ref{eq:noise_matching},
\ref{eq:velocity_matching_loss_reframed}). The straightness quality
of an SC flow can be improved by ``reflowing'' the original flow
\cite{liu2022flow}, i.e., retraining the velocity network $v_{\theta}$
with the deterministic coupling defined by $v_{\theta}$. Unfortunately,
this approach is resource-intensive, particularly with large models
and extensive training data. A more cost-effective strategy is to
devise better numerical solvers to enhance accuracy and speed of transitioning
along SC flows. This initiative begins by exploring well-established
higher-order ODE solvers such as Runge Kutta and linear multi-step
methods as alternatives to the first-order Euler solver. Notably,
integrating these high-order solvers into our framework requires little
\emph{yet non-trivial} modification to the update formula in Eq.~\ref{eq:Euler_update_time_adjust_2}.
For detailed mathematical derivations, please refer to Appdx.~\ref{sec:High-order-Numerical-Solvers-on-SC-Flows}.
Our proposed computationally-efficient high-order solvers, requiring
only a single model evaluation per update step, include the second-
and third-order Adams-Bashforth methods (SC-AB2, SC-AB3) and their
predictor-corrector variants (SC-AB1AM2, SC-AB2AM2, SC-AB2AM3, SC-AB3AM3).

\subsection{A general formula for straight stochastic processes\label{subsec:A-general-formula-for-straight}}

We can generalize our analysis in Section~\ref{subsec:Transforming-a-flow}
to derive a general formula for straight flows (processes). Clearly,
the velocity $v^{*}$ of a straight flow should have its direction
unchanged over time. This implies that $v^{*}$ should have the form
$v^{*}\left(x_{t},t\right)=\dot{\varphi}_{t}\Expect_{x_{0},x_{1}|x_{t}}\left[f\left(x_{0}\right)+h\left(x_{0},x_{1}\right)+g\left(x_{1}\right)\right]$
where $f$, $h$, $g$ are arbitrary functions. This posterior velocity
corresponds to the following stochastic process: 
\begin{equation}
X_{t}=\varphi_{t}\oplus\left(f\left(X_{0}\right)+h\left(X_{0},X_{1}\right)+g\left(X_{1}\right)\right)\label{eq:general_straight_process}
\end{equation}
where the operator $\oplus$ denotes the sum of the products between
$\varphi_{t}$ with added constants and all additive terms in $f$,
$h$, and $g$. For example, the expression $\varphi_{t}\oplus\left(X_{1}-2X_{0}^{2}X_{1}-3X_{0}^{3}\right)$
is equivalent to $\left(\varphi_{t}+b\right)X_{1}-2\left(\varphi_{t}+c\right)X_{0}^{2}X_{1}-3\left(\varphi_{t}+d\right)X_{0}^{3}$
where $b$, $c$, $d$ are constants w.r.t. $t$. Normally, we want
that at $t=0$, $X_{\ensuremath{t}}$ does not depend on $X_{1}$
and at $t=1$, $X_{\ensuremath{t}}$ does not depend on $X_{0}$.
Therefore, to simplify our formula, we discard the term $h\left(X_{0},X_{1}\right)$
in Eq.~\ref{eq:general_straight_process} and rewrite it as:
\begin{equation}
X_{t}=\varphi_{t}\oplus\left(f\left(X_{0}\right)+g\left(X_{1}\right)\right)\label{eq:simplified_straight_process}
\end{equation}
Eq.~\ref{eq:simplified_straight_process} provide a \emph{general}
formula for straight stochastic processes. If we further assume that
$f\left(X_{0}\right)\equiv-X_{0}$ and $g\left(X_{1}\right)\equiv X_{1}$,
Eq.~\ref{eq:simplified_straight_process} becomes:
\begin{align}
X_{t} & =\varphi_{t}\oplus\left(-X_{0}+X_{1}\right)\\
 & =\left(c-\varphi_{t}\right)X_{0}+\varphi_{t}X_{1}\label{eq:general_itpl_straight_process}
\end{align}
where $c$ is a constant w.r.t. $t$. It is reasonable to set $c$
to 1 because at $t=0$, $\varphi_{0}=0$ and the initial distribution
aligns with $p_{0}$. This means Eq.~\ref{eq:general_itpl_straight_process}
can be written as:
\begin{equation}
X_{t}=\left(1-\varphi_{t}\right)X_{0}+\varphi_{t}X_{1}
\end{equation}
which give us the same formula as Eq.~\ref{eq:straight_stochastic_1}.
We can also derive the formula in Eq.~\ref{eq:straight_stochastic_2}
by considering $v^{*}\left(x_{t},t\right)=\dot{\varphi}_{t}\Expect_{x_{0},x_{1}|x_{t}}\left[g\left(x_{1}\right)\right]$
with $g\left(X_{1}\right)\equiv X_{1}$.

\subsection{Flowing along arbitrary paths\label{subsec:Flowing-along-arbitrary-paths}}

An important extension of our proposed transformations in Section~\ref{subsec:Transforming-a-flow}
is that if we have learned the posterior flow of a linear stochastic
process $X_{t}=a_{t}X_{0}+\sigma_{t}X_{1}$, we can \emph{simulate}
the posterior flow of a \emph{different} linear stochastic process
described by $X_{t}=b_{t}X_{0}+\zeta_{t}X_{1}$ \emph{without training
a new model}. This can be done in two ways. In the first approach,
we simply perform updates along the posterior flow of the original
process and transform samples $x_{t}$ in this flow into samples $\hat{x}_{t}$
in the posterior flow of the target process. This transformation aligns
with the transformation of the original process $X_{t}=a_{t}X_{0}+\sigma_{t}X_{1}$
into the target process $\hat{X}_{t}=b_{t}X_{0}+\zeta_{t}X_{1}$ which
is described by the equation:
\begin{equation}
\hat{X}_{t}=\frac{b_{t}}{a_{t}}X_{t}+\left(\zeta_{t}-\frac{\sigma_{t}b_{t}}{a_{t}}\right)X_{1}\label{eq:original_process_to_target_process}
\end{equation}
Eq.~\ref{eq:original_process_to_target_process} suggests that $\hat{x}_{t}$
can be computed from $x_{t}$ as follows:
\begin{equation}
\hat{x}_{t}=\frac{b_{t}}{a_{t}}x_{t}+\left(\zeta_{t}-\frac{\sigma_{t}b_{t}}{a_{t}}\right)x_{1|t}\label{eq:original_sample_to_target_sample}
\end{equation}
where $x_{1|t}$ can be either modeled as $x_{1,\theta}\left(x_{t},t\right)$
or computed from $v_{\theta}\left(x_{t},t\right)$ and $x_{t}$ using
Eq.~\ref{eq:nonparam_velo_xt_x1}. The main drawback of this approach
is sampling inefficiency if sampling via the original flow requires
much more steps than via the target flow. The second approach can
overcome this limitation by first transforming the original flow into
a straight constant-speed (SC) flow, performing updates along the
SC flow and then mapping the updated samples $\bar{x}_{t'}$ in the
SC flow into samples $\hat{x}_{t'}$ in the target flow. A general
formula for computing $\hat{x}_{t'}$ from $\bar{x}_{t'}$ is:
\begin{equation}
\hat{x}_{t'}\approx k_{t'}\left(\bar{x}_{t'}-\left(t'-\frac{\zeta_{t'}}{k_{t'}}\right)v^{*}\left(\bar{x}_{t},t\right)\right)\label{eq:SC_sample_to_target_sample}
\end{equation}
where $k_{t}$ is $b_{t}+\zeta_{t}$ for the SC process $\bar{X}_{t}=\left(1-t\right)X_{0}+tX_{1}$
and $b_{t}$ for the SC process $\bar{X}_{t}=X_{0}+tX_{1}$.

Our study above also suggests that the SC process $X_{t}=\left(1-t\right)X_{0}+tX_{1}$
should be regarded as the \emph{standard form} of the class of linear
stochastic processes $X_{t}=a_{t}X_{0}+\sigma_{t}X_{1}$ because this
process can easily be converted into other processes in the class
via the combination of ``variable scaling'' and ``variable shifting''
transformations, in the same manner as the standard Gaussian distribution
$\Normal\left(0,\mathrm{I}\right)$ is transformed into other Gaussian
distributions of the form $\Normal\left(\mu,\sigma^{2}\mathrm{I}\right)$.
The SC process $X_{t}=X_{0}+tX_{1}$ is \emph{not} qualified for being
standard because the coefficient of $X_{0}$ is not 0 at $t=1$.

\section{Experiments}

Our experiment encompasses both toy and real image datasets. The former
serves to illustrate and validate our theoretical analyses presented
in Section~\ref{sec:Method}, while the latter demonstrates the effectiveness
of our proposed high-order solvers on SC flows. Due to space constraint,
we only provide a selection of results here and defer the remaining
results in Appdx.~\ref{sec:Additional-Experimental-Results}.

\subsection{2D Toy Dataset}

We created a simple 2D toy dataset in which $p_{0}$ and $p_{1}$
are mixtures of three and two Gaussian distributions respectively
as shown in Fig.~\ref{fig:Toy_ThirdDegree} (top row). Details about
the means and covariance matrices of these Gaussian distributions
can be found in Appdx.~\ref{subsec:2D-Toy-Dataset-Settings}. Our
experiment involves three stochastic processes of the form $X_{t}=a_{t}X_{0}+\sigma_{t}X_{1}$,
where the distributions of $X_{0}$ and $X_{1}$ correspond to $p_{0}$
and $p_{1}$, respectively. For clarity, we coin these processes as
``VP-SDE-like'', ``third-degree'', and ``fifth-degree'' processes.
The ``VP-SDE-like'' process share the same $a_{t}$ and $\sigma_{t}$
as VP-SDE (refer to Table~\ref{tab:velocity-formulas-summary}).
The ``third-degree'' process features $a_{t}$ and $\sigma_{t}$
as third degree polynomials of $t$, with $a_{t}=3\left(1-t\right)^{3}-6\left(1-t\right)^{2}+4\left(1-t\right)$
and $\sigma_{t}=2t^{3}-3t+2$. Similarly, the ``fifth-degree'' process
features $a_{t}$ and $\sigma_{t}$ as fifth degree polynomials of
$t$, with $a_{t}=\left(1-t\right)^{5}$ and $\sigma_{t}=t^{5}$.
We model the posterior flow of these processes by learning either
$v_{\theta}$ or $x_{1,\theta}$. The training details are provided
in Appdx.~\ref{subsec:2D-Toy-Dataset-Settings}. After training,
we generate samples of $X_{0}$ by starting with samples of $X_{1}$
and performing Euler updates along the \emph{posterior}, the transformed
\emph{straight} (SN), and \emph{straight constant-speed} (SC) flows
of the original process. Unless explicitly stated otherwise, the SN
and SC flows are characterized by the equations $\tilde{X}_{t}=\left(1-\varphi_{t}\right)X_{0}+\varphi X_{1}$
and $d\bar{X}_{t}=\left(X_{1}-X_{0}\right)d\varphi_{t}$, respectively,
with $\varphi_{t}=\frac{\sigma_{t}}{a_{t}+\sigma_{t}}$. The small
number $\epsilon$ for ensuring numerical stability (Section~\ref{subsec:Numerical-robustness})
is set to $1e^{-3}$.

\subsubsection{Empirical analysis of Euler solvers on posterior, SN, and SC flows}

\begin{figure*}
\begin{centering}
{\resizebox{\textwidth}{!}{%
\par\end{centering}
\begin{centering}
\begin{tabular}{ccccccc}
\multicolumn{7}{c}{\includegraphics[height=0.04\textwidth]{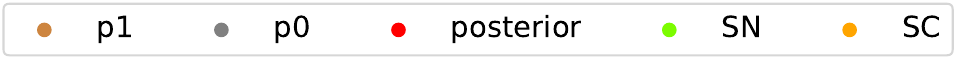}\hspace{0.02\textwidth}\includegraphics[height=0.04\textwidth]{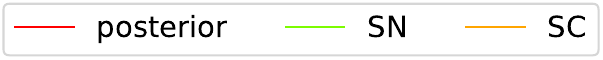}}\tabularnewline
\includegraphics[width=0.22\textwidth]{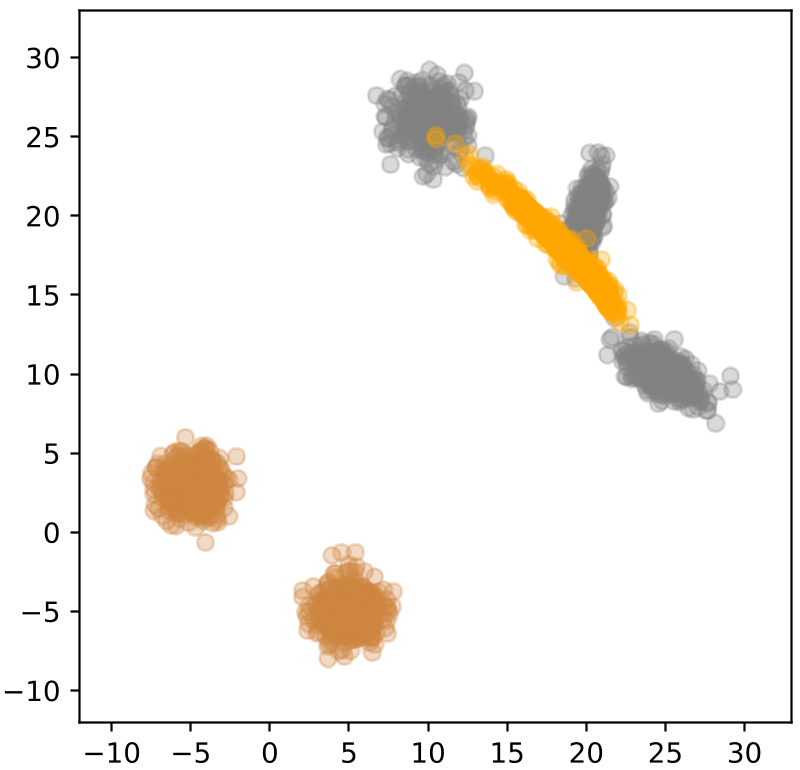} & \includegraphics[width=0.22\textwidth]{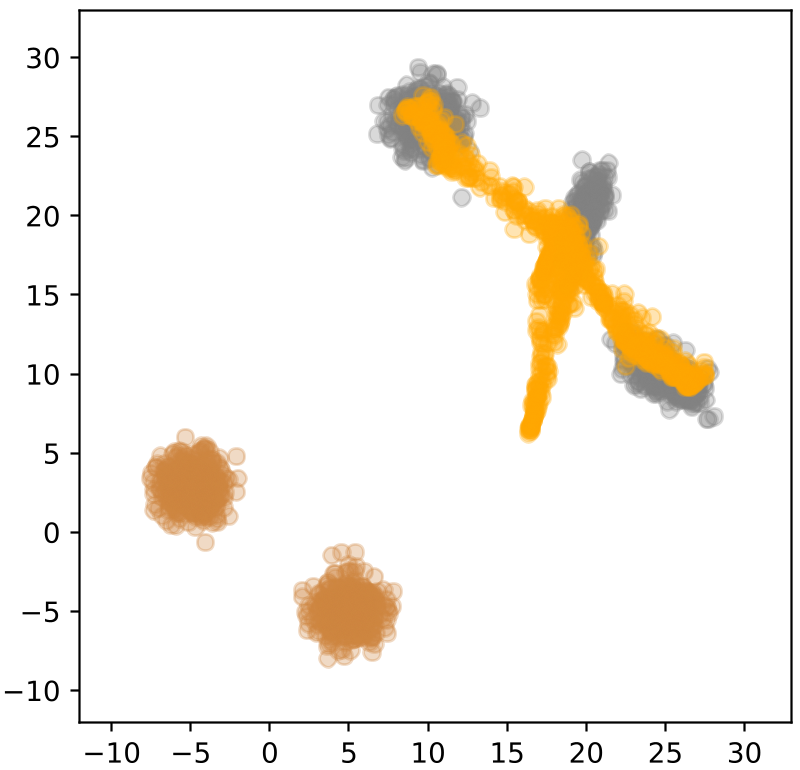} & \includegraphics[width=0.22\textwidth]{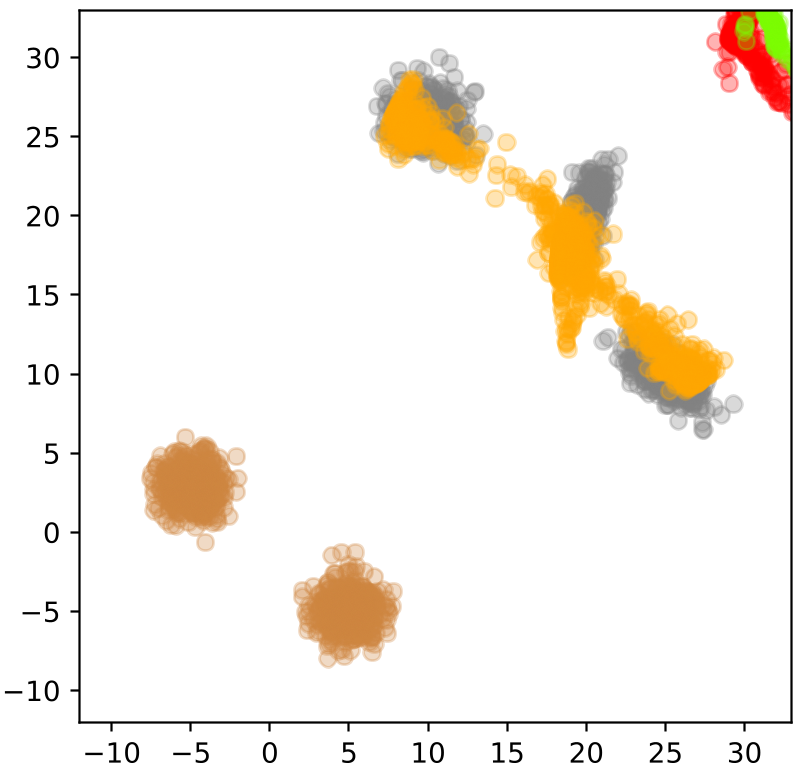} & \includegraphics[width=0.22\textwidth]{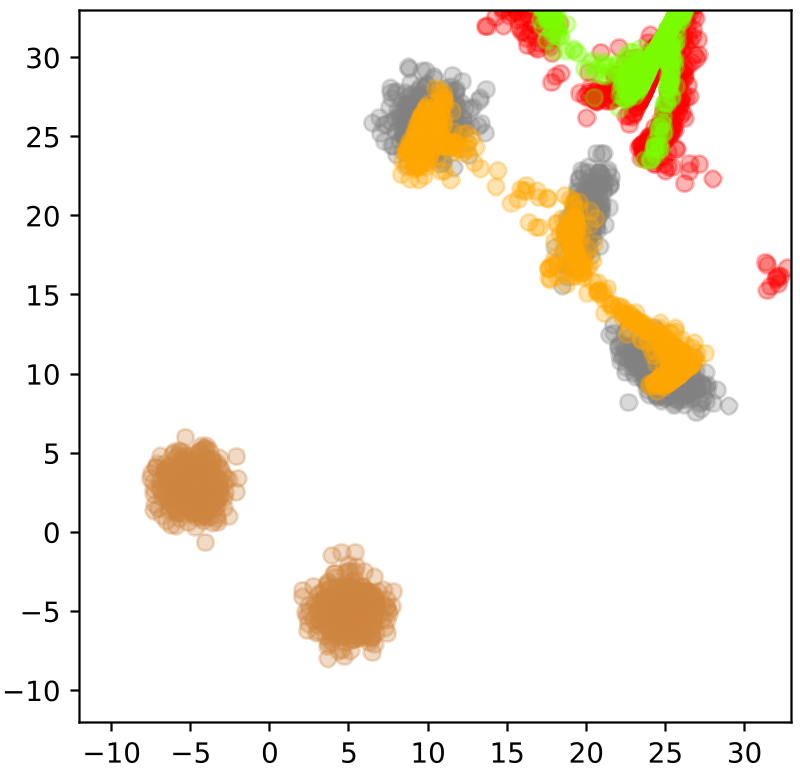} & \includegraphics[width=0.22\textwidth]{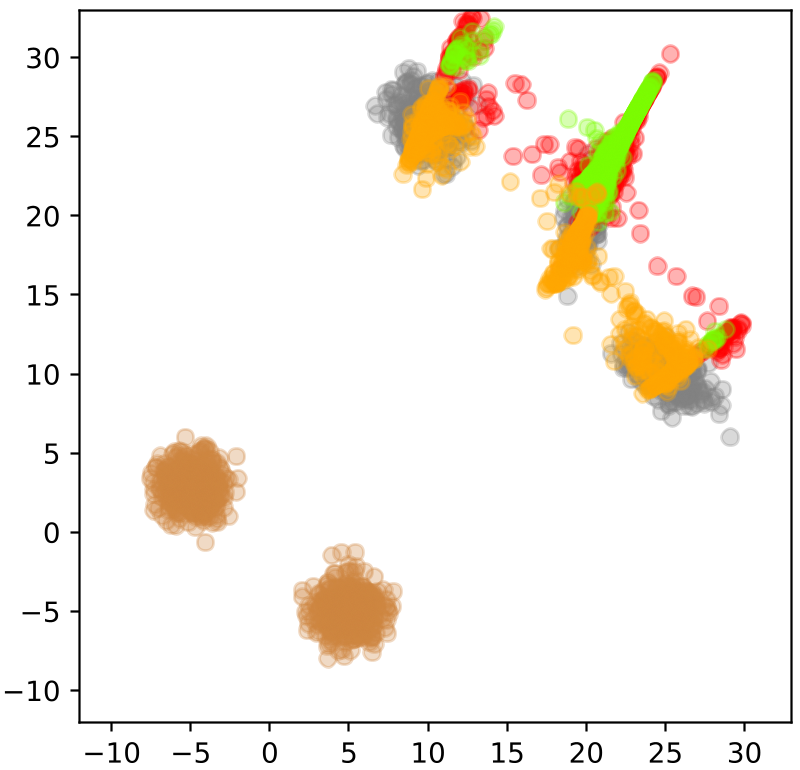} & \includegraphics[width=0.22\textwidth]{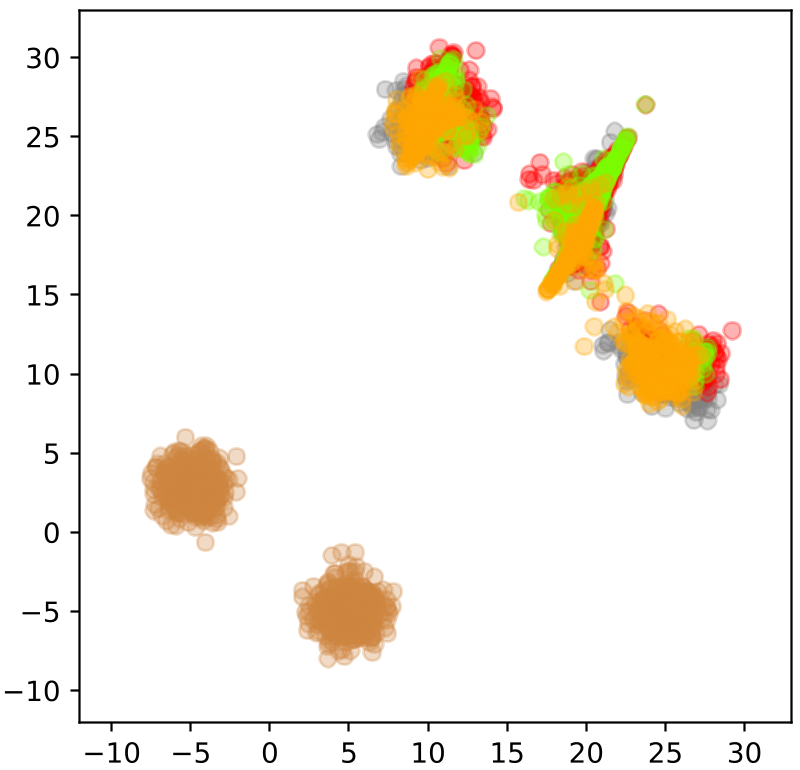} & \includegraphics[width=0.22\textwidth]{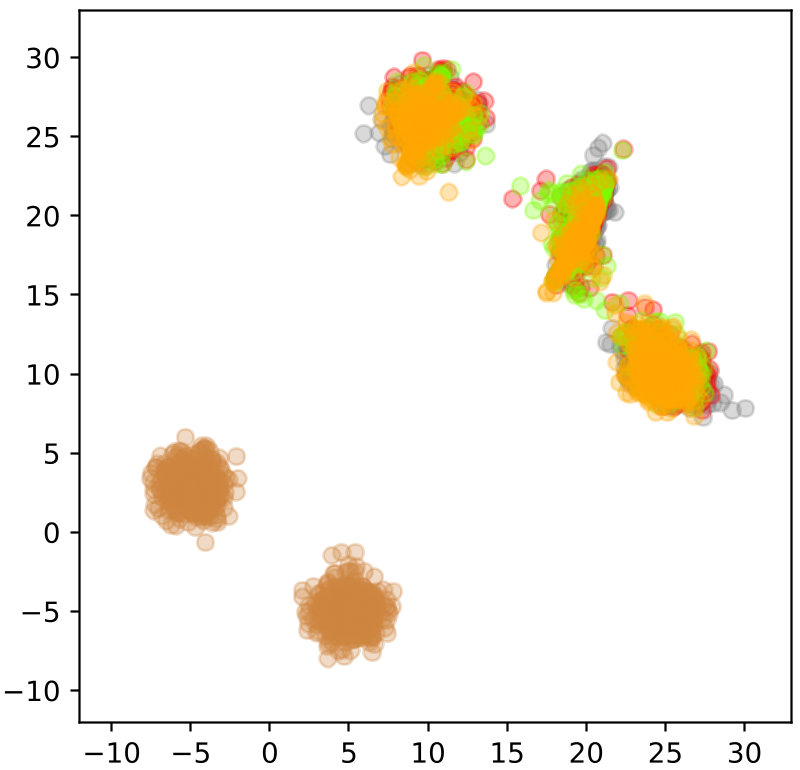}\tabularnewline
\includegraphics[width=0.22\textwidth]{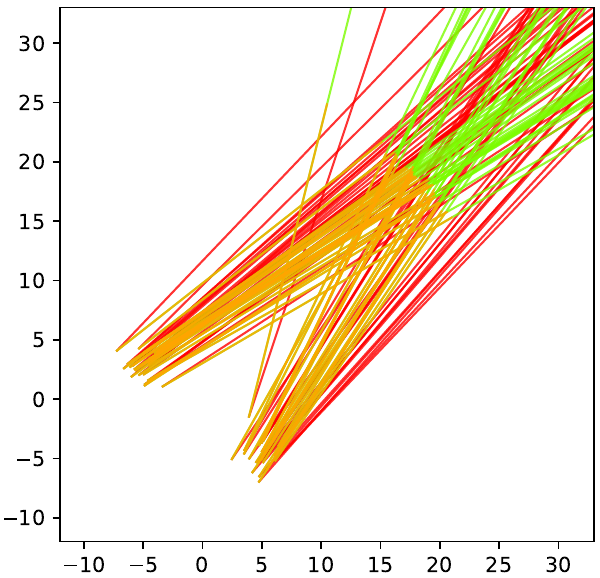} & \includegraphics[width=0.22\textwidth]{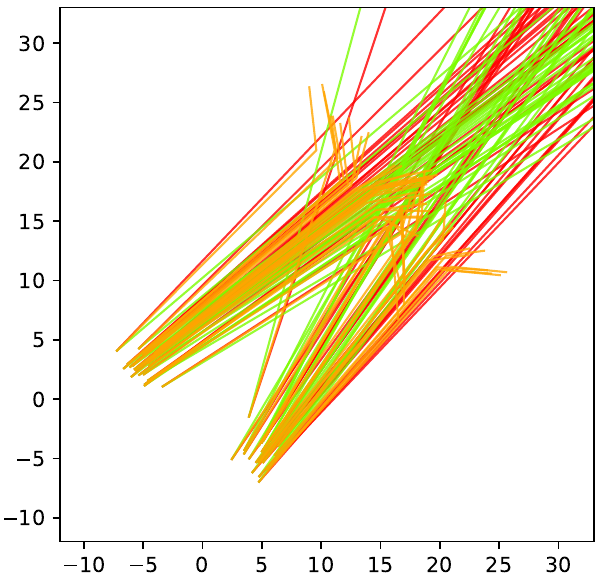} & \includegraphics[width=0.22\textwidth]{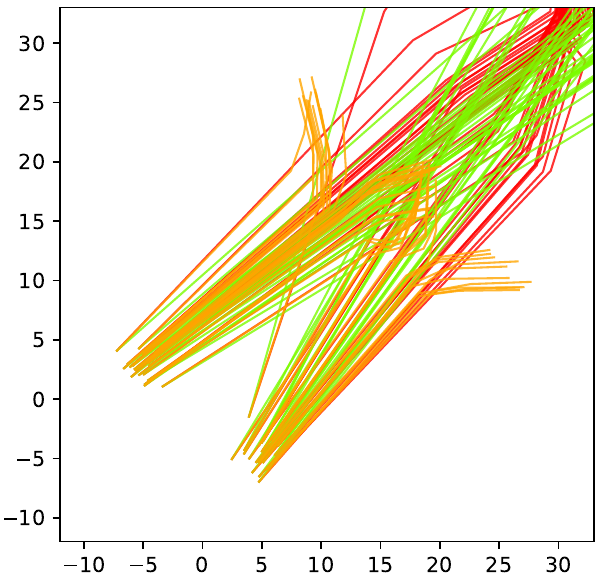} & \includegraphics[width=0.22\textwidth]{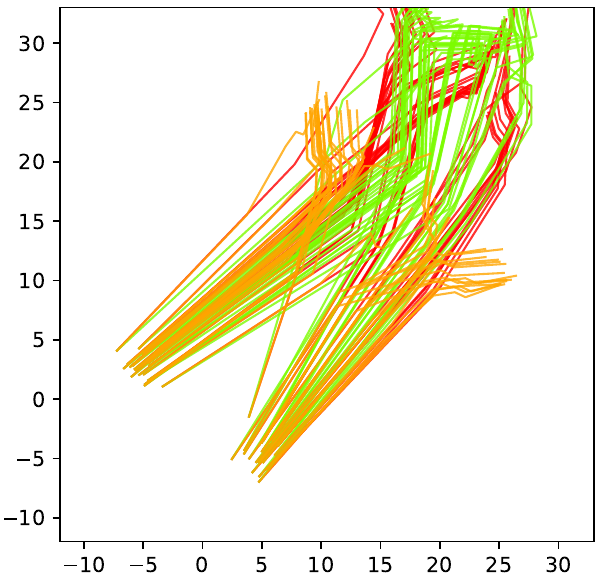} & \includegraphics[width=0.22\textwidth]{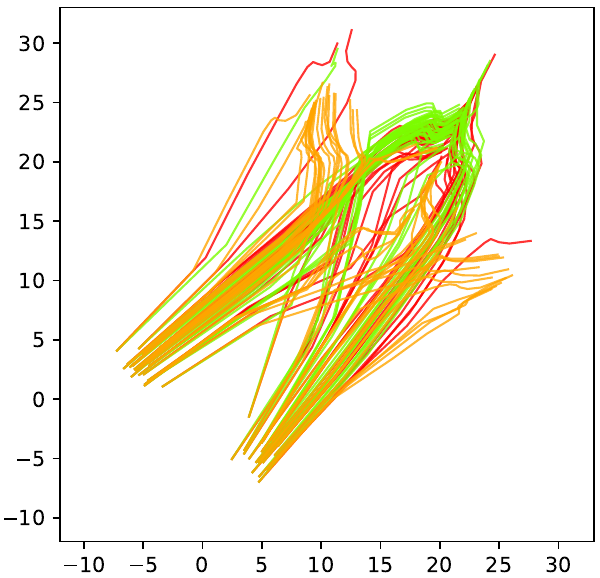} & \includegraphics[width=0.22\textwidth]{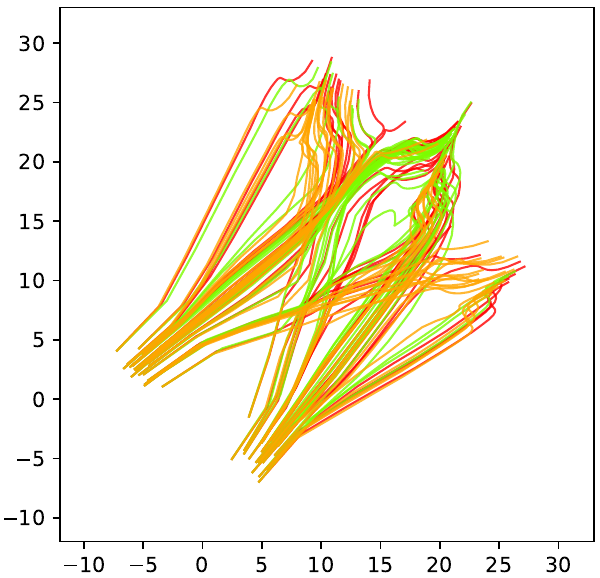} & \includegraphics[width=0.22\textwidth]{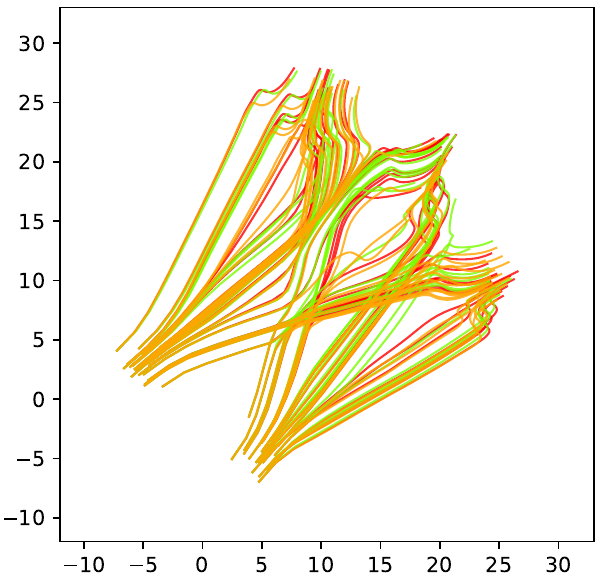}\tabularnewline
\includegraphics[width=0.22\textwidth]{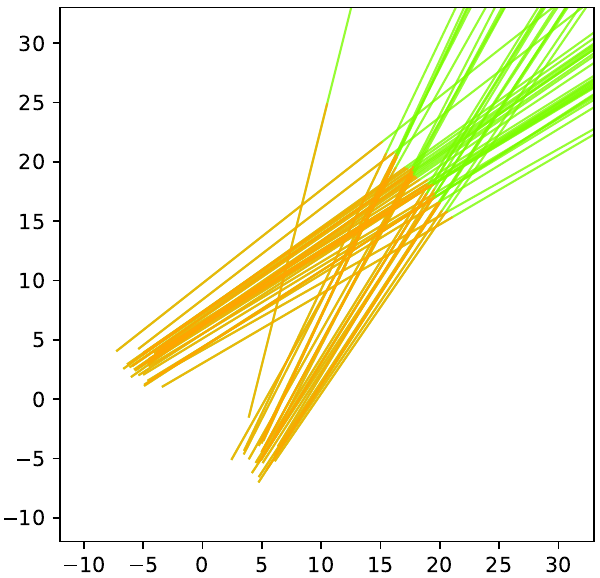} & \includegraphics[width=0.22\textwidth]{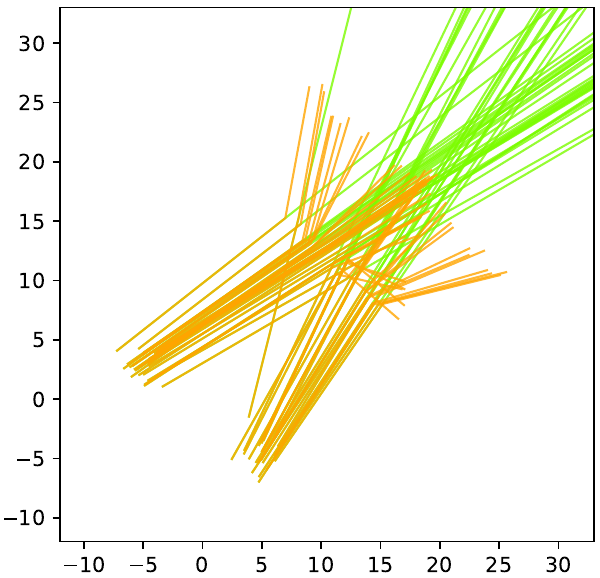} & \includegraphics[width=0.22\textwidth]{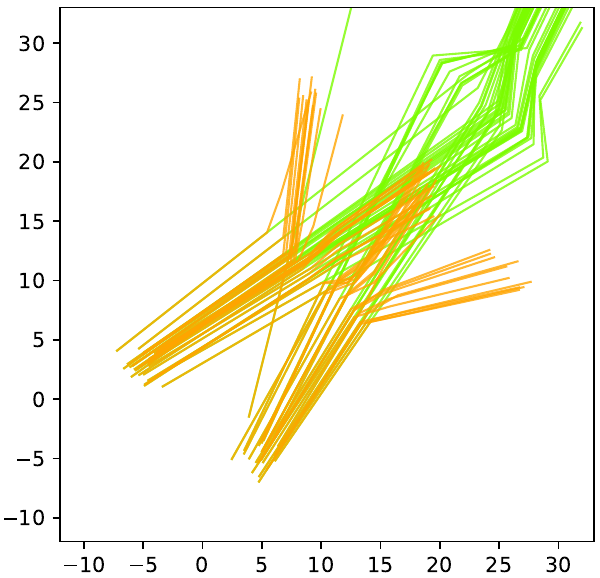} & \includegraphics[width=0.22\textwidth]{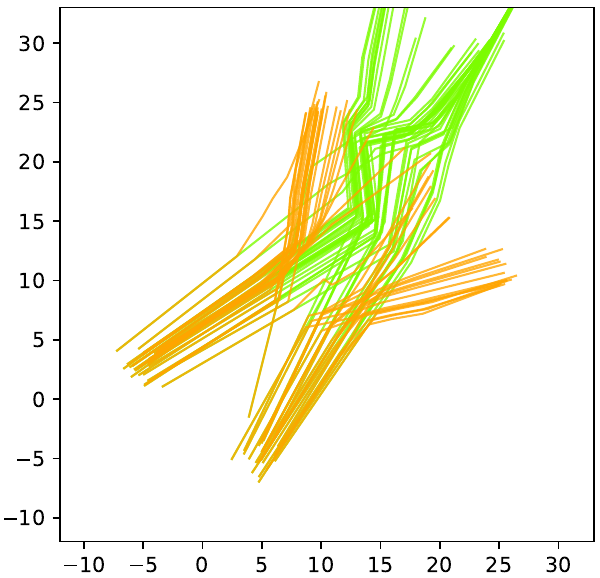} & \includegraphics[width=0.22\textwidth]{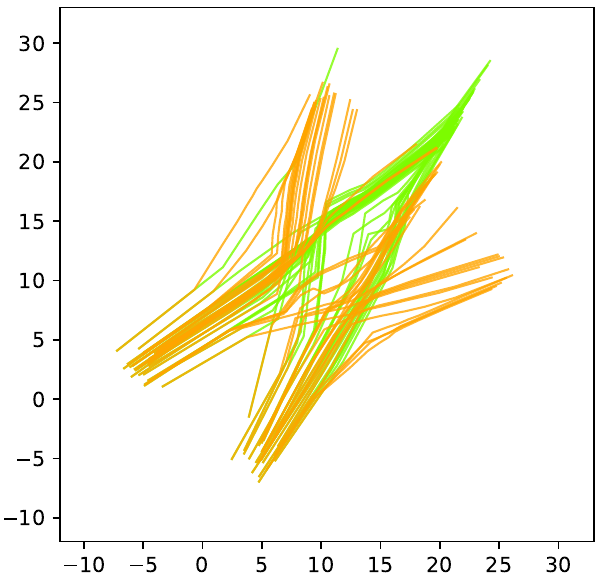} & \includegraphics[width=0.22\textwidth]{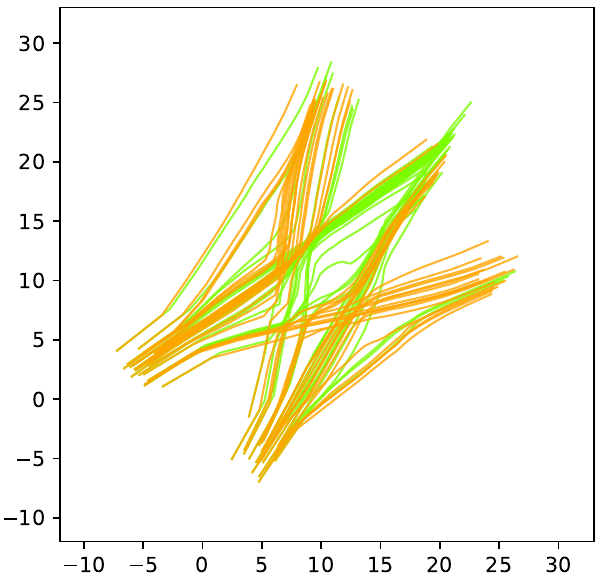} & \includegraphics[width=0.22\textwidth]{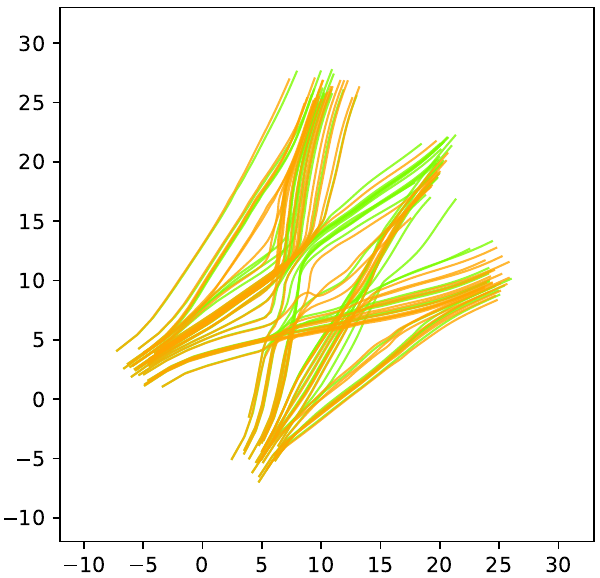}\tabularnewline
$N=1$ & $N=2$ & $N=3$ & $N=5$ & $N=10$ & $N=20$ & $N=50$\tabularnewline
\end{tabular}}}
\par\end{centering}
\caption{Starting from samples of \textcolor{brown}{$p_{1}$}, we generate
samples of \textcolor{gray}{$p_{0}$} by performing $N$ Euler updates
along the \textcolor{red}{posterior} flow and its transformed \textcolor{lawngreen}{SN},
\textcolor{orange}{SC} flows of the \emph{\textquotedblleft third-degree\textquotedblright}
stochastic process. The top and middle rows show generated samples
and $x_{t}$-trajectories of the three flows, respectively. The bottom
row shows the $\tilde{x}_{t}$-trajectories of the SN flow and the
$\bar{x}_{t}$-trajectories of the SC flow.\label{fig:Toy_ThirdDegree}}
\end{figure*}

\begin{figure*}
\begin{centering}
{\resizebox{\textwidth}{!}{%
\par\end{centering}
\begin{centering}
\begin{tabular}{ccccccc}
\multicolumn{7}{c}{\includegraphics[height=0.04\textwidth]{asset/ThreeGauss/legend_sample_with_p0_p1}\hspace{0.02\textwidth}\includegraphics[height=0.04\textwidth]{asset/ThreeGauss/legend_traj}}\tabularnewline
\includegraphics[width=0.22\textwidth]{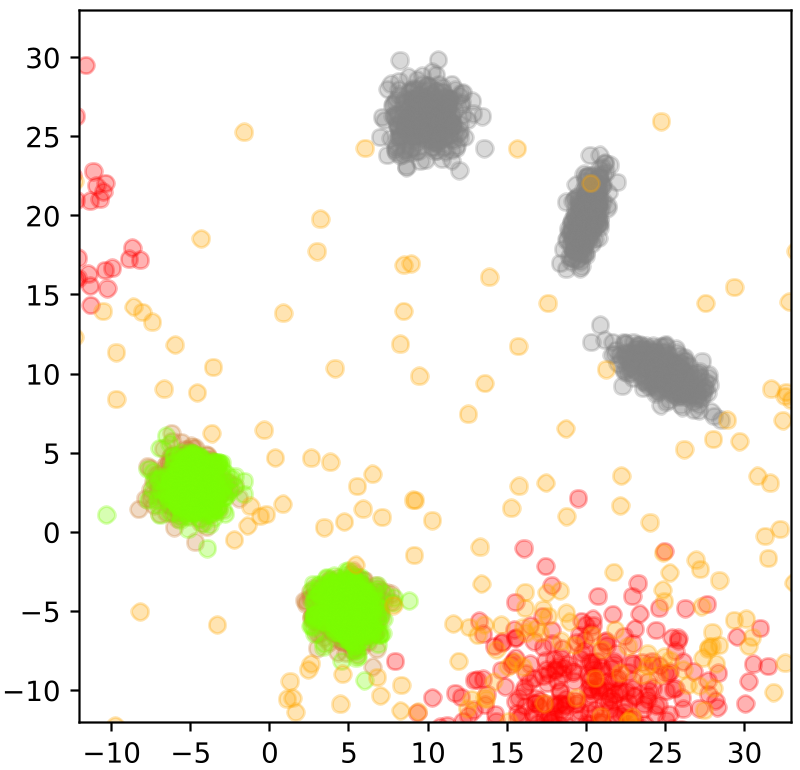} & \includegraphics[width=0.22\textwidth]{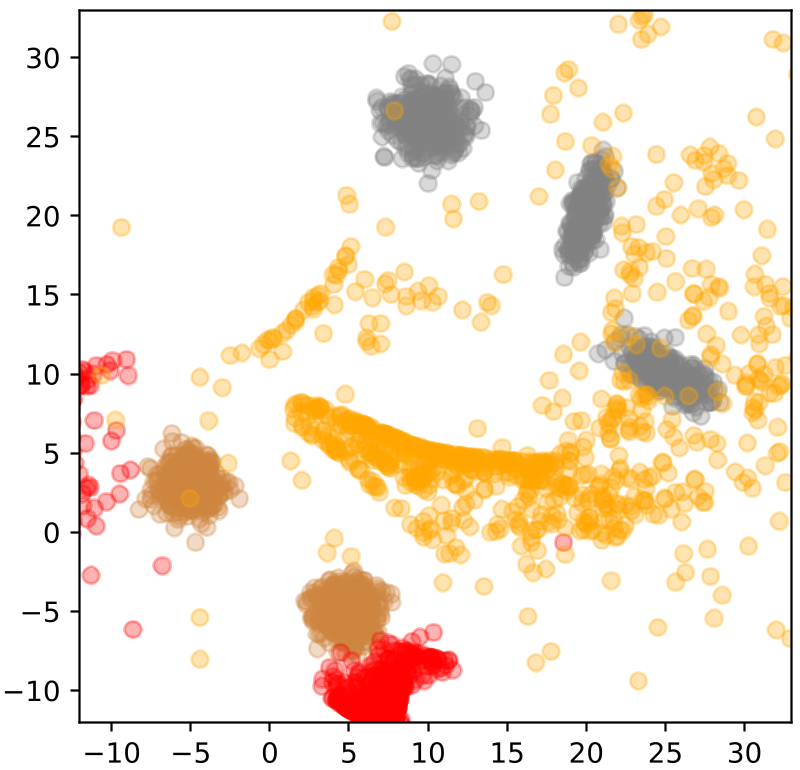} & \includegraphics[width=0.22\textwidth]{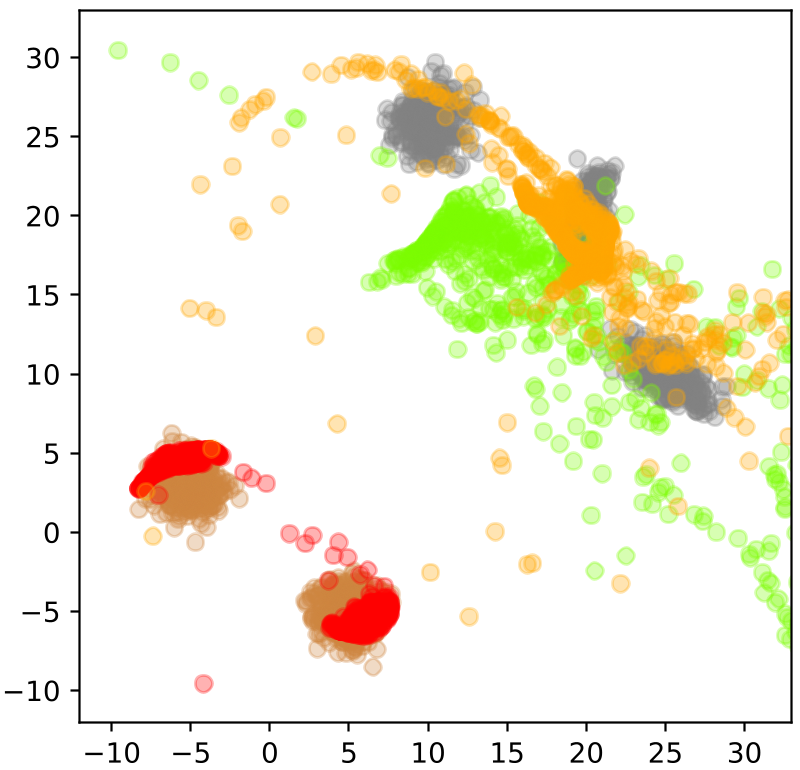} & \includegraphics[width=0.22\textwidth]{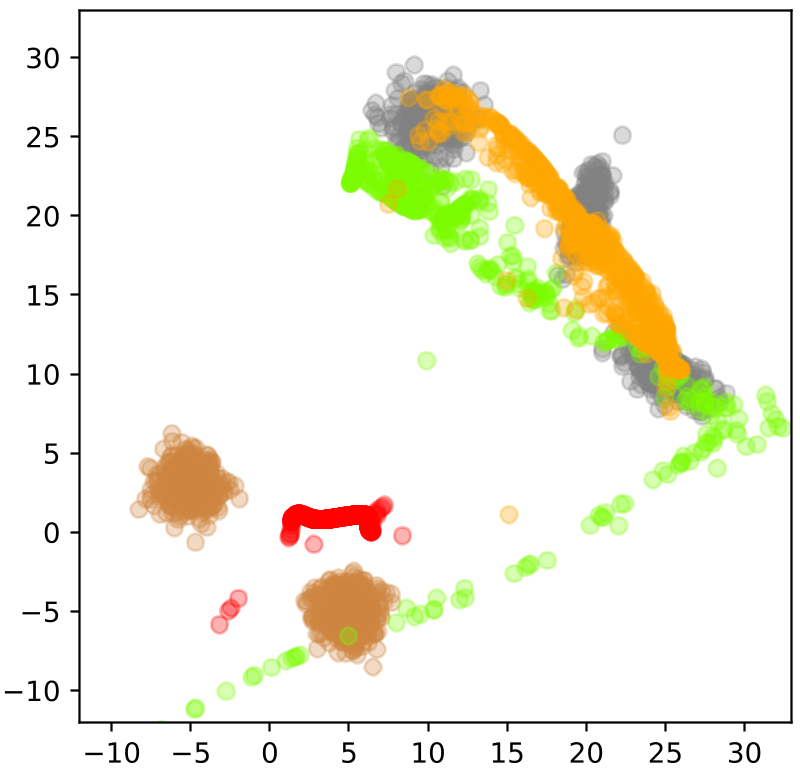} & \includegraphics[width=0.22\textwidth]{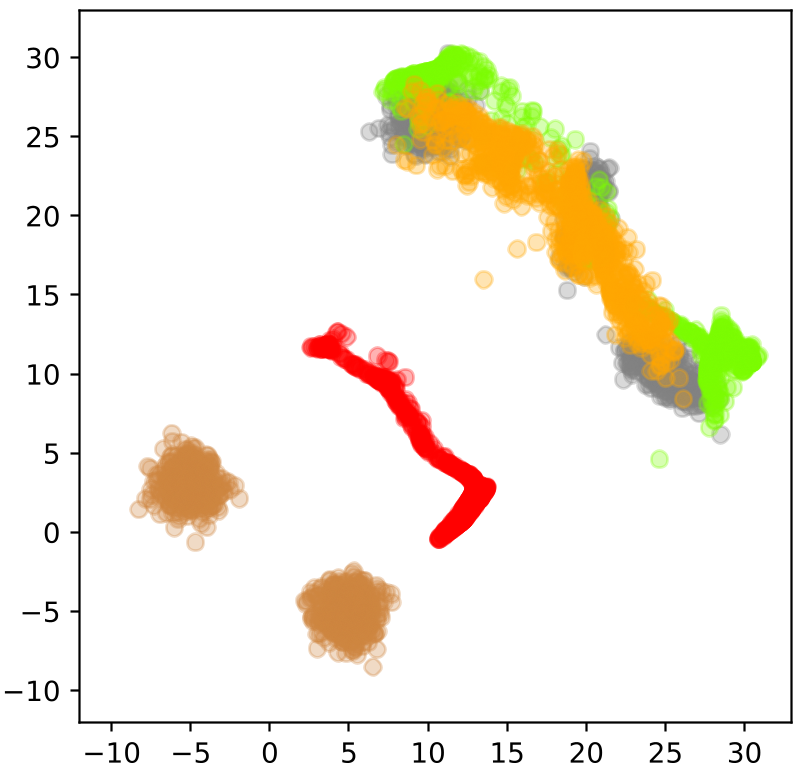} & \includegraphics[width=0.22\textwidth]{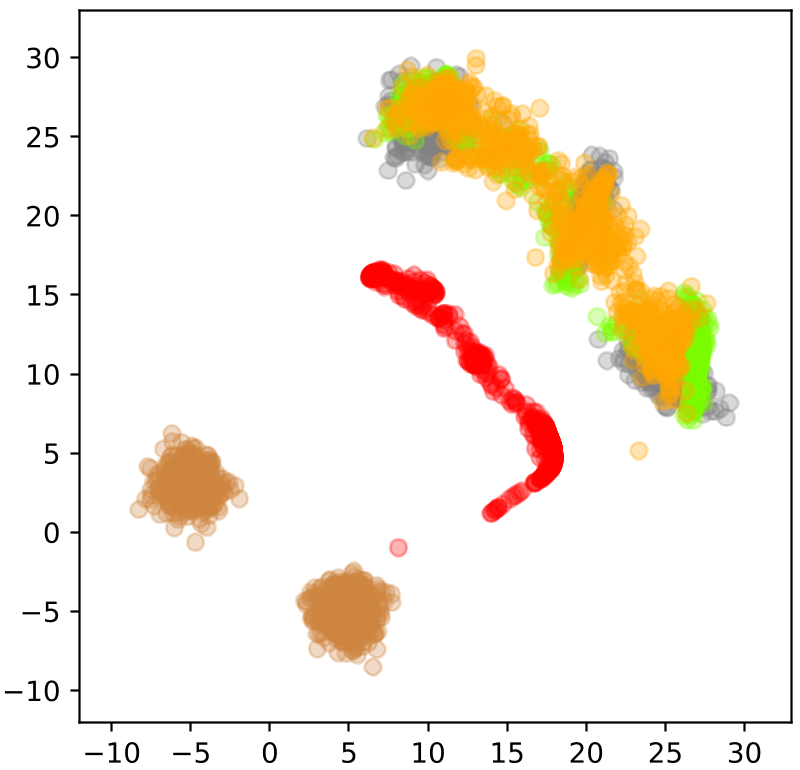} & \includegraphics[width=0.22\textwidth]{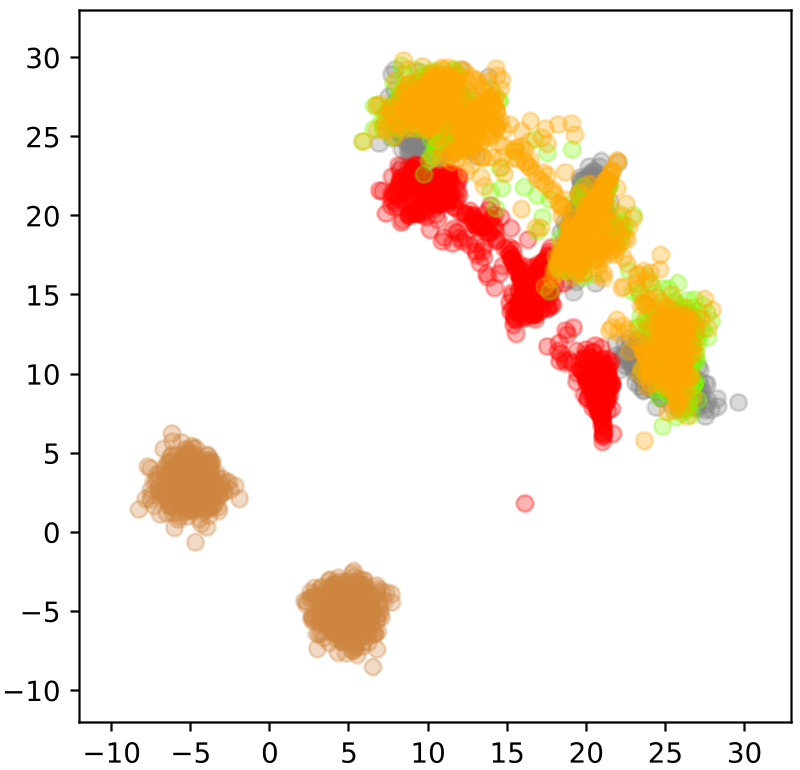}\tabularnewline
\includegraphics[width=0.22\textwidth]{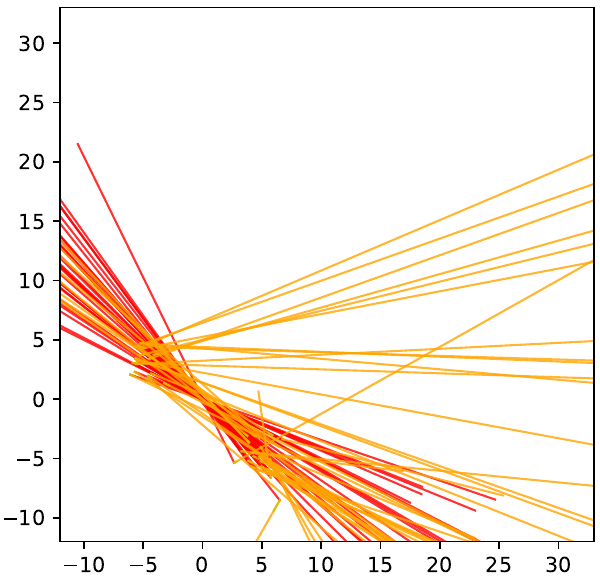} & \includegraphics[width=0.22\textwidth]{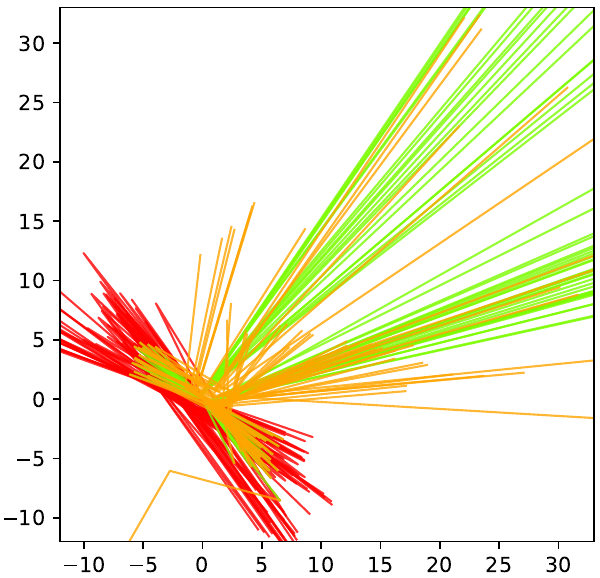} & \includegraphics[width=0.22\textwidth]{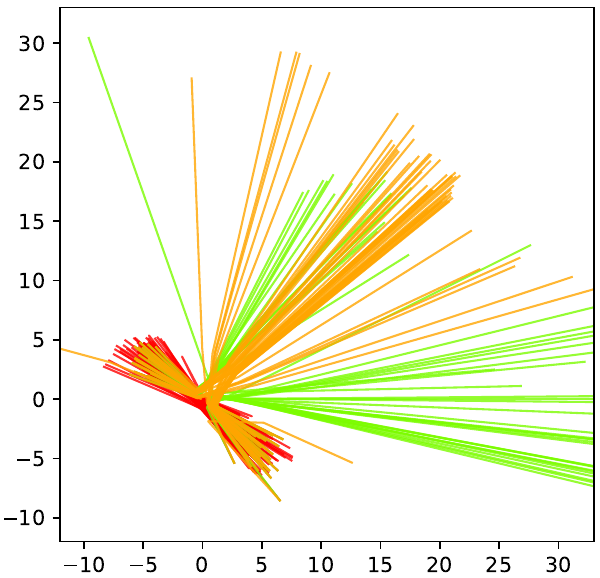} & \includegraphics[width=0.22\textwidth]{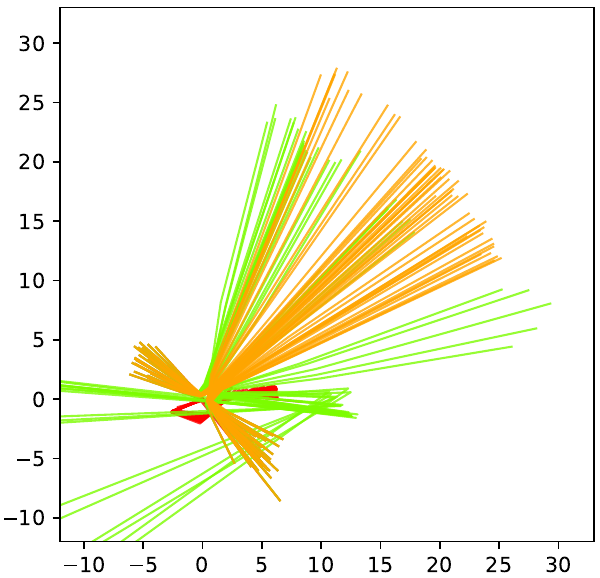} & \includegraphics[width=0.22\textwidth]{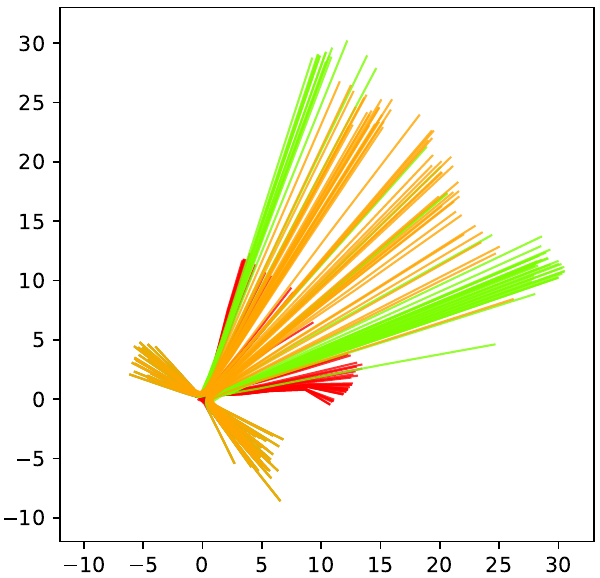} & \includegraphics[width=0.22\textwidth]{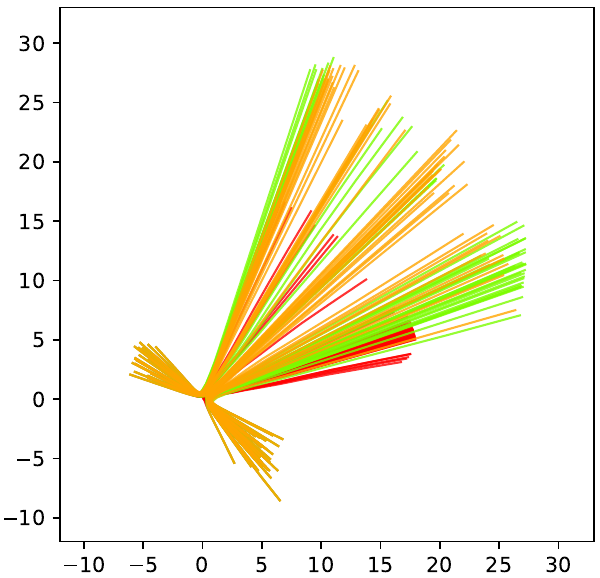} & \includegraphics[width=0.22\textwidth]{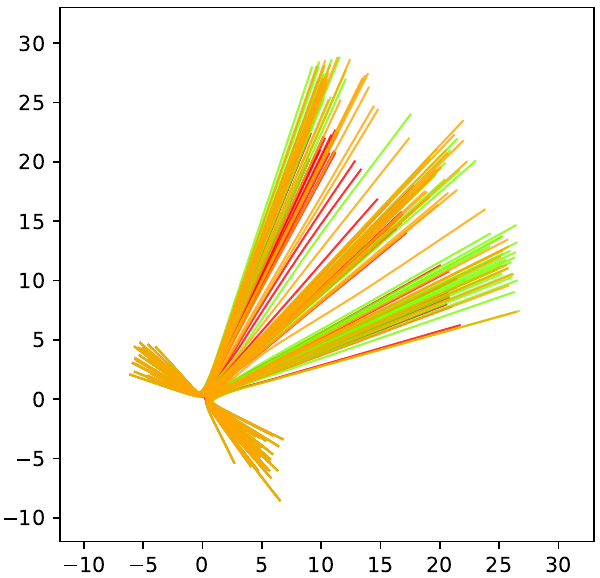}\tabularnewline
\includegraphics[width=0.22\textwidth]{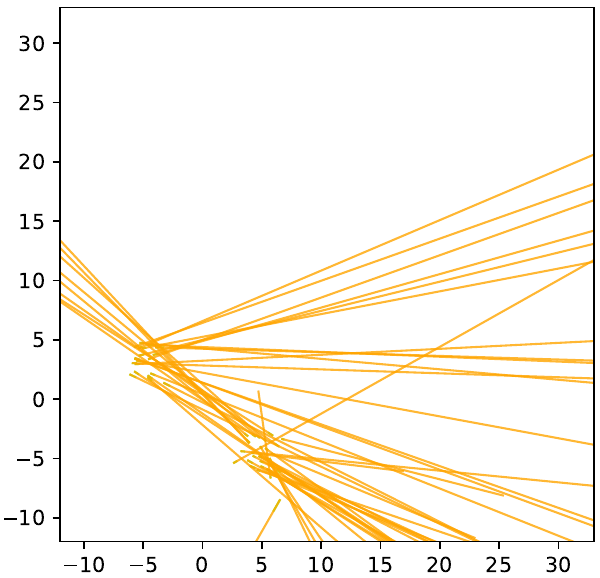} & \includegraphics[width=0.22\textwidth]{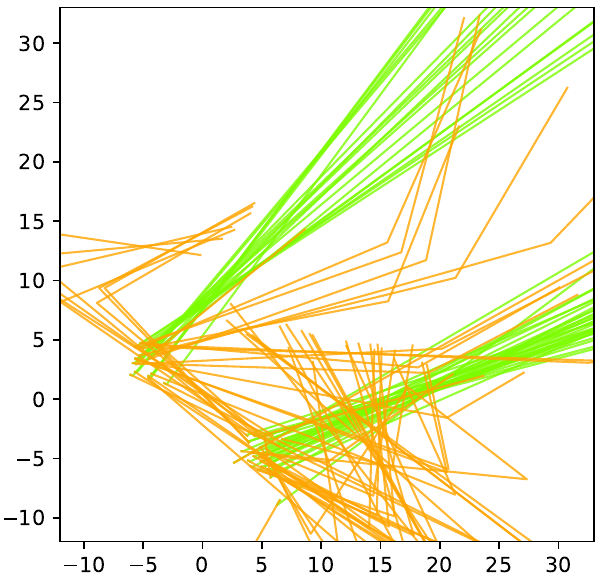} & \includegraphics[width=0.22\textwidth]{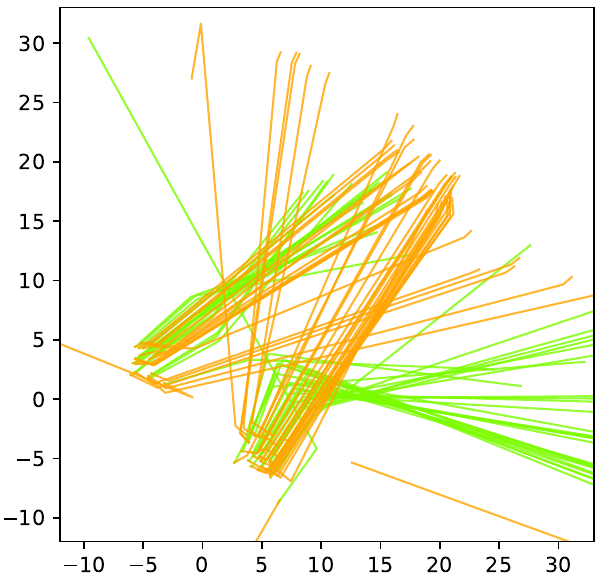} & \includegraphics[width=0.22\textwidth]{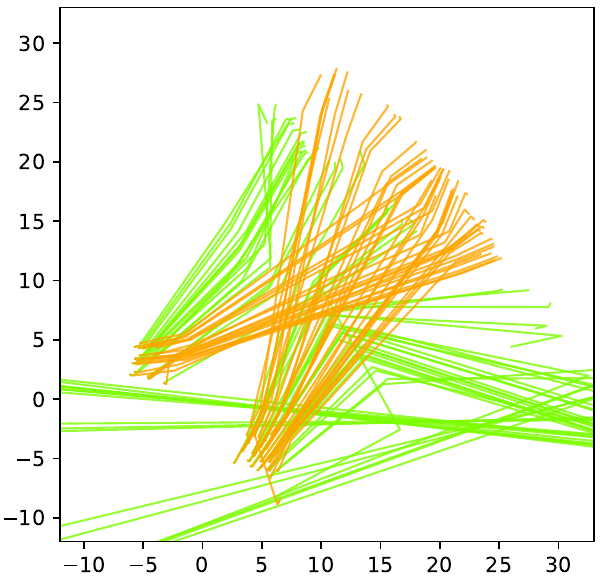} & \includegraphics[width=0.22\textwidth]{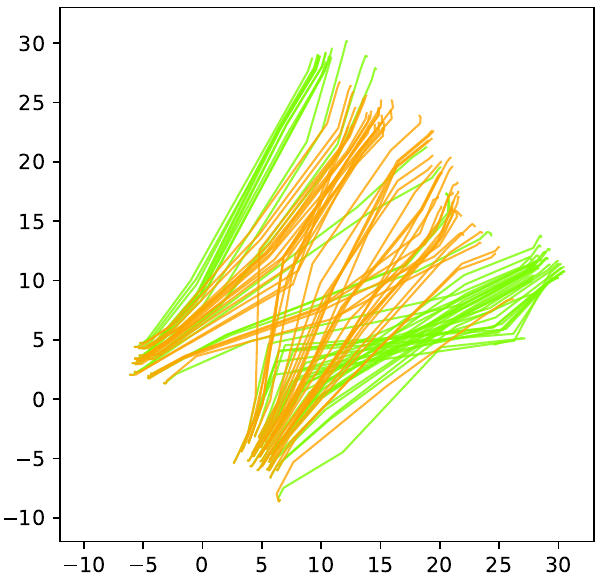} & \includegraphics[width=0.22\textwidth]{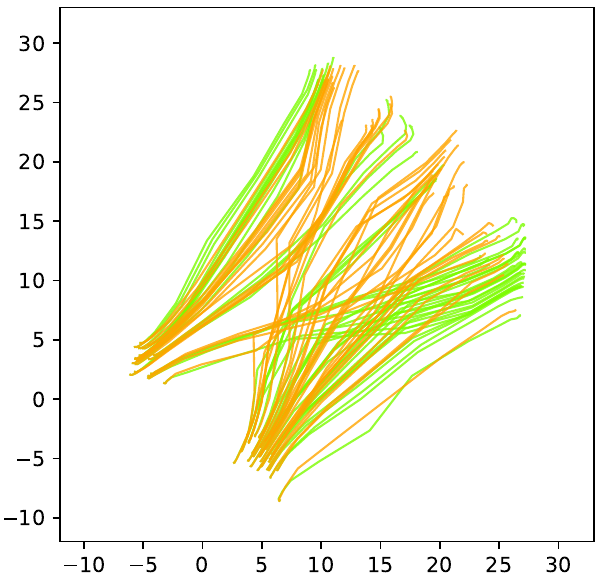} & \includegraphics[width=0.22\textwidth]{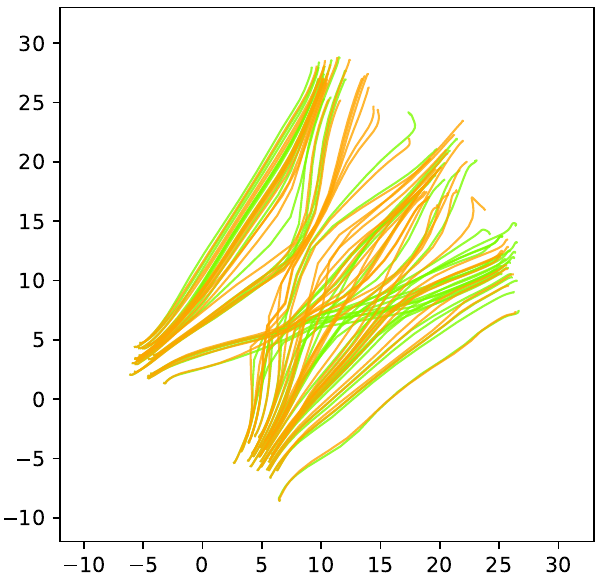}\tabularnewline
$N=1$ & $N=2$ & $N=3$ & $N=5$ & $N=10$ & $N=20$ & $N=50$\tabularnewline
\end{tabular}}}
\par\end{centering}
\caption{Generated samples (top), $x_{t}$-trajectories (middle), and $\tilde{x}_{t}$/$\bar{x}_{t}$-trajectories
(bottom) of the \textcolor{red}{posterior} flow and its transformed
\textcolor{lawngreen}{SN}, \textcolor{orange}{SC} flows of the
\emph{fifth-degree process}, with varying numbers of update steps
$N$.\label{fig:Toy_FifthDegree}}
\end{figure*}

\begin{figure*}
\begin{centering}
{\resizebox{\textwidth}{!}{%
\par\end{centering}
\begin{centering}
\begin{tabular}{ccccccc}
\multicolumn{7}{c}{\includegraphics[height=0.04\textwidth]{asset/ThreeGauss/legend_sample_with_p0_p1}\hspace{0.02\textwidth}\includegraphics[height=0.04\textwidth]{asset/ThreeGauss/legend_traj}}\tabularnewline
\includegraphics[width=0.22\textwidth]{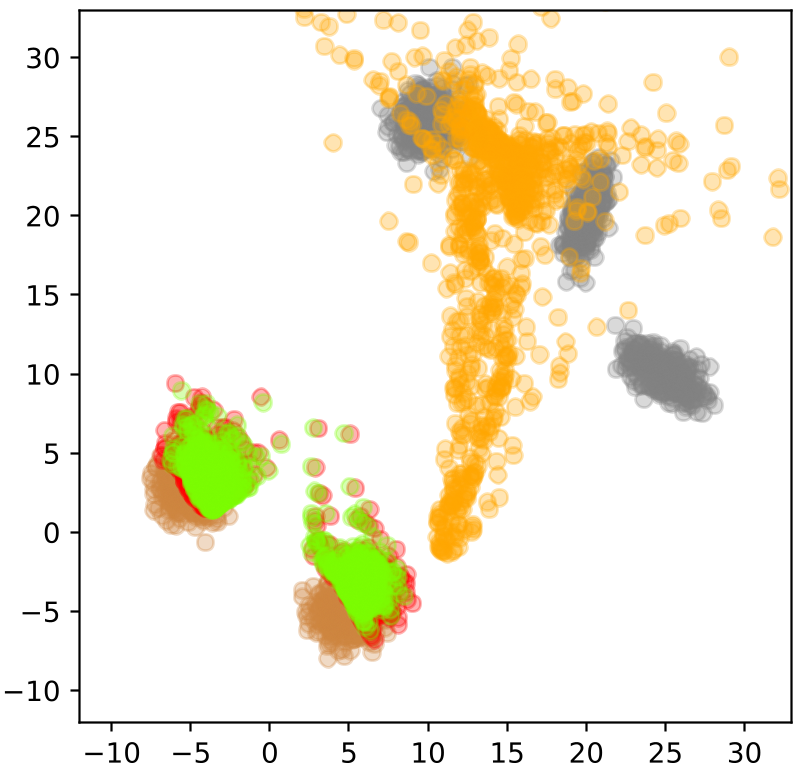} & \includegraphics[width=0.22\textwidth]{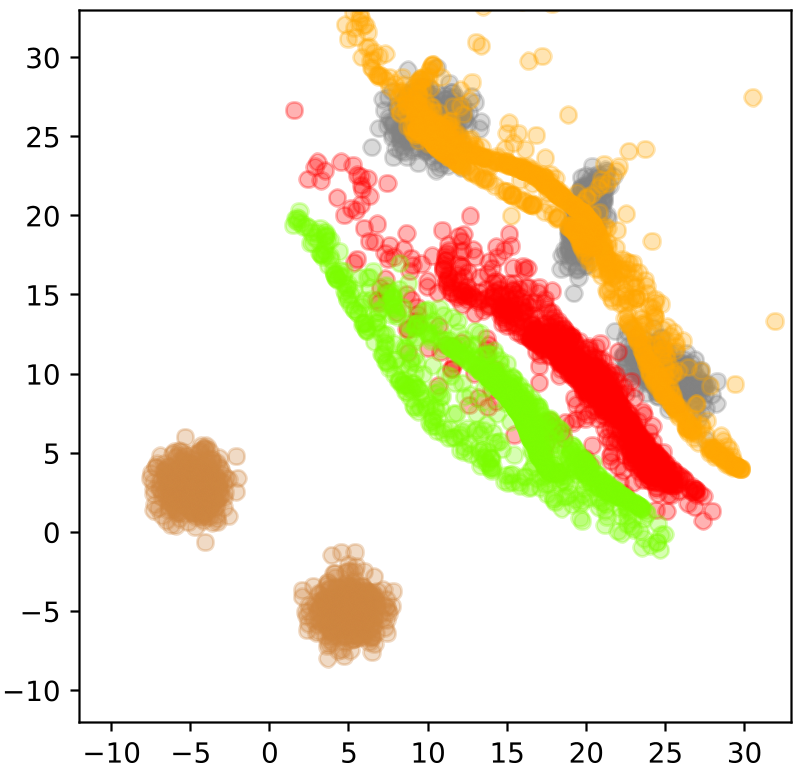} & \includegraphics[width=0.22\textwidth]{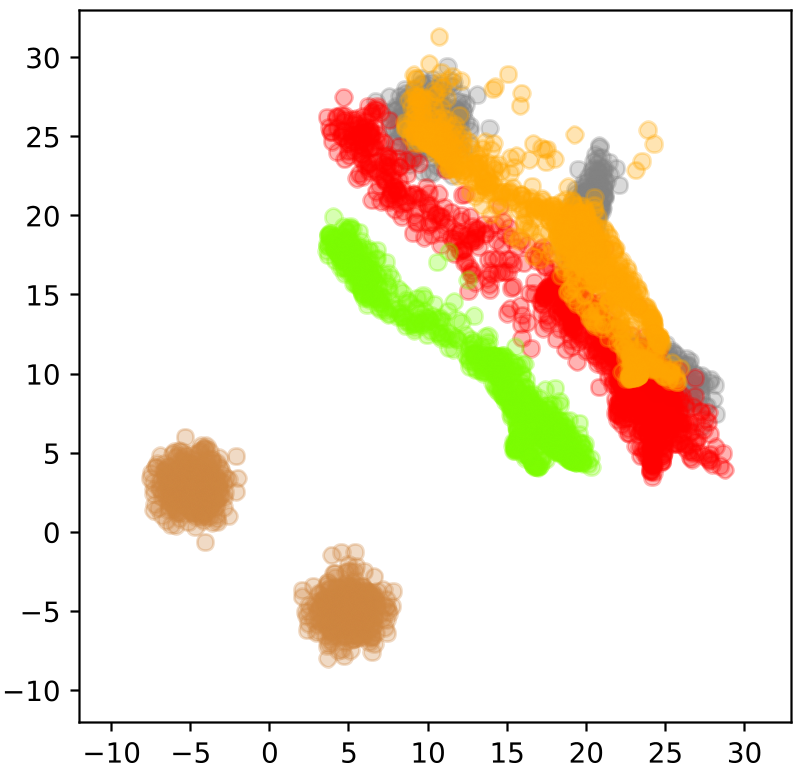} & \includegraphics[width=0.22\textwidth]{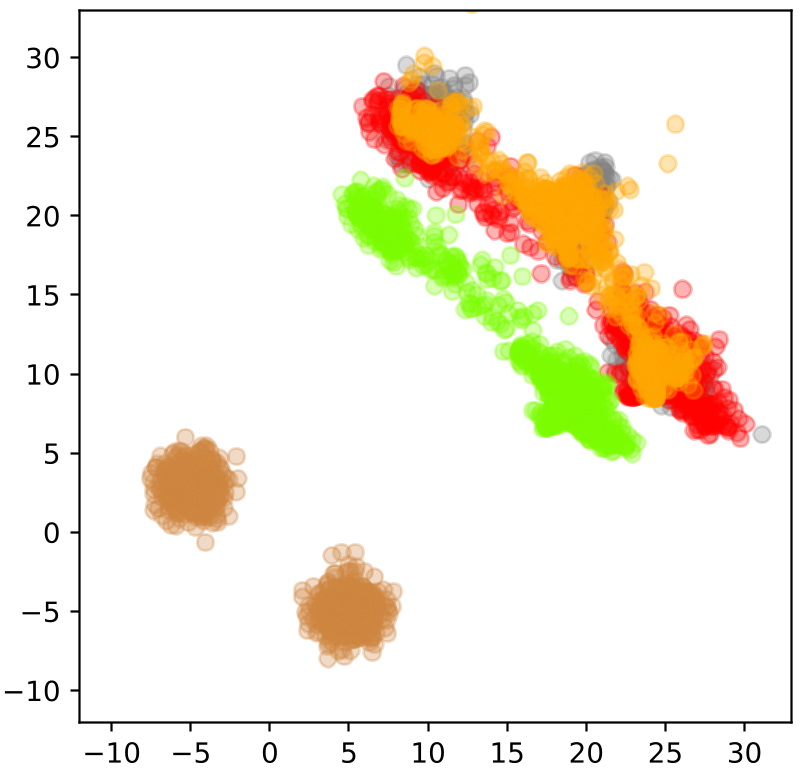} & \includegraphics[width=0.22\textwidth]{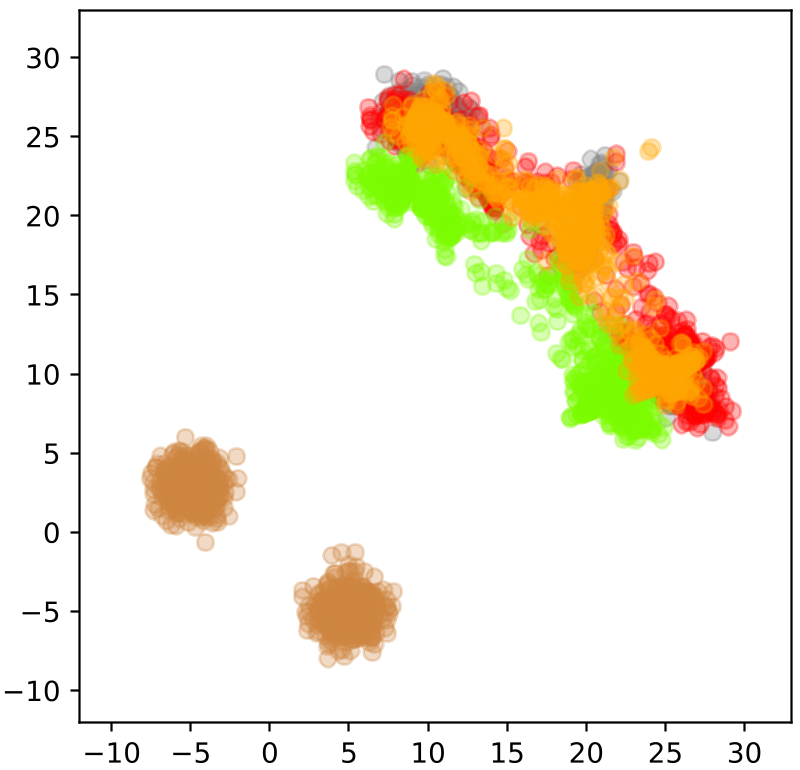} & \includegraphics[width=0.22\textwidth]{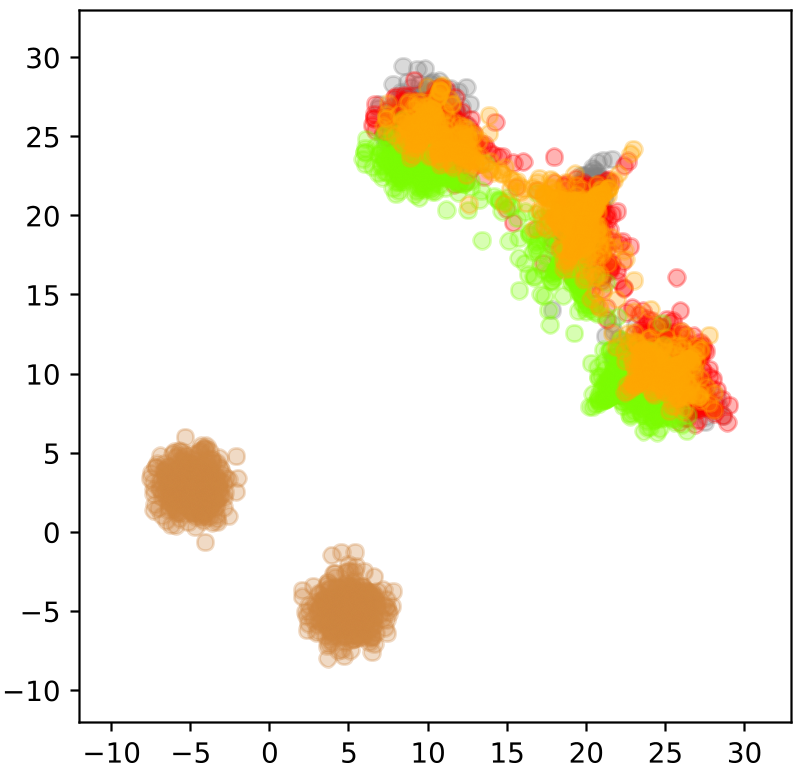} & \includegraphics[width=0.22\textwidth]{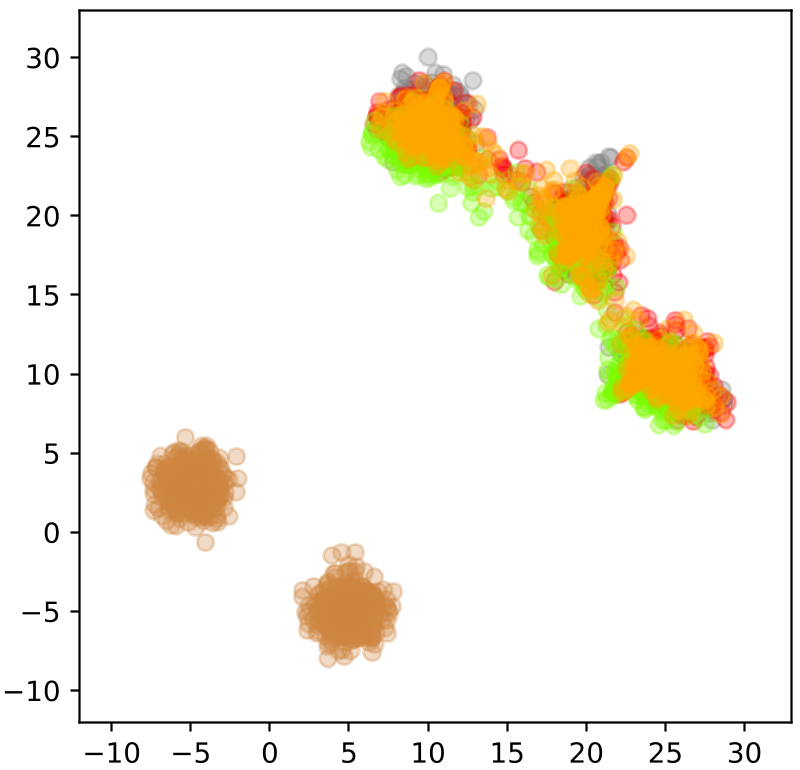}\tabularnewline
\includegraphics[width=0.22\textwidth]{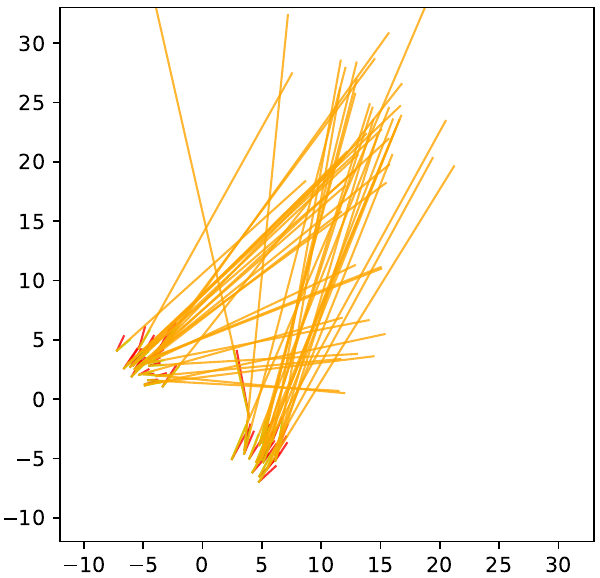} & \includegraphics[width=0.22\textwidth]{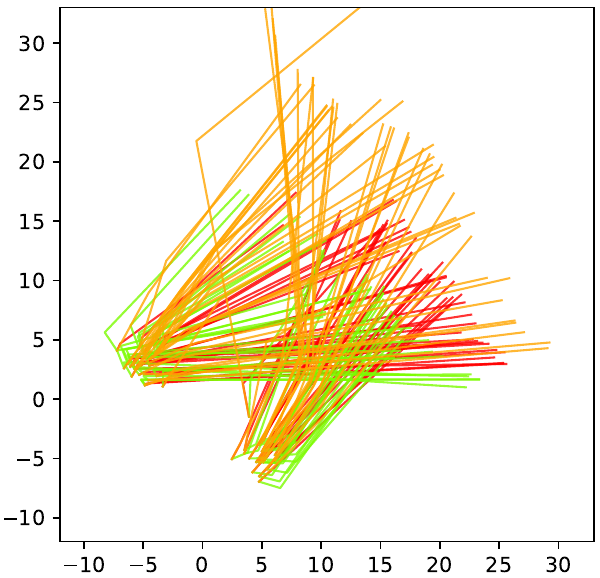} & \includegraphics[width=0.22\textwidth]{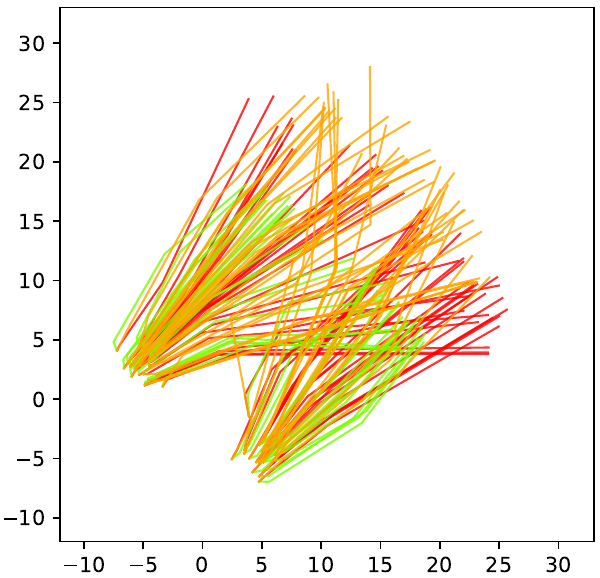} & \includegraphics[width=0.22\textwidth]{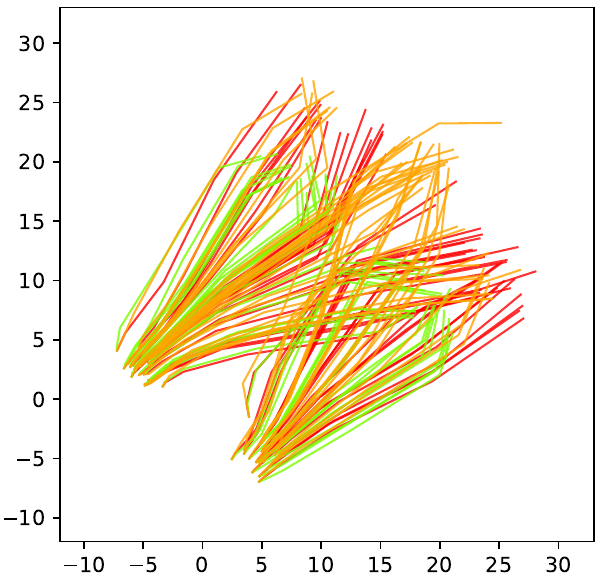} & \includegraphics[width=0.22\textwidth]{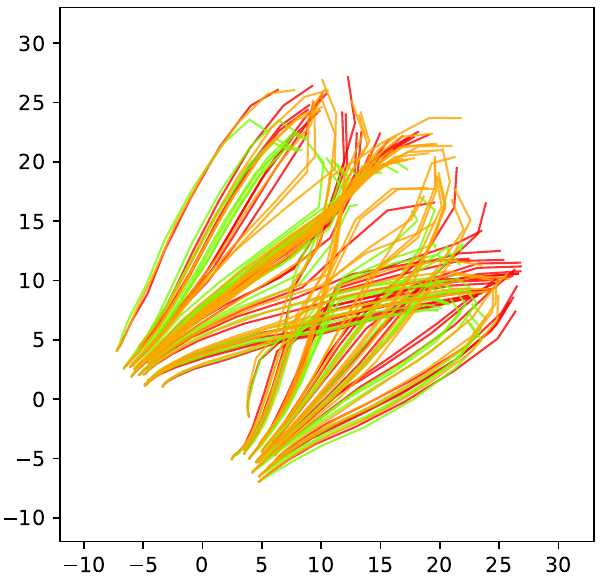} & \includegraphics[width=0.22\textwidth]{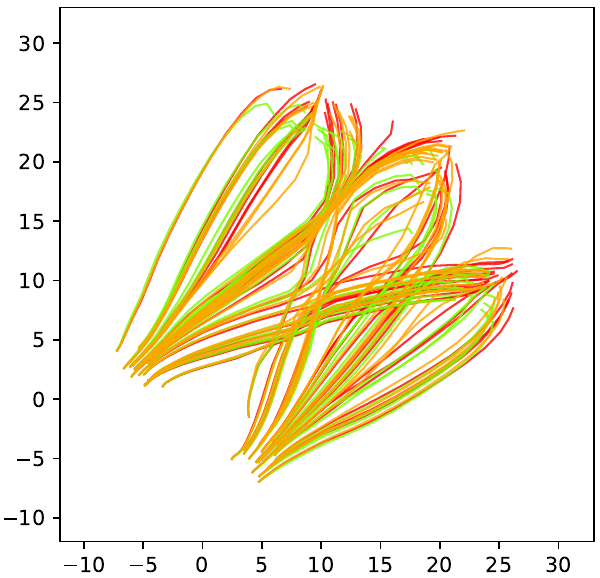} & \includegraphics[width=0.22\textwidth]{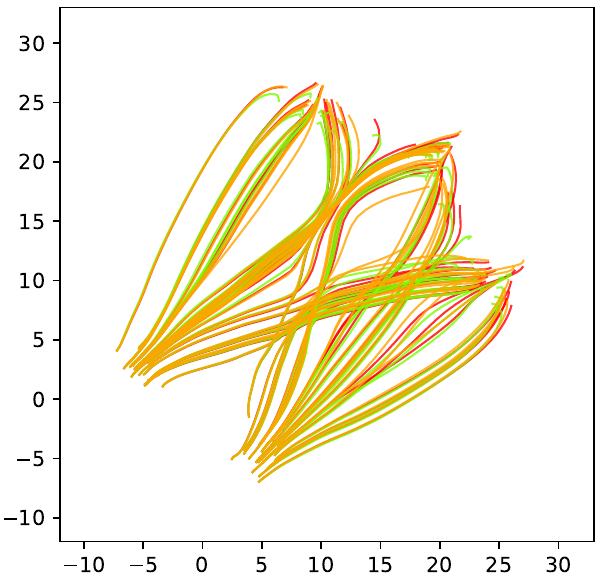}\tabularnewline
\includegraphics[width=0.22\textwidth]{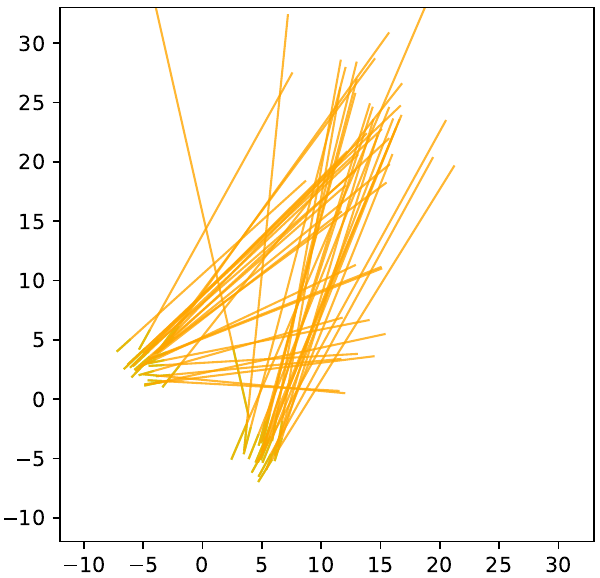} & \includegraphics[width=0.22\textwidth]{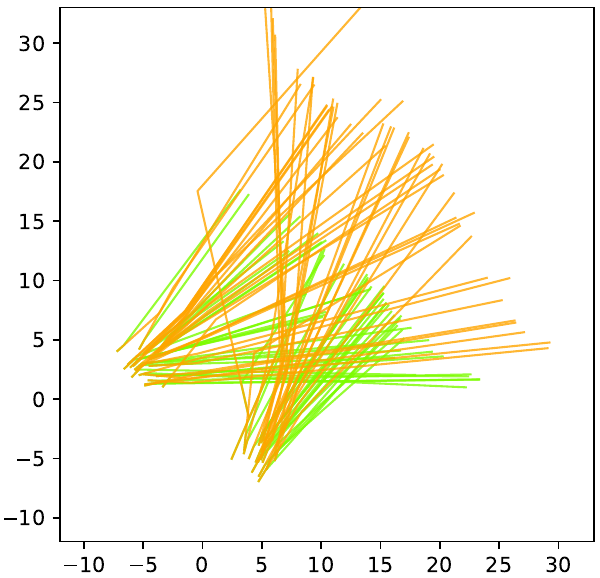} & \includegraphics[width=0.22\textwidth]{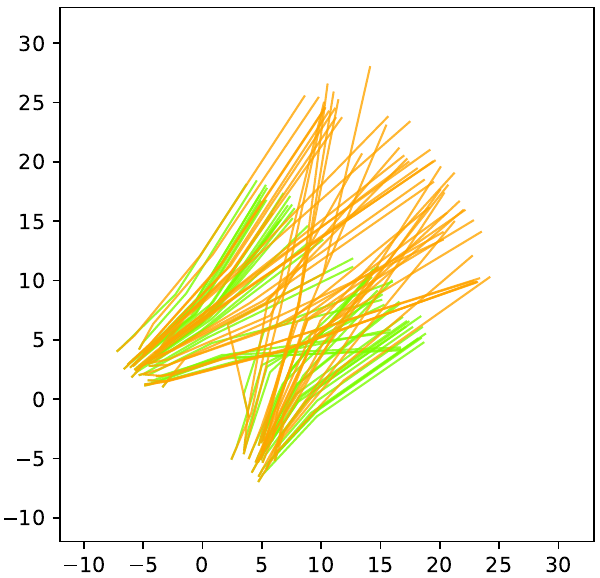} & \includegraphics[width=0.22\textwidth]{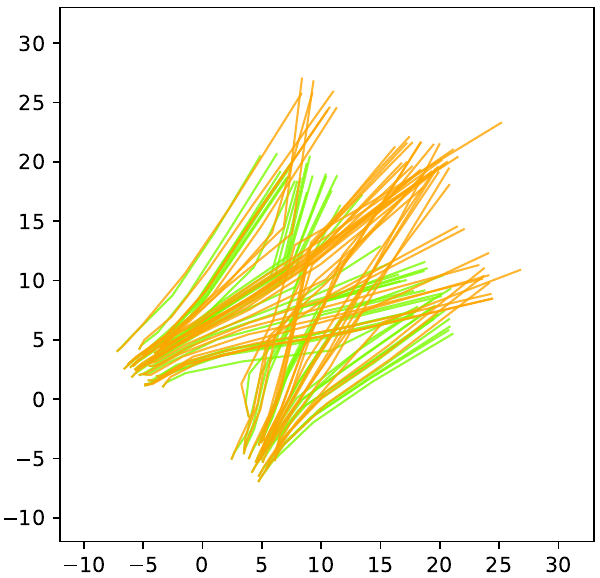} & \includegraphics[width=0.22\textwidth]{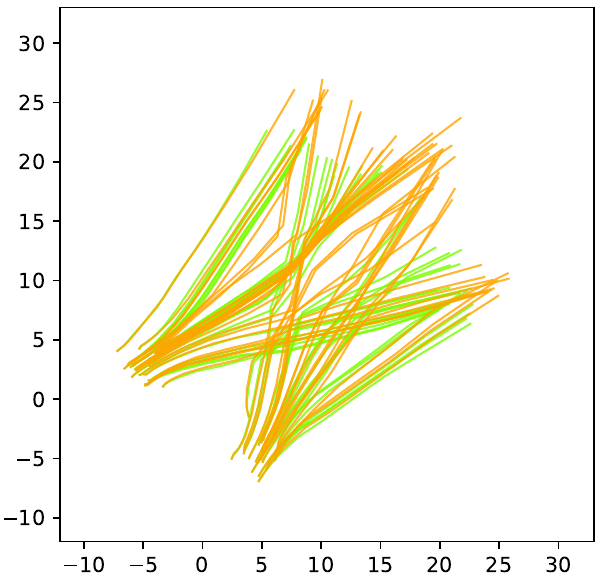} & \includegraphics[width=0.22\textwidth]{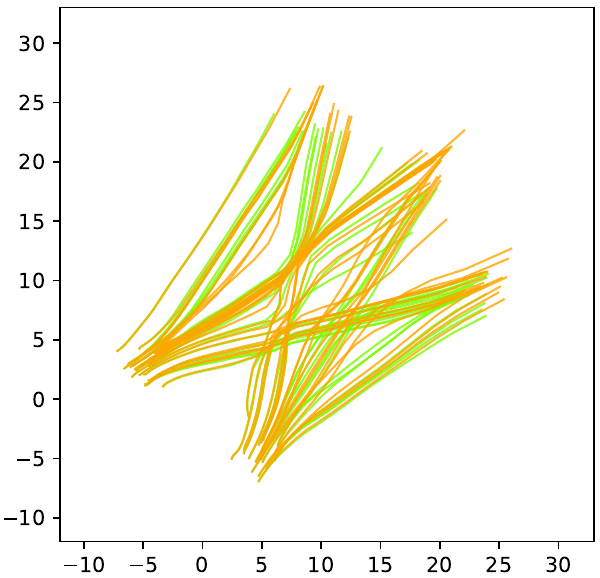} & \includegraphics[width=0.22\textwidth]{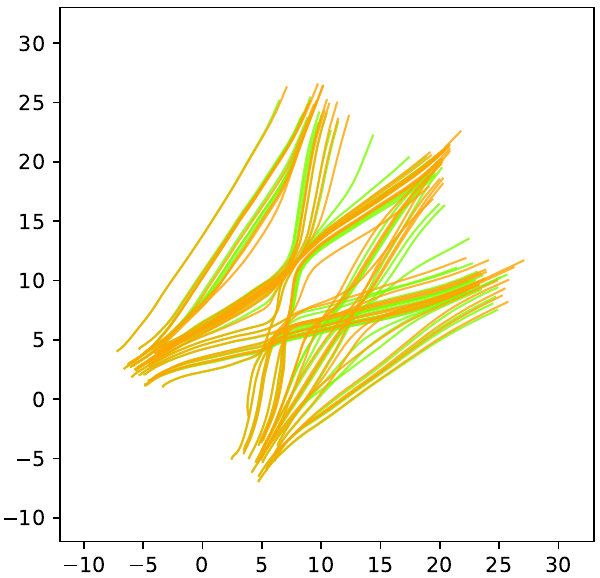}\tabularnewline
$N=1$ & $N=2$ & $N=3$ & $N=5$ & $N=10$ & $N=20$ & $N=50$\tabularnewline
\end{tabular}}}
\par\end{centering}
\caption{Generated samples (top), $x_{t}$-trajectories (middle), and $\tilde{x}_{t}$/$\bar{x}_{t}$-trajectories
(bottom) of the \textcolor{red}{posterior} flow and its transformed
\textcolor{lawngreen}{SN}, \textcolor{orange}{SC} flows of the
\emph{VP-SDE-like process}, with varying numbers of update steps $N$.\label{fig:Toy_VPSDELike}}
\end{figure*}

Fig.~\ref{fig:Toy_ThirdDegree} displays the generated samples and
trajectories associated with the posterior, SN, and SC flows of the
third-degree process. It is evident that sampling along the SC flow
significantly outperforms sampling along the posterior and SN flows
in terms of accuracy and speed. The former requires approximately
5 Euler updates to generate good-quality samples while the latter
two need at least 20 Euler updates. We observe similar phenomena for
the fifth-degree and VP-SDE-like processes, as depicted in Figs.~\ref{fig:Toy_FifthDegree}
and \ref{fig:Toy_VPSDELike}. For the third-degree process, the SN
flow does not exhibit a discernible improvement over the posterior
flow. However, for the fifth-degree process, where the posterior flow
is strongly curved, the SN flow demonstrates clear enhancement.

Furthermore, the middle row of Fig.~\ref{fig:Toy_ThirdDegree} illustrates
that the $x_{t}$-trajectories recovered from the SN and SC flows
closely align with the $x_{t}$-trajectories of the posterior flow
when the number of updates is substantial. This result empirically
validates the correctness of our transformations from the posterior
flow to the SN, SC flows and vice versa. 

\begin{figure*}
\begin{centering}
{\resizebox{\textwidth}{!}{%
\par\end{centering}
\begin{centering}
\begin{tabular}{ccc|ccc|c}
\multicolumn{6}{c|}{\includegraphics[height=0.04\textwidth]{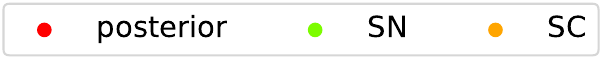}\hspace{0.02\textwidth}\includegraphics[height=0.04\textwidth]{asset/ThreeGauss/legend_traj}} & \includegraphics[width=0.16\textwidth]{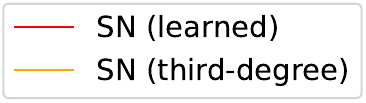}\tabularnewline
\includegraphics[width=0.22\textwidth]{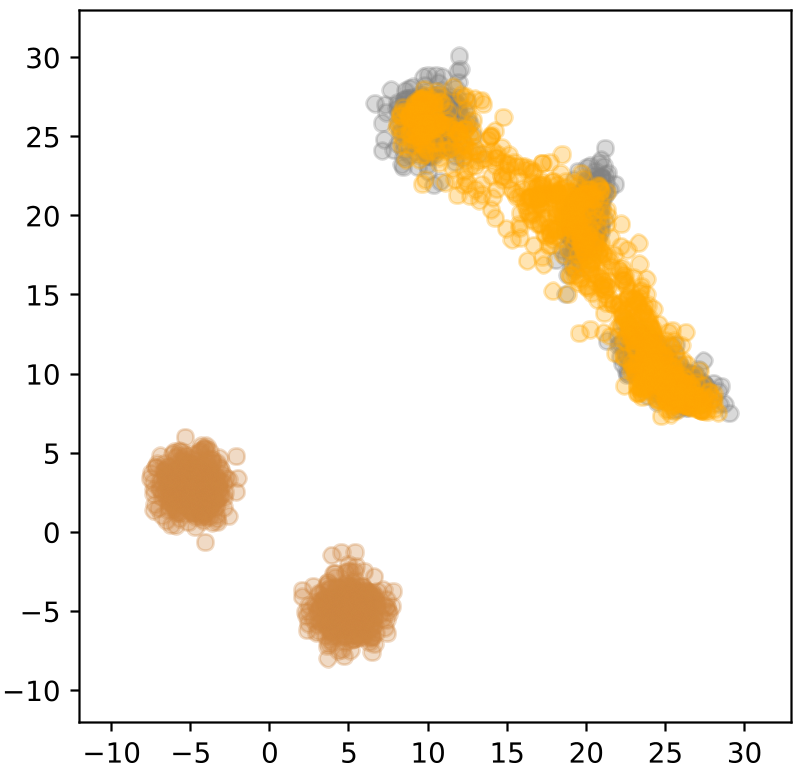} & \includegraphics[width=0.22\textwidth]{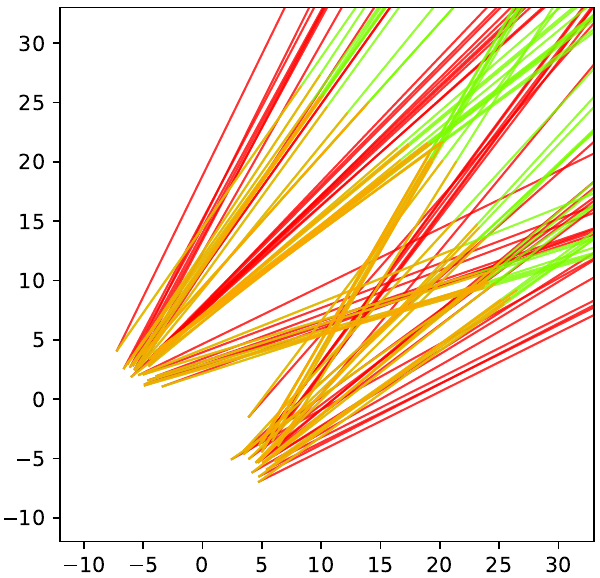} & \includegraphics[width=0.22\textwidth]{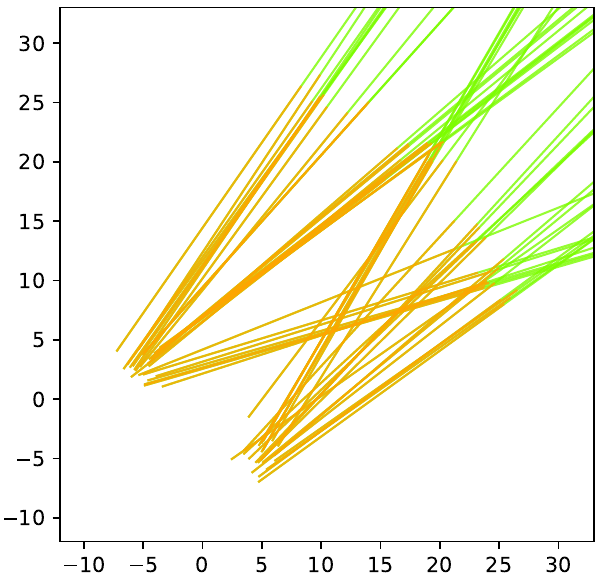} & \includegraphics[width=0.22\textwidth]{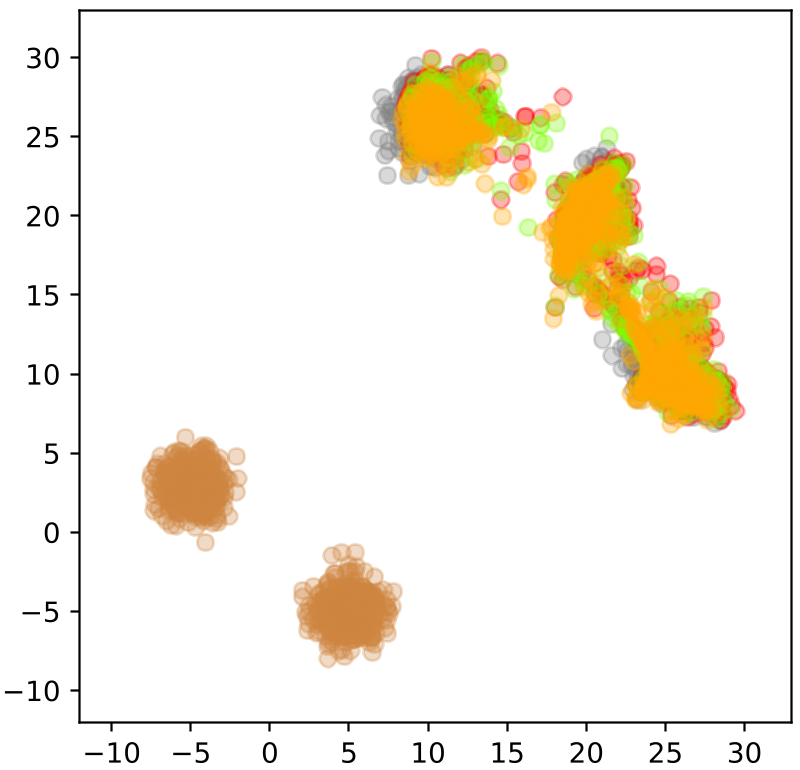} & \includegraphics[width=0.22\textwidth]{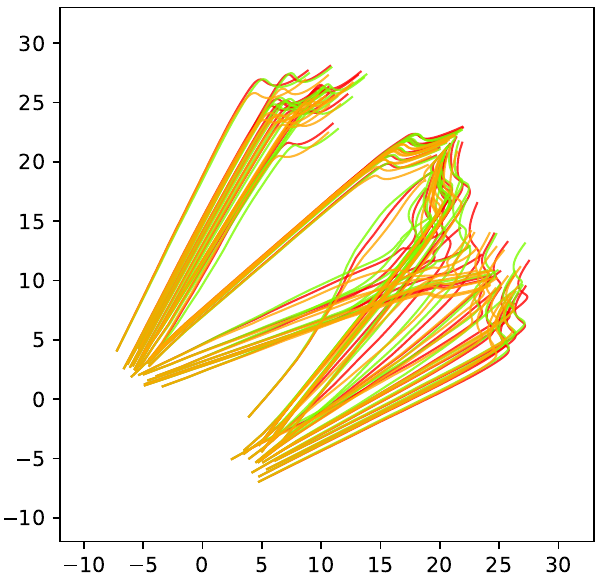} & \includegraphics[width=0.22\textwidth]{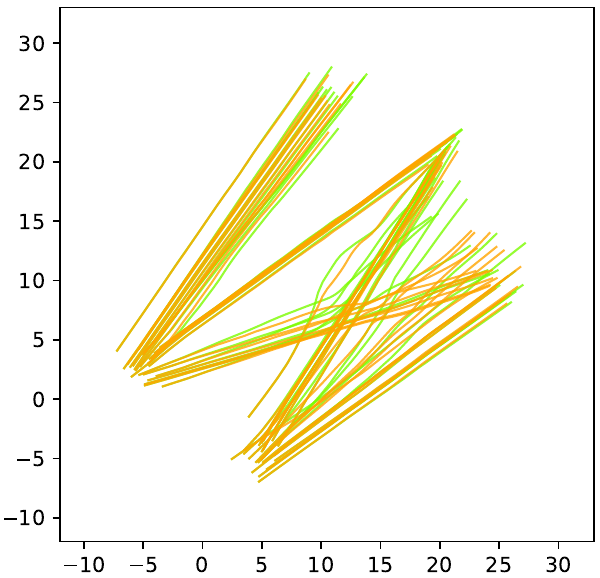} & \includegraphics[width=0.22\textwidth]{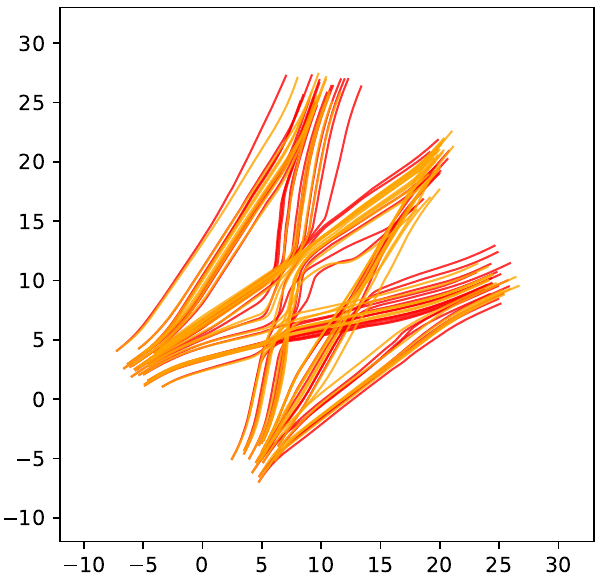}\tabularnewline
\multicolumn{3}{c}{$N=1$ (Reflowed)} & \multicolumn{3}{c|}{$N=50$ (Reflowed)} & $N=1000$\tabularnewline
\multicolumn{6}{c}{(a)} & (b)\tabularnewline
\end{tabular}}}
\par\end{centering}
\caption{\textbf{(a)} Generated samples (left), $x_{t}$-trajectories (middle)
and $\tilde{x}_{t}/\bar{x}_{t}$-trajectories (right) of the posterior,
SN, and SC flows of the third-degree process after being \emph{\textquotedblleft reflowed\textquotedblright{}
}with $N$=1 and $N$=50 Euler updates. \textbf{(b)} The $x_{t}$-trajectories
of the posterior flow of the process $X_{t}=\left(1-\varphi_{t}\right)X_{0}+\varphi_{t}X_{1}$
and the $\tilde{x}_{t}$-trajectories of the SN flow of the third-degree
process with $N$=1000 Euler updates.\label{fig:Toy_Reflow_ThirdDegree}}
\end{figure*}

Comparing trajectories in the middle and bottom rows of Figs.~\ref{fig:Toy_FifthDegree},
\ref{fig:Toy_VPSDELike} (and also Fig.~\ref{fig:Toy_ThirdDegree}),
it is apparent that the SN and SC flows exhibit straighter paths than
the posterior flow. However, the SN and SC flows are \emph{not} perfectly
straight due to randomness caused by the independent coupling used
in training. To enhance the straightness of the SN and SC flows, we
learn a \emph{new} velocity network $v_{\theta}$ using the deterministic
coupling induced by either the posterior flow or the SC flow. This
technique is inspired by \cite{liu2022flow} where it is termed ``reflowing''.
As illustrated in Fig.~\ref{fig:Toy_Reflow_ThirdDegree} (a), the
``reflowed'' SC flow is almost perfectly straight and can generate
good-quality samples with just 1 Euler update. Its trajectories still
exhibit crossovers because the original posterior flow is not an optimal
(transport) map between $p_{0}$ and $p_{1}$.

To empirically confirm that the stochastic process associated with
the SN flow indeed follows the equation $\tilde{X}_{t}=\left(1-\varphi_{t}\right)X_{0}+\varphi_{t}X_{1}$,
we directly learn the posterior flow of this process and show the
resemblance between the learned flow and the SN flow in Fig.~\ref{fig:Toy_Reflow_ThirdDegree}
(b). 

\subsubsection{Flowing along arbitrary paths}

\begin{figure*}
\begin{centering}
{\resizebox{0.92\textwidth}{!}{%
\par\end{centering}
\begin{centering}
\begin{tabular}{ccc|cc|cc}
\multirow{1}{*}[0.02\textwidth]{$N$} & \multicolumn{6}{c}{\includegraphics[height=0.035\textwidth]{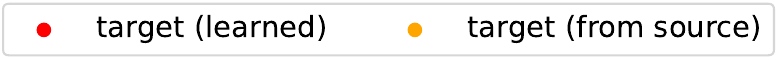}\hspace{0.02\textwidth}\includegraphics[height=0.035\textwidth]{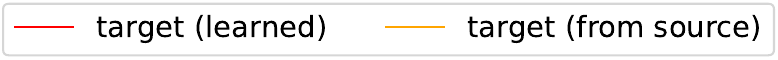}}\tabularnewline
\multirow{1}{*}[0.07\textwidth]{5} & \includegraphics[width=0.22\textwidth]{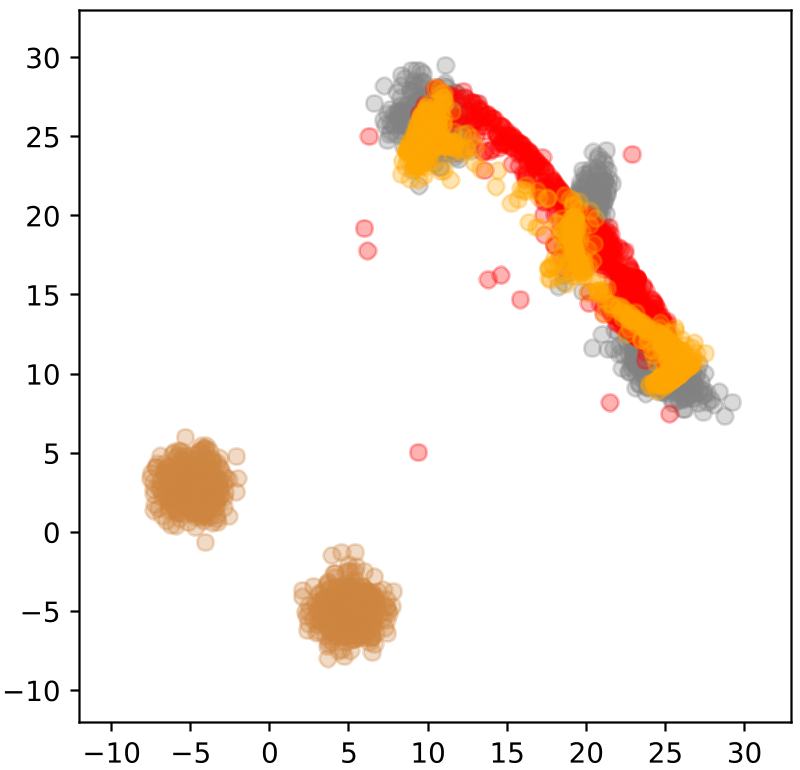} & \includegraphics[width=0.22\textwidth]{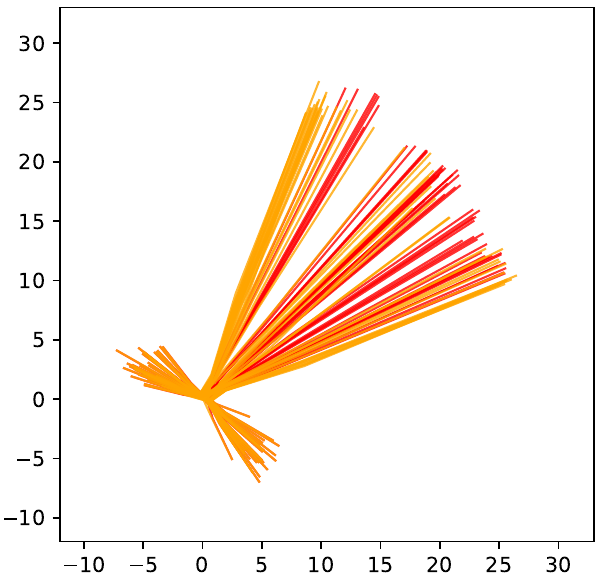} & \includegraphics[width=0.22\textwidth]{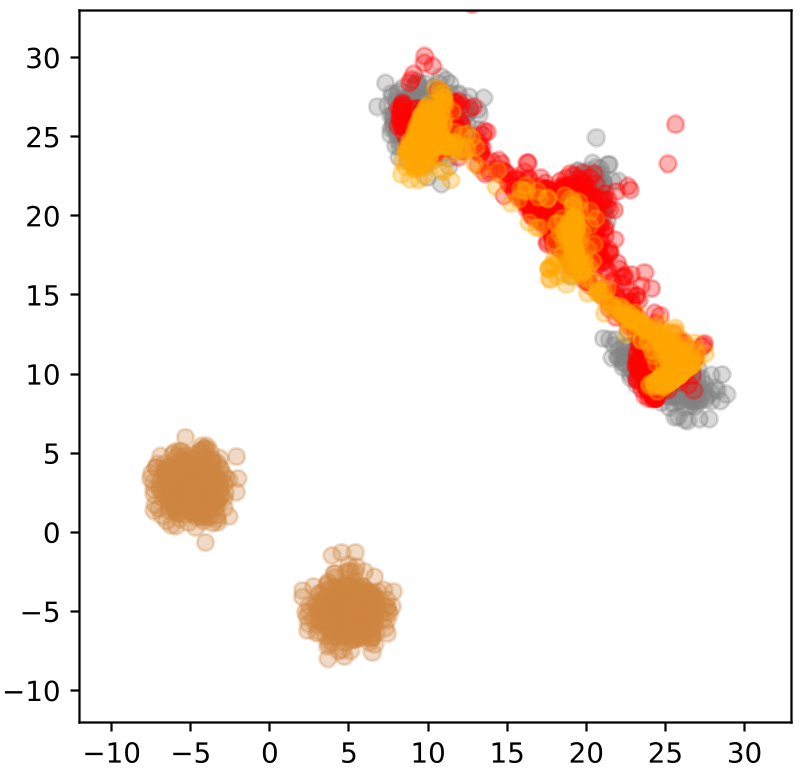} & \includegraphics[width=0.22\textwidth]{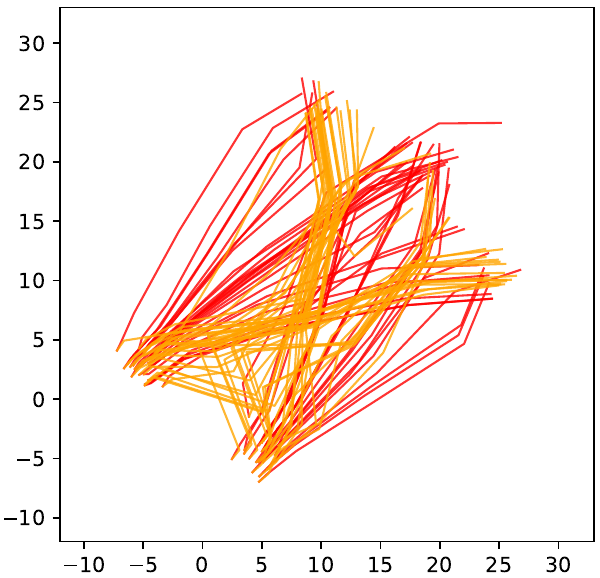} & \includegraphics[width=0.22\textwidth]{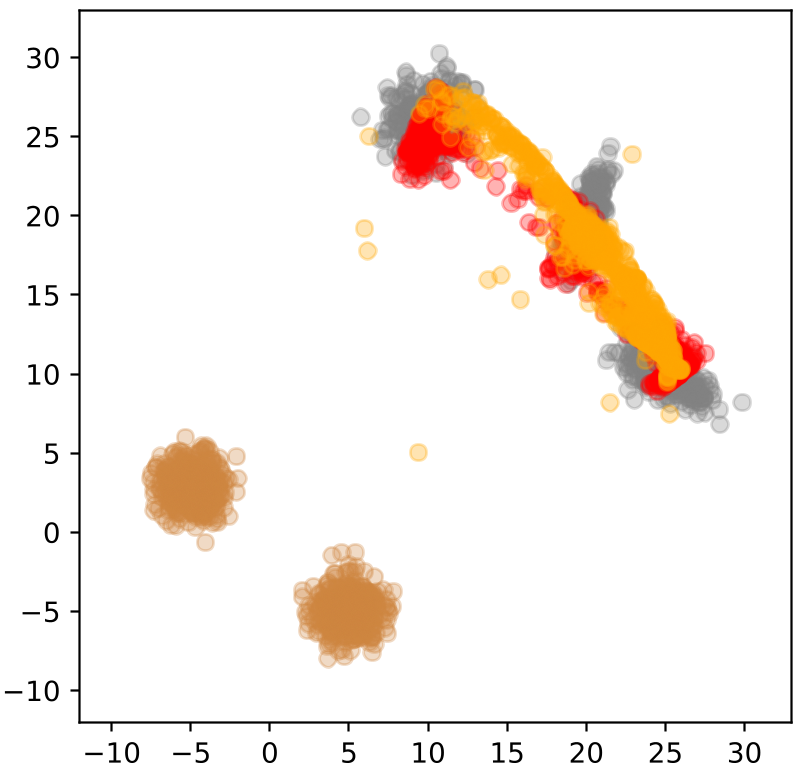} & \includegraphics[width=0.22\textwidth]{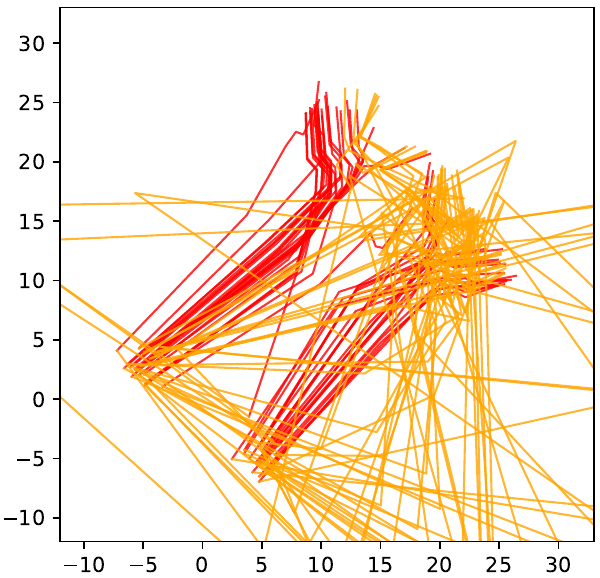}\tabularnewline
\multirow{1}{*}[0.07\textwidth]{50} & \includegraphics[width=0.22\textwidth]{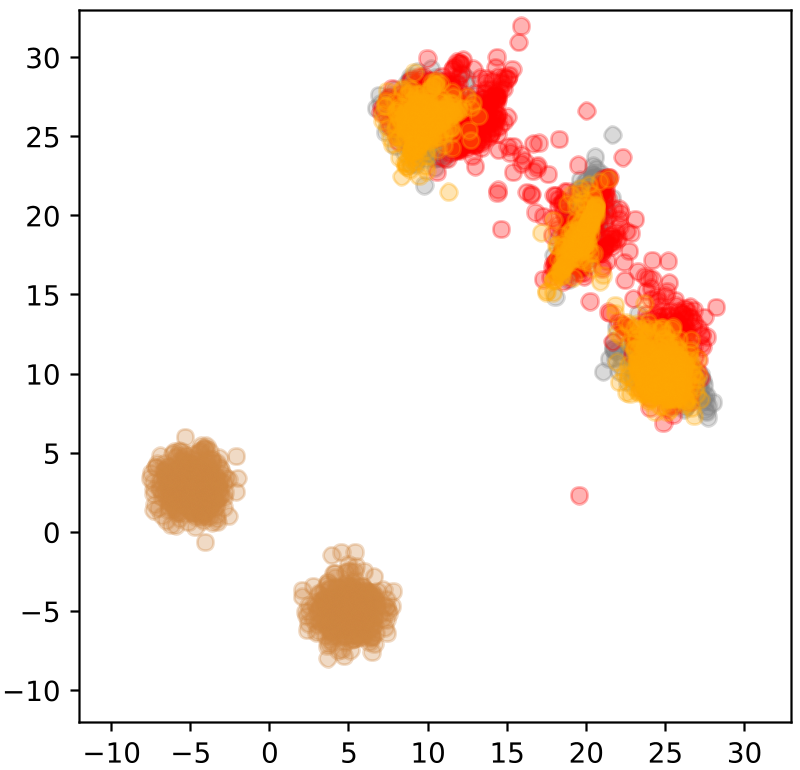} & \includegraphics[width=0.22\textwidth]{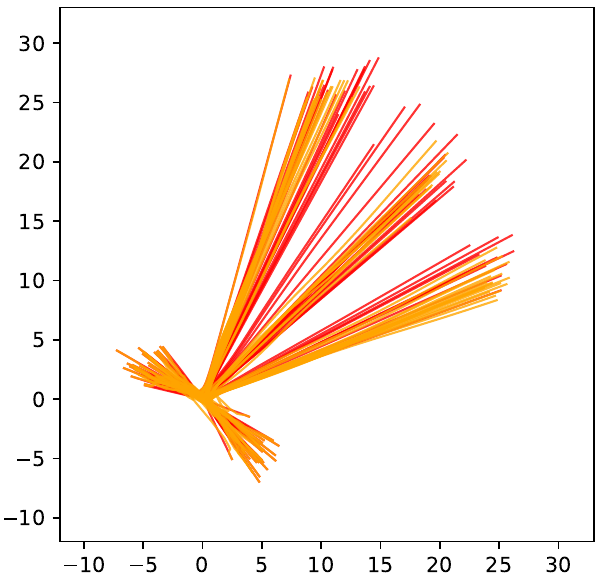} & \includegraphics[width=0.22\textwidth]{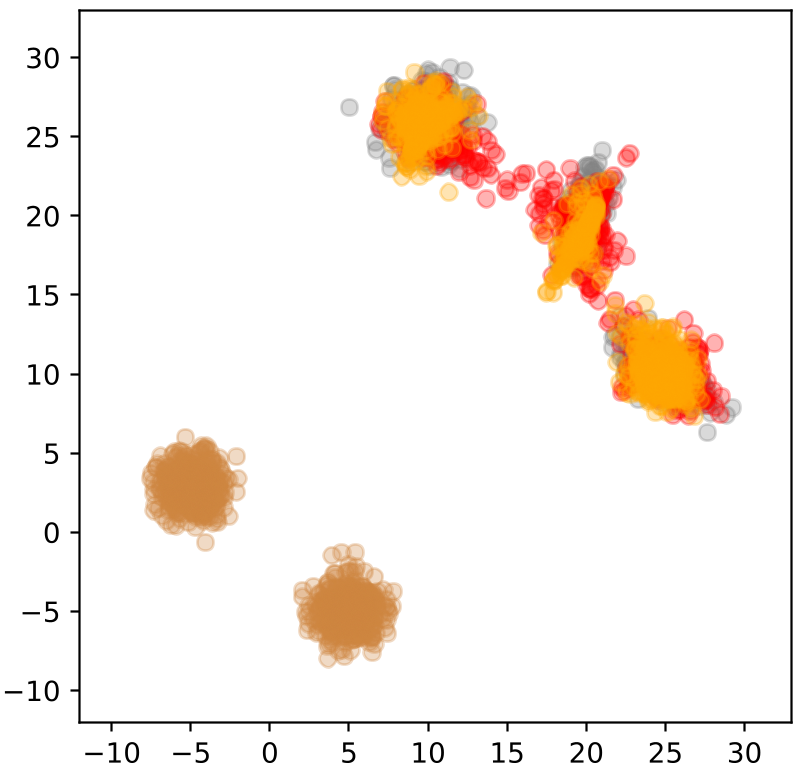} & \includegraphics[width=0.22\textwidth]{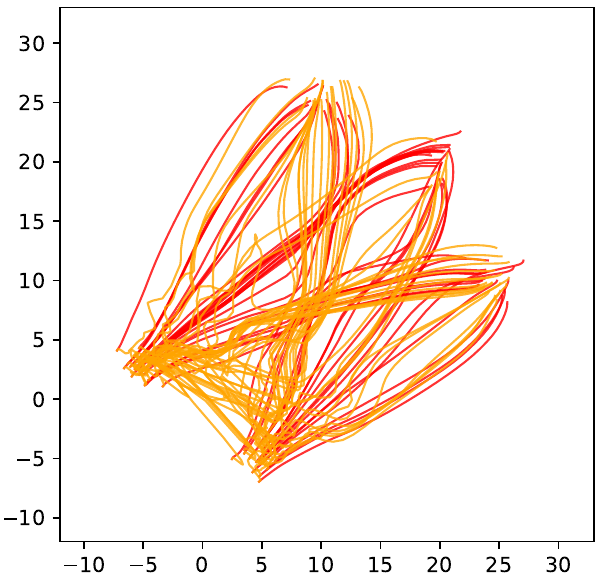} & \includegraphics[width=0.22\textwidth]{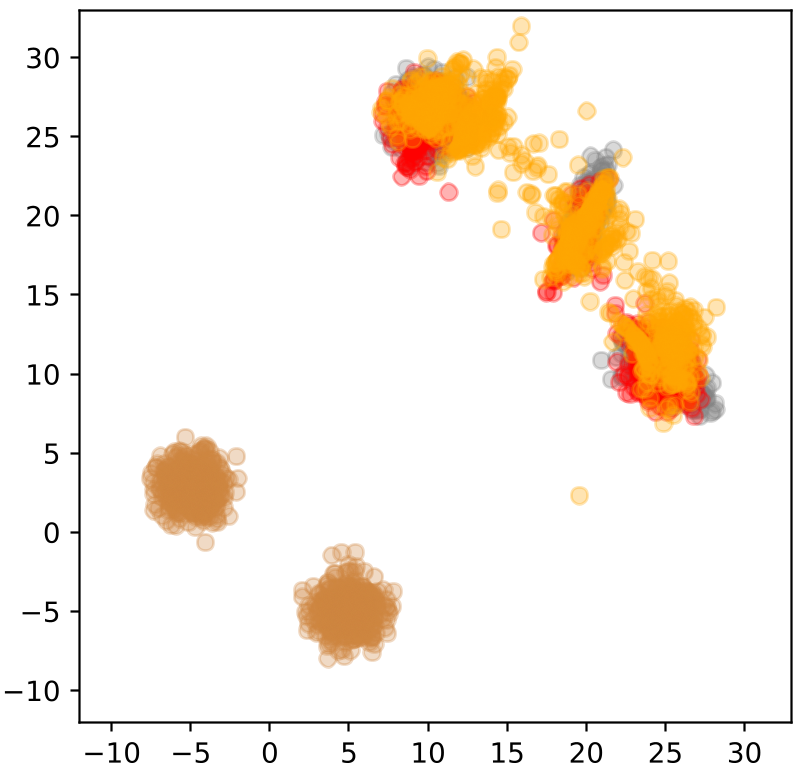} & \includegraphics[width=0.22\textwidth]{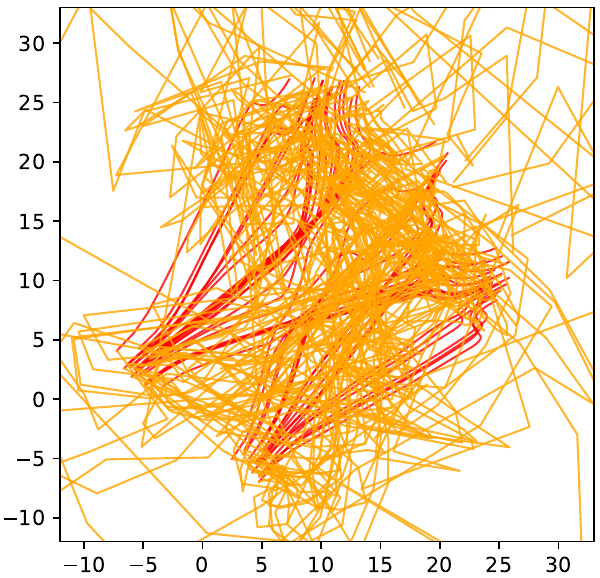}\tabularnewline
\multirow{1}{*}[0.07\textwidth]{1000} & \includegraphics[width=0.22\textwidth]{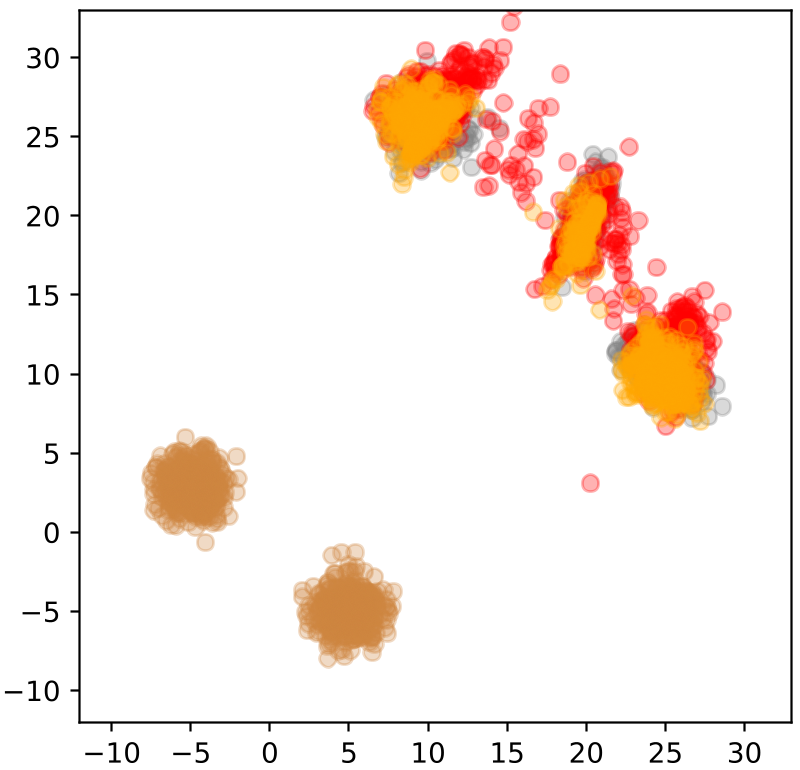} & \includegraphics[width=0.22\textwidth]{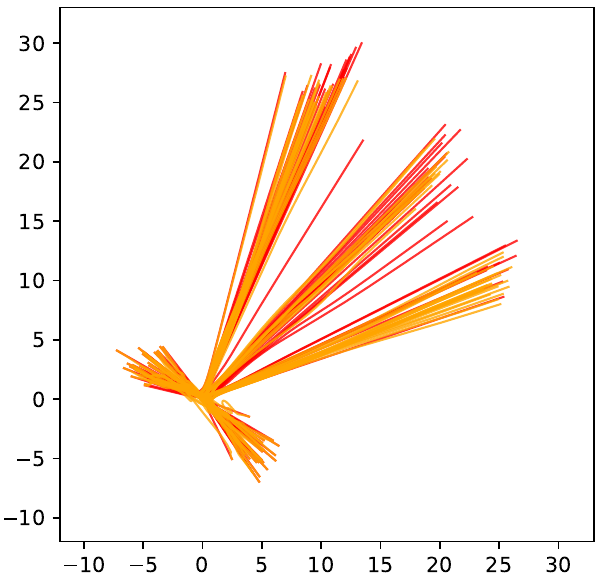} & \includegraphics[width=0.22\textwidth]{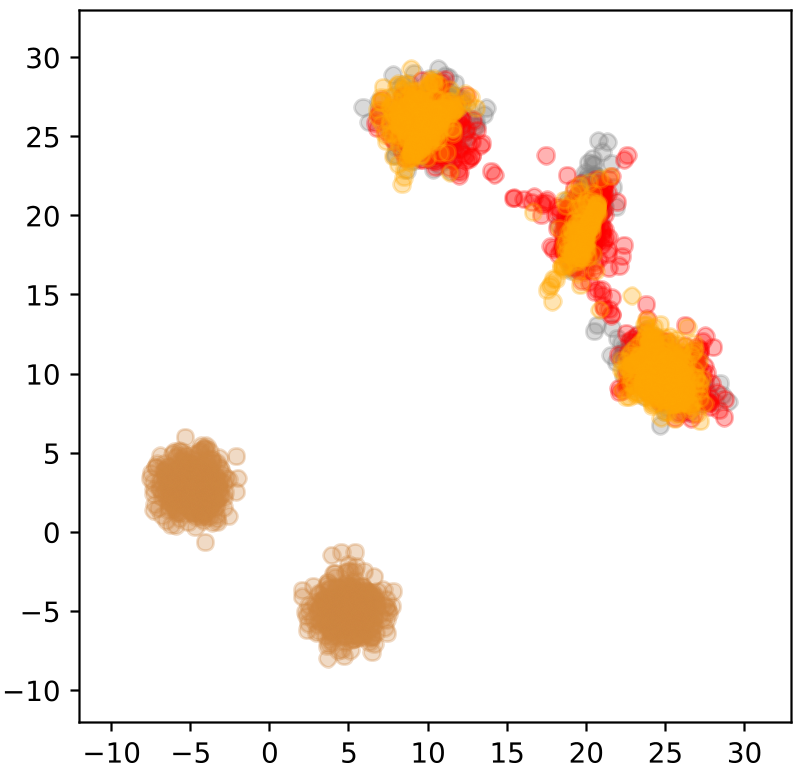} & \includegraphics[width=0.22\textwidth]{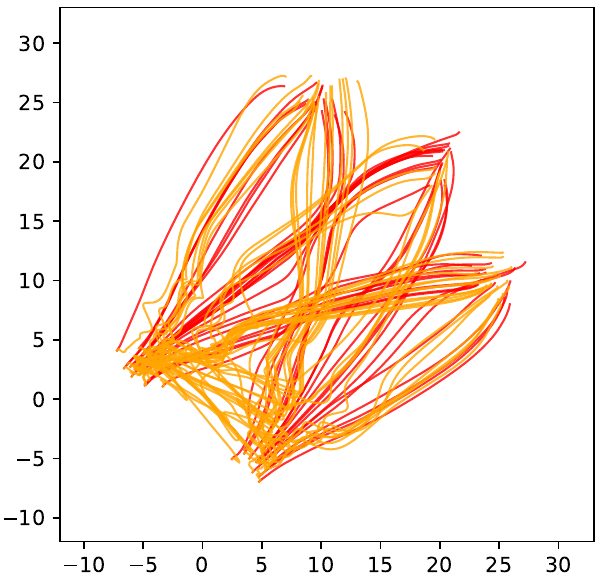} & \includegraphics[width=0.22\textwidth]{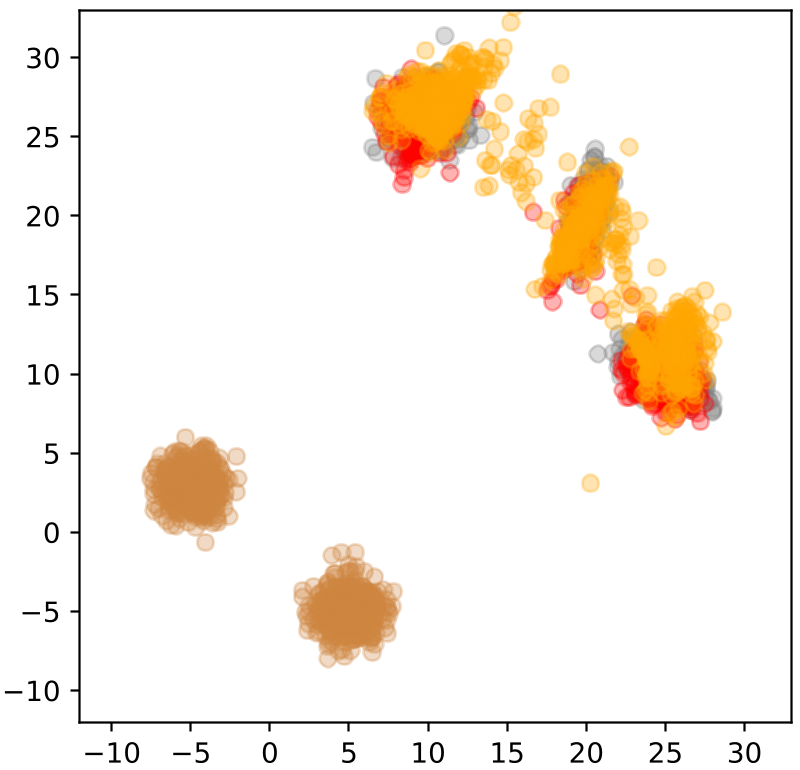} & \includegraphics[width=0.22\textwidth]{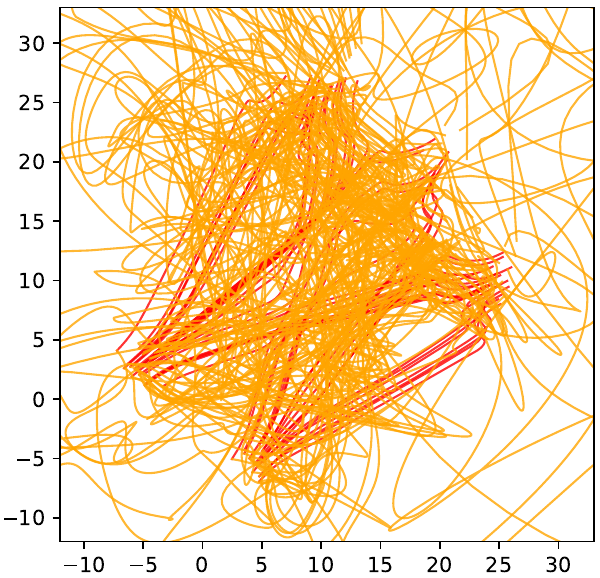}\tabularnewline
 & \multicolumn{2}{c}{third-degree $\rightarrow$ fifth-degree} & \multicolumn{2}{c}{third-degree $\rightarrow$ VPSDE-like} & \multicolumn{2}{c}{fifth-degree $\rightarrow$ third-degree}\tabularnewline
\end{tabular}}}
\par\end{centering}
\caption{Given the learned velocity corresponding to a source process, we simulate
the flow of a target process (labeled as \textcolor{orange}{target
(from source)}) and compare it with the learned flow of the target
process (labeled as \textcolor{red}{target (learned)}).\label{fig:Toy_Third-to-Fifth-VPSDELike}}
\end{figure*}

\begin{figure*}
\begin{centering}
{\resizebox{0.92\textwidth}{!}{%
\par\end{centering}
\begin{centering}
\begin{tabular}{ccc|cc|cc}
\multirow{1}{*}[0.02\textwidth]{$N$} & \multicolumn{6}{c}{\includegraphics[height=0.035\textwidth]{asset/ThreeGauss/legend_sample_origin_to_target}\hspace{0.02\textwidth}\includegraphics[height=0.035\textwidth]{asset/ThreeGauss/legend_traj_origin_to_target}}\tabularnewline
\multirow{1}{*}[0.07\textwidth]{5} & \includegraphics[width=0.22\textwidth]{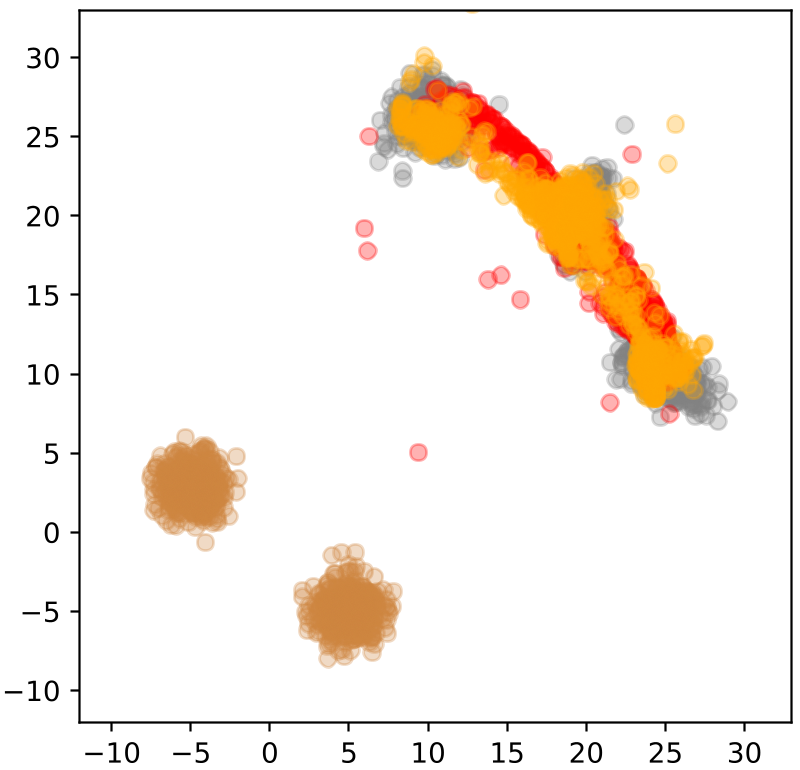} & \includegraphics[width=0.22\textwidth]{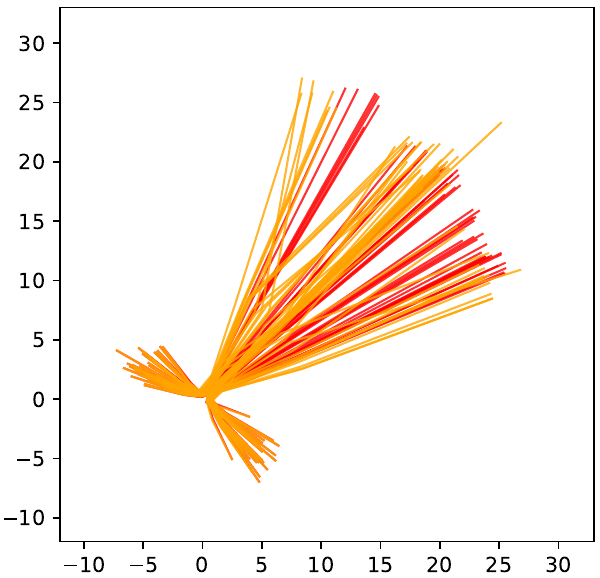} & \includegraphics[width=0.22\textwidth]{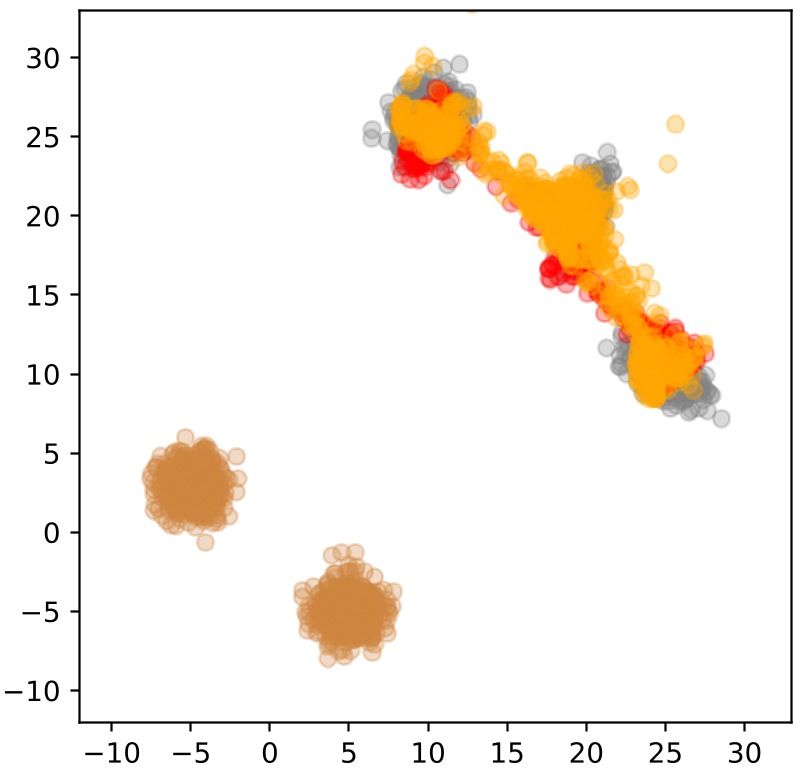} & \includegraphics[width=0.22\textwidth]{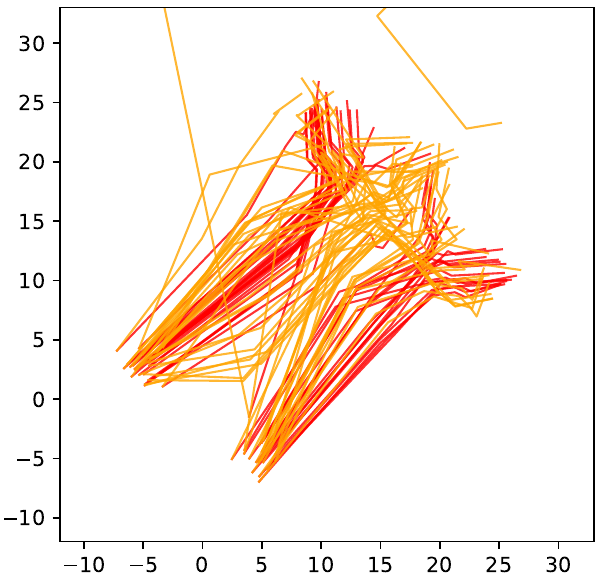} & \includegraphics[width=0.22\textwidth]{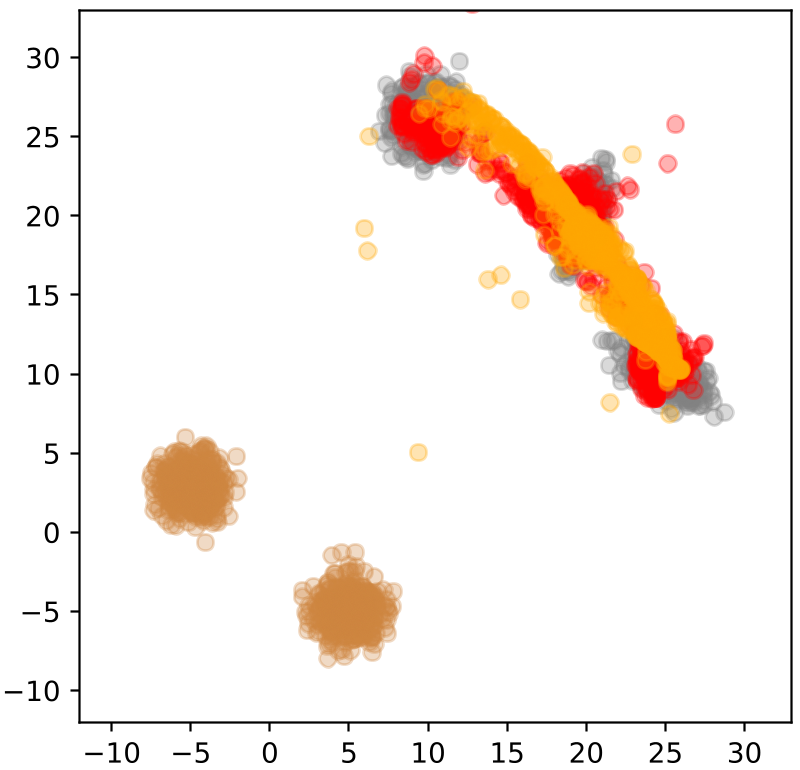} & \includegraphics[width=0.22\textwidth]{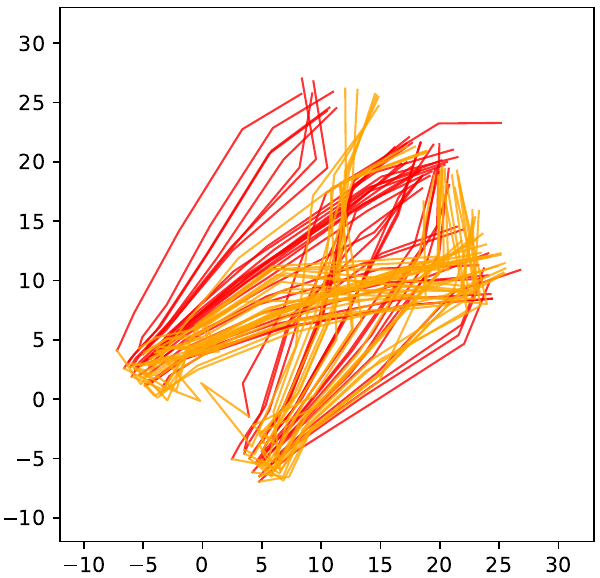}\tabularnewline
\multirow{1}{*}[0.07\textwidth]{50} & \includegraphics[width=0.22\textwidth]{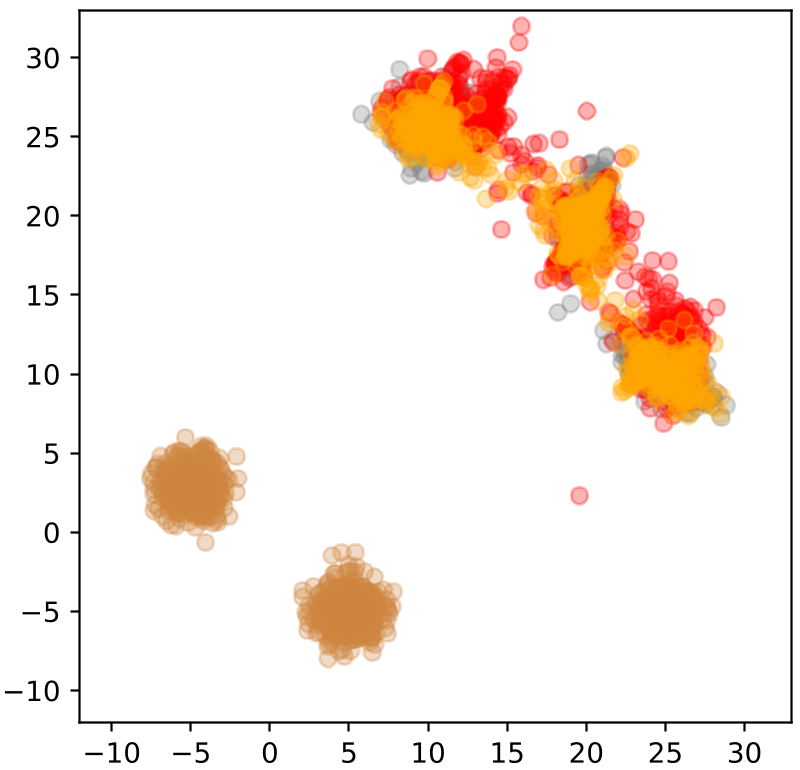} & \includegraphics[width=0.22\textwidth]{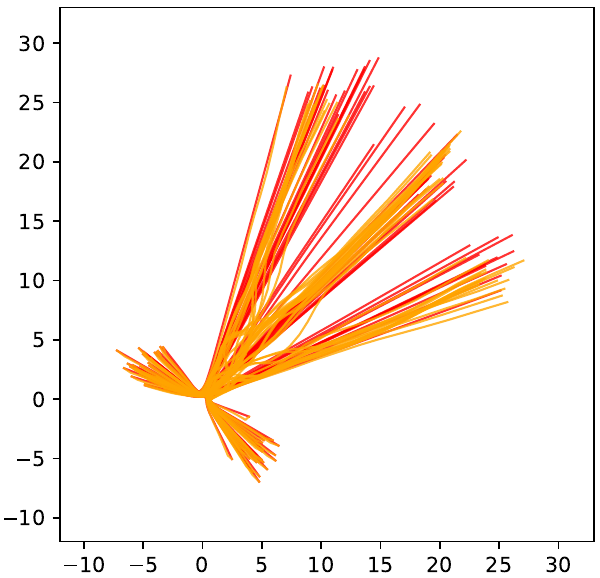} & \includegraphics[width=0.22\textwidth]{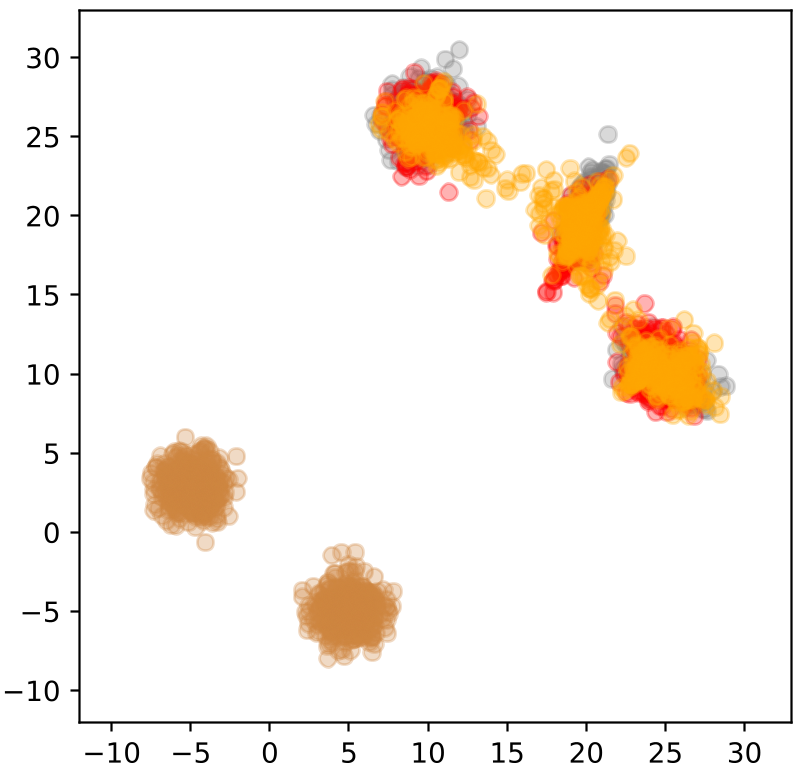} & \includegraphics[width=0.22\textwidth]{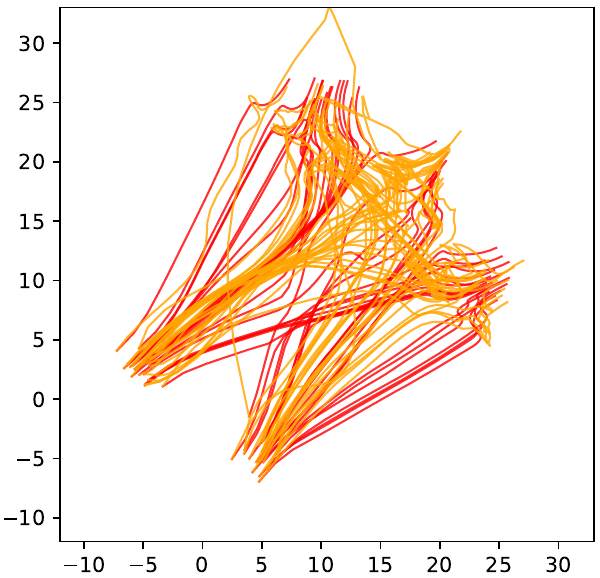} & \includegraphics[width=0.22\textwidth]{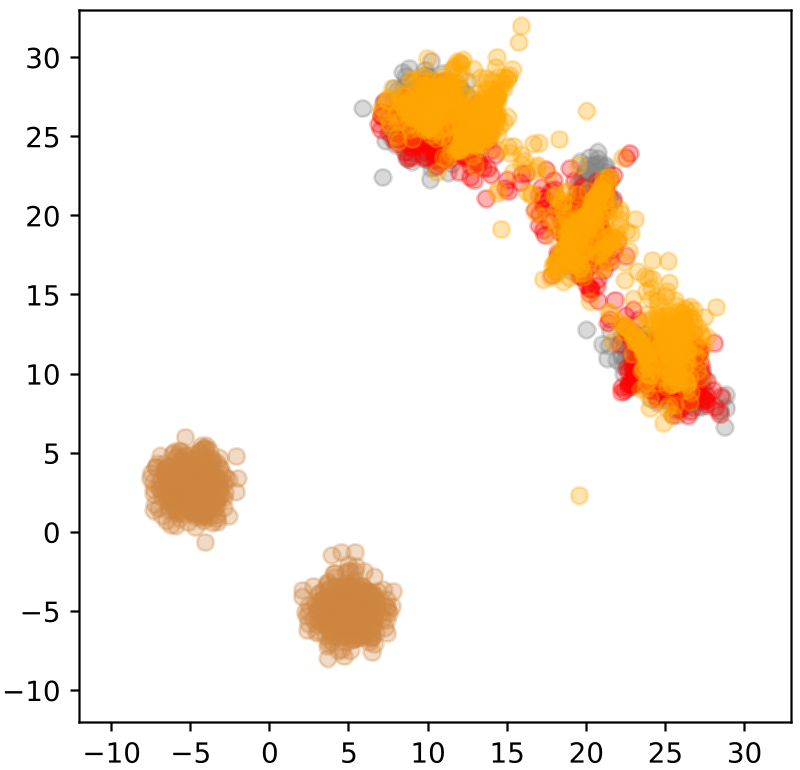} & \includegraphics[width=0.22\textwidth]{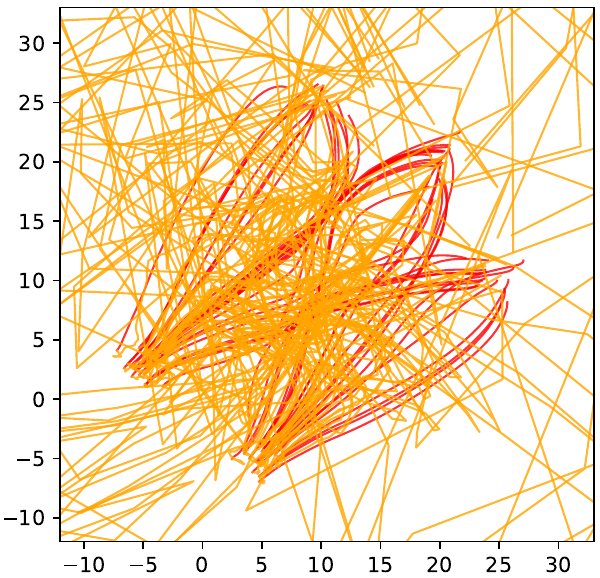}\tabularnewline
\multirow{1}{*}[0.07\textwidth]{1000} & \includegraphics[width=0.22\textwidth]{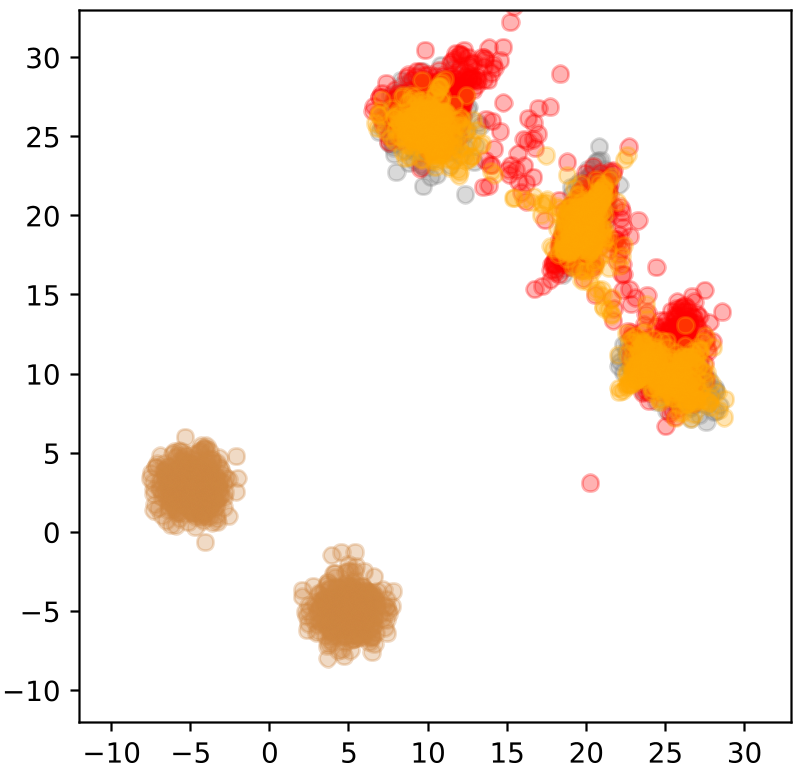} & \includegraphics[width=0.22\textwidth]{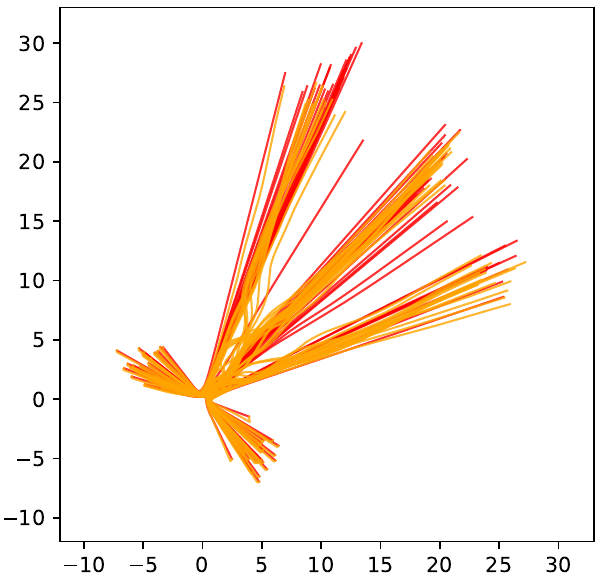} & \includegraphics[width=0.22\textwidth]{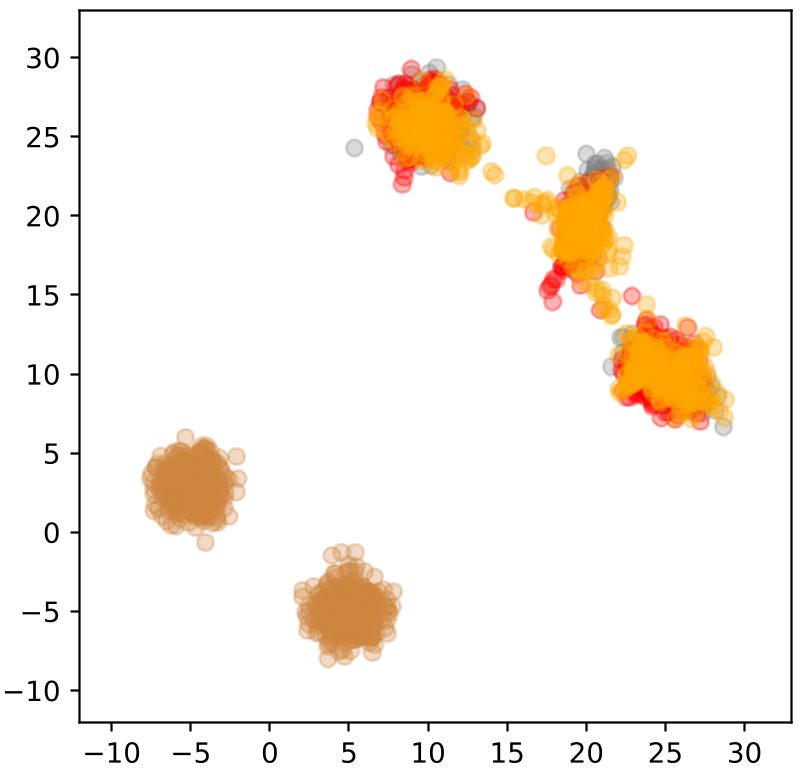} & \includegraphics[width=0.22\textwidth]{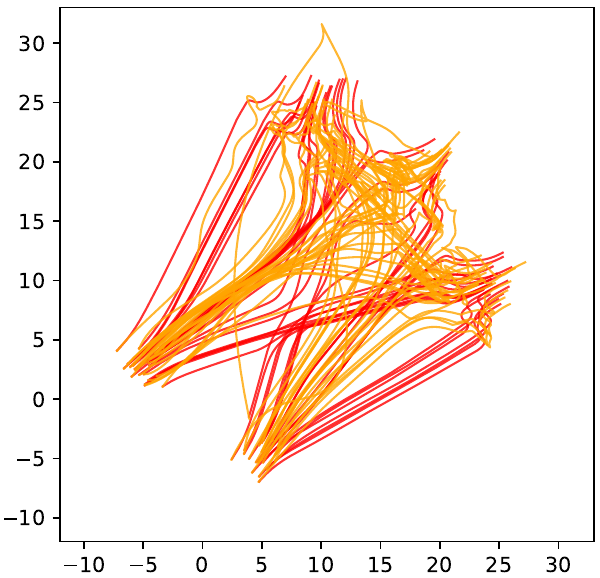} & \includegraphics[width=0.22\textwidth]{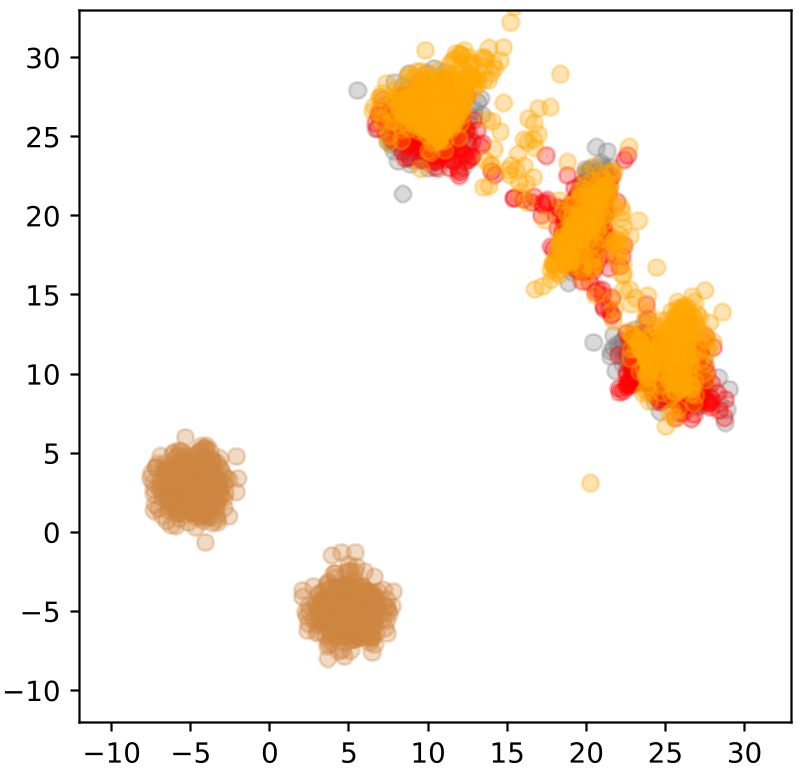} & \includegraphics[width=0.22\textwidth]{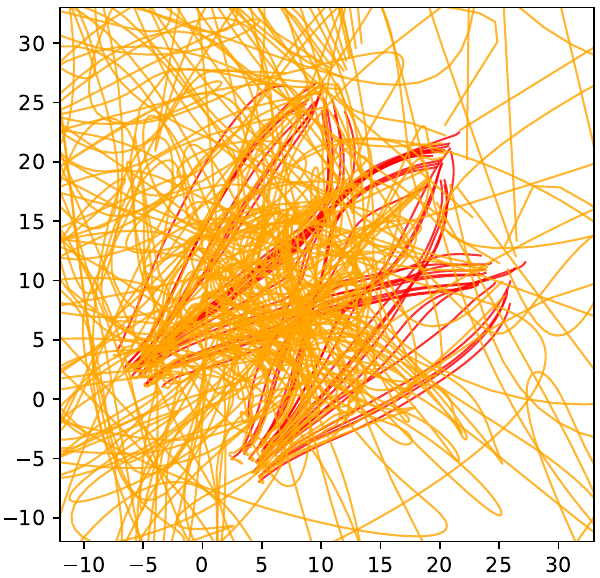}\tabularnewline
 & \multicolumn{2}{c}{VP-SDE-like $\rightarrow$ fifth-degree} & \multicolumn{2}{c}{VP-SDE-like $\rightarrow$ third-degree} & \multicolumn{2}{c}{fifth-degree $\rightarrow$ VP-SDE-like}\tabularnewline
\end{tabular}}}
\par\end{centering}
\caption{Given the learned velocity corresponding to a source process, we simulate
the flow of a target process (labeled as \textcolor{orange}{target
(from source)}) and compare it with the learned flow of the target
process (labeled as \textcolor{red}{target (learned)}).\label{fig:Toy_VPSDELike-to-Third-Fifth}}
\end{figure*}

\begin{figure*}
\begin{centering}
{\resizebox{0.92\textwidth}{!}{%
\par\end{centering}
\begin{centering}
\begin{tabular}{ccc|cc|cc}
\multirow{1}{*}[0.02\textwidth]{$N$} & \multicolumn{6}{c}{\includegraphics[height=0.035\textwidth]{asset/ThreeGauss/legend_sample_origin_to_target}\hspace{0.02\textwidth}\includegraphics[height=0.035\textwidth]{asset/ThreeGauss/legend_traj_origin_to_target}}\tabularnewline
\multirow{1}{*}[0.07\textwidth]{5} & \includegraphics[width=0.22\textwidth]{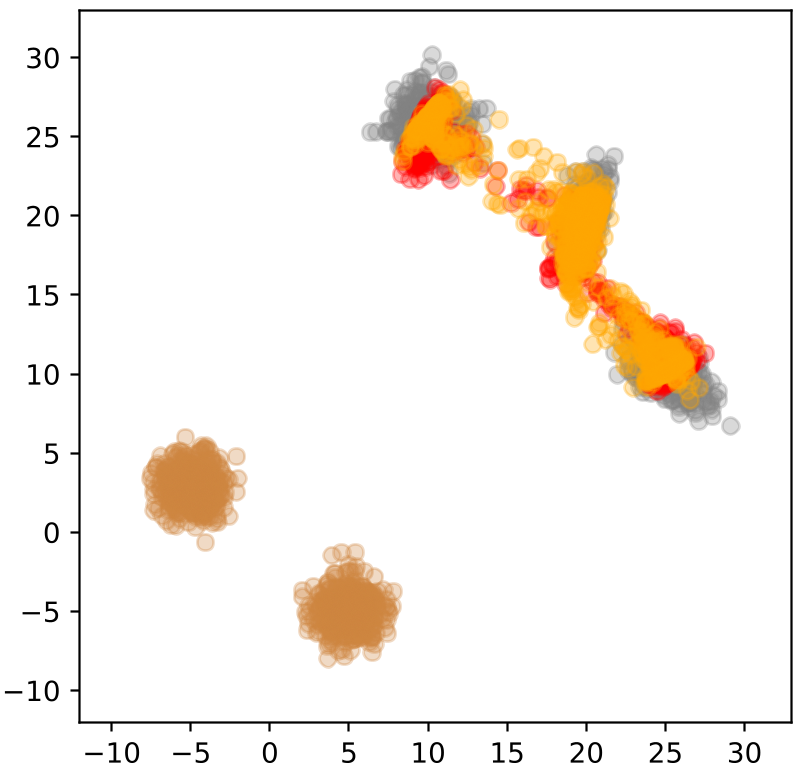} & \includegraphics[width=0.22\textwidth]{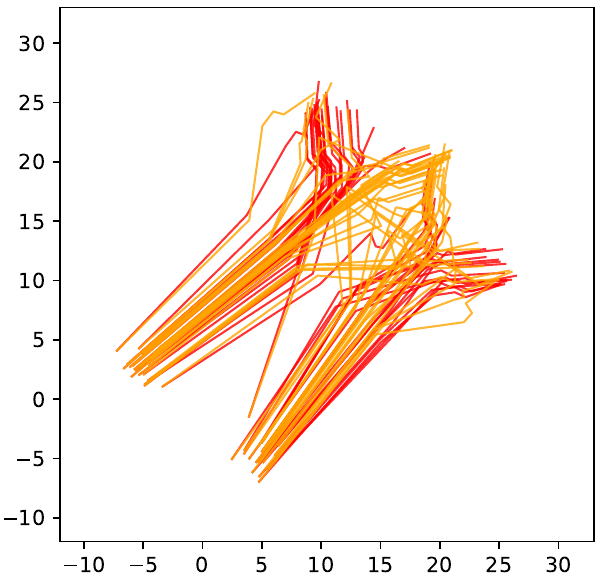} & \includegraphics[width=0.22\textwidth]{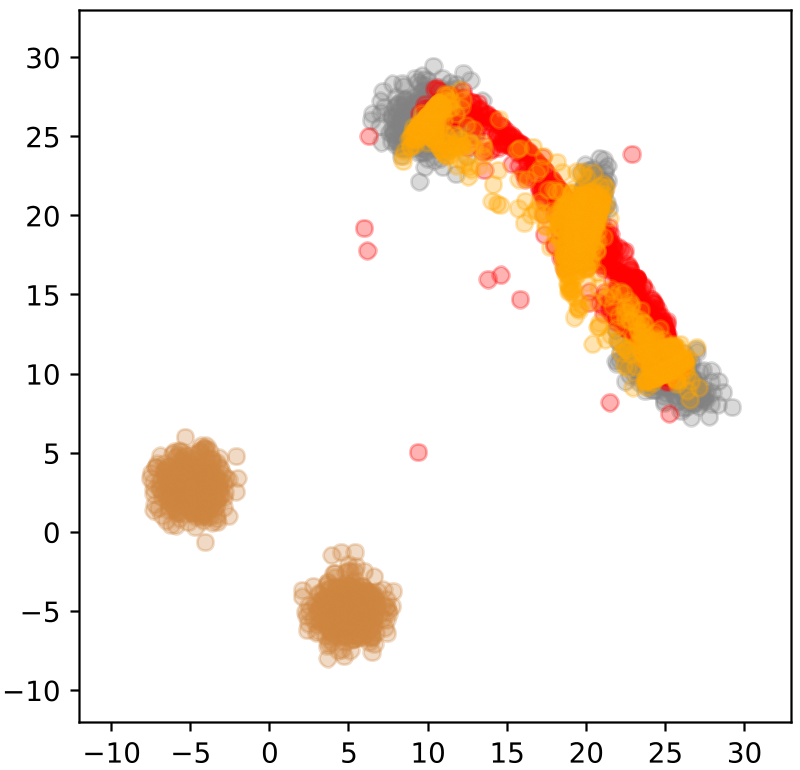} & \includegraphics[width=0.22\textwidth]{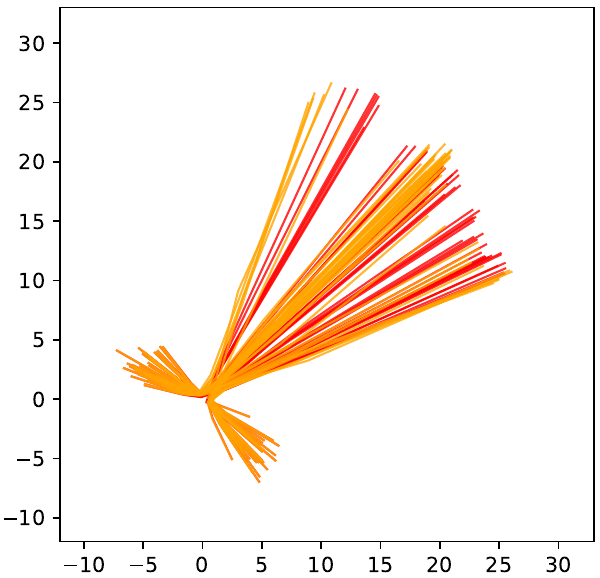} & \includegraphics[width=0.22\textwidth]{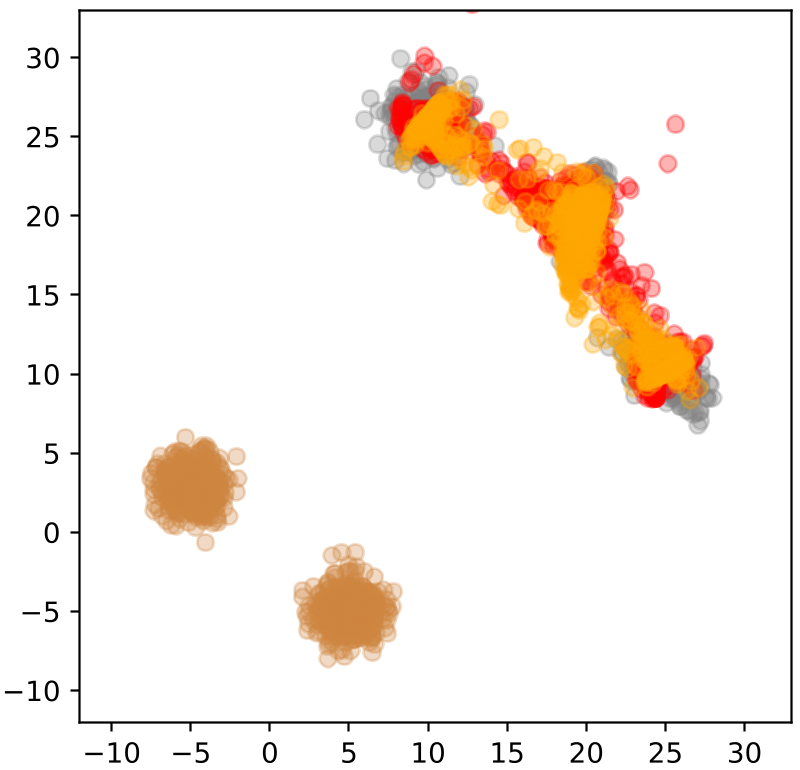} & \includegraphics[width=0.22\textwidth]{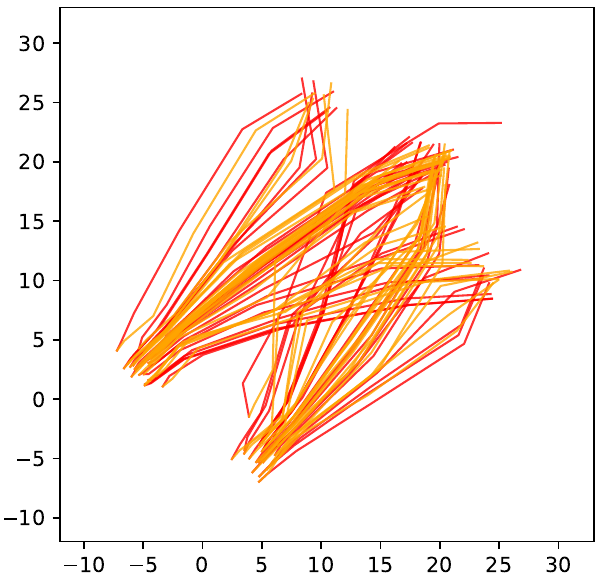}\tabularnewline
\multirow{1}{*}[0.07\textwidth]{50} & \includegraphics[width=0.22\textwidth]{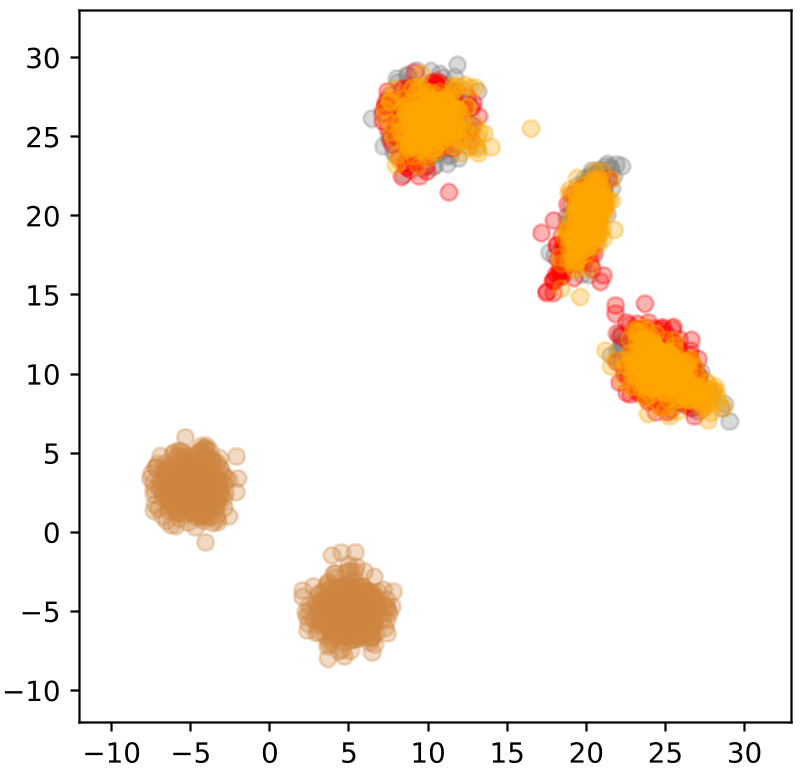} & \includegraphics[width=0.22\textwidth]{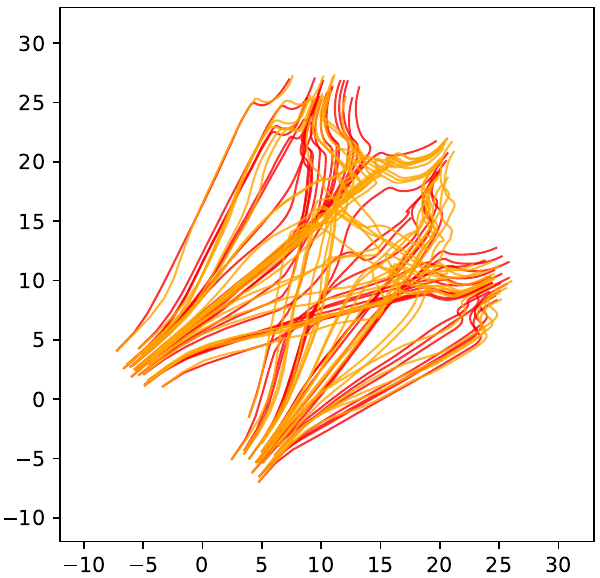} & \includegraphics[width=0.22\textwidth]{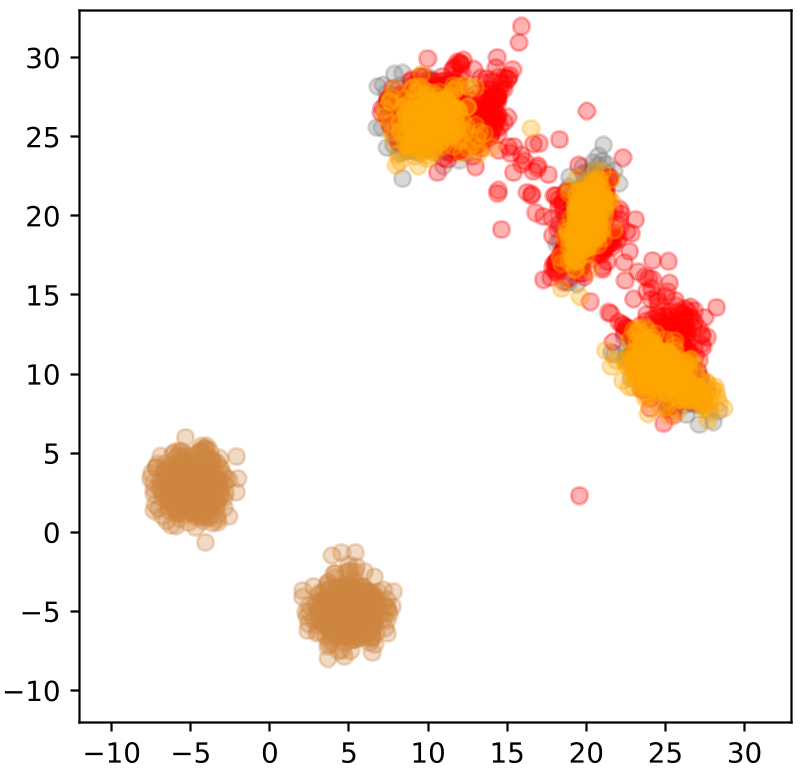} & \includegraphics[width=0.22\textwidth]{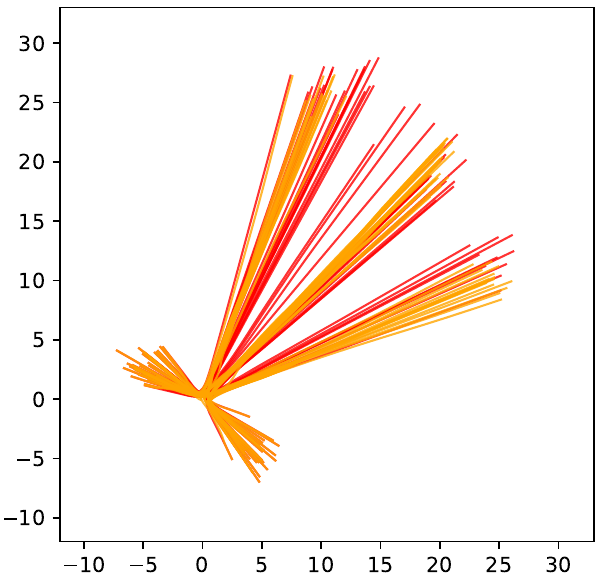} & \includegraphics[width=0.22\textwidth]{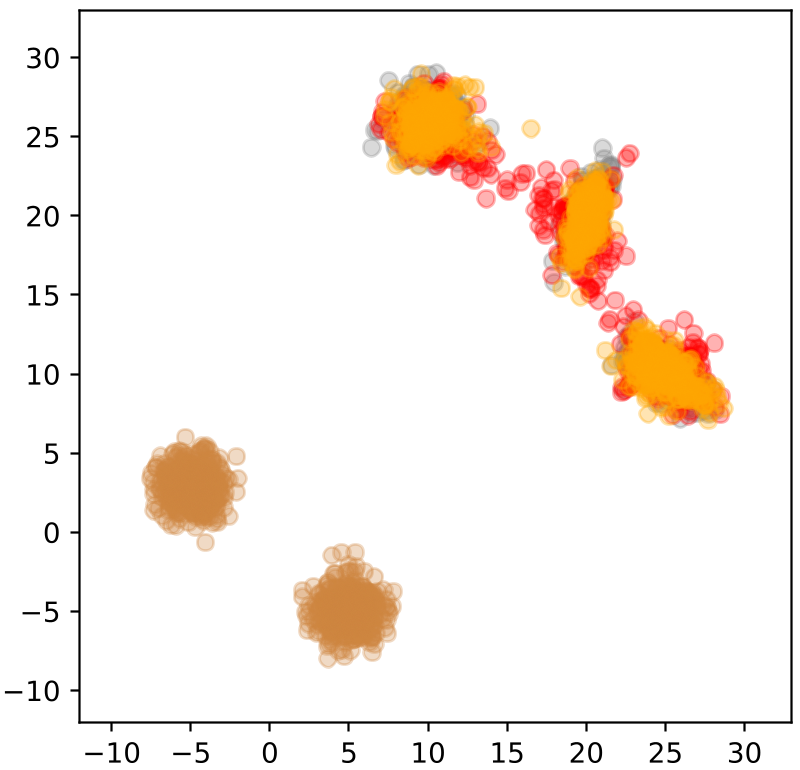} & \includegraphics[width=0.22\textwidth]{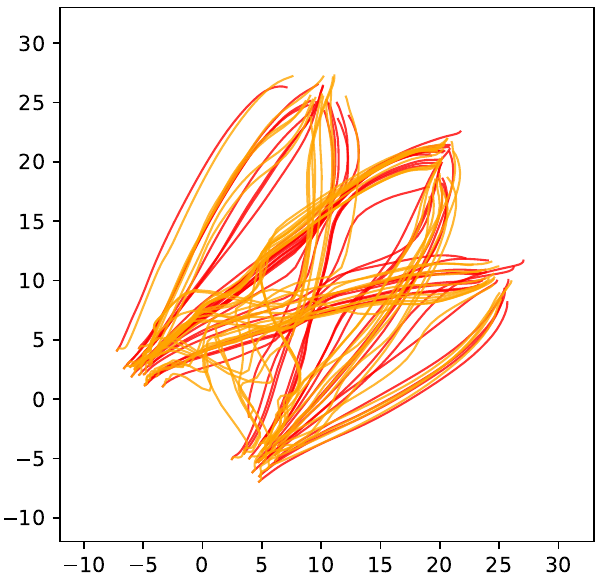}\tabularnewline
\multirow{1}{*}[0.07\textwidth]{1000} & \includegraphics[width=0.22\textwidth]{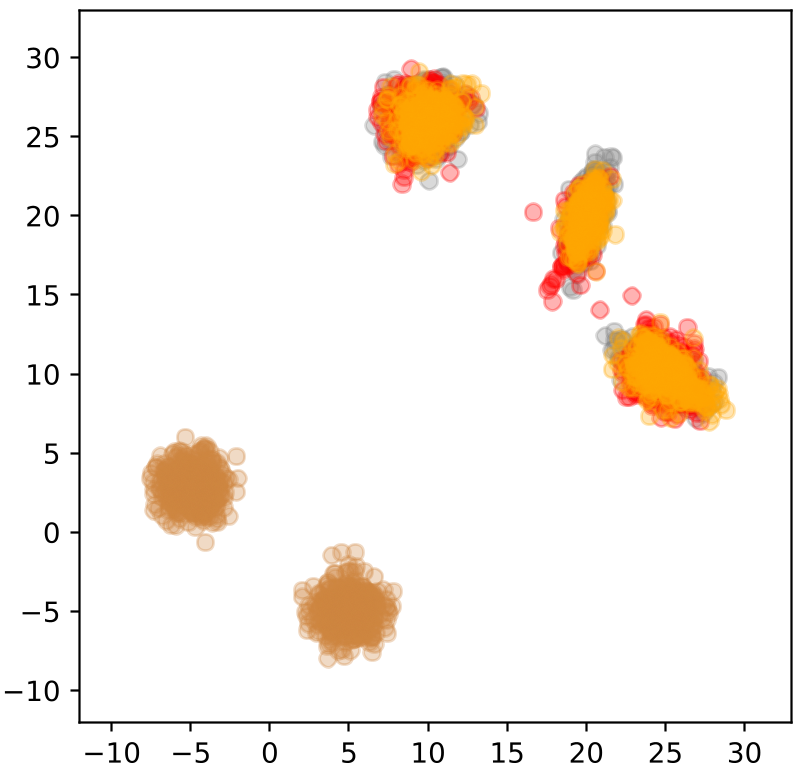} & \includegraphics[width=0.22\textwidth]{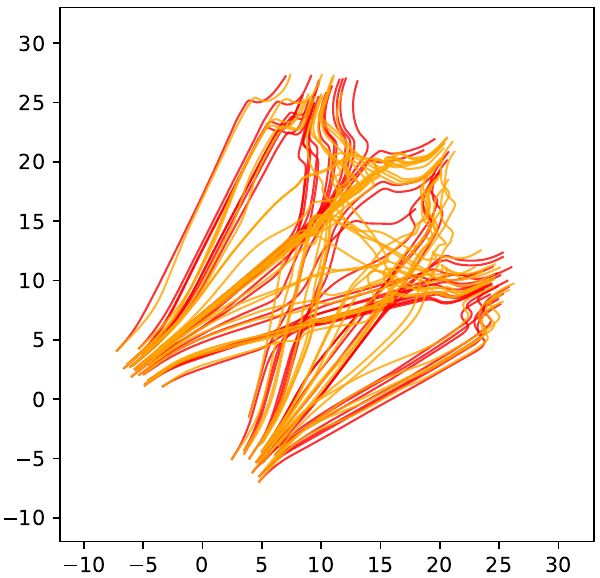} & \includegraphics[width=0.22\textwidth]{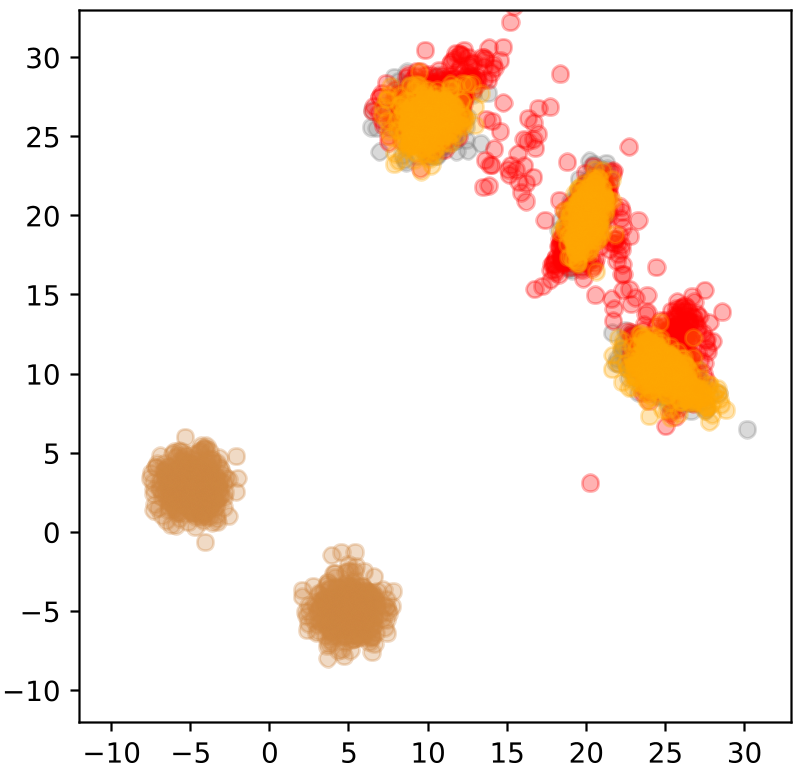} & \includegraphics[width=0.22\textwidth]{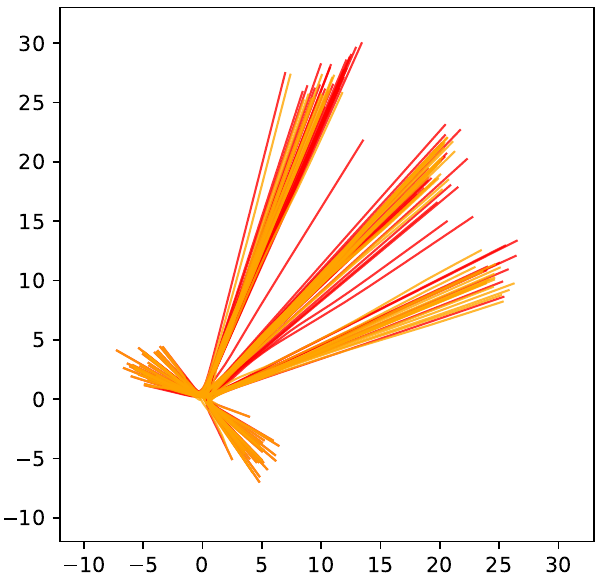} & \includegraphics[width=0.22\textwidth]{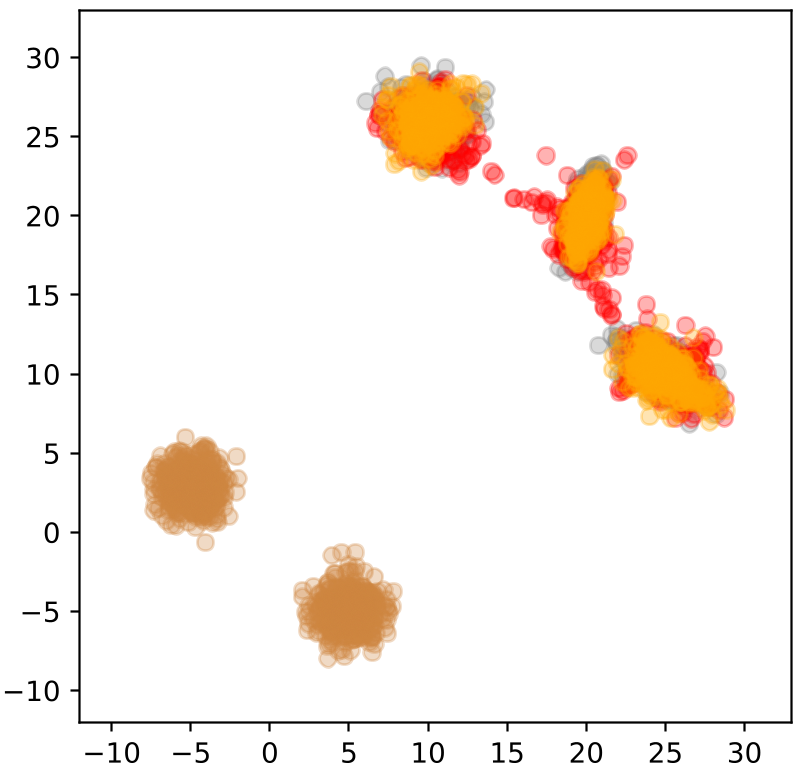} & \includegraphics[width=0.22\textwidth]{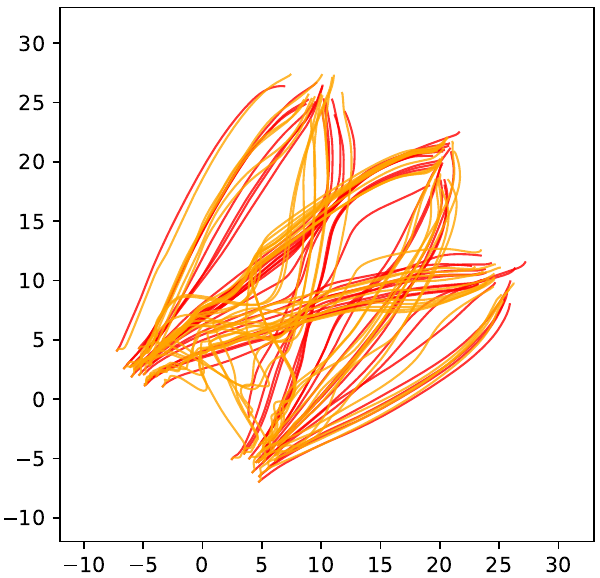}\tabularnewline
 & \multicolumn{2}{c}{rectified $\rightarrow$ third-degree} & \multicolumn{2}{c}{rectified $\rightarrow$ fifth-degree} & \multicolumn{2}{c}{rectified $\rightarrow$ VPSDE-like}\tabularnewline
\end{tabular}}}
\par\end{centering}
\caption{Given the learned velocity of a rectified flow, we simulate the flow
of a target process (labeled as \textcolor{orange}{target (from source)})
and compare it with the learned flow of the target process (labeled
as \textcolor{red}{target (learned)}).\label{fig:Toy_RF-to-others}}
\end{figure*}

By utilizing the second approach in Section~\ref{subsec:Flowing-along-arbitrary-paths},
given the trained velocity corresponding to the third-degree or the
VP-SDE-like process, we can simulate the posterior flows of the other
two processes with a fair quality, as depicted in Figs.~\ref{fig:Toy_Third-to-Fifth-VPSDELike},
\ref{fig:Toy_VPSDELike-to-Third-Fifth}. However, when the fifth-degree
process is used as source, we are unable to simulate the posterior
flows of the third-degree and VP-SDE-like processes even when the
number of updates is set to 1000. We guess a possible reason is that
for the fifth-degree process with $v^{*}\left(x_{t},t\right)$ being
modeled, $v^{*}\left(\bar{x}_{t},t\right)$ is computed as either
$\frac{\left(a_{t}+\sigma_{t}\right)v^{*}\left(x_{t},t\right)-\left(\dot{a}_{t}+\dot{\sigma}_{t}\right)x_{t}}{a_{t}\dot{\sigma}_{t}-\dot{a}_{t}\sigma_{t}}$
(Eq.~\ref{eq:straight_constant_velo_way_1_velo_net}) or $\frac{a_{t}v^{*}\left(x_{t},t\right)-\dot{a}_{t}x_{t}}{a_{t}\dot{\sigma}_{t}-\dot{a}_{t}\sigma_{t}}$
(Eq.~\ref{eq:straight_constant_velo_way_1_noise_net}) and has extremely
large values at $t=1$ and $t=0$ since $a_{t}\dot{\sigma}_{t}-\dot{a}_{t}\sigma_{t}=0$
for this process (see the image at row 3 column 4 in Fig.~\ref{fig:Toy_Coefficients}),
as shown in rows 1, 2 column (e) in Fig.~\ref{fig:Toy_NumericalRobust_FifthDegree}.
Consequently, the computation of samples in the target flow via Eq.~\ref{eq:original_sample_to_target_sample}
tend to be incorrect. The third-degree and VP-SDE-like process do
not exhibit this problem as the velocities of their corresponding
SC flows well behave (rows 1, 2 column (e) in Figs.~\ref{fig:Toy_NumericalRobust_ThirdDegree},
\ref{fig:Toy_NumericalRobust_VPSDELike}). By using the ``standard''
process $X_{t}=\left(1-t\right)X_{0}+tX_{1}$ as the source process,
we can simulate the posterior flows of all the above processes better,
which is shown in Fig.~\ref{fig:Toy_RF-to-others}.

\subsection{Large-scale Image Generation with Classifier-Guidance}

\begin{figure*}
\begin{centering}
{\resizebox{\textwidth}{!}{%
\par\end{centering}
\begin{centering}
\begin{tabular}{cccc}
\multirow{1}{*}[0.09\textwidth]{$s$ = 0} & \includegraphics[width=0.3\textwidth]{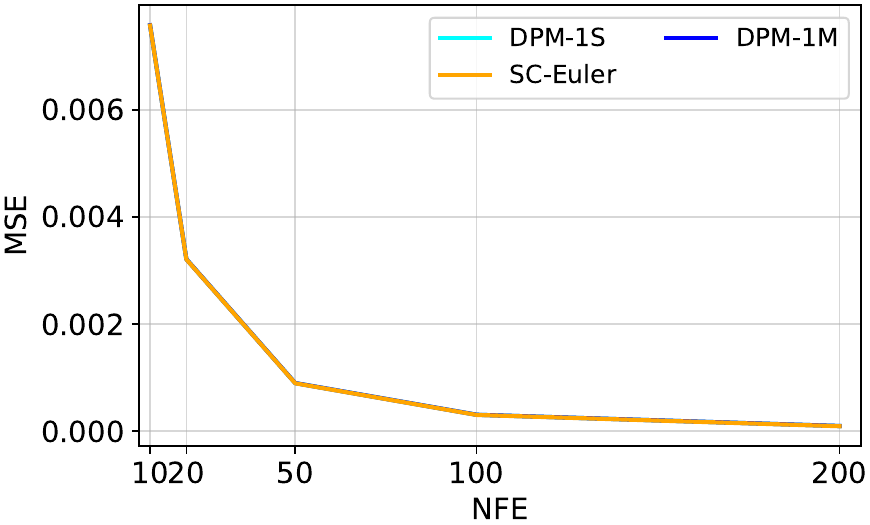} & \includegraphics[width=0.3\textwidth]{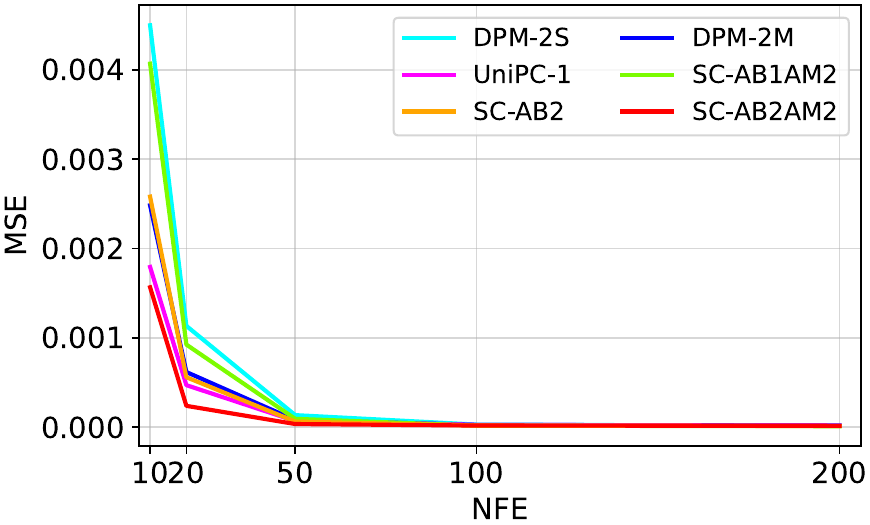} & \includegraphics[width=0.3\textwidth]{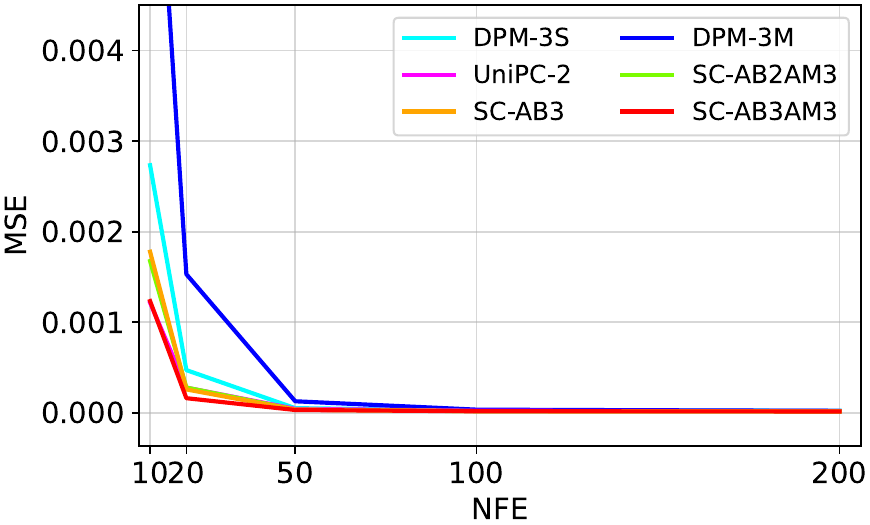}\tabularnewline
\multirow{1}{*}[0.09\textwidth]{$s$ = 8} & \includegraphics[width=0.3\textwidth]{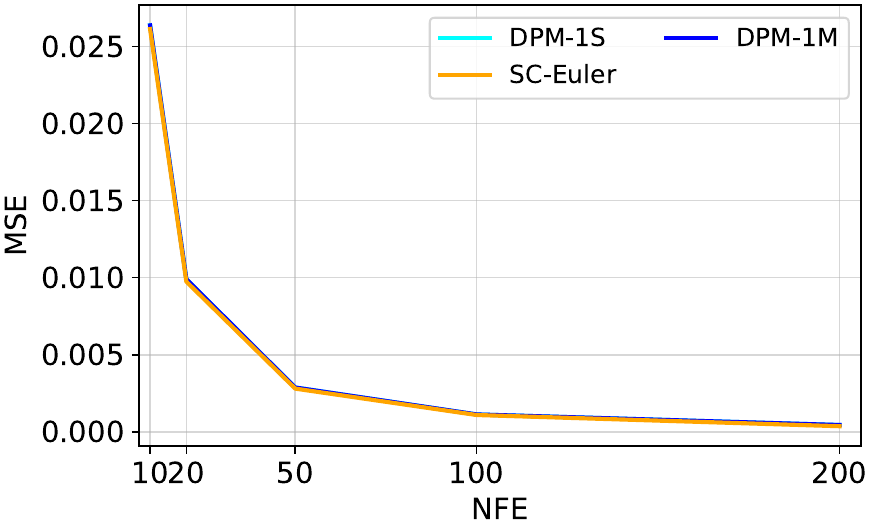} & \includegraphics[width=0.3\textwidth]{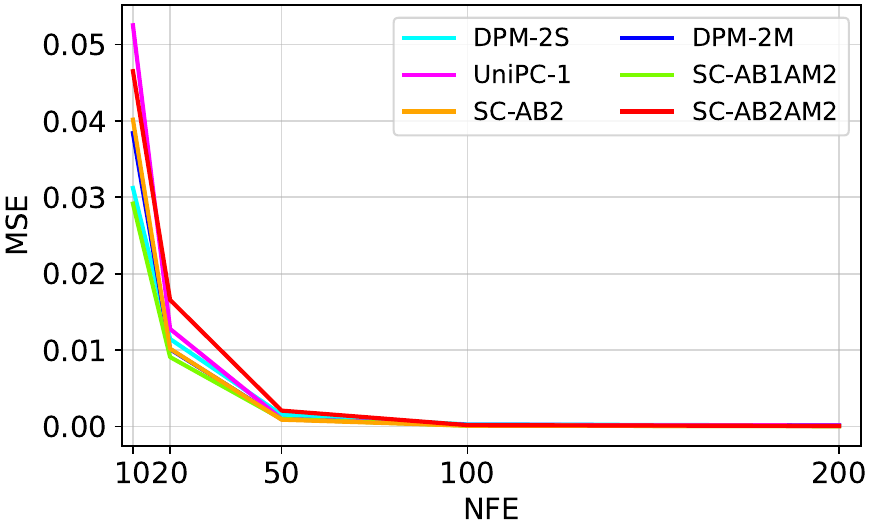} & \includegraphics[width=0.3\textwidth]{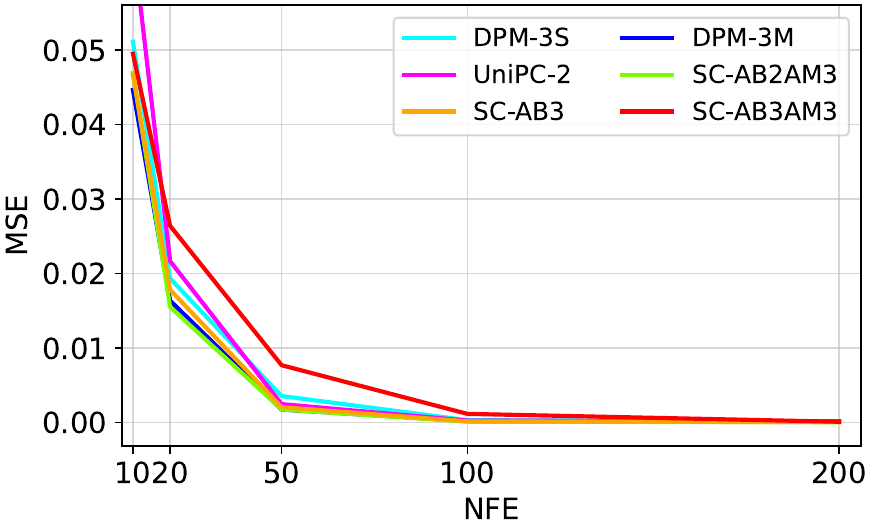}\tabularnewline
 & (a) First-order & (b) Second-order & (c) Third-order\tabularnewline
\end{tabular}}}
\par\end{centering}
\caption{MSEs between 1000 ImageNet 256$\times$256 samples generated by a
1000-step DDIM and those generated by our proposed solvers (SC-{*})
and baselines. The solvers are categorized according to their accuracy
orders. The number of function evaluations varies from 10 to 200.
DPM stands for DPM-Solver++. Top row: Classifier-guidance is not applied.
Bottom row: Classifier-guidance with the guidance scale of 8 is applied.\label{fig:ImageNet256-MSE-200-steps}}
\end{figure*}

\begin{figure*}
\begin{centering}
{\resizebox{\textwidth}{!}{%
\par\end{centering}
\begin{centering}
\begin{tabular}{cccc}
\multirow{1}{*}[0.09\textwidth]{$s$ = 0} & \includegraphics[width=0.3\textwidth]{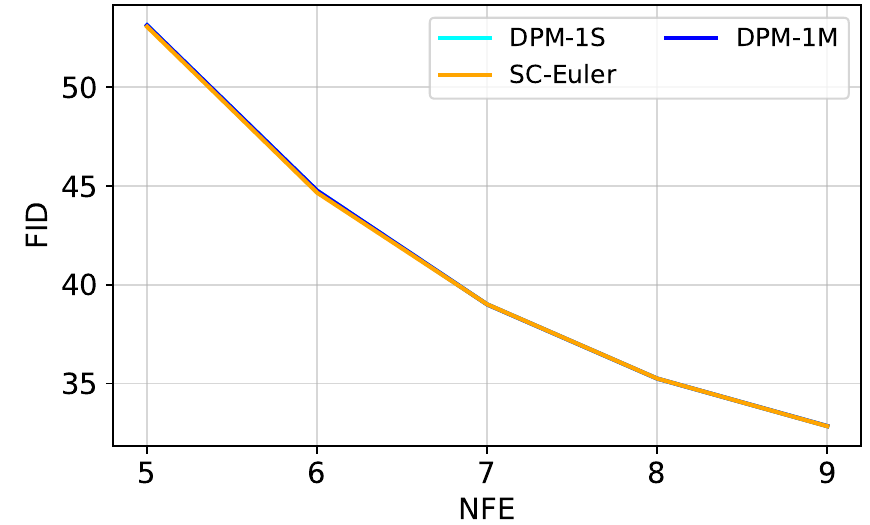} & \includegraphics[width=0.3\textwidth]{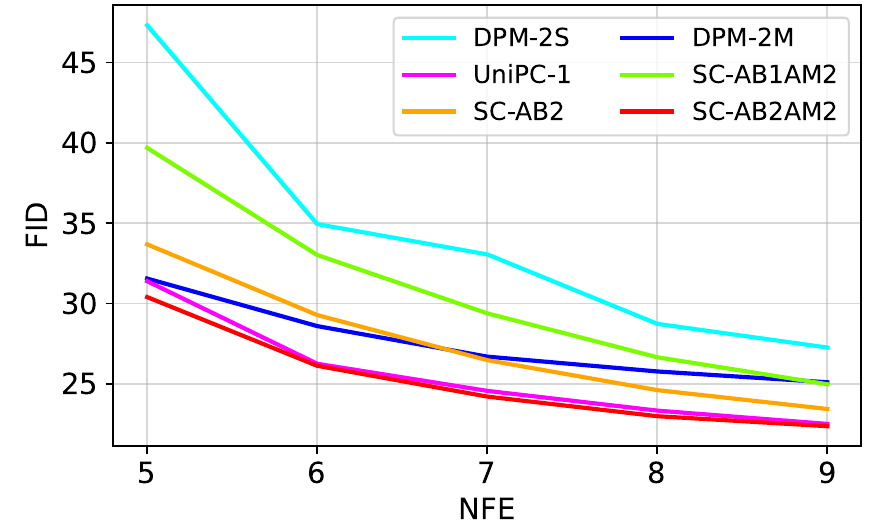} & \includegraphics[width=0.3\textwidth]{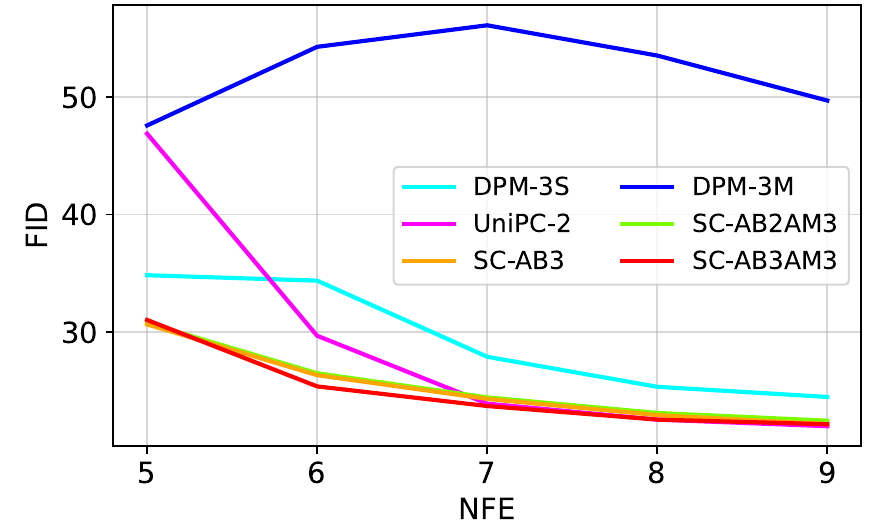}\tabularnewline
\multirow{1}{*}[0.09\textwidth]{$s$ = 8} & \includegraphics[width=0.3\textwidth]{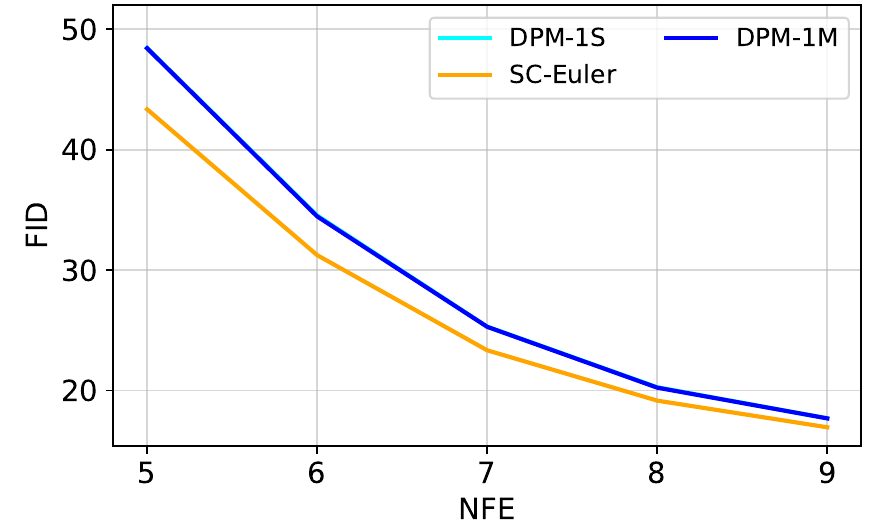} & \includegraphics[width=0.3\textwidth]{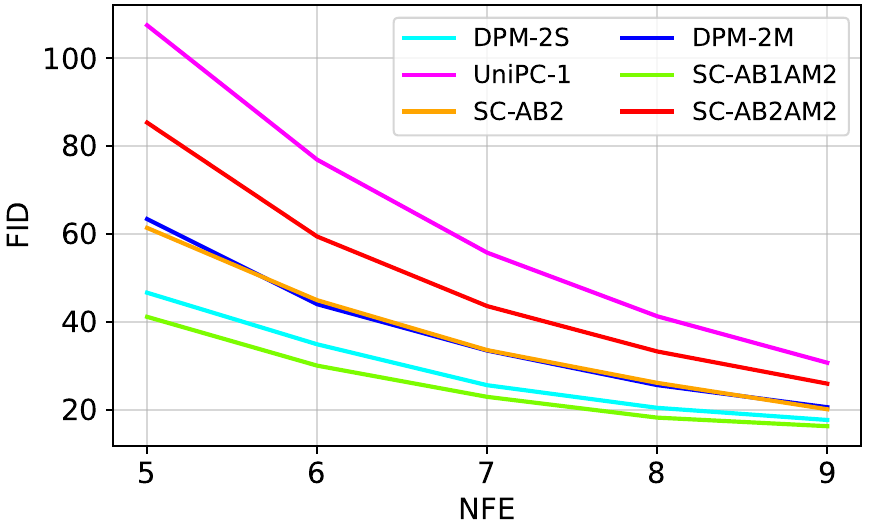} & \includegraphics[width=0.3\textwidth]{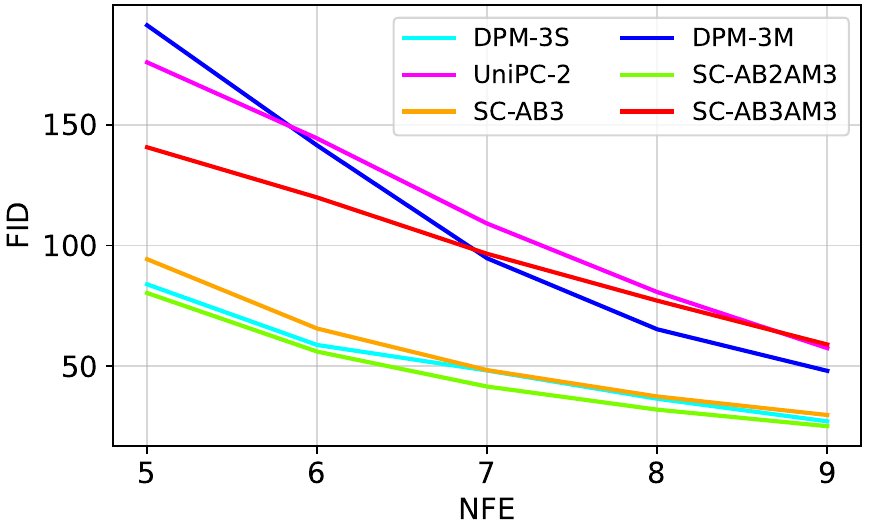}\tabularnewline
 & (a) First-order & (b) Second-order & (c) Third-order\tabularnewline
\end{tabular}}}
\par\end{centering}
\caption{FIDs of 5000 ImageNet 256$\times$256 samples generated by our proposed
solvers (SC-{*}) and baselines with varying accuracy orders (1-3)
and numbers of function evaluations (5-9). DPM stands for DPM-Solver++.
Top row: Classifier-guidance is not applied. Bottom row: Classifier-guidance
with the guidance scale of 8 is applied.\label{fig:ImageNet256-FID-few-steps}}
\end{figure*}

In this section, we compare the performances of our proposed high-order
ODE solvers (Section~\ref{subsec:Higher-order-numerical-solvers})
with current state-of-the-art ODE solvers for diffusion models \cite{lu2022dpm,lu2022dpmpp,zhao2023unipc}
in the context of large-scale image generation tasks. To streamline
presentation, we focus on showcasing the generation results specifically
for ImageNet 256$\times$256.

Our implementation builds upon the open-source code\footnote{https://github.com/openai/improved-diffusion}
provided by the authors of the paper \cite{dhariwal2021diffusion}.
In addition, we utilize their pretrained class-conditional DDPM on
ImageNet 256$\times$256 images for image generation. The baselines
are DPM-Solver++ \cite{lu2022dpmpp} and UniPC \cite{zhao2023unipc}.
They are run using their official source codes\footnote{https://github.com/LuChengTHU/dpm-solver}\footnote{https://github.com/wl-zhao/UniPC}.
We keep all default settings of the baselines except disabling dynamic
thresholding \cite{saharia2022photorealistic} since it has not been
implemented in our methods.

We conduct two different experiments. In the first experiment, we
aim to examine the convergence capability and speed of our methods
and the baselines. We vary the number of function evaluations $N$
from 10 to 200 and generate 1000 samples for each value of $N$. The
mean square error (MSE) between these samples and 1000 samples generated
by a 1000-step DDIM \cite{song2020denoising} serves as the convergence
error. To ensure a fair play ground, numerical methods are categorized
according to their order of accuracy and only methods in the same
category are compared. As illustrated in Fig.~\ref{fig:ImageNet256-MSE-200-steps},
our methods as well as the baselines converge as $N$ increases. Our
SC-Euler also performs similarly to the first-order variants of DPM-Solver++
as they all resemble DDIM. When classifier guidance is not applied
($s$ = 0), higher-order solvers tend to converge faster than lower-order
ones (i.e., requires fewer steps to reach the same MSE value). However,
when the classifier guidance is applied with $s$ = 8, third-order
solvers seem to converge slower than second-order solvers. In addition,
at $N$ = 10, first-order solvers achieve better MSE values than second-
and third-order solvers (about 0.025 vs 0.04, 0.05). The main reason
is that when the classifier guidance is not applied, the posterior
flow remains unchanged and updates along this flow or its associated
SC flow should follow the theoretical analysis on convergence rate.
By contrast, when the classifier guidance is applied, the posterior
flow is modified and the theory may not be applicable. In this case,
lower-order solvers, which are more robust to errors, tend to perform
better than higher-order solvers, especially when $N$ is small.

Therefore, in the second experiment, our target is to compare the
quality of images generated by our methods and the baselines when
using a small number of function evaluations. We generate 5000 images
for each value of $N$ from 5 to 9 and compute the Frechet Inception
Distance (FID) \cite{heusel2017gans} between these images and those
in the ImageNet training set using the pytorch-fid library\footnote{https://github.com/mseitzer/pytorch-fid}.
The ImageNet training images are resized and center-cropped to the
size 256$\times$256. Fig.~\ref{fig:ImageNet256-FID-few-steps} shows
the results of this trial. It is clear that the performances of our
methods follow the increasing order from bad to good: AB1-AM2 (AB2-AM3)
< AB2 (AB3) < AB2-AM2 (AB3-AM3) when there is no classifier guidance
and follow the reverse order when classifier guidance is employed.
It could possibly because the convergence rates of these methods follow
this order while the levels of robustness follow the reverse order.
For the baseline solvers, we also observe the same phenomenon. The
second- and third-order variants of UniPC perform better than the
second- and third-order variants of DPM-Solver++ M/S when $s$ = 0
but worse when $s$ = 8. These results suggest that there is no winning
solver in all cases but should be customized for certain scenarios.
In this experiment, we see that when $N$ is small, SC-AB1AM2 (SC-AB2AM3)
is more accurate and robust than DPM-Solver++ 2S (3S), SC-AB2 is on
par with DPM-Solver++ 2M regardless of classifier guidance, and SC-AB2AM2
(SC-AB3AM3) has quite similar level of accuracy to UniPC-1 (2) but
is slightly more robust.

\subsection{Text-to-Image Generation with Stable Diffusion}

\begin{figure*}
\begin{centering}
{\resizebox{\textwidth}{!}{%
\par\end{centering}
\begin{centering}
\begin{tabular}{ccc}
\includegraphics[width=0.3\textwidth]{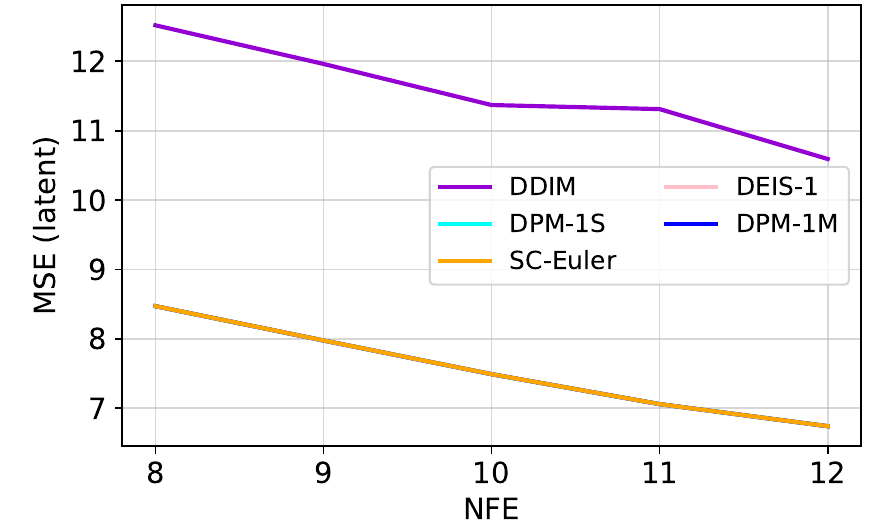} & \includegraphics[width=0.3\textwidth]{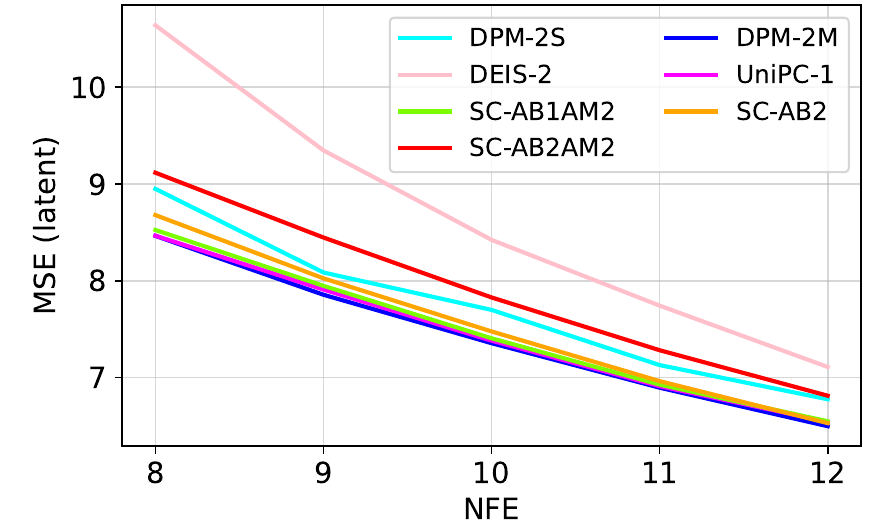} & \includegraphics[width=0.3\textwidth]{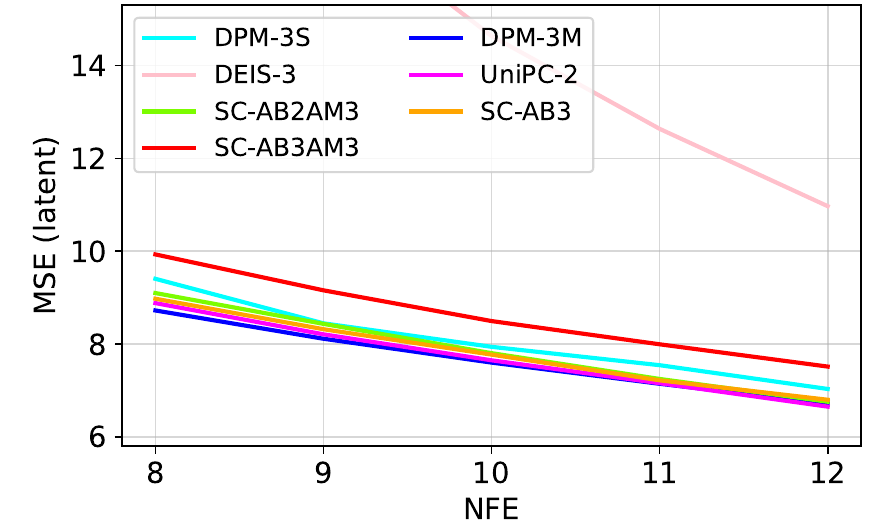}\tabularnewline
(a) First-order & (b) Second-order & (c) Third-order\tabularnewline
\end{tabular}}}
\par\end{centering}
\caption{MSEs between 1000 \emph{latent} samples generated by a 1000-step DDIM
and those generated by our proposed solvers (SC-{*}) and baseline
solvers. The model is Stable Diffusion and the text prompts are COCO2014
captions. DPM stands for DPM-Solver++.\label{fig:StableDiffusion-MSE-latent-few-steps}}
\end{figure*}

We continue comparing our proposed multi-step solvers with baseline
solvers \cite{lu2022dpmpp,zhang2023fast,zhao2023unipc} on the text-to-image
generation task with Stable Diffusion. We utilize the pretrained Stable
Diffusion from the Hugging Face repository CompVis/stable-diffusion-v1-4\footnote{https://huggingface.co/CompVis/stable-diffusion-v1-4},
which can also be found in the Github link\footnote{https://github.com/CompVis/stable-diffusion}.
The baseline solvers are taken from the Diffusers library \cite{von-platen-etal-2022-diffusers}
with the default configurations. Following \cite{lu2022dpmpp,zhao2023unipc},
we randomly sample 1000 image-caption pairs from the COCO2014 validation
set \cite{lin2014microsoft} and use the captions as text prompts
for the pretrained Stable Diffusion to generate 1000 images of size
512$\times$512. The default guidance scale $s$ = 7.5 is applied.
The mean square error (MSE) between the \emph{latent} samples generated
by a 1000-step DDIM and those generated by each numerical solvers
serves as a convergence error.

From Fig.~\ref{fig:StableDiffusion-MSE-latent-few-steps} (a), it
is expected that our Euler solver performs similarly to the first-order
variants of DPM-Solver++ and DEIS as they are all equivalent to DDIM.
However, the considerably worse results of DDIM surprises us since
in theory, they should be the same as ours and the baselines'. Our
hypothesis is that the implementation of DDIM in the Diffusers library
may not be the best version.

In terms of second- and third-order solvers, our SC-AB1AM2 (-AB2AM3)
and SC-AB2 (-AB3) are comparable to DPM-Solver++ 2M (3M) and UniPC-1
(-2) and slightly better than DPM-Solver++ 2S (3S) as shown in Fig.~\ref{fig:StableDiffusion-MSE-latent-few-steps}
(b, c). It is because both DPM-Solver++ and its derivative, UniPC,
implicitly perform an SC transformation in their implementations (Appdx.~\ref{sec:Analysis-of-Existing-High-Order-Solvers}).
Among all high-order solvers, DEIS achieves the worst performance,
possibly because of it operates on the posterior flow rather than
the SC flow. These results reflect the qualities of images generated
by our solvers and the baseline solvers in Figs.~\ref{fig:GenImages_StableDiffusion_Order1},
\ref{fig:GenImages_StableDiffusion_Order2}, \ref{fig:GenImages_StableDiffusion_Order3}.

\section{Related Work}

\subsection{Diffusion and Flow Models}

Diffusion models were proposed in \cite{ho2020denoising,sohl2015deep,song2019generative}
under the variational inference and score matching frameworks. Originally,
these models only supported discrete time and their sampling was primarily
based on Markov Chain Monte Carlo (MCMC) methods such as Langevin
dynamics \cite{Welling2011}. The research by Song et al. \cite{song2020score}
established an interesting connection between diffusion models and
stochastic differential equations (SDEs), allowing to represent diffusion
models in the continuous-time form. In addition, this work introduced
probability flow ordinary differential equations (PF ODEs), which
share the same marginal distributions of samples as the original diffusion
SDEs. These PF ODEs are easier to solve due to their deterministic
update steps, thereby accelerating the sampling process of diffusion
models without compromising the sample quality. Consequently, this
advancement has driven subsequent researches on a general class of
ODE-based flows connecting two arbitrary distributions \cite{albergo2022building,lipman2022flow,liu2022rectified,liu2022flow}.
Among various flow models, those that enable fast sampling are of
special interest. They include straight flows (e.g., Rectified Flow)
\cite{liu2022rectified,liu2022flow,pooladian2023multisample} and
flows derived from optimal transport maps \cite{finlay2020train,lipman2022flow,onken2021ot}.
However, these methods necessitate training a flow model with the
desired straightness property, which can be very expensive especially
for large models like Stable Diffusion. By contrast, our method can
straighten any pretrained non-straight flow models without training
a new one.

\subsection{Variational Interpretation of Diffusion and Flow Models}

As far as we are aware, the connection between score matching and
maximum likelihood was explored early by Lyu \cite{lyu2009interpretation}.
This study interprets the score matching loss as the Fisher divergence
and links it to the KL divergence in the case the diffusion process
is $X_{t}=X_{0}+\sqrt{t}X_{1}$ with $X_{1}\sim\mathcal{N}\left(0,\mathrm{I}\right)$.
\cite{sohl2015deep} proposed discrete-time diffusion models within
the variational inference framework. Ho et al. \cite{ho2020denoising}
utilized the noise matching loss $\Loss_{\text{NM}}$ to train diffusion
models, interpreting it as a weighted variational bound. Kingma et
al. \cite{kingma2021variational} extended the variational analysis
to a general class of diffusion processes characterized by the equation
$X_{t}=a_{t}X_{0}+\sigma_{t}X_{1}$ with $X_{1}\sim\mathcal{N}\left(0,\mathrm{I}\right)$.
They formally defined the signal-to-noise ratio (SNR) at each time
step $t$ of these processes and reformulated the variational bound
in terms of the SNR. An important finding in their work is that in
the continuous-time setting, the variational bound remains \emph{invariant}
to the noise schedule (i.e., the choice of $a_{t}$, $\sigma_{t}$),
except for the SNR values at $t=0$ and $t=1$. This result complements
our work by ensuring that given a learned flow, we can simulate any
flow as long as the SNRs of the two flows at the end points are the
same. Besides, their transformation of time from $t$ to $\text{SNR}\left(t\right)$
relates to our transformation from $t$ to $\varphi_{t}$ and was
utilized to derive fast numerical solvers in \cite{lu2022dpm,lu2022dpmpp}.
Song et al. \cite{song2021maximum} directly derived the variational
bound for diffusion models in their continuous-time SDE forms rather
than generalizing from the discrete-time setting as in \cite{kingma2021variational}.
Their approach involves computing the KL divergence between the ground-truth
and parameterized SDEs, which is upper-bounded by the KL divergence
between the two corresponding path measures. This KL divergence is
then transformed into the score matching loss using the Girsanov theorem
\cite{oksendal2013stochastic}. Kim et al. \cite{kim2022maximum}
expanded the result in \cite{song2021maximum} to nonlinear diffusion
processes. Albergo et al. \cite{albergo2023stochastic} derived the
variational bound for general flow models using the term $\frac{dD_{\text{KL}}}{dt}$,
which differs from previous approaches \cite{kingma2021variational,song2021maximum}
that utilize $D_{\text{KL}}$. The authors showed that $\frac{dD_{\text{KL}}}{dt}$
between two marginal distributions of samples at time $t$ can be
expressed analytically by applying the continuity equation.

\subsection{Fast Sampling Methods for Diffusion Models}

Improving the sampling process of diffusion models has been an important
research topic since the time this kind of models was proposed \cite{kong2021fast,nichol2021improved,song2020denoising,watson2021learningfast,watson2021learning}.
Improved samplers can generally be categorized as either training-based
\cite{luhman2021knowledge,salimans2022progressive,song2023consistency}
or training-free methods \cite{bao2022analytic,liu2022pseudo,lu2022dpm,lu2022dpmpp,song2020denoising,zhang2023fast,zhao2023unipc}.
Training-based methods typically involve in training another network
that directly transforms the latent sample $x_{1}$ to the data sample
$x_{0}$ in a single step via distilling knowledge from a pretrained
diffusion model. Some methods like \cite{luhman2021knowledge,salimans2022progressive}
need to generate $\left(x_{0},x_{1}\right)$ pairs for distillation
while others avoid this by leveraging the consistency loss \cite{song2023consistency}.
By contrast, training-free methods usually exploit special structures
in the generation process of diffusion models to reduce the number
of steps. Analytic-DDPM \cite{bao2022analytic} estimates the optimal
variance in the denoising distribution $q\left(x_{t_{i-1}}|x_{t_{i}}\right)$
of DDPM. DDIM \cite{song2020denoising} uses a different denoising
distribution $q\left(x_{t_{i-1}}|x_{t_{i}},x_{0|t_{i}}\right)$ that
utilizes the information in $x_{0|t_{i}}$ to improve sample quality.
The variance of this distribution can vary and when it equals 0, the
sampling process becomes deterministic. Karras et al. \cite{karras2022elucidating}
introduced several important modifications to the design of diffusion
models to improve the sampling speed and the quality of generated
images. These changes include training a ``straight'' process $X_{t}=X_{0}+\sigma_{t}X_{1}$
instead of the original diffusion process, modeling the denoised sample
(i.e., $x_{0|t}$) instead of noise (i.e., $x_{1|t}$), and employing
second-order Heun's solver for fast sampling. DEIS \cite{zhang2023fast}
and DPM-Solver \cite{lu2022dpm} exploit the semi-linear structure
of PF ODEs to apply exponential integrators. The later method employs
an additional change-of-variables operation that links to our SC transformation
(Appdx.~\ref{subsec:DPM-Solver-Analysis}). DPM-Solver++ \cite{lu2022dpmpp}
extends DPM-Solver by supporting the model of $x_{0|t}$ (\cite{karras2022elucidating})
besides the noise model as well as multi-step methods. UniPC \cite{zhao2023unipc}
integrates a corrector into DPM-Solver++, making it a predictor-corrector
method. Our work takes a different approach from these works by leveraging
straight, constant-speed flows to perform updates, which results in
a more generalizable framework that can be applied to a larger class
of stochastic processes rather than just diffusion processes.

\section{Discussion}

\begin{figure}
\begin{centering}
{\resizebox{\textwidth}{!}{%
\par\end{centering}
\begin{centering}
\begin{tabular}{cccc}
 & Euler & AB2 & AB2AM2\tabularnewline
\multirow{1}{*}[0.25\textwidth]{Posterior} & \includegraphics[width=0.5\textwidth]{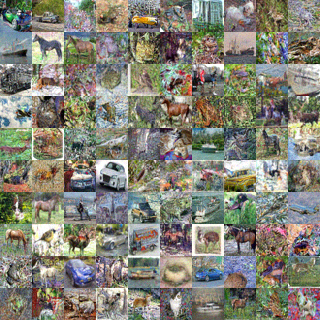} & \includegraphics[width=0.5\textwidth]{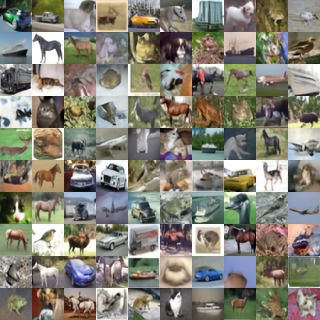} & \includegraphics[width=0.5\textwidth]{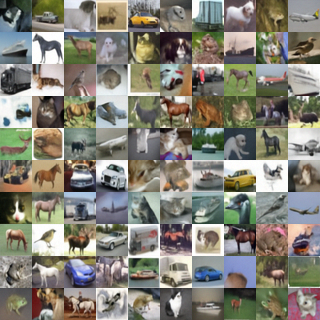}\tabularnewline
\multirow{1}{*}[0.25\textwidth]{SC} & \includegraphics[width=0.5\textwidth]{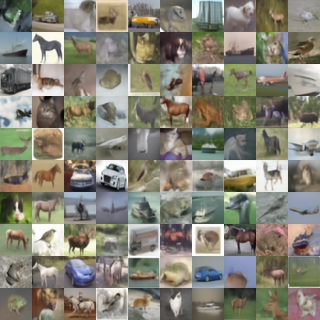} & \includegraphics[width=0.5\textwidth]{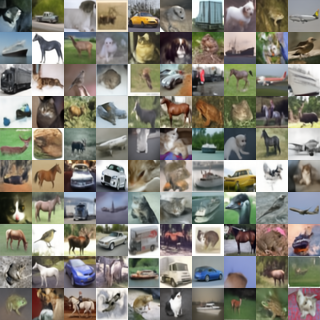} & \includegraphics[width=0.5\textwidth]{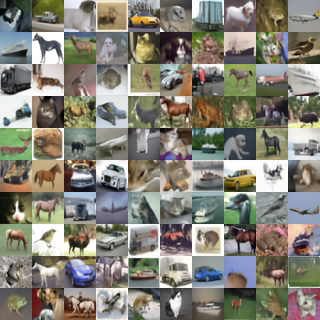}\tabularnewline
\end{tabular}}}
\par\end{centering}
\caption{Generated CIFAR10 samples from the \emph{continuous-time} VP-SDE model
officially pretrained by Song et al. \cite{song2020score} using different
numerical solvers (Euler, AB2, AB2AM2) along the original posterior
flow (top row) and the SC flow $d\bar{X}_{t}=\left(X_{1}-X_{0}\right)d\varphi_{t}$
(bottom row) with NFE=10. It is clear that when high-order solvers
(AB2, AB2AM2) are employed, the straightness and constant-speed advantages
of the SC flow over the posterior flow disappears, and generated samples
w.r.t. these flows have almost similar quality.\label{fig:Generated-CIFAR10-samples}}
\end{figure}

We have presented a training-free approach to convert the posterior
flows of ``linear'' stochastic processes defined by $X_{t}=a_{t}X_{0}+\sigma_{t}X_{1}$
into straight constant-speed flows that facilitate fast and more accurate
sampling using the Euler method. While our method can be applied to
any linear process and support high-order solvers, it has an inherent
limitation. Although first-order solvers like the Euler method benefit
significantly from straightness and constant-speed properties of a
flow, high-order solvers gain only minimal benefit. This is because
high-order solvers are designed to deal with the non-linearity and
curvature of a flow. Consequently, high-order solvers on the SC flow
tend to perform comparably to their counterparts on the posterior
flow, as demonstrated in Fig.~\ref{fig:Generated-CIFAR10-samples}.

\section*{}

\bibliographystyle{plain}

\addtocontents{toc}{\protect\setcounter{tocdepth}{2}}

\clearpage{}

\appendix
\tableofcontents{}

\section{Theoretical Results on Posterior Flows and Straight Constant-speed
Flows}

\subsection{Derivation of the posterior velocity $v^{*}\left(x_{t},t\right)$\label{subsec:Derivation-of-posterior-velocity}}

Let us recall that the posterior velocity $v^{*}$ is the solution
of minimizing the velocity matching loss $\Loss_{\text{VM}}$ (Eq.~\ref{eq:velocity_matching_loss_reframed}):
\begin{equation}
v^{*}:=\min_{v}\Loss_{\text{VM}}\left(v\right)
\end{equation}
where $\Loss_{\text{VM}}\left(v\right)$ can be written as follow:
\begin{align}
\Loss_{\text{VM}}\left(v\right)=\  & \Expect_{t}\left[\Expect_{x_{0},x_{1}}\Expect_{x_{t}|x_{0},x_{1}}\left[\left\Vert v_{\phi}\left(x_{0},x_{1},t\right)-v\left(x_{t},t\right)\right\Vert ^{2}\right]\right]\label{eq:VM_Loss_Decomp_1}\\
=\  & \Expect_{t}\Big[\Expect_{x_{t}}\Expect_{x_{0},x_{1}|x_{t}}\Big[\left\Vert v\left(x_{t},t\right)\right\Vert ^{2}-2v\left(x_{t},t\right)\cdot v_{\phi}\left(x_{0},x_{1},t\right)\nonumber \\
 & \ \ \ \ \ \ \ \ \ \ \ \ \ \ \ \ \ \ \ \ \ \ \ \ \ \ \ \ +\left\Vert v_{\phi}\left(x_{0},x_{1},t\right)\right\Vert ^{2}\Big]\Big]\label{eq:VM_Loss_Decomp_2}\\
=\  & \Expect_{t}\Expect_{x_{t}}\Big[\left\Vert v\left(x_{t},t\right)\right\Vert ^{2}-2v\left(x_{t},t\right)\cdot\Expect_{x_{0},x_{1}|x_{t}}\left[v_{\phi}\left(x_{0},x_{1},t\right)\right]\nonumber \\
 & \ \ \ \ \ \ \ \ \ \ \ \ +\Expect_{x_{0},x_{1}|x_{t}}\left[\left\Vert v_{\phi}\left(x_{0},x_{1},t\right)\right\Vert ^{2}\right]\Big]\label{eq:VM_Loss_Decomp_3}\\
=\  & \Expect_{t}\Expect_{x_{t}}\Big[\left\Vert v\left(x_{t},t\right)\right\Vert ^{2}-2v\left(x_{t},t\right)\cdot\Expect_{x_{0},x_{1}|x_{t}}\left[v_{\phi}\left(x_{0},x_{1},t\right)\right]\nonumber \\
 & \ \ \ \ \ \ \ \ \ \ \ \ +\left\Vert \Expect_{x_{0},x_{1}|x_{t}}\left[v_{\phi}\left(x_{0},x_{1},t\right)\right]\right\Vert ^{2}+\text{const}\Big]\label{eq:VM_Loss_Decomp_4}\\
=\  & \Expect_{t}\Expect_{x_{t}}\left[\left\Vert v\left(x_{t},t\right)-\Expect_{x_{0},x_{1}|x_{t}}\left[v_{\phi}\left(x_{0},x_{1},t\right)\right]\right\Vert ^{2}\right]+\text{const}\label{eq:VM_Loss_Decomp_5}
\end{align}

Eq.~\ref{eq:VM_Loss_Decomp_5} implies that:
\begin{align}
v^{*}\left(x_{t},t\right) & =\Expect_{x_{0},x_{1}|x_{t}}\left[v_{\phi}\left(x_{0},x_{1},t\right)\right]
\end{align}

If the stochastic process is $X_{t}=a_{t}X_{0}+\sigma_{t}X_{1}$,
we have:
\begin{align}
v_{\phi}\left(x_{0},x_{1},t\right) & =\dot{a}_{t}x_{0}+\dot{\sigma}_{t}x_{1}\\
v^{*}\left(x_{t},t\right) & =\Expect_{x_{0},x_{1}|x_{t}}\left[\dot{a}_{t}x_{0}+\text{\ensuremath{\dot{\sigma}_{t}}}x_{1}\right]
\end{align}

In this case, we can further decompose $v^{*}\left(x_{t},t\right)$
as follows:
\begin{align}
v^{*}\left(x_{t},t\right)=\  & \Expect_{x_{0},x_{1}|x_{t}}\left[\dot{a}_{t}x_{0}+\dot{\sigma}_{t}x_{1}\right]\\
=\  & \Expect_{x_{0}|x_{t}}\Expect_{x_{1}|x_{0},x_{t}}\left[\dot{a}_{t}x_{0}+\dot{\sigma}_{t}x_{1}\right]\\
=\  & \Expect_{x_{0}|x_{t}}\left[\dot{a}_{t}x_{0}\right]+\Expect_{x_{0}|x_{t}}\Expect_{x_{1}|x_{0},x_{t}}\left[\dot{\sigma}_{t}x_{1}\right]\label{eq:Posterior_velo_decomp_3}\\
=\  & \dot{a}_{t}\Expect_{x_{0}|x_{t}}\left[x_{0}\right]+\dot{\sigma}_{t}\Expect_{x_{0}|x_{t}}\Expect_{x_{1}|x_{0},x_{t}}\left[x_{1}\right]\label{eq:Posterior_velo_decomp_4}\\
=\  & \dot{a}_{t}\Expect_{x_{0}|x_{t}}\left[x_{0}\right]+\dot{\sigma}_{t}\Expect_{x_{0}|x_{t}}\left[\frac{x_{t}-a_{t}x_{0}}{\sigma_{t}}\right]\label{eq:Posterior_velo_decomp_5}\\
=\  & \frac{\dot{\sigma}_{t}}{\sigma_{t}}x_{t}+\left(\dot{a}_{t}-\frac{\dot{\sigma}_{t}a_{t}}{\sigma_{t}}\right)\Expect_{x_{0}|x_{t}}\left[x_{0}\right]\label{eq:Posterior_velo_decomp_xt_x0_6}\\
=\  & \frac{\dot{\sigma}_{t}}{\sigma_{t}}x_{t}+\left(\dot{a}_{t}-\frac{\dot{\sigma}_{t}a_{t}}{\sigma_{t}}\right)x_{0|t}\label{eq:Posterior_velo_decomp_xt_x0_7}\\
=\  & \frac{\dot{\sigma}_{t}}{\sigma_{t}}x_{t}+a_{t}\left(\frac{\dot{a}_{t}}{a_{t}}-\frac{\dot{\sigma}_{t}}{\sigma_{t}}\right)x_{0|t}\label{eq:Posterior_velo_decomp_xt_x0_8}
\end{align}
where $x_{0|t}:=\Expect_{x_{0}|x_{t}}\left[x_{0}\right]$. Eq.~\ref{eq:Posterior_velo_decomp_5}
is equivalent to Eq.~\ref{eq:Posterior_velo_decomp_4} because given
$x_{t}$ and $x_{0}$, $x_{1}$ is deterministically identified as
$\frac{x_{t}-a_{t}x_{0}}{\sigma_{t}}$ according to Eq.~\ref{eq:general_stochastic_process}.
$v^{*}\left(x_{t},t\right)$ can also be expressed as:
\begin{align}
v^{*}\left(x_{t},t\right) & =\frac{\dot{a}_{t}}{a_{t}}x_{t}+\left(\dot{\sigma}_{t}-\frac{\sigma_{t}\dot{a}_{t}}{a_{t}}\right)\Expect_{x_{1}|x_{t}}\left[x_{1}\right]\label{eq:Posterior_velo_decomp_xt_x1_6}\\
 & =\frac{\dot{a}_{t}}{a_{t}}x_{t}+\left(\dot{\sigma}_{t}-\frac{\sigma_{t}\dot{a}_{t}}{a_{t}}\right)x_{1|t}\label{eq:Posterior_velo_decomp_xt_x1_7}\\
 & =\frac{\dot{a}_{t}}{a_{t}}x_{t}+\sigma_{t}\left(\frac{\dot{\sigma}_{t}}{\sigma_{t}}-\frac{\dot{a}_{t}}{a_{t}}\right)x_{1|t}\label{eq:Posterior_velo_decomp_xt_x1_8}
\end{align}
due to the interchangeability between $x_{0}$ and $x_{1}$.

\subsection{Proof of the statistical equivalence between the stochastic process
$X_{t}=a_{t}X_{0}+\sigma_{t}X_{1}$ and its corresponding posterior
flow ODE\label{subsec:Statistical-equivalence-between-process-and-flow}}

For the stochastic process $X_{t}=a_{t}X_{0}+\sigma_{t}X_{1}$, $x_{t}$
is sampled from $p_{t}\left(x_{t}\right)=\Expect_{p_{0,1}\left(x_{0},x_{1}\right)}\left[p_{t}\left(x_{t}|x_{0},x_{1}\right)\right]$
where $p_{0,1}$ denotes the distribution of the coupling $\left(X_{0},X_{1}\right)$.
The infinitesimal change of $p_{t}\left(x_{t}\right)$ can be formulated
as follows: 

\begin{align}
\frac{\partial p_{t}\left(x_{t}\right)}{\partial t}=\  & \frac{\partial}{\partial t}\Expect_{p_{0,1}\left(x_{0},x_{1}\right)}\left[p_{t}\left(x_{t}|x_{0},x_{1}\right)\right]\\
=\  & \Expect_{p_{0,1}\left(x_{0},x_{1}\right)}\left[\frac{\partial p_{t}\left(x_{t}|x_{0},x_{1}\right)}{\partial t}\right]\label{eq:VFM_ContinuityEq_2}\\
=\  & -\Expect_{p_{0,1}\left(x_{0},x_{1}\right)}\left[\nabla\cdot\left(p_{t}\left(x_{t}|x_{0},x_{1}\right)v_{\phi}\left(x_{0},x_{1},t\right)\right)\right]\label{eq:VFM_ContinuityEq_3}\\
=\  & -\nabla\cdot\Expect_{p_{0,1}\left(x_{0},x_{1}\right)}\left[p_{t}\left(x_{t}|x_{0},x_{1}\right)v_{\phi}\left(x_{0},x_{1},t\right)\right]\label{eq:VFM_ContinuityEq_4}\\
=\  & -\nabla\cdot\int p_{0,1}\left(x_{0},x_{1}\right)p_{t}\left(x_{t}|x_{0},x_{1}\right)v_{\phi}\left(x_{0},x_{1},t\right)dx_{0}x_{1}\\
=\  & -\nabla\cdot\int p_{0,1}\left(x_{0},x_{1},x_{t}\right)v_{\phi}\left(x_{0},x_{1},t\right)dx_{0}x_{1}\\
=\  & -\nabla\cdot\int p_{t}\left(x_{t}\right)p_{t}\left(x_{0},x_{1}|x_{t}\right)v_{\phi}\left(x_{0},x_{1},t\right)dx_{0}x_{1}\\
=\  & -\nabla\cdot\left(p_{t}\left(x_{t}\right)\int p_{t}\left(x_{0},x_{1}|x_{t}\right)v_{\phi}\left(x_{0},x_{1},t\right)dx_{0}x_{1}\right)\\
=\  & -\nabla\cdot\left(p_{t}\left(x_{t}\right)\Expect_{p\left(x_{0},x_{1}|x_{t}\right)}\left[v_{\phi}\left(x_{0},x_{1},t\right)\right]\right)\\
=\  & -\nabla\cdot\left(p_{t}\left(x_{t}\right)v^{*}\left(x_{t},t\right)\right)\label{eq:VFM_ContinuityEq_10}
\end{align}
where Eq.~\ref{eq:VFM_ContinuityEq_3} is equivalent to Eq.~\ref{eq:VFM_ContinuityEq_2}
because $\frac{\partial p_{t}\left(x_{t}|x_{0},x_{1}\right)}{\partial t}=-\nabla\cdot\left(p_{t}\left(x_{t}|x_{0},x_{1}\right)v_{\phi}\left(x_{0},x_{1},t\right)\right)$
is the continuity equation for the ODE $\dot{x}_{t}=v_{\phi}\left(x_{t}\left(x_{0},x_{1},t\right),t\right)$
that connects two boundary points $x_{0}$, $x_{1}$; Eq.~\ref{eq:VFM_ContinuityEq_4}
is equivalent to Eq.~\ref{eq:VFM_ContinuityEq_3} due to the linearity
of the divergence operator; Eq.~\ref{eq:VFM_ContinuityEq_10} is
the continuity equation for the posterior flow ODE.

The above result implies that the changes in probability over time
for the stochastic process and its posterior flow ODE are identical.
Therefore, given the same initial distribution (either $p_{0}$ or
$p_{1}$), the stochastic process and its posterior flow ODE share
the same marginal distribution of samples. In other words, they are
statistically equivalent.

\subsection{Variational inference interpretation of the velocity matching loss\label{subsec:Variational-interpretation-of}}

\subsubsection{Variational inference interpretation for discrete-time diffusion
processes}

If we assume our stochastic process is a \emph{discrete-time diffusion}
process, we can base on the theoretical analyses in \cite{ho2020denoising,kingma2021variational,song2020denoising}
to show that the velocity matching loss in Eq.~\ref{eq:velocity_matching_loss_reframed}
serves as a variational upper bound on the negative log-likelihood.

Given a general diffusion process defined by $X_{t}=a_{t}X_{0}+\sigma_{t}X_{1}$
with $X_{1}\sim\Normal\left(0,\mathrm{I}\right)$, we have $p\left(x_{t}|x_{0}\right)=\Normal\left(x_{t}|a_{t}x_{0},\sigma_{t}^{2}\mathrm{I}\right)$
and $p\left(x_{s}|x_{0}\right)=\Normal\left(x_{s}|a_{s}x_{0},\sigma_{s}^{2}\mathrm{I}\right)$
for any two time steps $0\leq s<t\leq1$. Since $p\left(x_{t}|x_{0}\right)$
and $p\left(x_{s}|x_{0}\right)$ are both Gaussian distributions,
$p\left(x_{s}|x_{t},x_{0}\right)$ is also a Gaussian distribution
characterized by the following equation \cite{bishop2006pattern,song2020denoising}:
\begin{align}
p\left(x_{s}|x_{t},x_{0}\right)=\  & \Normal\left(x_{s}\bigg|\frac{\sqrt{\sigma_{s}^{2}-\gamma_{s}^{2}}}{\sigma_{t}}x_{t}+\left(a_{s}-a_{t}\frac{\sqrt{\sigma_{s}^{2}-\gamma_{s}^{2}}}{\sigma_{t}}\right)x_{0},\gamma_{s}^{2}\mathrm{I}\right)\label{eq:p(xs|xt,x0)_1}\\
=\  & \Normal\left(x_{s}\bigg|\frac{a_{s}}{a_{t}}\left(x_{t}-\sigma_{t}x_{1}\right)+\sqrt{\sigma_{s}^{2}-\gamma_{s}^{2}}x_{1},\gamma_{s}^{2}\mathrm{I}\right)\label{eq:p(xs|xt,x0)_2}
\end{align}
where $\gamma_{s}$ can take any value in the range $\left[0,\sigma_{s}\right]$.

According to \cite{ho2020denoising,kingma2021variational}, the variational
upper bound on the negative log-likelihood of a data sample $x_{0}$
is represented as follows:

\begin{align}
-\log p\left(x_{0}\right)\leq\  & \text{VUB}\left(x_{0}\right)\\
=\  & \Expect_{p\left(x_{t_{1}:1}|x_{0}\right)}\left[\log\frac{p\left(x_{t_{1}:1}|x_{0}\right)}{p_{\theta}\left(x_{0:1}\right)}\right]\\
=\  & \underbrace{\Expect_{p\left(x_{1}|x_{0}\right)}\left[D_{\text{KL}}\left(p\left(x_{1}|x_{0}\right)\|p\left(x_{1}\right)\right)\right]}_{\text{Prior loss}}\nonumber \\
 & +\underbrace{\Expect_{p\left(x_{t_{1}}|x_{0}\right)}\left[-\log p_{\theta}\left(x_{0}|x_{t_{1}}\right)\right]}_{\text{Reconstruction loss}}\nonumber \\
 & +\underbrace{\sum_{i=2}^{N}\Expect_{p\left(x_{t}|x_{0}\right)}\left[D_{\text{KL}}\left(p\left(x_{t_{i-1}}|x_{t_{i}},x_{0}\right)\|p_{\theta}\left(x_{t_{i-1}}|x_{t_{i}}\right)\right)\right]}_{\text{Diffusion loss}}\label{eq:VUB_discrete_diffusion_3}
\end{align}
where $0=t_{0}<t_{1}<...<t_{N}=1$.

Here, we focus on the KL divergence term in the diffusion loss in
Eq.~\ref{eq:VUB_discrete_diffusion_3}. If we model $p_{\theta}\left(x_{t_{i-1}}|x_{t_{i}}\right)$
as $p\left(x_{t_{i-1}}|x_{t_{i}},x_{0,\theta}\left(x_{t_{i}},t_{i}\right)\right)$
such that its formula is identical to the formula of $p\left(x_{t_{i-1}}|x_{t_{i}},x_{0}\right)$
in Eq.~\ref{eq:p(xs|xt,x0)_1} but with $x_{0}$ replaced by $x_{0,\theta}\left(x_{t_{i}},t_{i}\right)$,
we can express the KL divergence in Eq.~\ref{eq:VUB_discrete_diffusion_3}
analytically as follows:

\begin{align}
 & D_{\text{KL}}\left(p\left(x_{t_{i-1}}|x_{t_{i}},x_{0}\right)\|p_{\theta}\left(x_{t_{i-1}}|x_{t_{i}}\right)\right)\nonumber \\
=\  & D_{\text{KL}}\left(p\left(x_{t_{i-1}}|x_{t_{i}},x_{0}\right)\|p\left(x_{t_{i-1}}|x_{t_{i}},x_{0,\theta}\left(x_{t_{i}},t_{i}\right)\right)\right)\\
=\  & \frac{\left(a_{s}-a_{t}\frac{\sqrt{\sigma_{s}^{2}-\gamma_{s}^{2}}}{\sigma_{t}}\right)^{2}}{2\gamma_{s}^{2}}\left\Vert x_{0,\theta}\left(x_{t_{i}},t_{i}\right)-x_{0}\right\Vert _{2}^{2}+\text{const}\label{eq:KL_by_x0_2}
\end{align}
Following \cite{kingma2021variational}, we set $\gamma_{s}^{2}=\sigma_{s}^{2}\left(1-\frac{a_{t}^{2}}{\sigma_{t}^{2}}\frac{\sigma_{s}^{2}}{a_{s}^{2}}\right)=\sigma_{s}^{2}\left(1-\frac{\text{SNR}(t)}{\text{SNR}(s)}\right)$.
The coefficient in Eq.~\ref{eq:KL_by_x0_2} simplifies to $\frac{1}{2}\left(\frac{a_{s}^{2}}{\sigma_{s}^{2}}-\frac{a_{t}^{2}}{\sigma_{t}^{2}}\right)=\frac{1}{2}\left(\text{SNR}\left(s\right)-\text{SNR}\left(t\right)\right)$
and Eq.~\ref{eq:KL_by_x0_2} becomes:
\begin{align}
 & D_{\text{KL}}\left(p\left(x_{t_{i-1}}|x_{t_{i}},x_{0}\right)\|p_{\theta}\left(x_{t_{i-1}}|x_{t_{i}}\right)\right)\nonumber \\
=\  & \frac{1}{2}\left(\text{SNR}\left(s\right)-\text{SNR}\left(t\right)\right)\left\Vert x_{0,\theta}\left(x_{t_{i}},t_{i}\right)-x_{0}\right\Vert _{2}^{2}+\text{const}\label{eq:KL_by_x0_3}
\end{align}

When the time is continuous, Kingma et al. \cite{kingma2021variational}
have shown that Eq.~\ref{eq:KL_by_x0_3} can be written as:

\begin{equation}
\frac{1}{2}\frac{d\text{SNR}\left(t\right)}{dt}\left\Vert x_{0,\theta}\left(x_{t_{i}},t_{i}\right)-x_{0}\right\Vert _{2}^{2}+\text{const}\label{eq:KL_by_x0_continuous_1}
\end{equation}
where $\frac{d\text{SNR}\left(t\right)}{dt}=\frac{2a_{t}\left(\dot{a}_{t}\sigma_{t}-a_{t}\dot{\sigma}_{t}\right)}{\sigma_{t}^{3}}$.

It is worth noting that when the time is continuous, the KL divergence
in Eq.~\ref{eq:KL_by_x0_3} degenerates and Eq.~\ref{eq:KL_by_x0_continuous_1}
may no longer represent a KL divergence. Nevertheless, for a very
tiny \emph{discrete} time step $dt$, Eq.~\ref{eq:KL_by_x0_continuous_1}
can still serve as an estimate for the KL divergence. Therefore, in
practice, Eq.~\ref{eq:KL_by_x0_continuous_1} remains viable for
deriving the variational bound for the continuous-time case.

Since in the continuous-time case, $x_{0,\theta}\left(x_{t},t\right)$
and $x_{0}$ can be transformed into $v_{\theta}\left(x_{t},t\right)$
and $v_{\phi}\left(x_{t},t\right)$ respectively using Eq\@.~\ref{eq:Posterior_velo_decomp_xt_x0_8},
we can reformulate Eq.~\ref{eq:KL_by_x0_continuous_1} to incorporate
$v_{\theta}\left(x_{t},t\right)$ and $v_{\phi}\left(x_{t},t\right)$
as follows:
\begin{align}
 & \frac{1}{2}\frac{d\text{SNR}\left(t\right)}{dt}\left\Vert x_{0,\theta}\left(x_{t_{i}},t_{i}\right)-x_{0}\right\Vert _{2}^{2}+\text{const}\nonumber \\
=\  & \frac{1}{2}\frac{\frac{d\text{SNR}\left(t\right)}{dt}}{a_{t}^{2}\left(\frac{\dot{a}_{t}}{a_{t}}-\frac{\dot{\sigma}_{t}}{\sigma_{t}}\right)^{2}}\left\Vert v_{\theta}\left(x_{t_{i}},t_{i}\right)-v_{\phi}\left(x_{t_{i}},t_{i}\right)\right\Vert _{2}^{2}+\text{const}\\
=\  & \frac{1}{2}\frac{\frac{2a_{t}\left(\dot{a}_{t}\sigma_{t}-a_{t}\dot{\sigma}_{t}\right)}{\sigma_{t}^{3}}}{\frac{\left(\dot{a}_{t}\sigma_{t}-a_{t}\dot{\sigma}_{t}\right)^{2}}{\sigma_{t}^{2}}}\left\Vert v_{\theta}\left(x_{t_{i}},t_{i}\right)-v_{\phi}\left(x_{t_{i}},t_{i}\right)\right\Vert _{2}^{2}+\text{const}\\
=\  & \frac{a_{t}}{\sigma_{t}\left(\dot{a}_{t}\sigma_{t}-a_{t}\dot{\sigma}_{t}\right)}\left\Vert v_{\theta}\left(x_{t_{i}},t_{i}\right)-v_{\phi}\left(x_{t_{i}},t_{i}\right)\right\Vert _{2}^{2}+\text{const}\label{eq:KL_by_v_continuous_3}
\end{align}
Eq.~\ref{eq:KL_by_v_continuous_3} implies that the weighted velocity
matching loss can be considered as a variational upper bound on the
negative log-likelihood.

Interestingly, if we convert $v_{\theta}$ and $v_{\phi}$ into $x_{1,\theta}$
and $x_{1}$, respectively using Eq.~\ref{eq:nonparam_velo_xt_x0},
we can rewrite Eq.~\ref{eq:KL_by_v_continuous_3} as:
\begin{align}
 & \frac{1}{2}\frac{d\text{SNR}\left(t\right)}{dt}\left\Vert x_{0,\theta}\left(x_{t_{i}},t_{i}\right)-x_{0}\right\Vert _{2}^{2}+\text{const}\nonumber \\
=\  & \frac{1}{2}\frac{a_{t}\left(\frac{\dot{\sigma}_{t}a_{t}-\dot{a}_{t}\sigma_{t}}{a_{t}}\right)^{2}}{\sigma_{t}\left(\dot{a}_{t}\sigma_{t}-a_{t}\dot{\sigma}_{t}\right)}\left\Vert x_{1,\theta}\left(x_{t_{i}},t_{i}\right)-x_{1}\right\Vert _{2}^{2}+\text{const}\\
=\  & \frac{1}{2}\frac{\dot{a}_{t}\sigma_{t}-a_{t}\dot{\sigma}_{t}}{a_{t}\sigma_{t}}\left\Vert x_{1,\theta}\left(x_{t_{i}},t_{i}\right)-x_{1}\right\Vert _{2}^{2}+\text{const}\\
=\  & \frac{1}{2}\frac{a_{t}}{\sigma_{t}}\left(-\dot{\varphi}_{t}\right)\left\Vert x_{1,\theta}\left(x_{t_{i}},t_{i}\right)-x_{1}\right\Vert _{2}^{2}+\text{const}\\
=\  & -\frac{1}{2}\frac{\dot{\varphi}_{t}}{\varphi_{t}}\left\Vert x_{1,\theta}\left(x_{t_{i}},t_{i}\right)-x_{1}\right\Vert _{2}^{2}+\text{const}\\
=\  & \frac{1}{2}\frac{d\left(-\log\varphi_{t}\right)}{dt}\left\Vert x_{1,\theta}\left(x_{t_{i}},t_{i}\right)-x_{1}\right\Vert _{2}^{2}+\text{const}\label{eq:KL_by_x1_continuous_5}
\end{align}
where $\varphi_{t}=\frac{\sigma_{t}}{a_{t}}$ and $\dot{\varphi}_{t}=\frac{\dot{\sigma}_{t}a_{t}-\dot{a}_{t}\sigma_{t}}{a_{t}^{2}}$.
It is straightforward to show that:
\begin{align}
\varphi_{t} & =\sqrt{\frac{1}{\text{SNR}\left(t\right)}}\\
-\log\varphi_{t} & =\log\frac{a_{t}}{\sigma_{t}}=\frac{1}{2}\log\text{SNR}\left(t\right)
\end{align}

This implies that Eq.~\ref{eq:KL_by_x1_continuous_5} is equivalent
to:
\begin{align}
\frac{1}{4}\frac{d\log\text{SNR}\left(t\right)}{dt}\left\Vert x_{1,\theta}\left(x_{t_{i}},t_{i}\right)-x_{1}\right\Vert _{2}^{2}+\text{const}
\end{align}

Similar to $\text{SNR}\left(t\right)$, $-\log\varphi_{t}$ is a monotonic
function and can be used in the change-of-variables formula, leading
to the following expression of the variational bound:
\begin{align}
\text{VUB}\left(x_{0}\right)=\  & \int_{0}^{1}\Expect_{x_{1}}\left[\frac{1}{2}\frac{d\text{SNR}\left(t\right)}{dt}\left\Vert x_{0,\theta}\left(x_{t_{i}},t_{i}\right)-x_{0}\right\Vert _{2}^{2}\right]dt+\text{const}\\
=\  & \int_{0}^{1}\Expect_{x_{1}}\left[\frac{1}{2}\frac{d\left(-\log\varphi_{t}\right)}{dt}\left\Vert x_{1,\theta}\left(x_{t_{i}},t_{i}\right)-x_{1}\right\Vert _{2}^{2}\right]dt+\text{const}\\
=\  & \frac{1}{2}\Expect_{x_{1}}\left[\int_{0}^{1}\frac{d\psi_{t}}{dt}\left\Vert x_{1,\theta}\left(x_{t_{i}},t_{i}\right)-x_{1}\right\Vert _{2}^{2}dt\right]+\text{const}\\
=\  & \frac{1}{2}\Expect_{x_{1}}\left[\int_{\psi_{0}}^{\psi_{1}}\left\Vert \hat{x}_{1,\theta}\left(x_{\psi_{t}},\psi_{t}\right)-x_{1}\right\Vert _{2}^{2}d\psi_{t}\right]+\text{const}
\end{align}
where $\psi_{t}\equiv-\log\varphi_{t}=\frac{1}{2}\log\text{SNR}\left(t\right)$,
$\psi_{1}$, $\psi_{0}$ are the two values of $\psi_{t}$ at $t$
= 1 and 0, respectively.

\subsubsection{Variational proxy interpretation for general continuous-time stochastic
processes}

When the stochastic process is continuous-time and non-diffusion,
it is not straightforward to apply the above technique because $p\left(x_{s}|x_{t},x_{0}\right)$
is not a Gaussian distribution and the KL divergence is not well defined.
Informally, minimizing $\Loss_{\text{VM}}$ enhances the alignment
between the parameterized flow represented by $v_{\theta}$ and the
posterior flow characterized by $v^{*}$, thereby narrowing the gap
between $p_{0,\theta}$ and $p_{0}$. In this sense, $\Loss_{\text{VM}}$
can be viewed as a variational proxy for the negative log-likelihood.

\subsection{Explicit formulas of $v_{\phi}\left(x_{0},x_{1},t\right)$ and $v^{*}\left(x_{t},t\right)$
for diffusion models and Rectified Flow\label{subsec:Explicit-formulas-of}}

Below, we derive the explicit formulas of the target and posterior
velocities for well-known diffusion models \cite{kingma2021variational,song2020score}
and Rectified Flow \cite{liu2022flow} with the summary provided in
Table.~\ref{tab:velocity-formulas-summary}. In addition, we also
present the explicit formulas of the PF ODEs and SDEs of these models
in Table.~\ref{tab:sde_formulas_summary}. For Rectified Flow, we
assume that the latent distribution $p_{1}$ is $\Normal\left(0,\mathrm{I}\right)$
so that we can treat its corresponding stochastic process as an SDE.

\begin{table*}
\begin{centering}
{\resizebox{\textwidth}{!}{
\begin{tabular}{ccccc}
\hline 
Model & $a_{\ensuremath{t}}$ & $\sigma_{t}$ & PF ODE & Forward SDE\tabularnewline
\hline 
\hline 
VP SDE & $\sqrt{\overline{\alpha}_{t}}$ & $\sqrt{1-\overline{\alpha}_{t}}$ & $dx_{t}=\frac{-\beta_{t}}{2}\left(x_{t}+\nabla\log p_{t}\left(x_{t}\right)\right)dt$ & $dx_{t}=\frac{-\beta_{t}}{2}x_{t}\ dt+\sqrt{\beta_{t}}dw_{t}$\tabularnewline
sub-VP SDE & $\sqrt{\overline{\alpha}_{t}}$ & $1-\overline{\alpha}_{t}$ & $dx_{t}=\frac{-\beta_{t}}{2}\left(x_{t}+\left(1-\overline{\alpha}_{t}^{2}\right)\nabla\log p_{t}\left(x_{t}\right)\right)dt$ & $dx_{t}=\frac{-\beta_{t}}{2}x_{t}\ dt+\sqrt{\beta_{t}\left(1-\overline{\alpha}_{t}^{2}\right)}dw_{t}$\tabularnewline
VE SDE & $1$ & $\sigma_{t}$ & $dx_{t}=-\dot{\sigma}_{t}\sigma_{t}\nabla\log p_{t}\left(x_{t}\right)dt$ & $dx_{t}=\sqrt{2\dot{\sigma}_{t}\sigma_{t}}dw_{t}$\tabularnewline
Rect. Flow$^{*}$ & $1-t$ & $t$ & $dx_{t}=\frac{-1}{1-t}\left(x_{t}+t\nabla\log p_{t}\left(x_{t}\right)\right)dt$ & $dx_{t}=\frac{-1}{1-t}x_{t}dt+\sqrt{\frac{2t}{1-t}}dw_{t}$\tabularnewline
General & $a_{t}$ & $\sigma_{t}$ & $dx_{t}=\left(\frac{\dot{a}_{t}}{a_{t}}x_{t}-g_{t}\sigma_{t}\nabla\log p_{t}\left(x_{t}\right)\right)dt$ & $dx_{t}=\frac{\dot{a}_{t}}{a_{t}}x_{t}dt+\sqrt{2g_{t}\sigma_{t}}dw_{t}$\tabularnewline
\hline 
\end{tabular}}}
\par\end{centering}
\caption{The coefficients $a_{t}$, $\sigma_{t}$, PF ODE formulas, and forward
SDE formulas corresponding to well-known diffusion processes \cite{kingma2021variational,song2020score}
and Rectified Flow (Rect. Flow) \cite{liu2022flow}. In the case of
(sub-)VP SDE, $\beta_{t}=t\left(\protect\betamax-\protect\betamin\right)+\protect\betamin$
and $\overline{\alpha}_{t}=e^{-\frac{1}{2}t\left(\beta_{t}+\protect\betamin\right)}$.
For Rectified Flow, the latent distribution $p_{1}$ is assumed to
be $\protect\Normal\left(0,\mathrm{I}\right)$. For general diffusion
processes (General), $g_{t}=\sigma_{t}\left(\frac{\dot{\sigma}_{t}}{\sigma_{t}}-\frac{\dot{a}_{t}}{a_{t}}\right)$
similar to the coefficient of $x_{1|t}$ in Eq.~\ref{eq:Posterior_velo_decomp_xt_x1_8}.\label{tab:sde_formulas_summary}}
\end{table*}

\subsubsection{VP SDE\label{subsec:VP-SDE}}

The (forward) VP SDE \cite{song2020score} is an Ornstein-Uhlenbeck
process characterized by the equation:
\begin{equation}
dx_{t}=-\frac{\beta_{t}}{2}x_{t}\ dt+\sqrt{\beta_{t}}dw_{t}\label{eq:VP_SDE}
\end{equation}
where $w_{t}$ is the Wiener process at time $t,$ $\beta_{t}=t\left(\betamax-\betamin\right)+\betamin$.
Its corresponding Fokker-Planck equation (aka Kolmogorov forward equation)
is:
\begin{equation}
\frac{\partial p_{t}}{\partial t}=\frac{\beta_{t}}{2}\left(\nabla\cdot\left(p_{t}x_{t}\right)+\Delta p_{t}\right)
\end{equation}
where $\nabla\cdot$ and $\Delta$ denote the divergence and Laplace
operators, respectively. When the initial distribution is $p_{0}=\delta_{x_{0}}$
where $\delta_{x_{0}}$denotes the Dirac delta distribution which
has value zero everywhere except at $x_{0}$, this Fokker-Planck equation
has the following solution:
\begin{equation}
p\left(x_{t}|x_{0}\right)=\Normal\left(x_{0}e^{-\frac{1}{2}\int_{0}^{t}\beta_{\tau}\ d\tau},\left(1-e^{-\int_{0}^{t}\beta_{\tau}\ d\tau}\right)\mathrm{I}\right)
\end{equation}
The closed-form expression of $\int_{0}^{t}\beta_{\tau}\ d\tau$ exists
and is given by:
\begin{align}
\int_{0}^{t}\beta_{\tau}\ d\tau & =\int_{0}^{t}\left(\tau\left(\betamax-\betamin\right)+\betamin\right)d\tau\\
 & =\left(\frac{\tau^{2}}{2}\left(\betamax-\betamin\right)+\tau\betamin\right)\bigg|_{\tau=0}^{\tau=t}\\
 & =\frac{t^{2}}{2}\left(\betamax-\betamin\right)+t\betamin\\
 & =\frac{t}{2}\left(\beta_{t}+\betamin\right)
\end{align}
Thus, $p\left(x_{t}|x_{0}\right)$ can be expressed as:
\begin{align}
p\left(x_{t}|x_{0}\right) & =\Normal\left(x_{0}e^{-\frac{1}{2}\frac{t}{2}\left(\beta_{t}+\betamin\right)},\left(1-e^{-\frac{t}{2}\left(\beta_{t}+\betamin\right)}\right)\mathrm{I}\right)\label{eq:p(xt|x0)-VPSDE-our-1}\\
 & =\Normal\left(\sqrt{\overline{\alpha}_{t}}x_{0},\left(1-\overline{\alpha}_{t}\right)\mathrm{I}\right)\label{eq:p(xt|x0)-VPSDE-our-2}
\end{align}
where $\overline{\alpha}_{t}:=e^{-\frac{t}{2}\left(\beta_{t}+\betamin\right)}$.
Eq.~\ref{eq:p(xt|x0)-VPSDE-our-1} is the same as Eq. 33 in \cite{song2020score}
while Eq.~\ref{eq:p(xt|x0)-VPSDE-our-2} is similar to Eq.~4 in
\cite{ho2020denoising}. According to \cite{song2020score}, the probability
flow ODE of the VP SDE is:
\begin{equation}
dx_{t}=-\frac{\beta_{t}}{2}\left(x_{t}+\nabla\log p\left(x_{t}\right)\right)dt\label{eq:PFODE_VPSDE}
\end{equation}
Eq.~\ref{eq:p(xt|x0)-VPSDE-our-2} implies that we can represent
the VP SDE as:
\begin{equation}
X_{t}=\sqrt{\overline{\alpha}_{t}}X_{0}+\sqrt{1-\overline{\alpha}_{t}}X_{1}\label{eq:VPSDE_stochastic_process}
\end{equation}
where $X_{1}$ is the standard Gaussian. Eq.~\ref{eq:VPSDE_stochastic_process}
is an instance of the general class of stochastic processes described
by $X_{t}=a_{t}X_{0}+\sigma_{t}X_{1}$ (Eq.~\ref{eq:GRF_DiffusionProcess_v2})
with $a_{t}=\sqrt{\overline{\alpha}_{t}}$ and $\sigma_{t}=\sqrt{1-\overline{\alpha}_{t}}$. 

Taking the derivatives of $\overline{\alpha}_{t}$ w.r.t. $t$, we
have:
\begin{align}
\frac{d\overline{\alpha}_{t}}{dt} & =\frac{d}{dt}\left(e^{-\frac{1}{2}t\left(\beta_{t}+\betamin\right)}\right)\\
 & =e^{-\frac{1}{2}t\left(\beta_{t}+\betamin\right)}\frac{d}{dt}\left(-\frac{1}{2}t\left(\beta_{t}+\betamin\right)\right)\\
 & =\overline{\alpha}_{t}\left(-\frac{1}{2}\left(\left(\beta_{t}+\betamin\right)+t\frac{d\beta_{t}}{dt}\right)\right)\\
 & =\frac{-\overline{\alpha}_{t}}{2}\left(\beta_{t}+\betamin+t\left(\betamax-\betamin\right)\right)\\
 & =\frac{-\overline{\alpha}_{t}}{2}\left(2\beta_{t}\right)\\
 & =-\overline{\alpha}_{t}\beta_{t}\label{eq:alpha_bar_t_dot}
\end{align}
$\dot{a}_{t}$ and $\dot{\sigma}_{t}$ can be derived from $\frac{d\overline{\alpha}_{t}}{dt}$
as follows:
\begin{equation}
\dot{a}_{t}=\frac{d\left(\sqrt{\overline{\alpha}_{t}}\right)}{dt}=\frac{1}{2\sqrt{\overline{\alpha}_{t}}}\frac{d\overline{\alpha}_{t}}{dt}=\frac{-\beta_{t}\sqrt{\overline{\alpha}_{t}}}{2}
\end{equation}
\begin{equation}
\dot{\sigma}_{t}=\frac{d\left(\sqrt{1-\overline{\alpha}_{t}}\right)}{dt}=\frac{1}{2\sqrt{1-\overline{\alpha}_{t}}}\frac{-d\overline{\alpha}_{t}}{dt}=\frac{\beta_{t}\overline{\alpha}_{t}}{2\sqrt{1-\overline{\alpha}_{t}}}
\end{equation}
Thus, we can express $v_{\phi}\left(x_{0},x_{1},t\right)$ as:
\begin{align}
v_{\phi}\left(x_{0},x_{1},t\right) & =\dot{a}_{t}x_{0}+\dot{\sigma}_{t}x_{1}\\
 & =\frac{-\beta_{t}\sqrt{\overline{\alpha}_{t}}}{2}x_{0}+\frac{\beta_{t}\overline{\alpha}_{t}}{2\sqrt{1-\overline{\alpha}_{t}}}x_{1}\\
 & =\frac{\beta_{t}\overline{\alpha}_{t}}{2}\left(\frac{x_{1}}{\sqrt{1-\overline{\alpha}_{t}}}-\frac{x_{0}}{\sqrt{\overline{\alpha}_{t}}}\right)
\end{align}
$v^{*}\left(x_{t},t\right)$ can be represented based on Eq.~\ref{eq:Posterior_velo_decomp_xt_x1_6}
as:
\begin{align}
v^{*}\left(x_{t},t\right)=\  & \frac{\dot{a}_{t}}{a_{t}}x_{t}+\left(\dot{\sigma}_{t}-\frac{\dot{a}_{t}\sigma_{t}}{a_{t}}\right)\Expect_{x_{1}|x_{t}}\left[x_{1}\right]\\
=\  & \frac{-\beta_{t}\sqrt{\overline{\alpha}_{t}}}{2}\frac{1}{\sqrt{\overline{\alpha}_{t}}}x_{t}+\left(\frac{\beta_{t}\overline{\alpha}_{t}}{2\sqrt{1-\overline{\alpha}_{t}}}+\frac{\beta_{t}\sqrt{\overline{\alpha}_{t}}}{2}\frac{\sqrt{1-\overline{\alpha}_{t}}}{\sqrt{\overline{\alpha}_{t}}}\right)x_{1|t}\\
=\  & -\frac{\beta_{t}}{2}x_{t}+\left(\frac{\beta_{t}\overline{\alpha}_{t}}{2\sqrt{1-\overline{\alpha}_{t}}}+\frac{\beta_{t}\left(1-\overline{\alpha}_{t}\right)}{2\sqrt{1-\overline{\alpha}_{t}}}\right)x_{1|t}\\
=\  & -\frac{\beta_{t}}{2}x_{t}+\frac{\beta_{t}}{2\sqrt{1-\overline{\alpha}_{t}}}x_{1|t}\\
=\  & -\frac{\beta_{t}}{2}\left(x_{t}-\frac{x_{1|t}}{\sqrt{1-\overline{\alpha}_{t}}}\right)\label{eq:PosteriorVelo_x1t_VPSDE_5}
\end{align}
It is not straightforward to obtain an analytical formula of $x_{1|t}$.
However, if we assume that Eq.~\ref{eq:PosteriorVelo_x1t_VPSDE_5}
is correct and compare it with Eq.~\ref{eq:PFODE_VPSDE}, we can
guess $\nabla\log p\left(x_{t}\right)=\frac{-x_{1|t}}{\sqrt{1-\overline{\alpha}_{t}}}$.
In DDPM \cite{ho2020denoising}, $x_{1|t}$ is modeled by a noise
network $\epsilon_{\theta}\left(x_{t},t\right)$, which suggests that
we can approximate the score function in DDPM as $\frac{-\epsilon_{\theta}\left(x_{t},t\right)}{\sqrt{1-\overline{\alpha}_{t}}}$.

On the other hand, if we express $v^{*}\left(x_{t},t\right)$ according
to the formula in Eq.~\ref{eq:Posterior_velo_decomp_xt_x0_7}, we
have:
\begin{align}
v^{*}\left(x_{t},t\right)=\  & \frac{\dot{\sigma}_{t}}{\sigma_{t}}x_{t}+\left(\dot{a}_{t}-\frac{\dot{\sigma}_{t}a_{t}}{\sigma_{t}}\right)\Expect_{x_{0}|x_{t}}\left[x_{0}\right]\\
=\  & \frac{\beta_{t}\overline{\alpha}_{t}}{2\sqrt{1-\overline{\alpha}_{t}}}\frac{1}{\sqrt{1-\overline{\alpha}_{t}}}x_{t}\\
 & +\left(\frac{-\beta_{t}\sqrt{\overline{\alpha}_{t}}}{2}-\frac{\beta_{t}\overline{\alpha}_{t}}{2\sqrt{1-\overline{\alpha}_{t}}}\frac{\sqrt{\overline{\alpha}_{t}}}{\sqrt{1-\overline{\alpha}_{t}}}\right)x_{0|t}\\
=\  & \frac{\beta_{t}\overline{\alpha}_{t}}{2\left(1-\overline{\alpha}_{t}\right)}x_{t}-\frac{\beta_{t}\sqrt{\overline{\alpha}_{t}}}{2}\left(1+\frac{\overline{\alpha}_{t}}{1-\overline{\alpha}_{t}}\right)x_{0|t}\\
=\  & \frac{\beta_{t}\overline{\alpha}_{t}}{2\left(1-\overline{\alpha}_{t}\right)}x_{t}-\frac{\beta_{t}\sqrt{\overline{\alpha}_{t}}}{2\left(1-\overline{\alpha}_{t}\right)}x_{0|t}\\
=\  & \frac{\beta_{t}\overline{\alpha}_{t}}{2\left(1-\overline{\alpha}_{t}\right)}\left(x_{t}-\frac{x_{0|t}}{\sqrt{\overline{\alpha}_{t}}}\right)\label{eq:PosteriorVelo_DDPM_x0_xt_last}
\end{align}
Since $x_{t}$ is a Gaussian sample given $x_{0}$, $\Expect_{x_{0}|x_{t}}\left[x_{0}\right]$
can be interpreted as the posterior estimation of the ``mean'' $x_{0}$
given its sample $x_{t}$. The analytical expression of $\Expect_{x_{0}|x_{t}}\left[x_{0}\right]$
exists an is given by the so-called \emph{``Tweedie's formula''}
\cite{efron2011tweedie} (Appdx.~\ref{subsec:Tweedie's-formula-for})
as follows:
\begin{equation}
x_{0|t}:=\Expect_{x_{0}|x_{t}}\left[x_{0}\right]=\frac{x_{t}+\left(1-\overline{\alpha}_{t}\right)\nabla\log p_{t}\left(x_{t}\right)}{\sqrt{\overline{\alpha}_{t}}}
\end{equation}
Applying this formula of $x_{0|t}$ to Eq.~\ref{eq:PosteriorVelo_DDPM_x0_xt_last},
we have:
\begin{align}
v^{*}\left(x_{t},t\right) & =\frac{\beta_{t}\overline{\alpha}_{t}}{2\left(1-\overline{\alpha}_{t}\right)}\left(x_{t}-\frac{x_{t}+\left(1-\overline{\alpha}_{t}\right)\nabla\log p_{t}\left(x_{t}\right)}{\overline{\alpha}_{t}}\right)\\
 & =\frac{\beta_{t}\overline{\alpha}_{t}}{2\left(1-\overline{\alpha}_{t}\right)}\left(-\frac{1-\overline{\alpha}_{t}}{\overline{\alpha}_{t}}x_{t}-\frac{1-\overline{\alpha}_{t}}{\overline{\alpha}_{t}}\nabla\log p_{t}\left(x_{t}\right)\right)\\
 & =-\frac{\beta_{t}}{2}\left(x_{t}+\nabla\log p_{t}\left(x_{t}\right)\right)
\end{align}
which exactly matches the formula of the PF ODE velocity in Eq.~\ref{eq:PFODE_VPSDE}.

\subsubsection{VE SDE}

The (forward) VE SDE \cite{song2020score} is represented as follows:

\begin{equation}
dx_{t}=\sqrt{2\dot{\sigma}_{t}\sigma_{t}}dw
\end{equation}
Its corresponding probability flow ODE is characterized by the equation:

\begin{align}
dx_{t} & =-\dot{\sigma}_{t}\sigma_{t}\nabla\log p\left(x_{t}\right)dt\label{eq:PFODE_DDIM_1}\\
 & =-\sigma_{t}\nabla\log p\left(x_{t}\right)d\sigma_{t}\label{eq:PFODE_DDIM_2}\\
 & \approx\epsilon_{\theta}\left(x_{t},t\right)d\sigma_{t}\label{eq:PFODE_DDIM_3}
\end{align}
Solving the Fokker-Planck equation of the VE SDE, we get $p\left(x_{t}|x_{0}\right)=\Normal\left(x_{0},\sigma_{t}^{2}\mathrm{I}\right)$.
This means the VE SDE can be written in the following form:
\begin{equation}
X_{t}=X_{0}+\sigma_{t}X_{1}\label{eq:VESDE_stochastic_process}
\end{equation}
where the distribution of $X_{1}$ is $\Normal\left(0,\mathrm{I}\right)$.
Eq.~\ref{eq:VESDE_stochastic_process} is a variant of Eq.~\ref{eq:GRF_DiffusionProcess_v2}
with $a_{t}=1$ $\forall t\in\left[0,1\right]$. 

The formula of the target velocity is: 
\begin{equation}
v_{\phi}\left(x_{0},x_{1},t\right)=\dot{\sigma}_{t}x_{1}
\end{equation}

If we compute $v^{*}\left(x_{t},t\right)$ based on Eq.~\ref{eq:Posterior_velo_decomp_xt_x1_6},
we have:
\begin{align}
v^{*}\left(x_{t},t\right) & =\frac{\dot{a}_{t}}{a_{t}}x_{t}+\left(\dot{\sigma}_{t}-\frac{\dot{a}_{t}\sigma_{t}}{a_{t}}\right)\Expect_{x_{1}|x_{t}}\left[x_{1}\right]\\
 & =0+\left(\text{\ensuremath{\dot{\sigma}_{t}}}-0\right)x_{1|t}\\
 & =\dot{\sigma}_{t}x_{1|t}
\end{align}
On the other hand, if Eq.~\ref{eq:Posterior_velo_decomp_xt_x0_7}
is utilized, $v^{*}\left(x_{t},t\right)$ can be represented as follows:
\begin{align}
v^{*}\left(x_{t},t\right) & =\frac{\dot{\sigma}_{t}}{\sigma_{t}}x_{t}+\left(\dot{a}_{t}-\frac{\dot{\sigma}_{t}a_{t}}{\sigma_{t}}\right)\Expect_{x_{0}|x_{t}}\left[x_{0}\right]\\
 & =\frac{\dot{\sigma}_{t}}{\sigma_{t}}x_{t}+\left(0-\frac{\dot{\sigma}_{t}}{\sigma_{t}}\right)x_{0|t}\\
 & =\frac{\dot{\sigma}_{t}}{\sigma_{t}}\left(x_{t}-x_{0|t}\right)\label{eq:PosteriorVelo_xt_x0_3}
\end{align}
According to the Tweedie's formula \cite{efron2011tweedie} (Appdx.~\ref{subsec:Tweedie's-formula-for}),
we have:
\begin{equation}
x_{0|t}=\Expect_{x_{0}|x_{t}}\left[x_{0}\right]=x_{t}+\sigma_{t}^{2}\nabla\log p_{t}\left(x_{t}\right)
\end{equation}
Therefore, Eq.~\ref{eq:PosteriorVelo_xt_x0_3} is equivalent to:
\begin{equation}
v^{*}\left(x_{t},t\right)=-\dot{\sigma}_{t}\sigma_{t}\nabla\log p_{t}\left(x_{t}\right)
\end{equation}
which is identical to the velocity of the probability flow ODE in
Eq.~\ref{eq:PFODE_DDIM_1}.

\subsubsection{Sub-VP SDE}

The (forward) sub-VP SDE \cite{song2020score} has the following formula:
\begin{equation}
dx_{t}=-\frac{\beta_{t}}{2}x_{t}\ dt+\sqrt{\beta_{t}\left(1-e^{-2\int_{0}^{t}\beta_{\tau}\ d_{\tau}}\right)}dw_{t}
\end{equation}
The solution of the Fokker-Planck equation corresponding to the sub-VP
SDE is $p\left(x_{t}|x_{0}\right)=\Normal\left(\sqrt{\overline{\alpha}_{t}}x_{0},\left(1-\overline{\alpha}_{t}\right)^{2}\mathrm{I}\right)$.
This allows us to reformulate the sub-VP SDE as:
\begin{equation}
X_{t}=\sqrt{\overline{\alpha}_{t}}X_{0}+\left(1-\overline{\alpha}_{t}\right)X_{1}\label{eq:subVPSDE_stochastic_process}
\end{equation}
where $\overline{\alpha}_{t}:=e^{-\frac{t}{2}\left(\beta_{t}+\betamin\right)}$
is the same as in the VP SDE. In the above process, $a_{t}=\sqrt{\overline{\alpha}_{t}}$
and $\sigma_{t}=1-\overline{\alpha}_{t}$. Their derivatives w.r.t.
$t$ can be expressed as follows:

\begin{equation}
\dot{a}_{t}=\frac{d\left(\sqrt{\overline{\alpha}_{t}}\right)}{dt}=\frac{1}{2\sqrt{\overline{\alpha}_{t}}}\frac{d\overline{\alpha}_{t}}{dt}=\frac{-\beta_{t}\sqrt{\overline{\alpha}_{t}}}{2}
\end{equation}
\begin{equation}
\dot{\sigma}_{t}=\frac{d\left(1-\overline{\alpha}_{t}\right)}{dt}=\frac{-d\overline{\alpha}_{t}}{dt}=\overline{\alpha}_{t}\beta_{t}
\end{equation}
where $\frac{d\overline{\alpha}_{t}}{dt}=-\overline{\alpha}_{t}\beta_{t}$
according to Eq.~\ref{eq:alpha_bar_t_dot}. 

The target velocity is represented as: 
\begin{equation}
v_{\phi}\left(x_{0},x_{1},t\right)=-\frac{\beta_{t}\sqrt{\overline{\alpha}_{t}}}{2}x_{0}+\overline{\alpha}_{t}\beta_{t}x_{1}
\end{equation}
The posterior velocity $v^{*}\left(x_{t},t\right)$, according to
Eq.~\ref{eq:Posterior_velo_decomp_xt_x1_6}, is expressed as:
\begin{align}
v^{*}\left(x_{t},t\right)=\  & \frac{\dot{a}_{t}}{a_{t}}x_{t}+\left(\dot{\sigma}_{t}-\frac{\dot{a}_{t}\sigma_{t}}{a_{t}}\right)\Expect_{x_{1}|x_{t}}\left[x_{1}\right]\\
=\  & \frac{-\beta_{t}\sqrt{\overline{\alpha}_{t}}}{2\sqrt{\overline{\alpha}_{t}}}x_{t}+\left(\overline{\alpha}_{t}\beta_{t}+\frac{\beta_{t}\sqrt{\overline{\alpha}_{t}}}{2}\frac{1-\overline{\alpha}_{t}}{\sqrt{\overline{\alpha}_{t}}}\right)x_{1|t}\\
=\  & -\frac{\beta_{t}}{2}x_{t}+\left(\overline{\alpha}_{t}\beta_{t}+\frac{\beta_{t}\left(1-\overline{\alpha}_{t}\right)}{2}\right)x_{1|t}\\
=\  & -\frac{\beta_{t}}{2}x_{t}+\frac{\beta_{t}\left(1+\overline{\alpha}_{t}\right)}{2}x_{1|t}\\
=\  & -\frac{\beta_{t}}{2}\left(x_{t}-\left(1+\overline{\alpha}_{t}\right)x_{1|t}\right)
\end{align}
Meanwhile, if Eq.~\ref{eq:Posterior_velo_decomp_xt_x0_6} is used,
the posterior velocity has the following analytical expression:
\begin{align}
v^{*}\left(x_{t},t\right)=\  & \frac{\dot{\sigma}_{t}}{\sigma_{t}}x_{t}+\left(\dot{a}_{t}-\frac{\dot{\sigma}_{t}a_{t}}{\sigma_{t}}\right)\Expect_{x_{0}|x_{t}}\left[x_{0}\right]\\
=\  & \frac{\overline{\alpha}_{t}\beta_{t}}{1-\overline{\alpha}_{t}}x_{t}+\left(\frac{-\beta_{t}\sqrt{\overline{\alpha}_{t}}}{2}-\frac{\overline{\alpha}_{t}\beta_{t}\sqrt{\overline{\alpha}_{t}}}{1-\overline{\alpha}_{t}}\right)x_{0|t}\\
=\  & \frac{\beta_{t}\overline{\alpha}_{t}}{1-\overline{\alpha}_{t}}x_{t}-\frac{\beta_{t}\sqrt{\overline{\alpha}_{t}}\left(1-\overline{\alpha}_{t}\right)+2\beta_{t}\overline{\alpha}_{t}\sqrt{\overline{\alpha}_{t}}}{2\left(1-\overline{\alpha}_{t}\right)}x_{0|t}\\
=\  & \frac{\beta_{t}\overline{\alpha}_{t}}{1-\overline{\alpha}_{t}}x_{t}-\frac{\beta_{t}\sqrt{\overline{\alpha}_{t}}\left(1+\overline{\alpha}_{t}\right)}{2\left(1-\overline{\alpha}_{t}\right)}x_{0|t}\\
=\  & \frac{\beta_{t}\sqrt{\overline{\alpha}_{t}}}{1-\overline{\alpha}_{t}}\left(\sqrt{\overline{\alpha}_{t}}x_{t}-\frac{1+\overline{\alpha}_{t}}{2}x_{0|t}\right)
\end{align}

\subsubsection{Rectified Flow}

Rectified Flow \cite{liu2022flow} is the posterior flow of the following
stochastic process:
\begin{equation}
X_{t}=\left(1-t\right)X_{0}+tX_{1}\label{eq:stochastic_process_RF}
\end{equation}
Here, $a_{t}=1-t$ and $\sigma_{t}=t$. The target velocity is $v_{\phi}\left(x_{0},x_{1},t\right)=x_{1}-x_{0}$. 

The closed-form expression of $v^{*}\left(x_{t},t\right)$ according
to Eq.~\ref{eq:Posterior_velo_decomp_xt_x1_6} is:
\begin{align}
v^{*}\left(x_{t},t\right) & =\frac{\dot{a}_{t}}{a_{t}}x_{t}+\left(\dot{\sigma}_{t}-\frac{\dot{a}_{t}\sigma_{t}}{a_{t}}\right)\Expect_{x_{1}|x_{t}}\left[x_{1}\right]\\
 & =\frac{-1}{1-t}x_{t}+\left(1+\frac{t}{1-t}\right)x_{1|t}\\
 & =\frac{-1}{1-t}x_{t}+\frac{1}{1-t}x_{1|t}\\
 & =\frac{1}{1-t}\left(x_{1|t}-x_{t}\right)
\end{align}
By contrast, if we use Eq.~\ref{eq:Posterior_velo_decomp_xt_x0_6},
$v^{*}\left(x_{t},t\right)$ can be expressed as:

\begin{align}
v^{*}\left(x_{t},t\right) & =\frac{\dot{\sigma}_{t}}{\sigma_{t}}x_{t}+\left(\dot{a}_{t}-\frac{\dot{\sigma}_{t}a_{t}}{\sigma_{t}}\right)\Expect_{x_{0}|x_{t}}\left[x_{0}\right]\\
 & =\frac{1}{t}x_{t}+\left(-1-\frac{1-t}{t}\right)x_{0|t}\\
 & =\frac{1}{t}x_{t}-\frac{1}{t}x_{0|t}\\
 & =\frac{1}{t}\left(x_{t}-x_{0|t}\right)\label{eq:posterior_velo_RF_xtx0_4}
\end{align}
If we assume that $X_{1}$ is the standard Gaussian, we can represent
$x_{0|t}$ using the Tweedie's formula (Appdx.~\ref{subsec:Tweedie's-formula-for})
as follows: 

\begin{equation}
x_{0|t}=\frac{x_{t}+t^{2}\nabla\log p_{t}\left(x_{t}\right)}{1-t}
\end{equation}
Consequently, the posterior velocity in Eq.~\ref{eq:posterior_velo_RF_xtx0_4}
can be written as:

\begin{align}
v^{*}\left(x_{t},t\right) & =\frac{1}{t}\left(x_{t}-\frac{x_{t}+t^{2}\nabla\log p\left(x_{t}\right)}{1-t}\right)\\
 & =\frac{1}{t}\left(\frac{-t}{1-t}x_{t}-\frac{t^{2}}{1-t}\nabla\log p\left(x_{t}\right)\right)\\
 & =\frac{-1}{1-t}\left(x_{t}+t\nabla\log p\left(x_{t}\right)\right)
\end{align}
This implies that the probability flow ODE is defined by the equation:
\begin{equation}
dx=\left(-\frac{1}{1-t}x_{t}-\frac{t}{1-t}\nabla\log p\left(x_{t}\right)\right)dt
\end{equation}
This probability flow ODE corresponds to the following diffusion process:

\begin{equation}
dx=-\frac{1}{1-t}x_{t}dt+\sqrt{\frac{2t}{1-t}}dw_{t}
\end{equation}

\subsection{Tweedie's formula for posterior Gaussian mean estimation\label{subsec:Tweedie's-formula-for}}

Assume that $x$ is a sample from an isotropic Gaussian distribution
with an \emph{unknown} mean $\mu$ and a \emph{known} standard deviation
$\sigma$, i.e., $x\sim\mathcal{N}\left(\mu,\sigma^{2}\mathrm{I}\right)$.
The posterior expectation of $\mu$ given $x$ can computed as follows:
\begin{equation}
\hat{\mu}=\Expect_{p\left(\mu|x\right)}\left[\mu\right]=x+\sigma^{2}\nabla\log p\left(x\right)\label{eq:Tweedie-formula-gauss}
\end{equation}
where $p\left(x\right)=\sum_{\mu}p\left(x,\mu\right)=\Expect_{p\left(\mu\right)}\left[p\left(x|\mu\right)\right]$.
Eq.~\ref{eq:Tweedie-formula-gauss} is known as Tweedie's formula
\cite{efron2011tweedie}.

To prove the correctness of Eq.~\ref{eq:Tweedie-formula-gauss},
we begin with the score function $\nabla\log p(x)$ and transform
it as follows:
\begin{align}
\nabla\log p(x) & =\frac{1}{p(x)}\nabla p(x)\\
 & =\frac{1}{p(x)}\nabla\sum_{\mu}p(\mu)p(x|\mu)\\
 & =\frac{1}{p(x)}\sum_{\mu}p(\mu)\nabla p(x|\mu)\\
 & =\sum_{\mu}\frac{p(\mu)p(x|\mu)}{p(x)}\nabla\log p(x|\mu)\label{eq:Tweedie_score_4}\\
 & =\sum_{\mu}p(\mu|x)\left(\frac{\mu-x}{\sigma^{2}}\right)\label{eq:Tweedie_score_5}\\
 & =\Expect_{p(\mu|x)}\left[\frac{\mu-x}{\sigma^{2}}\right]\\
 & =\frac{\Expect_{p(\mu|x)}\left[\mu\right]-x}{\sigma^{2}}\label{eq:Tweedie_score_7}
\end{align}
Eq.~\ref{eq:Tweedie_score_4} uses the log trick $\nabla p(x|\mu)=\frac{\nabla\log p(x|\mu)}{p(x|\mu)}$,
and Eq.~\ref{eq:Tweedie_score_5} makes use of the fact that if $x\sim\Normal\left(\mu,\sigma^{2}\mathrm{I}\right)$
then $\nabla\log p(x|\mu)=\frac{\mu-x}{\sigma^{2}}$. We can rearrange
Eq.~\ref{eq:Tweedie_score_7} to obtain Eq.~\ref{eq:Tweedie-formula-gauss}.

\subsection{Explanation of why $x_{1|t}$ and $x_{0|1,t}$ should be invariant
w.r.t. $t$\label{subsec:Explanation-of-invariant-wrt-t}}

Because two samples $x_{t}$, $x_{s}$ in the same ODE trajectory
have the same (unknown) boundary points $x_{0}$, $x_{1}$, the distributions
$p\left(x_{0}|x_{t}\right)$ and $p\left(x_{1}|x_{t}\right)$, characterized
by the ODE, tend to be \emph{sharp} and similar to $p\left(x_{0}|x_{s}\right)$
and $p\left(x_{1}|x_{s}\right)$, respectively. This means $\Expect_{p\left(x_{0}|x_{t}\right)}\left[x_{0}\right]\approx\Expect_{p\left(x_{0}|x_{s}\right)}\left[x_{0}\right]$
and $\Expect_{p\left(x_{1}|x_{t}\right)}\left[x_{1}\right]\approx\Expect_{p\left(x_{1}|x_{s}\right)}\left[x_{1}\right]$,
or equivalently, $x_{0|t}\approx x_{0|s}$ and $x_{1|t}\approx x_{1|s}$.
Given this result, we can apply the same reasoning to show that $x_{1|0,t}\approx x_{1|0,s}$
and $x_{0|1,t}\approx x_{0|1,s}$.

\subsection{Proof of Proposition \ref{Prop_1}\label{subsec:Proof-of-Proposition-1}}

By definition, $v_{\phi}\left(\tilde{x}_{t},t\right):=\frac{d\tilde{x}_{t}}{dt}$
where $\tilde{x}_{t}=k_{t}x_{t}=k_{t}\left(a_{t}x_{0}+\sigma_{t}x_{1}\right)$
for some $x_{0}$, $x_{1}$. Thus, we can compute $v_{\phi}\left(\tilde{x}_{t},t\right)$
as follows:
\begin{align}
v_{\phi}\left(\tilde{x}_{t},t\right) & =\frac{d}{dt}\left(k_{t}x_{t}\right)\label{eq:TargetVelo_X_bar_t_1}\\
 & =\dot{k}_{t}x_{t}+k_{t}\dot{x}_{t}\label{eq:TargetVelo_X_bar_t_2}\\
 & =\dot{k}_{t}x_{t}+k_{t}v_{\phi}\left(x_{t},t\right)\label{eq:TargetVelo_X_bar_t_3}
\end{align}
Recalling that $v_{\phi}\left(x_{t},t\right)=\dot{a}_{t}x_{0}+\dot{\sigma}_{t}x_{1}$
(Eq.~\ref{eq:target_velo_1}), $v_{\phi}\left(\tilde{x}_{t},t\right)$
can be expressed as a function of $x_{0}$, $x_{1}$ and $t$ as follows:
\begin{align*}
v_{\phi}\left(\tilde{x}_{t},t\right) & =\dot{k}_{t}\left(a_{t}x_{0}+\sigma_{t}x_{1}\right)+k_{t}\left(\dot{a}_{t}x_{0}+\dot{\sigma}_{t}x_{1}\right)\\
 & =\left(\dot{k}_{t}a_{t}+k_{t}\dot{a}_{t}\right)x_{0}+\left(\dot{k}_{t}\sigma_{t}+k_{t}\dot{\sigma}_{t}\right)x_{1}\\
 & =\frac{d\left(k_{t}a_{t}\right)}{dt}x_{0}+\frac{d\left(k_{t}\sigma_{t}\right)}{dt}x_{1}
\end{align*}
The posterior velocity $v\left(\tilde{x}_{t},t\right)$ can be computed
as follows:
\begin{align}
v^{*}\left(\tilde{x}_{t},t\right) & :=\Expect_{x_{0},x_{1}|\tilde{x}_{t}}\left[v_{\phi}\left(\tilde{x}_{t},t\right)\right]\label{eq:PosteriorVelo_X_bar_t_1}\\
 & =\Expect_{x_{0},x_{1}|x_{t}}\left[\dot{k}_{t}x_{t}+k_{t}v_{\phi}\left(x_{t},t\right)\right]\label{eq:PosteriorVelo_X_bar_t_2}\\
 & =\dot{k}_{t}x_{t}+k_{t}\Expect_{x_{0},x_{1}|x_{t}}\left[v_{\phi}\left(x_{t},t\right)\right]\label{eq:PosteriorVelo_X_bar_t_3}\\
 & =\dot{k}_{t}x_{t}+k_{t}v^{*}\left(x_{t},t\right)\label{eq:PosteriorVelo_X_bar_t_4}
\end{align}
In Eq.~\ref{eq:PosteriorVelo_X_bar_t_2}, $p\left(x_{0},x_{1}|x_{t}\right)=p\left(x_{0},x_{1}|\tilde{x}_{t}\right)$
because $\tilde{x}_{t}=k_{t}x_{t}$ is a \emph{deterministic} function
of $x_{t}$ and map to the same $\left(x_{0},x_{1}\right)$ pair as
$x_{t}$.

\subsection{Mathematical derivations of the transformed velocities $v^{*}\left(\tilde{x}_{t},t\right)$
and $v^{*}\left(\bar{x}_{t},t\right)$\label{subsec:Derivations-of-SN-SC-velocities}}

Below, we derive the velocities of the transformed straight (SN) and
straight constant-speed (SC) flows corresponding to the stochastic
process $X_{t}=a_{t}X_{0}+\sigma_{t}X_{1}$. Our results are summarized
in Tables~\ref{tab:straight_transform_summary_x0x1_itpl} and \ref{tab:straight_transform_summary_x1scale}.

\subsubsection{Non-straight flows $\rightarrow$ Straight flows\label{subsec:Appdx_NonStraight2Straight}}

Applying the results in Proposition~\ref{Prop_1} to the SN process
$\tilde{X}_{t}=X_{0}+\varphi_{t}X_{1}$ with $\tilde{X}_{t}=\frac{X_{t}}{a_{t}}$
and $\varphi_{t}=\frac{\sigma_{t}}{a_{t}}$, we can compute $v_{\phi}\left(\tilde{x}_{t},t\right)$
and $v^{*}\left(\tilde{x}_{t},t\right)$ as follows:
\begin{equation}
v_{\phi}\left(\tilde{x}_{t},t\right)=\frac{a_{t}v_{\phi}\left(x_{t},t\right)-\dot{a}_{t}x_{t}}{a_{t}^{2}}
\end{equation}

\begin{align}
v^{*}\left(\tilde{x}_{t},t\right) & =\frac{a_{t}v^{*}\left(x_{t},t\right)-\dot{a}_{t}x_{t}}{a_{t}^{2}}\\
 & \approx\frac{a_{t}v_{\theta}\left(x_{t},t\right)-\dot{a}_{t}x_{t}}{a_{t}^{2}}
\end{align}

Moreover, in the case of the SN process $\tilde{X}_{t}=\left(1-\text{\ensuremath{\varphi_{t}}}\right)X_{0}+\varphi_{t}X_{1}$,
we can represent $v_{\phi}\left(\tilde{x}_{t},t\right)$ and $v^{*}\left(\tilde{x}_{t},t\right)$
via $x_{0}$, $x_{1}$ and $x_{0|1,t}$, $x_{1|t}$ respectively as
follows: 
\begin{align}
v_{\phi}\left(\tilde{x}_{t}\left(x_{0},x_{1},t\right),t\right):= & \frac{d\tilde{x}_{t}\left(x_{0},x_{1},t\right)}{dt}\label{eq:target_velo_SN_via_x0_x1_1}\\
= & \frac{d}{dt}\left(\left(1-\varphi_{t}\right)x_{0}+\varphi_{t}x_{1}\right)\\
= & \dot{\varphi}_{t}\left(x_{1}-x_{0}\right)
\end{align}

\begin{align}
v^{*}\left(\tilde{x}_{t}\left(x_{0},x_{1},t\right),t\right) & :=\Expect_{x_{0},x_{1}|\tilde{x}_{t}}\left[v_{\phi}\left(\tilde{x}_{t}\left(x_{0},x_{1},t\right),t\right)\right]\\
 & =\Expect_{x_{0},x_{1}|\tilde{x}_{t}}\left[\dot{\varphi}_{t}\left(x_{1}-x_{0}\right)\right]\\
 & =\dot{\varphi}_{t}\Expect_{x_{0},x_{1}|\tilde{x}_{t}}\left[x_{1}-x_{0}\right]\\
 & =\dot{\varphi}_{t}\Expect_{x_{0},x_{1}|x_{t}}\left[x_{1}-x_{0}\right]\\
 & =\dot{\varphi}_{t}\left(x_{1|t}-x_{0|1,t}\right)\label{eq:posterior_velo_SN_via_x0_x1_5}
\end{align}
where the analytic formula of $\dot{\varphi}_{t}$ is:
\begin{align}
\dot{\varphi}_{t} & =\frac{d}{dt}\left(\frac{\sigma_{t}}{a_{t}+\sigma_{t}}\right)\\
 & =\frac{\left(a_{t}+\sigma_{t}\right)\dot{\sigma}_{t}-\left(\dot{a}_{t}+\dot{\sigma}_{t}\right)\sigma_{t}}{\left(a_{t}+\sigma_{t}\right)^{2}}\\
 & =\frac{a_{t}\dot{\sigma}_{t}-\dot{a}_{t}\sigma_{t}}{\left(a_{t}+\sigma_{t}\right)^{2}}\label{eq:varphi_dot_3}
\end{align}
This means besides Eq.~\ref{eq:straight_velo_itpl}, $v^{*}\left(\tilde{x}_{t},t\right)$
can be computed as:
\begin{align}
v^{*}\left(\tilde{x}_{t},t\right) & =\frac{a_{t}\dot{\sigma}_{t}-\dot{a}_{t}\sigma_{t}}{\left(a_{t}+\sigma_{t}\right)^{2}}\left(x_{1|t}-x_{0|1,t}\right)\\
 & =\frac{a_{t}\dot{\sigma}_{t}-\dot{a}_{t}\sigma_{t}}{\left(a_{t}+\sigma_{t}\right)^{2}}\left(x_{1|t}-\frac{x_{t}-\sigma_{t}x_{1|t}}{a_{t}}\right)\\
 & =\frac{a_{t}\dot{\sigma}_{t}-\dot{a}_{t}\sigma_{t}}{\left(a_{t}+\sigma_{t}\right)^{2}}\left(\frac{\left(a_{t}+\sigma_{t}\right)x_{1|t}-x_{t}}{a_{t}}\right)\\
 & \approx\frac{a_{t}\dot{\sigma}_{t}-\dot{a}_{t}\sigma_{t}}{\left(a_{t}+\sigma_{t}\right)^{2}}\left(\frac{\left(a_{t}+\sigma_{t}\right)x_{1,\theta}-x_{t}}{a_{t}}\right)
\end{align}

Applying the same line of reasoning from Eqs.~\ref{eq:target_velo_SN_via_x0_x1_1}-\ref{eq:posterior_velo_SN_via_x0_x1_5}
to the SN process $\tilde{X}_{t}=X_{0}+\varphi_{t}X_{1}$, we have
$v_{\phi}\left(\tilde{x}_{t}\left(x_{0},x_{1},t\right),t\right)=\dot{\varphi}_{t}x_{1}$
and $v^{*}\left(\tilde{x}_{t}\left(x_{0},x_{1},t\right),t\right)=\dot{\varphi}_{t}x_{1|t}$
with $\dot{\varphi}_{t}=\frac{a_{t}\dot{\sigma}_{t}-\dot{a}_{t}\sigma_{t}}{a_{t}^{2}}$.

\subsubsection{Straight flows $\rightarrow$ Straight constant-speed flows\label{subsec:Appdx_Straight2StraightConstant}}

\paragraph{Time adjustment transformation}

Since the posterior velocity corresponding to the straight (SN) process
$\tilde{X}_{t}=\left(1-\varphi_{t}\right)X_{0}+\varphi_{t}X_{1}$
is $v^{*}\left(\tilde{x}_{t},t\right)=\dot{\varphi}_{t}\left(x_{1|t}-x_{0|1,t}\right)$
(Eq.~\ref{eq:posterior_velo_SN_via_x0_x1_5}), we can compute the
posterior velocity $v^{*}\left(\bar{x}_{t},t\right)$ corresponding
to the straight constant-speed (SC) process $d\bar{X}_{t}=\left(X_{1}-X_{0}\right)d\varphi_{t}$
derived from this SN process as follows:
\begin{align}
v^{*}\left(\bar{x}_{t},t\right) & =x_{1|t}-x_{0|1,t}\\
 & =\frac{v^{*}\left(\tilde{x}_{t},t\right)}{\dot{\varphi}_{t}}\\
 & =\frac{\left(a_{t}+\sigma_{t}\right)v^{*}\left(x_{t},t\right)-\left(\dot{a}_{t}+\dot{\sigma}_{t}\right)x_{t}}{a_{t}\dot{\sigma}_{t}-\dot{a}_{t}\sigma_{t}}
\end{align}
where $v^{*}\left(\tilde{x}_{t},t\right)=\frac{\left(a_{t}+\sigma_{t}\right)v^{*}\left(x_{t},t\right)-\left(\dot{a}_{t}+\dot{\sigma}_{t}\right)x_{t}}{\left(a_{t}+\sigma_{t}\right)^{2}}$
and $\dot{\varphi}_{t}=\frac{a_{t}\dot{\sigma}_{t}-\dot{a}_{t}\sigma_{t}}{\left(a_{t}+\sigma_{t}\right)^{2}}$
are given in Eq.~\ref{eq:straight_velo_itpl} and Eq.~\ref{eq:varphi_dot_3},
respectively.

Similarly, the posterior velocity $v^{*}\left(\bar{x}_{t},t\right)$
associated with the SC process $d\bar{X}_{t}=X_{1}d\varphi_{t}$ derived
from the SN process $\tilde{X}_{t}=X_{0}+\varphi_{t}X_{1}$ can be
expressed as follows:
\begin{align}
v^{*}\left(\bar{x}_{t},t\right) & =x_{1|t}\\
 & =\frac{v^{*}\left(\tilde{x}_{t},t\right)}{\dot{\varphi}_{t}}\\
 & =\frac{a_{t}v^{*}\left(x_{t},t\right)-\dot{a}_{t}x_{t}}{a_{t}\dot{\sigma}_{t}-\dot{a}_{t}\sigma_{t}}
\end{align}
where $v^{*}\left(\tilde{x}_{t},t\right)=\frac{a_{t}v^{*}\left(x_{t},t\right)-\dot{a}_{t}x_{t}}{a_{t}^{2}}$
and $\dot{\varphi}_{t}=\frac{a_{t}\dot{\sigma}_{t}-\dot{a}_{t}\sigma_{t}}{a_{t}^{2}}$.

\paragraph{Variable shifting transformation}

If we transform the SN process $\tilde{X}_{t}=\left(1-\varphi_{t}\right)X_{0}+\varphi_{t}X_{1}$
into the SC process $\bar{X}_{t}=\left(1-t\right)X_{0}+tX_{1}$ using
the ``variable shifting'' transformation $\bar{X}_{t}=\tilde{X}_{t}+\left(t-\varphi_{t}\right)\left(X_{1}-X_{0}\right)$,
we can represent the target and posterior velocities of this SC process
as follows:
\begin{align}
v_{\phi}\left(x_{0},x_{1},t\right) & :=\frac{d\bar{x}_{t}\left(x_{0},x_{1},t\right)}{dt}\\
 & =\frac{d}{dt}\left(\tilde{x}_{t}\left(x_{0},x_{1},t\right)+\left(t-\varphi_{t}\right)\left(x_{1}-x_{0}\right)\right)\\
 & =\frac{d\tilde{x}_{t}\left(x_{0},x_{1},t\right)}{dt}+\left(1-\dot{\varphi}_{t}\right)\left(x_{1}-x_{0}\right)\\
 & =v_{\phi}\left(\tilde{x}_{t},t\right)+\left(1-\dot{\varphi}_{t}\right)\left(x_{1}-x_{0}\right)\\
 & =\dot{\varphi}_{t}\left(x_{1}-x_{0}\right)+\left(1-\dot{\varphi}_{t}\right)\left(x_{1}-x_{0}\right)\\
 & =x_{1}-x_{0}
\end{align}
\begin{align}
v^{*}\left(\bar{x}_{t},t\right) & :=\Expect_{x_{0},x_{1}|\bar{x}_{t}}\left[v_{\phi}\left(x_{0},x_{1},t\right)\right]\\
 & =\Expect_{x_{0},x_{1}|\bar{x}_{t}}\left[x_{1}-x_{0}\right]\label{eq:posterior_velo_variable_shift_itpl_appdx_2}\\
 & =\Expect_{x_{0},x_{1}|\tilde{x}_{t}}\left[x_{1}-x_{0}\right]\label{eq:posterior_velo_variable_shift_iplt_appdx_3}\\
 & =\Expect_{x_{0},x_{1}|x_{t}}\left[x_{1}-x_{0}\right]\\
 & =x_{1|t}-x_{0|1,t}
\end{align}
where Eq.~\ref{eq:posterior_velo_variable_shift_iplt_appdx_3} is
equivalent to Eq.~\ref{eq:posterior_velo_variable_shift_itpl_appdx_2}
because $\bar{x}_{t}$ and $\tilde{x}_{t}$ are in the \emph{same
straight trajectory} and have the \emph{same boundary points} $x_{0}$
and $x_{1}$ at $t=0$ and $t=1$, respectively.

Similarly, if we transform the SN process $\tilde{X}_{t}=X_{0}+\varphi_{t}X_{1}$
into the SC process $\bar{X}_{t}=X_{0}+tX_{1}$ using the transformation
$\bar{X}_{t}=\tilde{X}_{t}+\left(t-\varphi_{t}\right)X_{1}$, we can
express the target and posterior velocities of this SC process as
follows:
\begin{align}
v_{\phi}\left(x_{0},x_{1},t\right) & =x_{1}\\
v^{*}\left(\bar{x}_{t},t\right) & =x_{1|t}
\end{align}

\paragraph{Equivalence between the \textquotedblleft time adjustment\textquotedblright{}
and \textquotedblleft variable shifting\textquotedblright{} transformations}

Here, we prove that applying the ``time adjustment'' (TA) and ``variable
shifting'' (VS) transformations to an SN flow leads to equivalent
SC flows when the Euler method is used for sampling. Our main idea
is to show that the SN samples recovered from the SC flows corresponding
to the TA and VS transformations at step $t'$ (i.e., $\tilde{x}_{t'}^{\text{TA}}$
and $\tilde{x}_{t'}^{\text{VS}}$), are the same. Without loss of
generality, we assume the SN flow corresponds to the process $\tilde{X}_{t}=\left(1-\varphi_{t}\right)X_{0}+\varphi_{t}X_{1}$.
Then, we have:

\begin{align}
 & \tilde{x}_{t'}^{\text{TA}}\\
=\  & \tilde{x}_{t}^{\text{TA}}+\left(\varphi_{t'}-\varphi_{t}\right)\left(x_{1|t}-x_{0|1,t}\right)\\
=\  & \tilde{x}_{t}+\left(t-\varphi_{t}\right)\left(x_{1|t}-x_{0|1,t}\right)\nonumber \\
 & +\left(t'-t\right)\left(x_{1|t}-x_{0|1,t}\right)\nonumber \\
 & -\left(t'-\varphi_{t'}\right)\left(x_{1|t}-x_{0|1,t}\right)\\
=\  & \bar{x}_{t}^{\text{VS}}+\left(t'-t\right)\left(x_{1|t}-x_{0|1,t}\right)\nonumber \\
 & +\left(\varphi_{t'}-t'\right)\left(x_{1|t}-x_{0|1,t}\right)\\
=\  & \bar{x}_{t'}^{\text{VS}}-\left(t'-\varphi_{t'}\right)\left(x_{1|t}-x_{0|1,t}\right)\\
=\  & \tilde{x}_{t'}^{\text{VS}}
\end{align}
which completes the proof.

\subsection{DDIM as a special case of our method\label{subsec:DDIM-as-a-special-case}}

It is worth noting that DDIM \cite{song2020denoising} can be viewed
as a special case of our method, which transforms the probability
flow associated with VP SDE / DDPM \cite{ho2020denoising,song2020score}
into a straight constant-speed counterpart by applying a ``variable
scaling'' transformation followed by a ``time adjustment'' transformation
to the original flow. A DDIM-styled Euler update for a general class
of stochastic processes $X_{t}=a_{t}X_{0}+\sigma_{t}X_{1}$ is given
as follows:
\begin{align}
 & \bar{x}_{t'}=\bar{x}_{t}+\left(\varphi_{t'}-\varphi_{t}\right)x_{1|t}\\
\Leftrightarrow\  & \frac{x_{t'}}{a_{t'}}=\frac{x_{t}}{a_{t}}+\left(\frac{\sigma_{t'}}{a_{t'}}-\frac{\sigma_{t}}{a_{t}}\right)x_{1|t}\\
\Leftrightarrow\  & x_{t'}=a_{t'}\left(\frac{x_{t}-\sigma_{t}x_{1|t}}{a_{t}}+\frac{\sigma_{t'}}{a_{t'}}x_{1|t}\right)\label{eq:DDIM_styled_SC_3}\\
\Leftrightarrow\  & x_{t'}=a_{t'}x_{0|1,t}+\sigma_{t'}x_{1|t}
\end{align}
where $\bar{x}_{t}:=\frac{x_{t}}{a_{t}}$ and $\varphi_{t}:=\frac{\sigma_{t}}{a_{t}}$.
When the original stochastic process is VP SDE (i.e., $a_{t}=\sqrt{\overline{\alpha}_{t}}$
and $\sigma_{t}=\sqrt{1-\overline{\alpha}_{t}}$), $t$ is converted
to an integer in $\left[1,T\right]$, $t'=t-1$, and $x_{1|t}$ is
approximated by $\epsilon_{\theta}\left(x_{t},t\right)$, Eq.~\ref{eq:DDIM_styled_SC_3}
can be written as follows:
\begin{align}
x_{t-1}=\  & \sqrt{\overline{\alpha}_{t-1}}\left(\frac{x_{t}-\sqrt{1-\overline{\alpha}_{t}}\epsilon_{\theta}\left(x_{t},t\right)}{\sqrt{\overline{\alpha}_{t}}}\right)\nonumber \\
 & +\sqrt{1-\overline{\alpha}_{t-1}}\epsilon_{\theta}\left(x_{t},t\right)
\end{align}
which matches the deterministic sampling step in DDIM \cite{song2020denoising}.

\section{High-order Numerical Solvers on Straight Constant-speed Flows\label{sec:High-order-Numerical-Solvers-on-SC-Flows}}

\subsection{General formulations}

In this section, we provide mathematical derivations for the well-known
Runge-Kutta and linear multi-step methods on SC flows. We exclusively
focus on SC flows obtained via the ``time adjustment'' transformation
because applying high-order numerical solvers to this type of SC flows
is more straightforward compared to the counterpart obtained through
the ``variable shift'' transformation. The primary advantage of
the former is the exact back-and-forth mapping between $x_{t'}$ and
$\bar{x}_{t'}$ without the need for $v^{*}\left(\bar{x}_{t'},t'\right)$.
The exploration of high-order numerical solvers for the latter type
of SC flows is reserved for future work.

To streamline our presentation, we adopt the notations $t_{i}$ and
$t_{i+1}$ to denote the current and next time steps, respectively,
instead of $t$ and $t'$. We also use $t_{i-k}$ to indicate the
$k$-th step before the current step $t_{i}$. The posterior velocity
$v^{*}\left(\bar{x}_{t_{i}},t_{i}\right)$ of a sample $\bar{x}_{t_{i}}$
in a SC flow will be denoted as $\bar{v}_{t_{i}}^{*}$. 

Our proposed high-order solvers on SC flows adhere to the general
update formula below:
\begin{equation}
\bar{x}_{t_{i+1}}=\bar{x}_{t_{i}}+\Delta\varphi_{t_{i+1}}\bar{v}_{t_{i}:t_{i+1}}^{\circ}\label{eq:general_update_formula}
\end{equation}
where $\Delta\varphi_{t_{i+1}}:=\varphi_{t_{i+1}}-\varphi_{t_{i}}$,
and $\bar{v}_{t_{i}:t_{i+1}}^{\circ}$ represents the \emph{average}
velocity along SC flows within the time interval from $t_{i}$ to
$t_{i+1}$. Different numerical methods yield distinct approximations
of $\bar{v}_{t_{i}:t_{i+1}}^{\circ}$.

\subsubsection{Runge-Kutta methods}

\begin{table}[H]
\begin{centering}
\begin{tabular}{c|cccc}
$0$ & $0$ & $0$ & $\cdots$ & $0$\tabularnewline
$c_{2}$ & $a_{21}$ & $0$ & $\cdots$ & $0$\tabularnewline
$\vdots$ & $\vdots$ & $\vdots$ & $\ddots$ & $\vdots$\tabularnewline
$c_{k}$ & $a_{k1}$ & $a_{k2}$ & $\cdots$ & $0$\tabularnewline
\hline 
 & $b_{1}$ & $b_{2}$ & $\cdots$ & $b_{k}$\tabularnewline
\end{tabular}
\par\end{centering}
\caption{The Butcher tableau for explicit Runge-Kutta methods\label{tab:Butcher-tableau}}
\end{table}

The general update formula of explicit Runge-Kutta (RK) methods \emph{on
SC flows} is given by:
\begin{equation}
\bar{x}_{t_{i+1}}=\bar{x}_{t_{i}}+\Delta\varphi_{t_{i+1}}\sum_{j=1}^{k}b_{j}f_{j}\label{eq:General_RK_SC_final}
\end{equation}
where $f_{1}$, $f_{2}$,..., $f_{k}$ are computed as follows:
\begin{align}
f_{1} & =v^{*}\left(\bar{x}_{t_{i}},t_{i}\right)\label{eq:General_RK_SC_f1}\\
f_{2} & =v^{*}\left(\bar{x}_{t_{i}}+\Delta\varphi_{\tau_{2}}\left(\frac{a_{2,1}}{c_{2}}f_{1}\right),\tau_{2}\right);\ \tau_{2}=t_{i}+c_{2}\Delta t_{i+1}\label{eq:General_RK_SC_f2}\\
\vdots\nonumber \\
f_{k} & =v^{*}\left(\bar{x}_{t_{i}}+\Delta\varphi_{\tau_{k}}\left(\sum_{j=1}^{k-1}\frac{a_{k,j}}{c_{k}}f_{j}\right),\tau_{k}\right);\ \tau_{k}=t_{i}+c_{k}\Delta t_{i+1}\label{eq:General_RK_SC_fk}
\end{align}
Here, $\Delta\varphi_{t_{i+1}}:=\varphi_{t_{i+1}}-\varphi_{t_{i}}$
and $\Delta\varphi_{\tau_{k}}:=\varphi_{\tau_{k}}-\varphi_{t_{i}}$
for all $k\geq2$. The coefficients $b_{1}$,..., $b_{k}$, $c_{2}$,...,
$c_{k}$, $a_{2,1}$,..., $a_{k,k-1}$ should satisfy $\sum_{j=1}^{k}b_{j}=1$,
$\sum_{j=1}^{k-1}a_{k,j}=c_{k}$, and other conditions which ensure
that Runge-Kutta methods of order $k$ closely replicate the Taylor
series method of the same order. Typically, these coefficients are
structured within the Butcher tableau, as illustrated in Table~\ref{tab:Butcher-tableau},
for convenient reference.

When comparing Eq.~\ref{eq:General_RK_SC_final} with Eq.~\ref{eq:general_update_formula},
it becomes evident that the average velocity $\bar{v}_{t_{i}:t_{i+1}}^{\circ}$
can be interpreted as the linear interpolation of intermediate velocities
$f_{1}$, $f_{2}$,..., $f_{k}$ due to the condition $\sum_{j=1}^{k}b_{j}=1$.
Moreover, $f_{k}$ can be seen as the linear interpolation of $f_{1}$,...,
$f_{k-1}$ since $\sum_{j=1}^{k-1}\frac{a_{k,j}}{c_{k}}=1$.

The main advantage of the update formulas presented in Eqs.~\ref{eq:General_RK_SC_f1}-\ref{eq:General_RK_SC_fk}
lies in the separation of \emph{sample and time updates} into \emph{two
distinct time spaces}. Specifically, the sample update is performed
in the scaled time space (i.e., $\varphi$-space) while the time update
is conducted in the original time space (i.e., the $t$-space). 

If both the sample and time updates are executed in the $\varphi$-space,
explicit Runge-Kutta methods will exhibit the following general formulas:
\begin{equation}
\bar{x}_{\varphi_{i+1}}=\bar{x}_{\varphi_{i}}+\Delta\varphi_{i+1}\sum_{j=1}^{k}b_{j}f_{j}\label{eq:RK_SC_varphi_final}
\end{equation}
where $f_{1}$, $f_{2}$,..., $f_{k}$ are computed as follows:
\begin{align}
f_{1} & =v^{*}\left(\bar{x}_{\varphi_{i}},\varphi_{i}\right)\label{eq:RK_SC_varphi_f1}\\
f_{2} & =v^{*}\left(\bar{x}_{\varphi_{i}}+c_{2}\Delta\varphi_{i+1}\left(\frac{a_{2,1}}{c_{2}}f_{1}\right),\varphi_{i}+c_{2}\Delta\varphi_{i+1}\right)\label{eq:RK_SC_varphi_f2}\\
\vdots\nonumber \\
f_{k} & =v^{*}\left(\bar{x}_{\varphi_{i}}+c_{k}\Delta\varphi_{i+1}\left(\sum_{j=1}^{k-1}\frac{a_{k,j}}{c_{k}}f_{j}\right),\varphi_{i}+c_{k}\Delta\varphi_{i+1}\right)\label{eq:RK_SC_varphi_fk}
\end{align}
Here, $\Delta\varphi_{i+1}:=\varphi_{i+1}-\varphi_{i}$. In the above
equations, we delibrately use the notation $\varphi_{i}$ instead
of $\varphi_{t_{i}}$ to emphasize that we explicitly specify $\varphi_{i}$
during sampling, rather than specifying $t_{i}$ and then computing
$\varphi_{t_{i}}$ as a function of $t_{i}$.

However, since the velocity/noise network $v_{\theta}$/$x_{1,\theta}$
is trained in the $t$-space rather than the $\varphi$-space, we
need to map $\varphi_{i}+c_{k}\Delta\varphi_{i+1}$ back to the $t$-space
for computing $f_{k}$. This requires knowing the analytical form
of the inverse mapping $t\left(\varphi\right):=\varphi^{-1}\left(t\right)$,
which is often challenging to obtain for many stochastic processes.
Moreover, $t\left(\varphi_{i}+c_{k}\Delta\varphi_{i+1}\right)$ is
usually a real number different from the integer time steps used in
discrete-time models like DDPM \cite{ho2020denoising} or Stable Diffusion
\cite{rombach2022high}), leading to the approximation error when
$t\left(\varphi_{i}+c_{k}\Delta\varphi_{i+1}\right)$ is approximated
with the closest discrete time step. By performing time updates in
the $t$-space as shown in Eqs.~\ref{eq:General_RK_SC_f1}-\ref{eq:General_RK_SC_fk},
these challenges can be effectively mitigated.

In Appdx.~\ref{subsec:Runge-Kutta-methods-Appdx}, we delve into
the specific constraints governing the coefficients of explicit Runge-Kutta
(RK) methods of order 2 and 3. These contraints have some degrees
of freedom that give us flexibility in applying RK methods to discrete-time
models. Specifically, we can tailor $c_{k}$ to ensure that $t_{i}+c_{k}\Delta t_{i+1}$
aligns with a predefined discrete time step. Additionally, we provide
the update formulas for several well-known RK methods such as the
midpoint method, Kutta's third-order method (RK3), and the classic
Runge-Kutta method (RK4) on SC flows.

Below is the update formula for Heun's method, a second-order RK method,
on SC flows:
\begin{align}
\bar{x}_{t_{i+1}}^{\text{a}} & =\bar{x}_{t_{i}}+\Delta\varphi_{t_{i+1}}\bar{v}_{t_{i}}^{*}\label{eq:Heun_predictor}\\
\bar{x}_{t_{i+1}} & =\bar{x}_{t_{i}}+\Delta\varphi_{t_{i+1}}\frac{v^{*}\left(\bar{x}_{t_{i+1}}^{\text{a}},t_{i+1}\right)+\bar{v}_{t_{i}}^{*}}{2}\label{eq:Heun_corrector}
\end{align}

We note that $\bar{v}_{t_{i}}^{*}$ equals $x_{1|t_{i}}-x_{0|t_{i}}$
and can be computed directly via Eq.~\ref{eq:straight_constant_velo_way_1_velo_net}
or Eq.~\ref{eq:straight_constant_velo_way_1_noise_net}. To compute
$v^{*}\left(\bar{x}_{t_{i+1}}^{\text{a}},t_{i+1}\right)$ in Eq.~\ref{eq:Heun_corrector},
we first recover $x_{t_{i+1}}^{\text{a}}$ from $\bar{x}_{t_{i+1}}^{\text{a}}$
as $x_{t_{i+1}}^{\text{a}}=\left(a_{t_{i+1}}+\sigma_{t_{i+1}}\right)\bar{x}_{t_{i+1}}^{\text{a}}$
or $x_{t_{i+1}}^{\text{a}}=a_{t_{i+1}}\bar{x}_{t_{i+1}}^{\text{a}}$
depending on whether the straight process is $\tilde{X}_{t}=\left(1-\varphi_{t}\right)X_{0}+\varphi_{t}X_{1}$
(Eq.~\ref{eq:straight_stochastic_1}) or $\tilde{X}_{t}=X_{0}+\varphi_{t}X_{1}$
(Eq.~\ref{eq:straight_stochastic_2}). Then, we feed the tuple ($x_{t_{i+1}}^{\text{a}}$,
$t_{i+1}$) to the velocity/noise network $v_{\theta}$/$x_{1,\theta}$
to compute $v^{*}\left(\bar{x}_{t_{i+1}}^{\text{a}},t_{i+1}\right)$.

\subsubsection{Linear multi-step methods}

High-order RK methods typically compute intermediate velocities $f_{1}$,
$f_{2}$,... at intermediate steps between $t_{i}$ and $t_{i+1}$
to derive the final velocity $\bar{v}_{t_{i}:t_{i+1}}^{\circ}$ for
update. However, they do not reuse these intermediate velocities in
subsequent updates, resulting in computational inefficiency. To address
this limitation of RK methods, linear multi-step methods such as Adams-Bashforth
(AB) and Adams-Moulton (AM) \cite{atkinson2009numerical,griffiths2010numerical}
utilize cached velocities $\bar{v}_{t_{i-1}}^{*}$, $\bar{v}_{t_{i-2}}^{*}$,...
from prior updates. This allows them to achieve a high-order approximation
of $\bar{x}_{t_{i+1}}$ with minimal or no additional model evaluations.
For both AB and AM updates on SC flows, we propose the following approximations
of $\bar{v}_{t_{i}:t_{i+1}}^{\circ}$:
\begin{equation}
\bar{v}_{t_{i}:t_{i+1}}^{\circ}\approx\frac{\int_{t_{i}}^{t_{i+1}}\bar{v}\left(t\right)dt}{\Delta t_{i+1}}\label{eq:multistep_update_in_t_space}
\end{equation}
and
\begin{equation}
\bar{v}_{t_{i}:t_{i+1}}^{\circ}\approx\frac{\int_{\varphi_{t_{i}}}^{\varphi_{t_{i+1}}}\bar{v}\left(\varphi_{t}\right)d\varphi_{t}}{\Delta\varphi_{t_{i+1}}}\label{eq:multistep_update_in_varphi_space}
\end{equation}
where $\Delta t_{i+1}:=t_{i+1}-t_{i}$, $\Delta\varphi_{t_{i+1}}:=\varphi_{t_{i+1}}-\varphi_{t_{i}}$;
$\bar{v}\left(t\right)$ and $\bar{v}\left(\varphi_{t}\right)$ are
\emph{Lagrange interpolating polynomial functions} of $t$ and $\varphi_{t}$,
respectively. 

The main difference between Eq.~\ref{eq:multistep_update_in_t_space}
and Eq.~\ref{eq:multistep_update_in_varphi_space} is that in the
former, velocity interpolation is conducted in the $t$-space with
\emph{evenly-spaced} time steps while in the latter, interpolation
is performed in the $\varphi$-space with \emph{unevenly-spaced} time
steps. Given that $\varphi_{t}$ can be extremely large at $t=0$
and $t=1$ when the straight process is $\tilde{X}_{t}=\varphi_{t}X_{0}+X_{1}$
(i.e., $\varphi_{0}=\frac{a_{0}}{\sigma_{0}}$ with $\sigma_{0}=0$)
and $\tilde{X}_{t}=X_{0}+\varphi_{t}X_{1}$ (i.e., $\varphi_{1}=\frac{\sigma_{1}}{a_{1}}$
with $a_{1}=0$), respectively, interpolation in the $\varphi$-space
with unequal time gaps tends to yield inaccurate results in these
cases. By contrast, this problem does not occur with the interpolation
in the $t$-space (Eq.~\ref{eq:multistep_update_in_t_space}) thanks
to the well normalization of $t$ within $[0,1]$ and the use of uniform
time intervals.

This problem can alternatively be addressed if we perform linear multi-step
updates at evenly-spaced time steps in the $\varphi$-space as follows:
\begin{align}
\bar{x}_{\varphi_{i+1}} & =\bar{x}_{\varphi_{i}}+\Delta\varphi_{i+1}\bar{v}_{\varphi_{i}:\varphi_{i+1}}^{\circ}\label{eq:general_update_on_varphi_space}\\
\bar{v}_{\varphi_{i}:\varphi_{i+1}}^{\circ} & \approx\frac{\int_{\varphi_{i}}^{\varphi_{i+1}}\bar{v}\left(\varphi\right)d\varphi}{\Delta\varphi_{i+1}}\label{eq:multistep_update_in_varphi_space_evenly}
\end{align}
where $\Delta\varphi_{i+1}:=\varphi_{i+1}-\varphi_{i}$ is assumed
to be \emph{(approximately) the same} for different $i$. The notation
$\varphi_{i}$ in Eqs.~\ref{eq:general_update_on_varphi_space},
\ref{eq:multistep_update_in_varphi_space_evenly} has the same interpretation
as the courterpart in Eqs.~\ref{eq:RK_SC_varphi_final}-\ref{eq:RK_SC_varphi_fk}.

Below, we delve into the details of the Adams-Bashforth (AB) and Adams-Moulton
(AM) methods corresponding to Eq.~\ref{eq:multistep_update_in_t_space}.
At the core of these methods lie the \emph{interpolating coefficients}
($L_{0}$, $L_{1}$, ...) and \emph{velocities} ($\bar{v}_{t_{i}}^{*}$,
$\bar{v}_{t_{i-1}}^{*}$, ...). By examining the formulas of the AB
and AM methods w.r.t Eq.~\ref{eq:multistep_update_in_t_space}, one
can easily derive the formulas of the counterparts related to Eqs.~\ref{eq:multistep_update_in_varphi_space}
and \ref{eq:multistep_update_in_varphi_space_evenly}. Specifically,
the interpolating velocities w.r.t Eq.~\ref{eq:multistep_update_in_varphi_space}
mirror those associated with Eq.~\ref{eq:multistep_update_in_t_space},
while the interpolating coefficients w.r.t. Eq.~\ref{eq:multistep_update_in_varphi_space}
can be obtained from the counterparts corresponding to Eq.~\ref{eq:multistep_update_in_t_space}
by replacing $t_{i}$ with $\varphi_{t_{i}}$ in the formulas. The
interpolating coefficients w.r.t Eq.~\ref{eq:multistep_update_in_varphi_space_evenly}
closely resemble their counterparts w.r.t. Eq.~\ref{eq:multistep_update_in_t_space}
as both cases involve evenly-spaced time steps, albeit in different
time spaces. As for the interpolating velocities w.r.t Eq.~\ref{eq:multistep_update_in_varphi_space_evenly},
although they differ from those w.r.t. Eq.~\ref{eq:multistep_update_in_t_space},
they can be computed by mapping $\varphi_{i}$ back to $t_{\varphi_{i}}$
in the $t$-space.

\paragraph{Adams-Bashforth methods}

In an Adams-Bashforth method of order $k$ (AB$k$), $\bar{v}\left(t\right)$
is a Lagrange interpolating polynomial of order $k$ passing through
$k$ velocities $\bar{v}_{t_{i}}^{*},\bar{v}_{t_{i-1}}^{*},...,\bar{v}_{t_{i-(k-1)}}^{*}$
at time steps $t_{i}$, $t_{i-1}$,..., $t_{i-(k-1)}$, respectively.
The general update formula of an AB$k$ method is expressed as follows:
\begin{equation}
\bar{x}_{t_{i+1}}=\bar{x}_{t_{i}}+\Delta\varphi_{t_{i+1}}\left(\sum_{j=1}^{k-1}L_{j}\bar{v}_{t_{i-j}}^{*}\right)
\end{equation}
where $L_{0},...,L_{k-1}$ are referred to as interpolating coefficients.

In particular, the AB2 method has the following update:

\begin{equation}
\bar{x}_{t_{i+1}}=\bar{x}_{t_{i}}+\Delta\varphi_{t_{i+1}}\left(L_{0}\bar{v}_{t_{i}}^{*}+L_{1}\bar{v}_{t_{i-1}}^{*}\right)\label{eq:AB2_update}
\end{equation}
Here, $L_{0}=\frac{t_{i+1}+t_{i}-2t_{i-1}}{2\left(t_{i}-t_{i-1}\right)}$
and $L_{1}=\frac{-\left(t_{i+1}-t_{i}\right)}{2\left(t_{i}-t_{i-1}\right)}$.
Meanwhile, the AB3 update is given by:

\begin{equation}
\bar{x}_{t_{i+1}}=\bar{x}_{t_{i}}+\Delta\varphi_{t_{i+1}}\left(L_{0}\bar{v}_{t_{i}}^{*}+L_{1}\bar{v}_{t_{i-1}}^{*}+L_{2}\bar{v}_{t_{i-2}}^{*}\right)\label{eq:AB3_update}
\end{equation}
where $L_{0}$, $L_{1}$, and $L_{2}$ are computed as follows:
\begin{align}
L_{0} & =\frac{2\left(t_{i+1}^{2}+t_{i+1}t_{i}+t_{i}^{2}\right)-3\left(t_{i-1}+t_{i-2}\right)\left(t_{i+1}+t_{i}\right)+6t_{i-1}t_{i-2}}{6\left(t_{i}-t_{i-1}\right)\left(t_{i}-t_{i-2}\right)}\\
L_{1} & =\frac{\left(t_{i+1}-t_{i}\right)\left(2t_{i+1}+t_{i}-3t_{i-2}\right)}{6\left(t_{i-1}-t_{i}\right)\left(t_{i-1}-t_{i-2}\right)}\\
L_{2} & =\frac{\left(t_{i+1}-t_{i}\right)\left(2t_{i+1}+t_{i}-3t_{i-1}\right)}{6\left(t_{i-2}-t_{i}\right)\left(t_{i-2}-t_{i-1}\right)}
\end{align}

Detailed mathematical derivations of these interpolating coefficients
are provided in Appdx.~\ref{subsec:Linear-multi-step-methods-Appdx}.
If time steps are evenly spaced (i.e., $\Delta t_{i+1}=\Delta t_{i}=...=\Delta t$),
the AB2 and AB3 updates can be written in simpler forms below:
\begin{equation}
\bar{x}_{t_{i+1}}=\bar{x}_{t_{i}}+\Delta\varphi_{t_{i+1}}\left(\frac{3}{2}\bar{v}_{t_{i}}^{*}-\frac{1}{2}\bar{v}_{t_{i-1}}^{*}\right),\ \text{and}\label{eq:AB2_update_simple}
\end{equation}
\begin{equation}
\bar{x}_{t_{i+1}}=\bar{x}_{t_{i}}+\Delta\varphi_{t_{i+1}}\left(\frac{23}{12}\bar{v}_{t_{i}}^{*}-\frac{16}{12}\bar{v}_{t_{i-1}}^{*}+\frac{5}{12}\bar{v}_{t_{i-2}}^{*}\right)\label{eq:AB3_update_simple}
\end{equation}
Eqs.~\ref{eq:AB2_update_simple}, \ref{eq:AB3_update_simple} resemble
the update formulas of the well-known second- and third-order AB methods
\cite{griffiths2010numerical}.

In an AB method of order $k$, we should use a numerical method of
order at least $k-1$ to perform its initial $k-1$ updates to ensure
its order of accuracy is $k$ \cite{atkinson2009numerical}. Consequently,
we can use either the Euler solver or Heun's solver for the first
update of the AB2 method. The latter option incurs 1 additional model
evaluation. Similarly, for the first two updates of the AB3 method,
we can use either the AB2 method, Heun's method, or the RK3 method,
incurring at most 1, 2, and 4 additional model evaluations, respectively.
In general, to maintain the order of accuracy with minimal additional
model evaluations, we should conduct the first $k-1$ updates of an
AB method of order $k$ using AB methods of increasing orders from
1 to $k-1$.

\paragraph{Adams-Moulton methods}

Unlike explicit AB methods, AM methods are \emph{implicit}. Therefore,
in an AM method of order $k$ (AM$k$), $\bar{v}\left(t\right)$ is
a Lagrange interpolating polynomial of order $k$ going through $k$
velocities $\bar{v}_{t_{i+1}}^{*},\bar{v}_{t_{i}}^{*},...,\bar{v}_{t_{i+1-(k-1)}}^{*}$
at time steps $t_{i+1}$, $t_{i-1}$,..., $t_{i+1-(k-1)}$, respectively.

The AM2 update is expressed as:
\begin{equation}
\bar{x}_{t_{i+1}}=\bar{x}_{t_{i}}+\Delta\varphi_{t_{i+1}}\frac{v^{*}\left(\bar{x}_{t_{i+1}},t_{i+1}\right)+\bar{v}_{t_{i}}^{*}}{2}\label{eq:AM2_update}
\end{equation}
which is essentially the trapezoidal method. Instead of solving Eq.~\ref{eq:AM2_update}
using a root finding algorithm (e.g., Newton's method), we can approximate
$v^{*}\left(\bar{x}_{t_{i+1}},t_{i+1}\right)$ based on a reasonable
prediction of $\bar{x}_{t_{i+1}}$. If we use the Euler method as
the predictor, we obtain Heun's method. On the other hand, if we use
the AB2 method as the predictor, we get the following predictor-corrector
method, referred to as AB2AM2:
\begin{align}
\bar{x}_{t_{i+1}}^{\text{a}} & =\bar{x}_{t_{i}}+\Delta\varphi_{t_{i+1}}\left(L_{0}\bar{v}_{t_{i}}^{*}+L_{1}\bar{v}_{t_{i-1}}^{*}\right)\\
\bar{x}_{t_{i+1}} & =\bar{x}_{t_{i}}+\Delta\varphi_{t_{i+1}}\frac{v^{*}\left(\bar{x}_{t_{i+1}}^{\text{a}},t_{i+1}\right)+\bar{v}_{t_{i}}^{*}}{2}
\end{align}
with $L_{0}$, $L_{1}$ similar to those in Eq.~\ref{eq:AB2_update}. 

In general, we can combine an AB method of either order $k-1$ or
$k$ with an AM method of order $k$ to create a predictor-corrector
method of order $k$ \cite{griffiths2010numerical,moulton1926new}
as follows:
\begin{align}
\bar{x}_{t_{i+1}}^{\text{a}} & =\mathrm{AB}\left(\bar{x}_{t_{i}},\bar{v}_{t_{i}}^{*},...,\left[\bar{v}_{t_{i-(k-1)}}^{*}\right]\right)\label{eq:AB_predictor}\\
\bar{x}_{t_{i+1}} & =\mathrm{AM}\left(\bar{x}_{t_{i}},\bar{v}_{t_{i+1}}^{*,\text{a}},\bar{v}_{t_{i}}^{*},...,\bar{v}_{t_{i+1-(k-1)}}^{*}\right)\label{eq:AM_corrector}
\end{align}
Here, $\bar{v}_{t_{i+1}}^{*,\text{a}}$ in Eq.~\ref{eq:AM_corrector}
denotes $v^{*}\left(\bar{x}_{t_{i+1}}^{\text{a}},t_{i+1}\right)$
and the notion $\left[v_{t_{i-(k-1)}}^{*}\right]$ in Eq.~\ref{eq:AB_predictor}
implies that $v_{t_{i-(k-1)}}^{*}$ can be either provided or not.
If $v_{t_{i-(k-1)}}^{*}$ is available, the predictor is an AB method
of order $k$; otherwise it is an AB method of order $k-1$.

However, the aforementioned predictor-corrector method requires 2
function evaluations per update: one for computing the current velocity
$v_{t_{i}}^{*}$ and the other for computing the predicted future
velocity $\bar{v}_{t_{i+1}}^{*,\text{a}}$. Therefore, compared to
an AB method of the same order with a similar number of function evaluations
(NFE), the predictor-corrector method exhibits almost no gain in accuracy.
Fortunately, we can reduce the NFE to 1 per update by reusing $\bar{v}_{t_{i}}^{*,\text{a}}$
from the previous step $t_{i-1}$ in place of $\bar{v}_{t_{i}}^{*}$
to predict $\bar{x}_{t_{i+1}}^{\text{a}}$ in Eq.~\ref{eq:AB_predictor}.
This technique draws inspiration from UniPC \cite{zhao2023unipc}.
Although it introduces a small error when predicting $\bar{x}_{t_{i+1}}^{\text{a}}$
since $\bar{v}_{t_{i}}^{*,\text{a}}\neq\bar{v}_{t_{i}}^{*}$, this
error will be corrected in the correction step in Eq.~\ref{eq:AM_corrector}.

\subsection{Runge-Kutta methods\label{subsec:Runge-Kutta-methods-Appdx}}

\subsubsection{Second-order Runge-Kutta methods}

Below, we explicitly derive the relationships between the coefficients
$b_{1},b_{2},c_{2},a_{2,1}$ in the formulas of the second-order RK
methods.

The Taylor series of order 2 for $x\left(t_{i+1}\right)$ is given
by:
\begin{align}
x\left(t_{i+1}\right) & =x\left(t_{i}+h\right)\\
 & =x\left(t_{i}\right)+hx'\left(t_{i}\right)+\frac{h^{2}}{2}x''\left(t_{i}\right)+\mathcal{O}\left(h^{3}\right)\label{eq:Taylor_series_2}
\end{align}
where $h:=\Delta t_{i+1}=t_{i+1}-t_{i}$. $\mathcal{O}\left(h^{3}\right)$
means that the truncation error is a third-degree polynomial function
of $h$.

We can express $x''\left(t\right)$ as follows:
\begin{align}
x''\left(t\right) & =\frac{d}{dt}\left(x'\left(t\right)\right)\\
 & =\frac{d}{dt}v\left(x,t\right)\\
 & =\frac{\partial v\left(x,t\right)}{\partial x}v\left(x,t\right)+\frac{\partial v\left(x,t\right)}{\partial t}\\
 & =v_{x}\left(x,t\right)v\left(x,t\right)+v_{t}\left(x,t\right)\label{eq:second_grad_of_xt_4}
\end{align}
where $v_{x}\left(x,t\right)$ and $v_{t}\left(x,t\right)$ denote
$\frac{\partial v\left(x,t\right)}{\partial x}$ and $\frac{\partial v\left(x,t\right)}{\partial t}$,
respectively. Applying the formula of $x''\left(t\right)$ in Eq.~\ref{eq:second_grad_of_xt_4}
to Eq.~\ref{eq:Taylor_series_2}, we have:
\begin{align}
x\left(t_{i+1}\right)=\  & x\left(t_{i}\right)+hv\left(x_{t_{i}},t_{i}\right)\nonumber \\
 & +\frac{h^{2}}{2}\left(v_{x}\left(x_{t_{i}},t_{i}\right)v\left(x_{t_{i}},t_{i}\right)+v_{t}\left(x_{t_{i}},t_{i}\right)\right)\nonumber \\
 & +\mathcal{O}\left(h^{3}\right)\\
=\  & x\left(t_{i}\right)+h\left(v\left(x_{t_{i}},t_{i}\right)+\frac{h}{2}v_{x}\left(x_{t_{i}},t_{i}\right)v\left(x_{t_{i}},t_{i}\right)+\frac{h}{2}v_{t}\left(x_{t_{i}},t_{i}\right)\right)\nonumber \\
 & +\mathcal{O}\left(h^{3}\right)\label{eq:TS2_general_update}
\end{align}

The Taylor series (TS) method of order 2 requires knowing $v_{x}$
and $v_{t}$, which are usually unavailable in practice. To overcome
this problem, we approximate it with Runge-Kutta methods of the same
order described below:
\begin{equation}
x\left(t_{i+1}\right)=x\left(t_{i}\right)+hV\left(x_{t_{i}},t_{i},h\right)\label{eq:RK_general_update_with_V}
\end{equation}
where $V\left(x_{t_{i}},t_{i},h\right)$ is defined as:
\begin{equation}
V\left(x_{t_{i}},t_{i},h\right):=b_{1}v\left(x_{t_{i}},t_{i}\right)+b_{2}v\left(x_{t_{i}}+a_{2,1}hv\left(x_{t_{i}},t_{i}\right),t_{i}+c_{2}h\right)\label{eq:RK_general_V}
\end{equation}

The second term in the RHS of Eq.~\ref{eq:RK_general_V} can be approximated
using the first-order Taylor expansion for functions of two variables
$x$, $t$ as follows:
\begin{align}
 & v\left(x_{t_{i}}+a_{2,1}hv\left(x_{t_{i}},t_{i}\right),t_{i}+c_{2}h\right)\nonumber \\
=\  & v\left(x_{t_{i}},t_{i}\right)+a_{2,1}hv\left(x_{t_{i}},t_{i}\right)v_{x}\left(x_{t_{i}},t_{i}\right)+c_{2}hv_{t}\left(x_{t_{i}},t_{i}\right)+\mathcal{O}\left(h^{2}\right)\label{eq:RK_general_V_2}
\end{align}

From Eqs.~\ref{eq:RK_general_update_with_V}, \ref{eq:RK_general_V},
\ref{eq:RK_general_V_2}, we have a Runge-Kutta method of order 2
can be written as:
\begin{align}
x\left(t_{i+1}\right)=\  & x\left(t_{i}\right)+h\Big(\left(b_{1}+b_{2}\right)v\left(x_{t_{i}},t_{i}\right)\nonumber \\
 & \ \ \ \ \ \ \ \ \ \ +b_{2}a_{2,1}hv_{x}\left(x_{t_{i}},t_{i}\right)v\left(x_{t_{i}},t_{i}\right)\nonumber \\
 & \ \ \ \ \ \ \ \ \ \ +b_{2}c_{2}hv_{t}\left(x_{t_{i}},t_{i}\right)\Big)+\mathcal{O}\left(h^{3}\right)\label{eq:RK2_general_update}
\end{align}

For Eq.~\ref{eq:RK2_general_update} to be similar to Eq.~\ref{eq:TS2_general_update},
the coefficients $b_{1},b_{2},c_{2},a_{2,1}$ should satisfy the following
conditions:
\begin{align}
a_{2,1} & =c_{2}\\
b_{1}+b_{2} & =1\\
b_{2}c_{2} & =1/2
\end{align}

Since there are 4 variables but only 3 equations, we can let $c_{2}$
vary arbitrarily in the range $(0,1]$ and compute other coefficients
based on $c_{2}$ as follows:
\begin{align}
a_{2,1} & =c_{2}\label{eq:RK2_coeff_constraints_1}\\
b_{2} & =\frac{1}{2c_{2}}\label{eq:RK2_coeff_constraints_2}\\
b_{1} & =1-\frac{1}{2c_{2}}\label{eq:RK2_coeff_constraints_3}
\end{align}

\paragraph{Heun's method}

In Heun's method, $c_{2}=1$, which means $\tau_{2}=t_{i}+\Delta t_{i+1}=t_{i+1}$.
Since $t_{i+1}$ is always specified before sampling, this method
has \emph{no} problem working with discrete-time models.

\paragraph{Explicit midpoint method}

The explicit midpoint method corresponds to $c_{2}=\frac{1}{2}$.
Its coefficients are given in Table~\ref{tab:midpoint_butcher} and
its update on SC flows is represented as follows:
\begin{equation}
\bar{x}_{t_{i+1}}=\bar{x}_{t_{i}}+\Delta\varphi_{t_{i+1}}f_{2}\label{eq:midpoint_update}
\end{equation}
where $f_{1}$, $f_{2}$ are given by:
\begin{align}
f_{1} & =v^{*}\left(\bar{x}_{t_{i}},t_{i}\right)\\
f_{2} & =v^{*}\left(\bar{x}_{t_{i}}+\Delta\varphi_{\tau_{2}}f_{1},\tau_{2}\right);\ \tau_{2}=t_{i}+\frac{1}{2}\Delta t_{i+1}
\end{align}

\begin{table}[H]
\begin{centering}
\begin{tabular}{c|cc}
0 & 0 & 0\tabularnewline
1/2 & 1/2 & 0\tabularnewline
\hline 
 & 0 & 1\tabularnewline
\end{tabular}
\par\end{centering}
\caption{The Butcher tableau for the explicit midpoint method\label{tab:midpoint_butcher}}
\end{table}

Different from Heun's method, the explicit midpoint method is not
always applicable to discrete-time models when $c_{2}$ is \emph{fixed}
at $\frac{1}{2}$. This is because $\tau_{2}$ can be a \emph{non-integer}
if $t_{i}$ and $t_{i+1}$ have \emph{different parities} (i.e., $t_{i}$
is odd and $t_{i+1}$ is even, or vice versa), resulting in an unreliable
computation of $f_{2}$. We can overcome this problem by ensuring
that all $N$ discrete sampling time steps $t_{1}$,..., $t_{N}$
have the same parity but this is very tedious. A much easier way is
rounding $\tau_{2}$ to the closest integer, then recomputing $c_{2}$
as $c_{2}=\tau_{2}-t_{i}$ and other coefficients via Eqs.~\ref{eq:RK2_coeff_constraints_1}-\ref{eq:RK2_coeff_constraints_3}.

\subsubsection{Third-order Runge-Kutta methods}

For the third-order RK methods, we present the set of constraints
governing their coefficients without explicit derivation. They are
as follows:
\begin{align}
a_{2,1} & =c_{2}\\
a_{3,1}+a_{3,2} & =c_{3}\\
b_{1}+b_{2}+b_{3} & =1\\
b_{2}c_{2}+b_{3}c_{3} & =1/2\\
b_{2}c_{2}^{2}+b_{3}c_{3}^{2} & =1/3\\
c_{2}a_{3,2}b_{3} & =1/6
\end{align}
This system comprises 8 variables but only 6 equations, which means
two variables can be chosen arbitrarily. Since our focuses are $c_{2}$
and $c_{3}$, we set $c_{3}=1$ and let $c_{2}$ vary. Other variables
can be computed based on $c_{2}$ as follows: $a_{3,1}=\frac{3c_{2}-3c_{2}^{2}-1}{c_{2}\left(2-3c_{2}\right)}$,
$a_{3,2}=\frac{1-c_{2}}{c_{2}\left(2-3c_{2}\right)}$, $a_{2,1}=c_{2}$,
$b_{3}=\frac{2-3c_{2}}{6\left(1-c_{2}\right)}$, $b_{2}=\frac{1}{6c_{2}\left(1-c_{2}\right)}$,
and $b_{1}=\frac{3c_{2}-1}{6c_{2}}$.

\paragraph{Kutta's third-order method (RK3)}

The RK3 method corresponds to $c_{3}=1$ and $c_{2}=1/2$. Its coefficients
are shown in Table~\ref{tab:RK3_butcher} and its update on SC flows
is given below:
\begin{equation}
\bar{x}_{t_{i+1}}=\bar{x}_{t_{i}}+\Delta\varphi_{t_{i+1}}\left(\frac{f_{1}+4f_{2}+f_{3}}{6}\right)\label{eq:RK3_update}
\end{equation}
where $f_{1}$, $f_{2}$, $f_{3}$ are computed as:
\begin{align}
f_{1} & =v^{*}\left(\bar{x}_{t_{i}},t_{i}\right)\\
f_{2} & =v^{*}\left(\bar{x}_{t_{i}}+\Delta\varphi_{\tau_{2}}f_{1},\tau_{2}\right);\ \tau_{2}=t_{i}+\frac{1}{2}\Delta t_{i+1}\\
f_{3} & =v^{*}\left(\bar{x}_{t_{i}}+\Delta\varphi_{\tau_{3}}\left(2f_{2}-f_{1}\right),\tau_{3}\right);\ \tau_{3}=t_{i}+\Delta t_{i+1}
\end{align}

\begin{table}[H]
\begin{centering}
\begin{tabular}{c|ccc}
0 & 0 & 0 & 0\tabularnewline
1/2 & 1/2 & 0 & 0\tabularnewline
1 & -1 & 2 & 0\tabularnewline
\hline 
 & 1/6 & 2/3 & 1/6\tabularnewline
\end{tabular}
\par\end{centering}
\caption{The Butcher tableau for the RK3 method\label{tab:RK3_butcher}}
\end{table}

\subsubsection{Fourth-order Runge-Kutta methods}

Since it is challenging to derive the formulas of the coefficients
of fourth-order RK methods with one free variable. Below, we only
provide the coefficients for the well-known classic Runge-Kutta method
(RK4).

\paragraph{Classic Runge-Kutta method (RK4)}

The classic Runge-Kutta method (RK4) on SC flows is characterized
by the following equation:
\begin{equation}
\bar{x}_{t_{i+1}}=\bar{x}_{t_{i}}+\Delta\varphi_{t_{i+1}}\left(\frac{f_{1}+2f_{2}+2f_{3}+f_{4}}{6}\right)\label{eq:RK4_update}
\end{equation}
where $f_{1}$, $f_{2}$, $f_{3}$ are expressed as:
\begin{align}
f_{1} & =v^{*}\left(\bar{x}_{t_{i}},t_{i}\right)\\
f_{2} & =v^{*}\left(\bar{x}_{t_{i}}+\Delta\varphi_{\tau_{2}}f_{1},\tau_{2}\right);\ \tau_{2}=t_{i}+\frac{1}{2}\Delta t_{i+1}\\
f_{3} & =v^{*}\left(\bar{x}_{t_{i}}+\Delta\varphi_{\tau_{3}}f_{2},\tau_{3}\right);\ \tau_{3}=t_{i}+\frac{1}{2}\Delta t_{i+1}\\
f_{4} & =v^{*}\left(\bar{x}_{t_{i}}+\Delta\varphi_{\tau_{4}}f_{3},\tau_{4}\right);\ \tau_{4}=t_{i}+\Delta t_{i+1}
\end{align}
The Butcher tableau for the RK4 method is shown in Table~\ref{tab:RK4_butcher}.

\begin{table}[H]
\begin{centering}
\begin{tabular}{c|cccc}
0 & 0 & 0 & 0 & 0\tabularnewline
1/2 & 1/2 & 0 & 0 & 0\tabularnewline
1/2 & 0 & 1/2 & 0 & 0\tabularnewline
1 & 0 & 0 & 1 & 0\tabularnewline
\hline 
 & 1/6 & 1/3 & 1/3 & 1/6\tabularnewline
\end{tabular}
\par\end{centering}
\caption{The Butcher tableau for the RK4 method\label{tab:RK4_butcher}}
\end{table}

\subsection{Linear multi-step methods\label{subsec:Linear-multi-step-methods-Appdx}}

In an AB method of order $k$ (AB$k$), $\bar{v}\left(t\right)$
is characterized by the following equation:
\begin{align}
\bar{v}\left(t\right) & =\sum_{j=0}^{k-1}\frac{\prod_{m\neq j}^{0\leq m\leq k-1}\left(t-t_{i-m}\right)}{\prod_{m\neq j}^{0\leq m\leq k-1}\left(t_{i-j}-t_{i-m}\right)}\bar{v}_{t_{i-j}}^{*}\label{eq:Lagrange_approx_AB_1}\\
 & =\sum_{j=0}^{k-1}\ell_{j}\left(t\right)\bar{v}_{t_{i-j}}^{*}\label{eq:Lagrange_approx_AB_2}
\end{align}
Integrating both sides of Eq.~\ref{eq:Lagrange_approx_AB_2}, we
have:
\begin{align}
\frac{\int_{t_{i}}^{t_{i+1}}\bar{v}\left(t\right)dt}{\Delta t_{i+1}} & =\frac{1}{\Delta t_{i+1}}\int_{t_{i}}^{t_{i+1}}\left(\sum_{j=0}^{k-1}\ell_{j}\left(t\right)\bar{v}_{t_{i-j}}^{*}\right)dt\\
 & =\sum_{j=0}^{k-1}\left(\frac{\int_{t_{i}}^{t_{i+1}}\ell_{j}\left(t\right)dt}{\Delta t_{i+1}}\right)\bar{v}_{t_{i-j}}^{*}\\
 & =\sum_{j=0}^{k-1}L_{j}\bar{v}_{t_{i-j}}^{*}
\end{align}
where $L_{j}:=\frac{\int_{t_{i}}^{t_{i+1}}\ell_{j}\left(t\right)dt}{\Delta t_{i+1}}$.
This implies that the update formula of an AB method of order $k$
is:
\begin{align}
\bar{x}_{t_{i+1}} & =\bar{x}_{t_{i}}+\Delta\varphi_{t_{i+1}}\left(\sum_{j=0}^{k-1}L_{j}\bar{v}_{t_{i-j}}^{*}\right)\label{eq:ABk_update}
\end{align}

On the other hand, in an AM method of order $k$ (AM$k$), $\bar{v}\left(t\right)$
is defined as follows:
\begin{equation}
\bar{v}\left(t\right)=\sum_{j=0}^{k-1}\frac{\prod_{m\neq j}^{0\leq m\leq k-1}\left(t-t_{i+1-m}\right)}{\prod_{m\neq j}^{0\leq m\leq k-1}\left(t_{i+1-j}-t_{i+1-m}\right)}\bar{v}_{t_{i+1-j}}^{*}
\end{equation}
which leads to the following update formula:
\begin{align}
\bar{x}_{t_{i+1}} & =\bar{x}_{t_{i}}+\Delta\varphi_{t_{i+1}}\left(\sum_{j=0}^{k-1}\tilde{L}_{j}\bar{v}_{t_{i+1-j}}^{*}\right)
\end{align}
where $\tilde{\ell}_{j}\left(t\right):=\frac{\prod_{m\neq j}^{0\leq m\leq k-1}\left(t-t_{i+1-m}\right)}{\prod_{m\neq j}^{0\leq m\leq k-1}\left(t_{i+1-j}-t_{i+1-m}\right)}$
and $\tilde{L}_{j}:=\frac{\int_{t_{i}}^{t_{i+1}}\ell_{j}\left(t\right)dt}{\Delta t_{i+1}}$.

\paragraph{AB2}

In the AB2 method, $\bar{v}\left(t\right)$ is expressed as follows:
\begin{equation}
\bar{v}\left(t\right)=\ell_{0}\left(t\right)\bar{v}_{t_{i}}^{*}+\ell_{1}\left(t\right)\bar{v}_{t_{i-1}}^{*}
\end{equation}
where $\ell_{0}\left(t\right)=\frac{t-t_{i-1}}{t_{i}-t_{i-1}}$ and
$\ell_{1}\left(t\right)=\frac{t-t_{i}}{t_{i-1}-t_{i}}$. The formula
of $L_{0}$ can be derived as follows:
\begin{align}
L_{0} & =\frac{1}{\Delta t_{i+1}}\int_{t_{i}}^{t_{i+1}}\ell_{0}\left(t\right)dt\\
 & =\frac{1}{C_{0}\Delta t_{i+1}}\int_{t_{i}}^{t_{i+1}}\left(t-t_{i-1}\right)dt\\
 & =\frac{1}{C_{0}\Delta t_{i+1}}\left(\frac{t^{2}}{2}-t_{i-1}t\bigg|_{t=t_{i}}^{t=t_{i+1}}\right)\\
 & =\frac{1}{C_{0}\Delta t_{i+1}}\left(\frac{t_{i+1}^{2}-t_{i}^{2}}{2}-t_{i-1}\left(t_{i+1}-t_{i}\right)\right)\\
 & =\frac{1}{2C_{0}\Delta t_{i+1}}\left(t_{i+1}-t_{i}\right)\left(t_{i+1}+t_{i}-2t_{i-1}\right)\label{eq:AB2_L0_5}\\
 & =\frac{t_{i+1}+t_{i}-2t_{i-1}}{2\left(t_{i}-t_{i-1}\right)}\label{eq:AB2_L0_6}
\end{align}
where $C_{0}=t_{i}-t_{i-1}$. By replacing $C_{0}$ with $C_{1}=-C_{0}$
and $t_{i-1}$ with $t_{i}$ in Eq.~\ref{eq:AB2_L0_5}, we obtain
the formula of $L_{1}$, which is:
\begin{align}
L_{1} & =\frac{t_{i+1}-t_{i}}{2C_{1}}\label{eq:AB2_L1_1}\\
 & =\frac{-\left(t_{i+1}-t_{i}\right)}{2\left(t_{i}-t_{i-1}\right)}\label{eq:AB2_L1_2}
\end{align}

Therefore, the AB2 update on SC flows is:
\begin{equation}
\bar{x}_{t_{i+1}}=\bar{x}_{t_{i}}+\Delta\varphi_{t_{i+1}}\left(L_{0}\bar{v}_{t_{i}}^{*}+L_{1}\bar{v}_{t_{i-1}}^{*}\right)
\end{equation}
where $L_{0}$ and $L_{1}$ are provided in Eqs.~\ref{eq:AB2_L0_6}
and \ref{eq:AB2_L1_2}, respectively.

\paragraph{AM2}

The velocity $\bar{v}\left(t\right)$ in the AM2 method is represented
as:
\begin{equation}
\bar{v}\left(t\right)=\tilde{\ell}_{0}\left(t\right)v_{t_{i+1}}^{*}+\tilde{\ell}_{1}\left(t\right)v_{t_{i}}^{*}
\end{equation}
where $\tilde{\ell}_{0}\left(t\right)=\frac{t-t_{i}}{t_{i+1}-t_{i}}$
and $\tilde{\ell}_{1}\left(t\right)=\frac{t-t_{i+1}}{t_{i}-t_{i+1}}$. 

We can compute $\tilde{L}_{0}$ by replacing $C_{1}$ in Eq.~\ref{eq:AB2_L1_1}
with $\tilde{C}_{0}=t_{i+1}-t_{i}$, leading to the formula:
\begin{equation}
\tilde{L}_{0}=\frac{t_{i+1}-t_{i}}{2\left(t_{i+1}-t_{i}\right)}=\frac{1}{2}
\end{equation}

The formula of $\tilde{L}_{1}$ can be obtained by replacing $t_{i-1}$
in Eq.~\ref{eq:AB2_L0_6} with $t_{i+1}$ and is given below:
\begin{equation}
\tilde{L}_{1}=\frac{t_{i}-t_{i+1}}{2\left(t_{i}-t_{i+1}\right)}=\frac{1}{2}
\end{equation}

Thus, the AM2 update on SC flows is:
\begin{equation}
\bar{x}_{t_{i+1}}=\bar{x}_{t_{i}}+\Delta\varphi_{t_{i+1}}\left(\frac{\bar{v}_{t_{i+1}}^{*}+\bar{v}_{t_{i}}^{*}}{2}\right)
\end{equation}

\paragraph{AB3}

In the AB3 method, $\bar{v}\left(t\right)$ is represented as follows:
\begin{equation}
\bar{v}\left(t\right)=\ell_{0}\left(t\right)\bar{v}_{t_{i}}^{*}+\ell_{1}\left(t\right)\bar{v}_{t_{i-1}}^{*}+\ell_{2}\left(t\right)\bar{v}_{t_{i-2}}^{*}
\end{equation}
where $\ell_{0}\left(t\right)$, $\ell_{1}\left(t\right)$, $\ell_{2}\left(t\right)$
are given by:
\begin{align}
\ell_{0}\left(t\right) & =\frac{\left(t-t_{i-1}\right)\left(t-t_{i-2}\right)}{\left(t_{i}-t_{i-1}\right)\left(t_{i}-t_{i-2}\right)}\\
\ell_{1}\left(t\right) & =\frac{\left(t-t_{i}\right)\left(t-t_{i-2}\right)}{\left(t_{i-1}-t_{i}\right)\left(t_{i-1}-t_{i-2}\right)}\\
\ell_{2}\left(t\right) & =\frac{\left(t-t_{i}\right)\left(t-t_{i-1}\right)}{\left(t_{i-2}-t_{i}\right)\left(t_{i-2}-t_{i-1}\right)}
\end{align}

The coefficient $L_{0}$ has the following formula:
\begin{align}
L_{0}=\  & \int_{t_{i}}^{t_{i+1}}\ell_{0}\left(t\right)dt\\
=\  & \frac{1}{C_{0}\Delta t_{i+1}}\int_{t_{i}}^{t_{i+1}}\left(t-t_{i-1}\right)\left(t-t_{i-2}\right)dt\\
=\  & \frac{1}{C_{0}\Delta t_{i+1}}\int_{t_{i}}^{t_{i+1}}\left(t^{2}-\left(t_{i-1}+t_{i-2}\right)t+t_{i-1}t_{i-2}\right)dt\\
=\  & \frac{1}{C_{0}\Delta t_{i+1}}\left(\frac{t^{3}}{3}-\left(t_{i-1}+t_{i-2}\right)\frac{t^{2}}{2}+t_{i-1}t_{i-2}t\bigg|_{t_{i}}^{t_{i+1}}\right)\\
=\  & \frac{1}{C_{0}\Delta t_{i+1}}\left(\frac{t_{i+1}^{3}-t_{i}^{3}}{3}-\left(t_{i-1}+t_{i-2}\right)\frac{t_{i+1}^{2}-t_{i}^{2}}{2}+t_{i-1}t_{i-2}\left(t_{i+1}-t_{i}\right)\right)\\
=\  & \frac{\left(t_{i+1}-t_{i}\right)}{C_{0}\Delta t_{i+1}}\left(\frac{t_{i+1}^{2}+t_{i+1}t_{i}+t_{i}^{2}}{3}-\frac{\left(t_{i-1}+t_{i-2}\right)\left(t_{i+1}+t_{i}\right)}{2}+t_{i-1}t_{i-2}\right)\\
=\  & \frac{1}{6C_{0}}\left(2\left(t_{i+1}^{2}+t_{i+1}t_{i}+t_{i}^{2}\right)-3\left(t_{i-1}+t_{i-2}\right)\left(t_{i+1}+t_{i}\right)+6t_{i-1}t_{i-2}\right)\label{eq:AB3_L0_7}\\
=\  & \frac{1}{6C_{0}}\Big(2\left(t_{i}-t_{i-2}\right)\left(t_{i+1}+t_{i}-2t_{i-1}\right)+2\left(t_{i+1}-t_{i-1}\right)\left(t_{i+1}-t_{i-2}\right)\nonumber \\
 & \ \ \ \ \ \ \ \ \ -\left(t_{i+1}-t_{i}\right)\left(t_{i-1}-t_{i-2}\right)\Big)\label{eq:AB3_L0_8}
\end{align}
where $C_{0}=\left(t_{i}-t_{i-1}\right)\left(t_{i}-t_{i-2}\right)$. 

We can quickly compute $L_{1}$ by replacing $t_{i-1}$ and $C_{0}$
in Eq.~\ref{eq:AB3_L0_7} with $t_{i}$ and $C_{1}=\left(t_{i-1}-t_{i}\right)\left(t_{i-1}-t_{i-2}\right)$,
respectively. This leads to the formula:
\begin{align}
L_{1}=\  & \frac{1}{6C_{1}}\left(2\left(t_{i+1}^{2}+t_{i+1}t_{i}+t_{i}^{2}\right)-3\left(t_{i}+t_{i-2}\right)\left(t_{i+1}+t_{i}\right)+6t_{i}t_{i-2}\right)\\
=\  & \frac{1}{6C_{1}}\left(2t_{i+1}^{2}-t_{i+1}t_{i}-t_{i}^{2}-3t_{i+1}t_{i-2}+3t_{i}t_{i-2}\right)\\
=\  & \frac{\left(t_{i+1}-t_{i}\right)\left(2t_{i+1}+t_{i}-3t_{i-2}\right)}{6C_{1}}\\
=\  & \frac{\left(t_{i+1}-t_{i}\right)\left(2t_{i+1}+t_{i}-3t_{i-2}\right)}{6\left(t_{i-1}-t_{i}\right)\left(t_{i-1}-t_{i-2}\right)}\label{eq:AB3_L1_4}
\end{align}

Similarly, to compute $L_{2}$, we replace $t_{i-2}$ and $C_{0}$
in Eq.~\ref{eq:AB3_L0_7} with $t_{i}$ and $C_{2}=\left(t_{i-2}-t_{i}\right)\left(t_{i-2}-t_{i-1}\right)$,
respectively. The resulting formula is given below:
\begin{align}
L_{2}=\  & \frac{1}{6C_{2}}\left(2\left(t_{i+1}^{2}+t_{i+1}t_{i}+t_{i}^{2}\right)-3\left(t_{i}+t_{i-1}\right)\left(t_{i+1}+t_{i}\right)+6t_{i}t_{i-1}\right)\\
=\  & \frac{1}{6C_{2}}\left(2t_{i+1}^{2}-t_{i+1}t_{i}-t_{i}^{2}-3t_{i+1}t_{i-1}+3t_{i}t_{i-1}\right)\\
=\  & \frac{\left(t_{i+1}-t_{i}\right)\left(2t_{i+1}+t_{i}-3t_{i-1}\right)}{6C_{2}}\label{eq:AB3_L2_3}\\
=\  & \frac{\left(t_{i+1}-t_{i}\right)\left(2t_{i+1}+t_{i}-3t_{i-1}\right)}{6\left(t_{i-2}-t_{i}\right)\left(t_{i-2}-t_{i-1}\right)}\label{eq:AB3_L2_4}
\end{align}

Consequently, the AB3 update on SC flows is:
\begin{equation}
\bar{x}_{t_{i+1}}=\bar{x}_{t_{i}}+\Delta\varphi_{t_{i+1}}\left(L_{0}\bar{v}_{t_{i}}^{*}+L_{1}\bar{v}_{t_{i-1}}^{*}+L_{2}\bar{v}_{t_{i-2}}^{*}\right)
\end{equation}
where $L_{0}$, $L_{1}$, $L_{2}$ are provided in Eqs.~\ref{eq:AB3_L0_8},
\ref{eq:AB3_L1_4}, \ref{eq:AB3_L2_4}, respectively

\paragraph{AM3}

$\bar{v}\left(t\right)$ of the AM3 method is expressed as:
\begin{equation}
\bar{v}\left(t\right)=\tilde{\ell}_{0}\left(t\right)\bar{v}_{t_{i+1}}^{*}+\tilde{\ell}_{1}\left(t\right)\bar{v}_{t_{i}}^{*}+\tilde{\ell}_{2}\left(t\right)\bar{v}_{t_{i-1}}^{*}
\end{equation}
where $\tilde{\ell}_{0}\left(t\right)$, $\tilde{\ell}_{1}\left(t\right)$,
$\tilde{\ell}_{2}\left(t\right)$ have the following expressions:
\begin{align}
\tilde{\ell}_{0}\left(t\right) & =\frac{\left(t-t_{i}\right)\left(t-t_{i-1}\right)}{\left(t_{i+1}-t_{i}\right)\left(t_{i+1}-t_{i-1}\right)}\\
\tilde{\ell}_{1}\left(t\right) & =\frac{\left(t-t_{i+1}\right)\left(t-t_{i-1}\right)}{\left(t_{i}-t_{i+1}\right)\left(t_{i}-t_{i-1}\right)}\\
\tilde{\ell}_{2}\left(t\right) & =\frac{\left(t-t_{i+1}\right)\left(t-t_{i}\right)}{\left(t_{i-1}-t_{i+1}\right)\left(t_{i-1}-t_{i}\right)}
\end{align}

$\tilde{L}_{0}$ can be computed by replacing $C_{2}$ in Eq.~\ref{eq:AB3_L2_3}
with $\tilde{C}_{0}=\left(t_{i+1}-t_{i}\right)\left(t_{i+1}-t_{i-1}\right)$.
This results in the formula:
\begin{align}
L_{0} & =\frac{\left(t_{i+1}-t_{i}\right)\left(2t_{i+1}+t_{i}-3t_{i-1}\right)}{6\left(t_{i+1}-t_{i}\right)\left(t_{i+1}-t_{i-1}\right)}\\
 & =\frac{2t_{i+1}+t_{i}-3t_{i-1}}{6\left(t_{i+1}-t_{i-1}\right)}\label{eq:AM3_L0_2}
\end{align}

To compute $\tilde{L}_{1}$, we replace $t_{i-2}$ and $C_{0}$ in
Eq.~\ref{eq:AB3_L0_7} with $t_{i+1}$ and $\tilde{C}_{1}=\left(t_{i}-t_{i+1}\right)\left(t_{i}-t_{i-1}\right)$,
respectively. The analytical expression is given below:
\begin{align}
\tilde{L}_{1}=\  & \frac{1}{6C_{1}}\left(2\left(t_{i+1}^{2}+t_{i+1}t_{i}+t_{i}^{2}\right)-3\left(t_{i+1}+t_{i-1}\right)\left(t_{i+1}+t_{i}\right)+6t_{i+1}t_{i-1}\right)\\
=\  & \frac{1}{6C_{1}}\left(-t_{i+1}^{2}-t_{i+1}t_{i}+2t_{i}^{2}+3t_{i+1}t_{i-1}-3t_{i}t_{i-1}\right)\\
=\  & \frac{1}{6C_{1}}\left(-(t_{i+1}-t_{i})\left(t_{i+1}+t_{i}\right)-t_{i}\left(t_{i+1}-t_{i}\right)+3t_{i-1}\left(t_{i+1}-t_{i}\right)\right)\\
=\  & \frac{\left(t_{i+1}-t_{i}\right)\left(3t_{i-1}-2t_{i}-t_{i+1}\right)}{6C_{1}}\\
=\  & \frac{\left(t_{i+1}-t_{i}\right)\left(3t_{i-1}-2t_{i}-t_{i+1}\right)}{6\left(t_{i}-t_{i+1}\right)\left(t_{i}-t_{i-1}\right)}\\
=\  & \frac{t_{i+1}+2t_{i}-3t_{i-1}}{6\left(t_{i}-t_{i-1}\right)}\label{eq:AM3_L1_5}
\end{align}

We compute $\tilde{L}_{2}$ by replace $t_{i-1}$, $t_{i-2}$, and
$C_{0}$ in Eq.~\ref{eq:AB3_L0_7} with $t_{i+1}$, $t_{i}$, and
$\tilde{C}_{2}=\left(t_{i-1}-t_{i+1}\right)\left(t_{i-1}-t_{i}\right)$
respectively, leading to the following equation:
\begin{align}
\tilde{L}_{2}=\  & \frac{1}{6C_{2}}\left(2\left(t_{i+1}^{2}+t_{i+1}t_{i}+t_{i}^{2}\right)-3\left(t_{i+1}+t_{i}\right)\left(t_{i+1}+t_{i}\right)+6t_{i+1}t_{i}\right)\\
=\  & \frac{1}{6C_{2}}\left(-t_{i+1}^{2}+2t_{i+1}t_{i}-t_{i}^{2}\right)\\
=\  & \frac{-\left(t_{i+1}-t_{i}\right)^{2}}{6C_{2}}\\
=\  & \frac{-\left(t_{i+1}-t_{i}\right)^{2}}{6\left(t_{i-1}-t_{i+1}\right)\left(t_{i-1}-t_{i}\right)}\label{eq:AM3_L2_4}\\
=\  & \frac{-\left(t_{i+1}-t_{i}\right)^{2}}{6\left(t_{i+1}-t_{i-1}\right)\left(t_{i}-t_{i-1}\right)}
\end{align}

Thus, the AM3 update on SC flows is:
\begin{equation}
\bar{x}_{t_{i+1}}=\bar{x}_{t_{i}}+\Delta\varphi_{t_{i+1}}\left(\tilde{L}_{0}\bar{v}_{t_{i+1}}^{*}+\tilde{L}_{1}\bar{v}_{t_{i}}^{*}+\tilde{L}_{2}\bar{v}_{t_{i-1}}^{*}\right)
\end{equation}
where $\tilde{L}_{0}$, $\tilde{L}_{1}$, $\tilde{L}_{2}$ are given
in Eqs.~\ref{eq:AM3_L0_2}, \ref{eq:AM3_L1_5}, \ref{eq:AM3_L2_4},
respectively.

\section{Analysis of Existing High-order Solvers for Diffusion Models\label{sec:Analysis-of-Existing-High-Order-Solvers}}

In this section, we aim to analyze current state-of-the-art (SOTA)
high-order solvers for diffusion models based on our framework and
notations. Our analysis will focus exclusively on DPM-Solver \cite{lu2022dpm},
as other solvers either have connections to or derive from this solver
\cite{lu2022dpmpp,zhang2023fast,zhao2023unipc}.

\subsection{DPM-Solver\label{subsec:DPM-Solver-Analysis}}

The main idea behind DPM-Solver and DEIS is to leverage the semi-linear
structure in the probability flow ODEs of diffusion models to reduce
the approximation error. Here, we consider the posterior flow ODE
of a general stochastic process $X_{t}=a_{t}X_{0}+\sigma_{t}X_{1}$,
represented as follows:
\begin{align}
\dot{x}_{t} & =v^{*}\left(x_{t},t\right)\\
 & =\frac{\dot{a}_{t}}{a_{t}}x_{t}+\sigma_{t}\left(\frac{\dot{\sigma}_{t}}{\sigma_{t}}-\frac{\dot{a}_{t}}{a_{t}}\right)x_{1|t}\\
 & =f\left(t\right)x_{t}+g\left(t\right)x_{1,\theta}\left(x_{t},t\right)\label{eq:semi_linear_3}
\end{align}
where $f\left(t\right):=\frac{\dot{a}_{t}}{a_{t}}$, $g\left(t\right):=\sigma_{t}\left(\frac{\dot{\sigma}_{t}}{\sigma_{t}}-\frac{\dot{a}_{t}}{a_{t}}\right)$.

Eq.~\ref{eq:semi_linear_3} is a \emph{semi-linear} ODE because the
velocity $v^{*}\left(x_{t},t\right)$ is the sum of a linear term
$f\left(t\right)x_{t}$ and a non-linear term $g\left(t\right)x_{1,\theta}\left(x_{t},t\right)$.
The exact solution for this type of ODEs is:
\begin{equation}
x_{t'}=e^{\int_{t}^{t'}f\left(\tau\right)d\tau}x_{t}+\int_{t}^{t'}\left(e^{\int_{\tau}^{t'}f\left(r\right)dr}g\left(\tau\right)x_{1,\theta}\left(x_{\tau},\tau\right)\right)d\tau
\end{equation}
Since $f\left(t\right)=\frac{\dot{a}_{t}}{a_{t}}=\frac{d\log a_{t}}{dt}$,
we have $e^{\int_{t}^{t'}f\left(\tau\right)d\tau}=e^{\log a_{t'}-\log a_{t}}=\frac{a_{t'}}{a_{t}}$
and $e^{\int_{\tau}^{t'}f\left(r\right)dr}=\frac{a_{t'}}{a_{\tau}}$.
The above equation can be rewritten as:
\begin{equation}
x_{t'}=\frac{a_{t'}}{a_{t}}x_{t}+a_{t'}\int_{t}^{t'}\frac{g\left(\tau\right)}{a_{\tau}}x_{1,\theta}\left(x_{\tau},\tau\right)d\tau\label{eq:semi-linear-SC}
\end{equation}
Interestingly, we can represent $\frac{g\left(\tau\right)}{a_{\tau}}$
as follows:
\begin{align}
\frac{g\left(\tau\right)}{a_{\tau}} & =\frac{\sigma_{\tau}}{a_{\tau}}\left(\frac{\dot{\sigma}_{\tau}}{\sigma_{\tau}}-\frac{\dot{a}_{\tau}}{a_{\tau}}\right)\\
 & =\frac{\dot{\sigma}_{\tau}}{a_{\tau}}-\frac{\dot{a}_{\tau}\sigma_{\tau}}{a_{\tau}^{2}}\\
 & =\frac{a_{\tau}\dot{\sigma}_{\tau}-\dot{a}_{\tau}\sigma_{\tau}}{a_{\tau}^{2}}\\
 & =\dot{\varphi}_{\tau}
\end{align}
where $\varphi_{t}=\frac{\sigma_{t}}{a_{t}}$.

Therefore, Eq.~\ref{eq:semi-linear-SC} is equivalent to:
\begin{align}
x_{t'} & =\frac{a_{t'}}{a_{t}}x_{t}+a_{t'}\int_{t}^{t'}\dot{\varphi}_{\tau}x_{1,\theta}\left(x_{\tau},\tau\right)d\tau\\
\Leftrightarrow\frac{x_{t'}}{a_{t'}} & =\frac{x_{t}}{a_{t}}+\int_{t}^{t'}\dot{\varphi}_{\tau}x_{1,\theta}\left(x_{\tau},\tau\right)d\tau\label{eq:semi-linear-SC-COV-2}\\
\Leftrightarrow\bar{x}_{t'} & =\bar{x}_{t}+\int_{\varphi_{t}}^{\varphi_{t'}}\hat{x}_{1,\theta}\left(x_{\varphi_{\tau}},\varphi_{\tau}\right)d\varphi_{\tau}\label{eq:semi-linear-SC-COV-3}
\end{align}
where $\bar{x}_{t}=\frac{x_{t}}{a_{t}}$ and $\bar{x}_{t'}=\frac{x_{t'}}{a_{t'}}$.
In Eq.~\ref{eq:semi-linear-SC-COV-3}, we change the time variable
from $\tau$ to $\varphi_{\tau}$.

Eq.~\ref{eq:semi-linear-SC-COV-3} is exactly an update on the SC
flow obtained by applying the ``variable scaling'' transformation
$\tilde{X}_{t}=\frac{X_{t}}{a_{t}}$ followed by the ``time adjustment''
transformation $d\bar{X}_{t}=X_{1}d\varphi_{t}$ (Table~\ref{tab:straight_transform_summary_x1scale}).
The term $\int_{\varphi_{t}}^{\varphi_{t'}}\hat{x}_{1,\theta}\left(x_{\varphi_{\tau}},\varphi_{\tau}\right)d\varphi_{\tau}$
can be approximated using different numerical methods such as Taylor
expansion, Runge-Kutta, or linear multi-step methods. 

The authors of DPM-Solver, however, use a \emph{different} transformation
of $t$ inspired by the work \cite{kingma2021variational}. They set
$\lambda_{t}:=\log\frac{a_{t}}{\sigma_{t}}$, which leads to a different
expression of $\frac{g\left(\tau\right)}{a_{\tau}}$ given below:
\begin{equation}
\frac{d\lambda_{t}}{dt}=\frac{\sigma_{t}}{a_{t}}\frac{\dot{a}_{t}\sigma_{t}-a_{t}\dot{\sigma}_{t}}{\sigma_{t}^{2}}=\frac{\dot{a}_{t}}{a_{t}}-\frac{\dot{\sigma}_{t}}{\sigma_{t}}
\end{equation}

\begin{equation}
\frac{g\left(\tau\right)}{a_{\tau}}=\frac{\dot{\sigma}_{\tau}}{a_{\tau}}-\frac{\dot{a}_{\tau}\sigma_{\tau}}{a_{\tau}^{2}}=-\frac{d\lambda_{\tau}}{d\tau}\frac{\sigma_{\tau}}{a_{\tau}}
\end{equation}
Consequently, in their work, Eq.~\ref{eq:semi-linear-SC} is reformulated
as follows:
\begin{align}
x_{t'} & =\frac{a_{t'}}{a_{t}}x_{t}-a_{t'}\int_{t}^{t'}\frac{d\lambda_{\tau}}{d\tau}\frac{\sigma_{\tau}}{a_{\tau}}x_{1,\theta}\left(x_{\tau},\tau\right)d\tau\label{eq:semi-linear-DPMSolver-1}\\
 & =\frac{a_{t'}}{a_{t}}x_{t}-a_{t'}\int_{\lambda_{t}}^{\lambda_{t'}}e^{-\lambda}\hat{x}_{1,\theta}\left(x_{\lambda},\lambda\right)d\lambda\label{eq:semi-linear-DPMSolver-2}
\end{align}
where $\hat{x}_{1,\theta}\left(x_{\lambda},\lambda\right)$ is equivalent
to $x_{1,\theta}\left(x_{\tau},\tau\right)$ but with the time variable
changed from $\tau$ to $\lambda$. The integral in Eq.~\ref{eq:semi-linear-DPMSolver-2}
is similar to the counterpart in Eq.~\ref{eq:semi-linear-SC-COV-3}
except for the presence of the coefficient $e^{-\lambda}$ due to
using a different change of variables. However, this coefficient does
not cause any trouble when approximating the integral using numerical
methods.

To simplify our presentation, we rewrite Eq.~\ref{eq:semi-linear-DPMSolver-2}
with $t$ and $t'$ replaced with $t_{i}$ and $t_{i+1}$, respectively:
\begin{equation}
x_{t_{i+1}}=\frac{a_{t_{i+1}}}{a_{t_{i}}}x_{t_{i}}-a_{t_{i+1}}\int_{\lambda_{t_{i}}}^{\lambda_{t_{i+1}}}e^{-\lambda}\hat{x}_{1,\theta}\left(x_{\lambda},\lambda\right)d\lambda\label{eq:semi-linear-DPMSolver-3}
\end{equation}

In the DPM-Solver paper, the authors approximate $\hat{x}_{1,\theta}\left(x_{\lambda},\lambda\right)$
using the $\left(k-1\right)$-th order Taylor expansion as follows:
\begin{align}
\hat{x}_{1,\theta}\left(x_{\lambda},\lambda\right)=\  & \sum_{j=0}^{k-1}\frac{\left(\lambda-\lambda_{t_{i}}\right)^{j}}{j!}\hat{x}_{1,\theta}^{\left(j\right)}\left(x_{\lambda_{t_{i}}},\lambda_{t_{i}}\right)\nonumber \\
 & +\mathcal{O}\left(\left(\lambda-\lambda_{t_{i}}\right)^{k}\right)
\end{align}
where $\hat{x}_{1,\theta}^{\left(j\right)}\left(x_{\lambda_{t_{i-1}}},\lambda_{t_{i-1}}\right)$
denotes the $j$-th derivative of $\hat{x}_{1,\theta}\left(x_{\lambda},\lambda\right)$
at $\lambda=\lambda_{t_{i-1}}$. Substituting this expression of $\hat{x}_{1,\theta}\left(x_{\lambda},\lambda\right)$
into Eq.~\ref{eq:semi-linear-DPMSolver-3}, we have:
\begin{align}
x_{t_{i+1}}=\  & \frac{a_{t_{i+1}}}{a_{t_{i}}}x_{t_{i}}-a_{t_{i+1}}\sum_{j=0}^{k-1}\ell_{j}\left(\lambda\right)\hat{x}_{1,\theta}^{\left(j\right)}\left(x_{\lambda_{t_{i-1}}},\lambda_{t_{i-1}}\right)\nonumber \\
 & +\mathcal{O}\left(\left(\lambda_{t_{i+1}}-\lambda_{t_{i}}\right)^{k+1}\right)
\end{align}
where $\ell_{j}\left(\lambda\right)=\int_{\lambda_{t_{i}}}^{\lambda_{t_{i+1}}}e^{-\lambda}\frac{\left(\lambda-\lambda_{t_{i}}\right)^{j}}{j!}$
which can be expressed in closed-form \cite{lu2022dpm}. To approximate
$\hat{x}_{1,\theta}^{\left(j\right)}\left(x_{\lambda_{t_{i-1}}},\lambda_{t_{i-1}}\right)$,
the authors leverage Runge-Kutta methods with the time $\lambda$.
For example, the DPM-Solver-2 update (Algorithm 1 in \cite{lu2022dpm})
is given below:
\begin{align}
\tau_{2} & =t_{\lambda}\left(\frac{\lambda_{t_{i+1}}+\lambda_{t_{i}}}{2}\right)\label{eq:DPMSolver-2S-update-1}\\
x_{\tau_{2}} & =\frac{a_{\tau_{2}}}{a_{t_{i}}}x_{t_{i}}-\sigma_{\tau_{2}}\left(e^{\frac{\Delta\lambda_{t_{i+1}}}{2}}-1\right)\hat{x}_{1,\theta}\left(x_{t_{i}},t_{i}\right)\label{eq:DPMSolver-2S-update-2}\\
x_{t_{i+1}} & =\frac{a_{t_{i+1}}}{a_{\tau_{2}}}x_{t_{i}}-\sigma_{t_{i+1}}\left(e^{\Delta\lambda_{t_{i+1}}}-1\right)\hat{x}_{1,\theta}\left(x_{\tau_{2}},\tau_{2}\right)\label{eq:DPMSolver-2S-update-3}
\end{align}
This is indeed the \emph{explicit midpoint method} in which the intermediate
time step is computed in the $\lambda$-time space and then mapped
back to the original time space using the time inverse function $t_{\lambda}\left(\cdot\right)$
(Eq.~\ref{eq:DPMSolver-2S-update-1}). This update can be view as
a special case of the general update in Eqs.~\ref{eq:RK_SC_varphi_final}-\ref{eq:RK_SC_varphi_fk}.

\section{Experimental Settings}

\subsection{2D Toy Dataset\label{subsec:2D-Toy-Dataset-Settings}}

In this dataset, $p_{0}$ is a mixture of three 2D Gaussians with
the mean vectors and covariance matrices given below:
\begin{align*}
\mu_{0} & =\begin{bmatrix}20\\
20
\end{bmatrix},\ \begin{bmatrix}25\\
10
\end{bmatrix},\ \begin{bmatrix}10\\
26
\end{bmatrix}\\
\Sigma_{0} & =\begin{bmatrix}0.6^{2} & 0.7^{2}\\
0.7^{2} & 1.4^{2}
\end{bmatrix},\ \begin{bmatrix}1.3^{2} & -0.9^{2}\\
-0.9^{2} & 1.0^{2}
\end{bmatrix},\ \begin{bmatrix}1.2^{2} & 0.0\\
0.0 & 1.2^{2}
\end{bmatrix}
\end{align*}

$p_{1}$ is a mixture of two 2D Gaussian with the mean vectors and
covariance matrices given below:

\begin{align*}
\mu_{1} & =\begin{bmatrix}5\\
-5
\end{bmatrix},\ \begin{bmatrix}-5\\
3
\end{bmatrix}\\
\Sigma_{1} & =\begin{bmatrix}1.0^{2} & 0.0\\
0.0 & 1.0^{2}
\end{bmatrix},\ \begin{bmatrix}1.0^{2} & 0.0\\
0.0^{2} & 1.0^{2}
\end{bmatrix}
\end{align*}

We employ a simple three-layer MLP, similar to the one in \cite{liu2022flow},
to model either the posterior velocity $v^{*}\left(x_{t},t\right)$
or the predicted latent sample $x_{1|t}$ of each process. The network
architecture is outlined in Table~\ref{tab:velocity_network_architecture}.
During training, we minimize the velocity matching loss if the network
models $v^{*}\left(x_{t},t\right)$, and the noise matching loss if
it models $x_{1|t}$. Training is conducted on 30,000 samples drawn
from $p_{0}$ and $p_{1}$ for 20000 iterations, utilizing a batch
size of 2048. We employ the Adam optimizer with a learning rate of
0.003.

\begin{table}
\begin{centering}
\begin{tabular}{c}
Concat($\left[x_{t},t\right]$)\tabularnewline
Linear($d_{x}$ + 1, 100, bias=True)\tabularnewline
Tanh()\tabularnewline
Linear(100, 100, bias=True)\tabularnewline
Tanh()\tabularnewline
Linear(100, $d_{x}$, bias=True)\tabularnewline
\end{tabular}
\par\end{centering}
\caption{Architecture of the velocity network for experiments on the 2D toy
dataset.\label{tab:velocity_network_architecture}}
\end{table}

\begin{figure*}
\begin{centering}
{\resizebox{\textwidth}{!}{%
\par\end{centering}
\begin{centering}
\begin{tabular}{cc|cc|cc}
\includegraphics[width=0.18\textwidth]{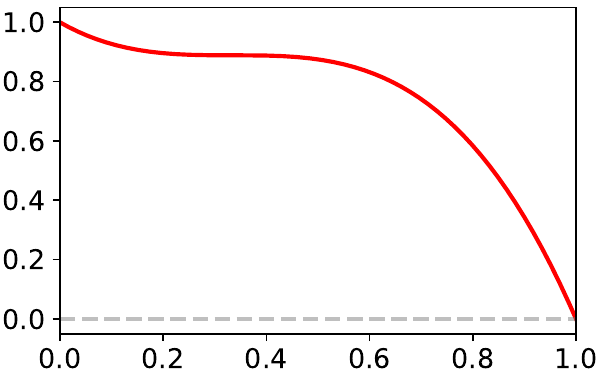} & \includegraphics[width=0.18\textwidth]{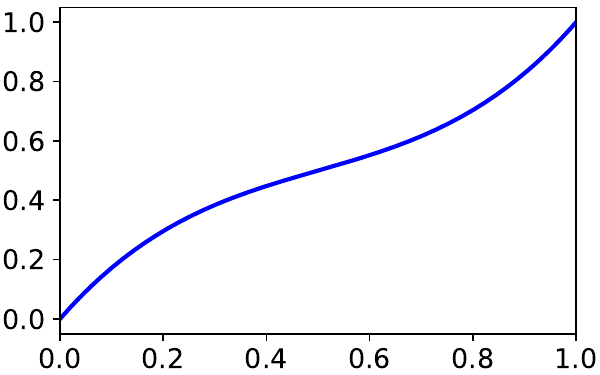} & \includegraphics[width=0.18\textwidth]{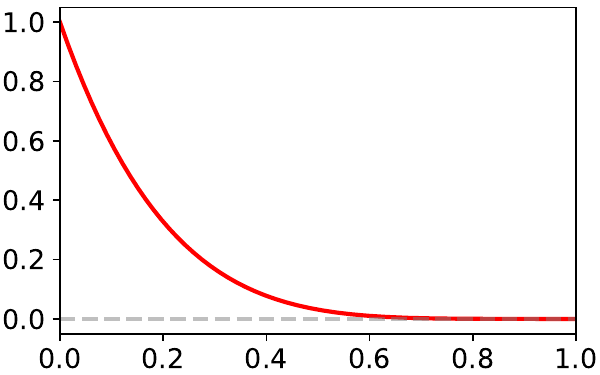} & \includegraphics[width=0.18\textwidth]{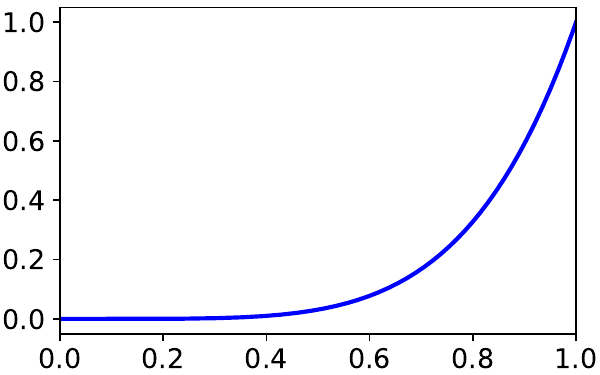} & \includegraphics[width=0.18\textwidth]{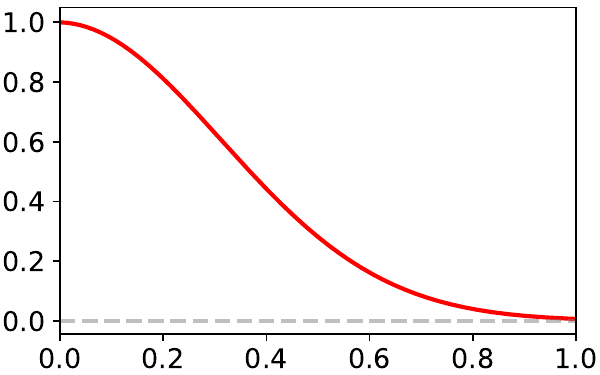} & \includegraphics[width=0.18\textwidth]{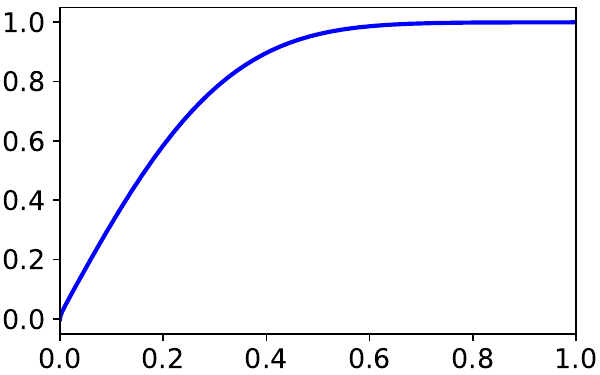}\tabularnewline
$a_{t}$ & $\sigma_{t}$ & $a_{t}$ & $\sigma_{t}$ & $a_{t}$ & $\sigma_{t}$\tabularnewline
\includegraphics[width=0.18\textwidth]{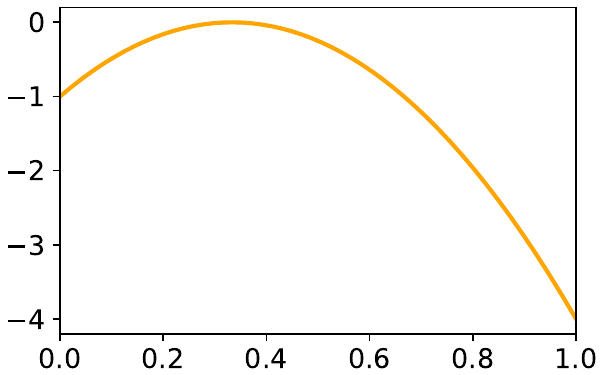} & \includegraphics[width=0.18\textwidth]{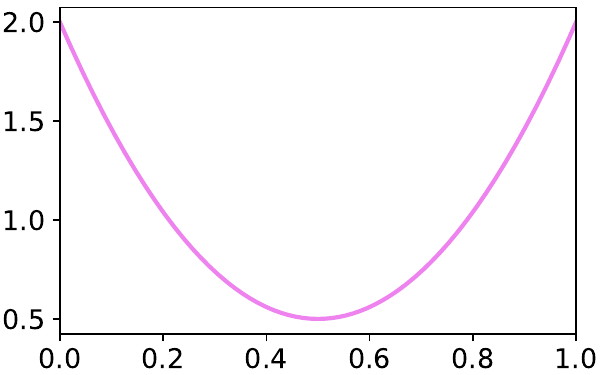} & \includegraphics[width=0.18\textwidth]{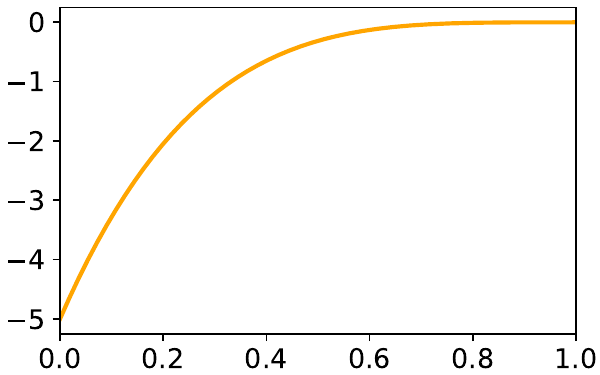} & \includegraphics[width=0.18\textwidth]{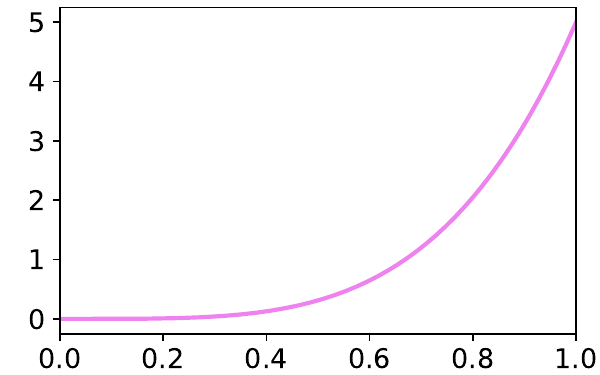} & \includegraphics[width=0.18\textwidth]{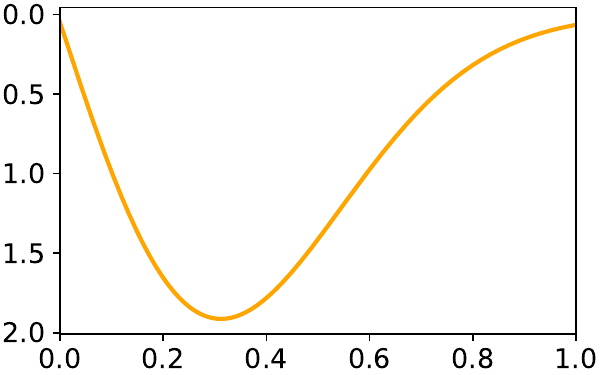} & \includegraphics[width=0.18\textwidth]{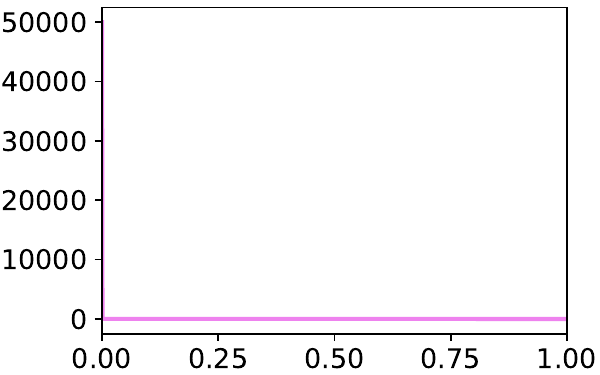}\tabularnewline
$\dot{a}_{t}$ & $\dot{\sigma}_{t}$ & $\dot{a}_{t}$ & $\dot{\sigma}_{t}$ & $\dot{a}_{t}$ & $\dot{\sigma}_{t}$\tabularnewline
\includegraphics[width=0.18\textwidth]{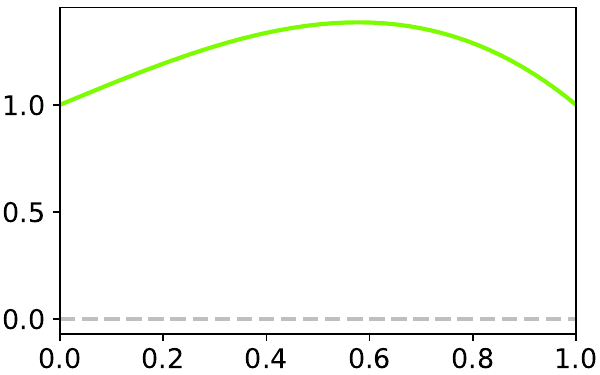} & \includegraphics[width=0.18\textwidth]{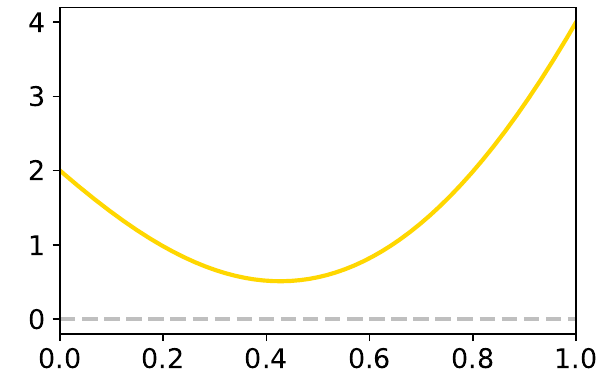} & \includegraphics[width=0.18\textwidth]{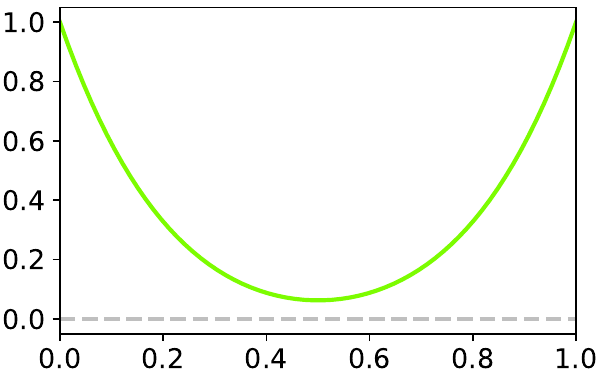} & \includegraphics[width=0.18\textwidth]{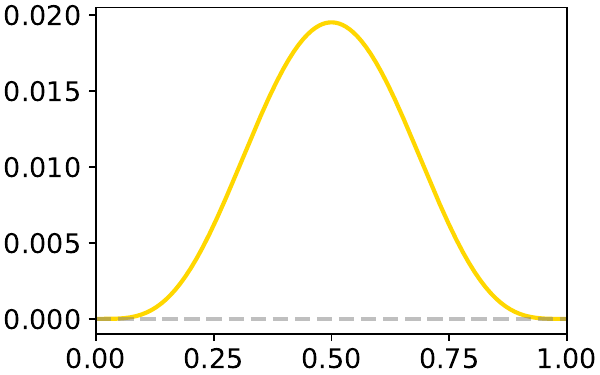} & \includegraphics[width=0.18\textwidth]{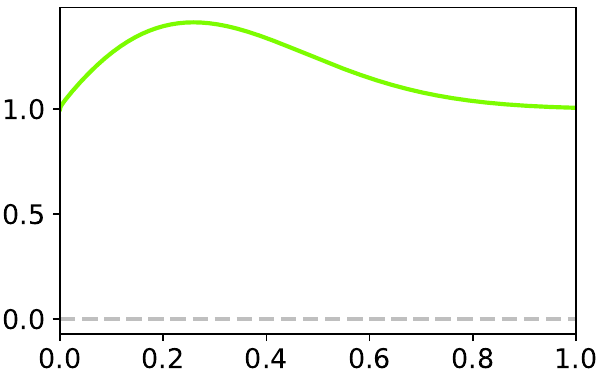} & \includegraphics[width=0.18\textwidth]{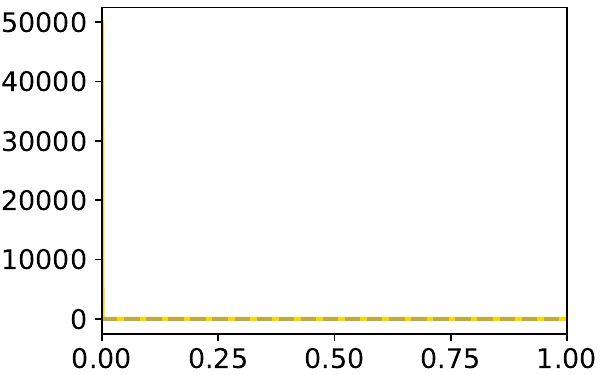}\tabularnewline
$a_{t}+\sigma_{t}$ & $a_{t}\dot{\sigma}_{t}-\dot{a}_{t}\sigma_{t}$ & $a_{t}+\sigma_{t}$ & $a_{t}\dot{\sigma}_{t}-\dot{a}_{t}\sigma_{t}$ & $a_{t}+\sigma_{t}$ & $a_{t}\dot{\sigma}_{t}-\dot{a}_{t}\sigma_{t}$\tabularnewline
\multicolumn{2}{c}{third-degree} & \multicolumn{2}{c}{fifth-degree} & \multicolumn{2}{c}{VP-SDE-like}\tabularnewline
\end{tabular}}}
\par\end{centering}
\caption{Coefficients associated with the \emph{third-degree} (left), \emph{fifth-degree}
(middle), and \emph{VP-SDE-like} (right) processes.\label{fig:Toy_Coefficients}}
\end{figure*}

In Fig.~\ref{fig:Toy_Coefficients}, we plot the coefficients associated
with the third-degree, fifth-degree, and VP-SDE-like processes as
functions of time. Besides $a_{1}$ and $\sigma_{0}$ equal 0 for
all processes, $a_{t}\dot{\sigma}_{t}-\dot{a}_{t}\sigma_{t}$ is 0
at $t=0$ and $t=1$ for the fifth-degree process.

\section{Additional Experimental Results\label{sec:Additional-Experimental-Results}}

\subsection{2D Toy Dataset}

\subsubsection{Empirical analysis on the numerical robustness of our method}

\begin{figure*}
\begin{centering}
{\resizebox{\textwidth}{!}{%
\par\end{centering}
\begin{centering}
\begin{tabular}{c|ccccccccc}
\multicolumn{1}{c}{model} & SC type & $\epsilon$ & $N$ & samples & $x_{t}$ & $\bar{x}_{t}$ & $\left(\varphi_{t'}-\varphi_{t}\right)v_{\theta}\left(\bar{x}_{t},t\right)$ & $v_{\theta}\left(\bar{x}_{t},t\right)$ & $\varphi_{t'}-\varphi_{t}$\tabularnewline
\multirow{2}{*}{$v_{\theta}\left(x_{t},t\right)$} & \multirow{1}{*}[0.075\textwidth]{Eq.~\ref{eq:straight_constant_speed_stochastic_1}} & \multirow{1}{*}[0.075\textwidth]{1e-6} & \multirow{1}{*}[0.075\textwidth]{20} & \includegraphics[height=0.15\textwidth]{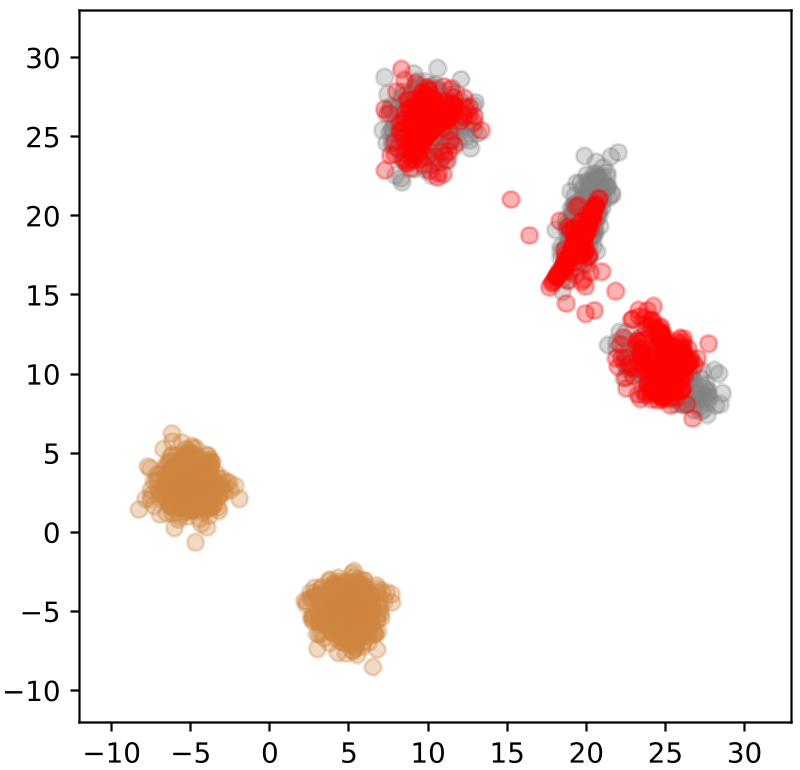} & \includegraphics[height=0.15\textwidth]{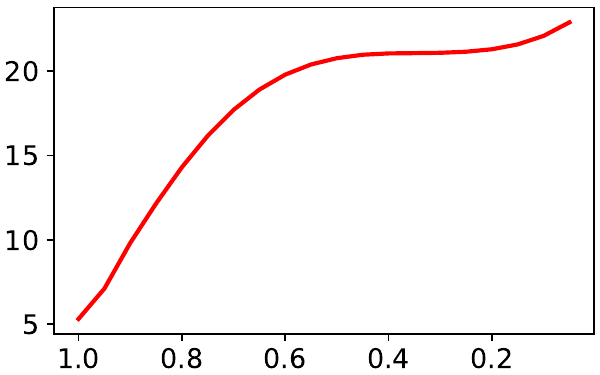} & \includegraphics[height=0.15\textwidth]{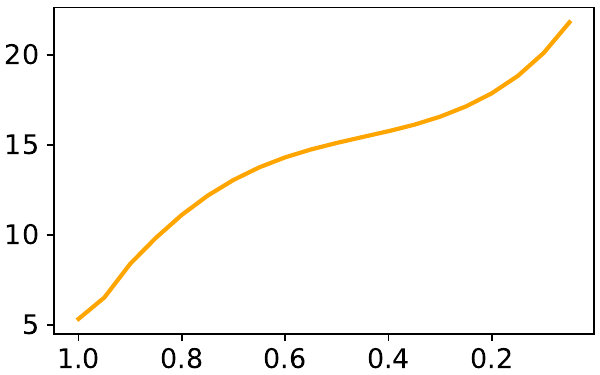} & \includegraphics[height=0.15\textwidth]{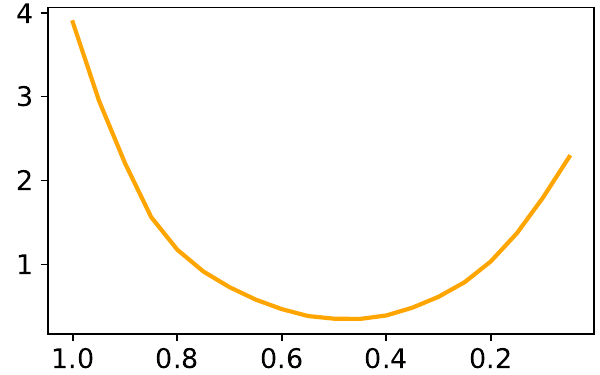} & \includegraphics[height=0.15\textwidth]{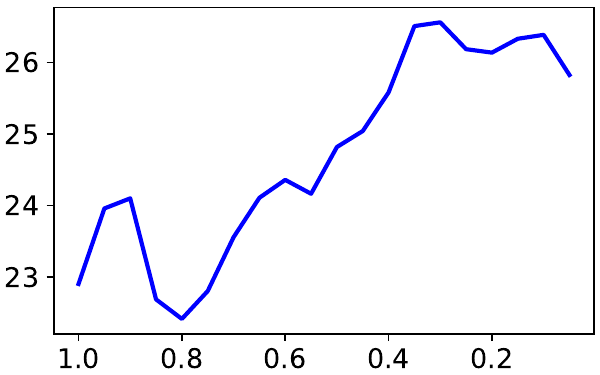} & \includegraphics[height=0.15\textwidth]{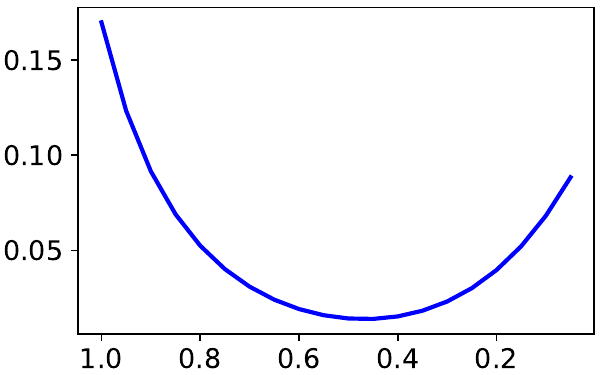}\tabularnewline
 & \multirow{1}{*}[0.075\textwidth]{Eq.~\ref{eq:straight_constant_speed_stochastic_2}} & \multirow{1}{*}[0.075\textwidth]{1e-6} & \multirow{1}{*}[0.075\textwidth]{20} & \includegraphics[height=0.15\textwidth]{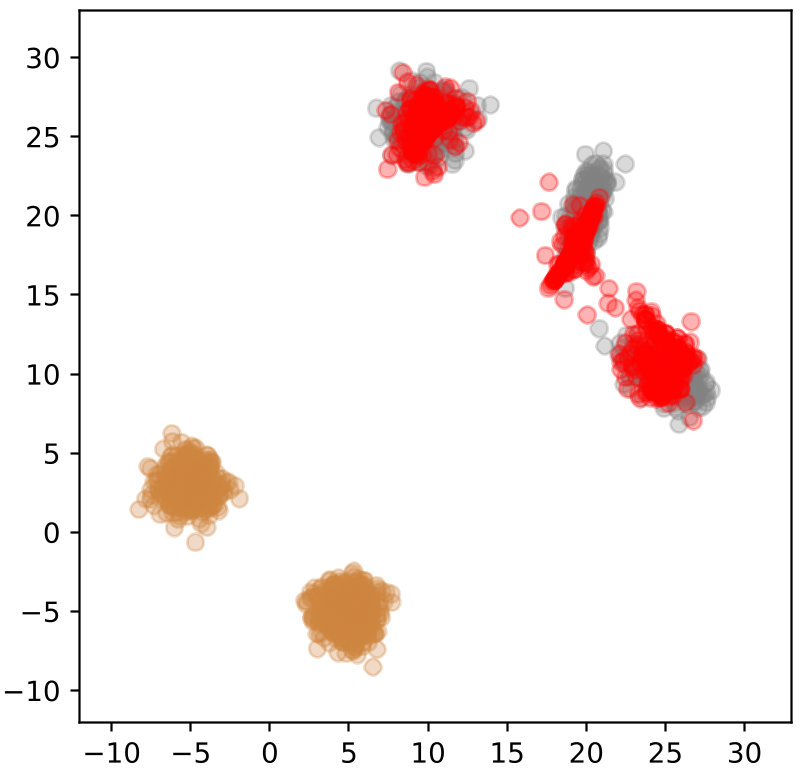} & \includegraphics[height=0.15\textwidth]{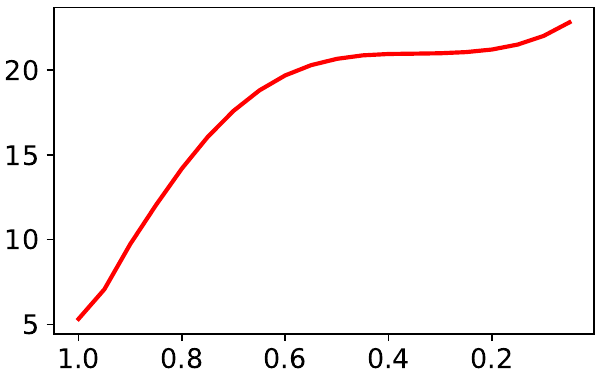} & \includegraphics[height=0.15\textwidth]{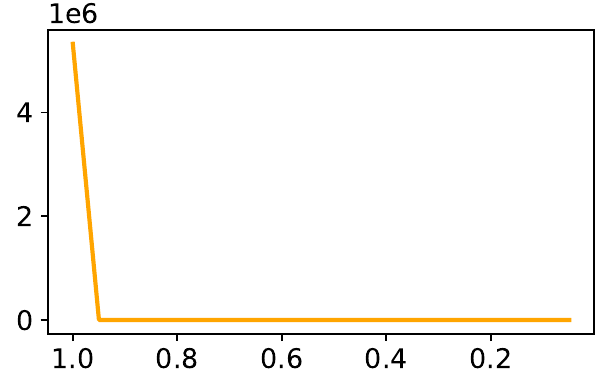} & \includegraphics[height=0.15\textwidth]{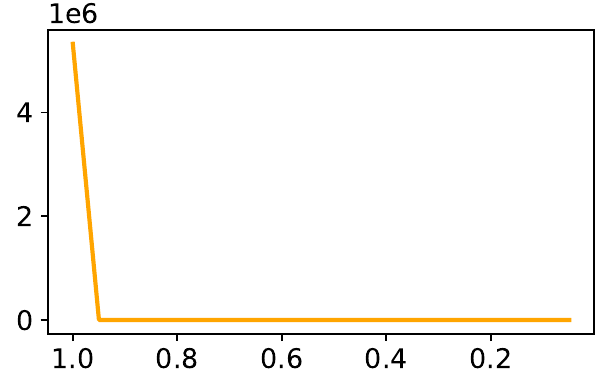} & \includegraphics[height=0.15\textwidth]{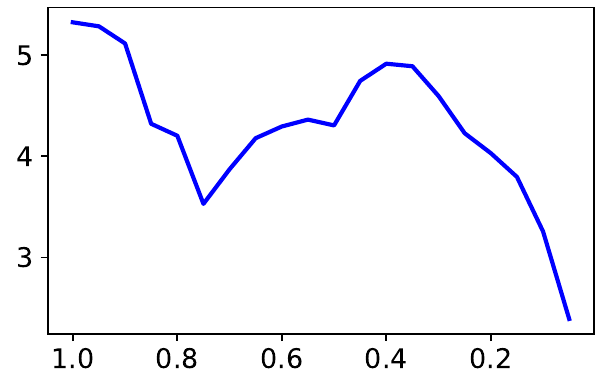} & \includegraphics[height=0.15\textwidth]{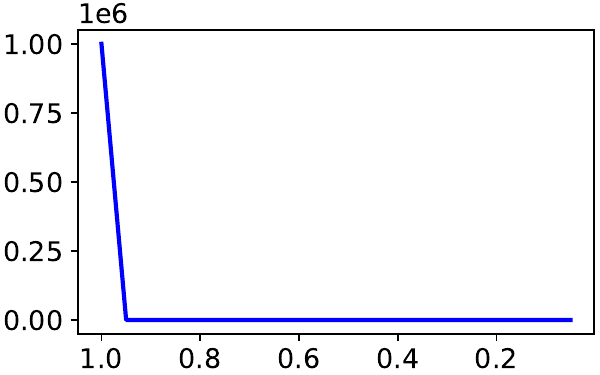}\tabularnewline
\multicolumn{1}{c}{} &  &  &  &  &  &  &  &  & \tabularnewline
\multirow{4}{*}[-0.12\textwidth]{$x_{1,\theta}\left(x_{t},t\right)$} & \multirow{1}{*}[0.075\textwidth]{Eq.~\ref{eq:straight_constant_speed_stochastic_1}} & \multirow{1}{*}[0.075\textwidth]{1e-3} & \multirow{1}{*}[0.075\textwidth]{20} & \includegraphics[height=0.15\textwidth]{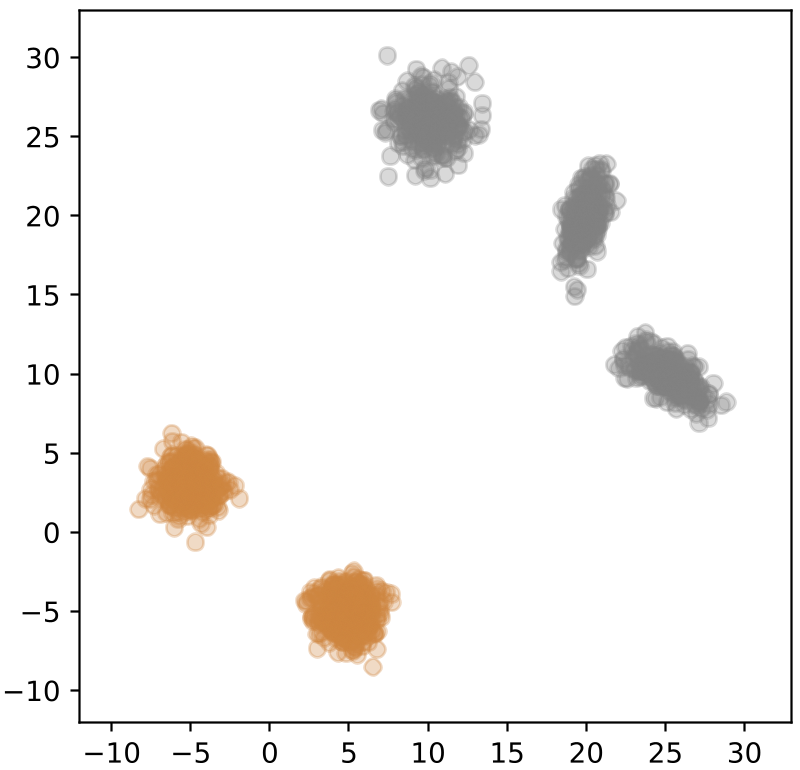} & \includegraphics[height=0.15\textwidth]{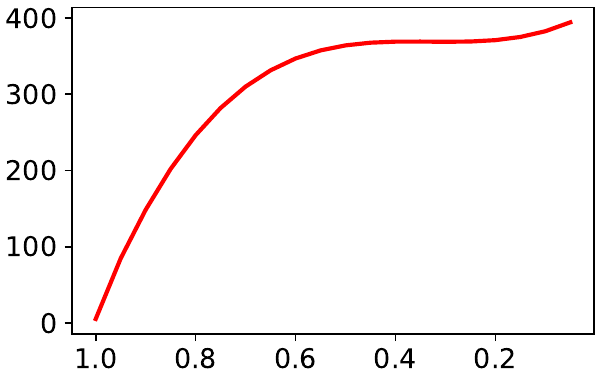} & \includegraphics[height=0.15\textwidth]{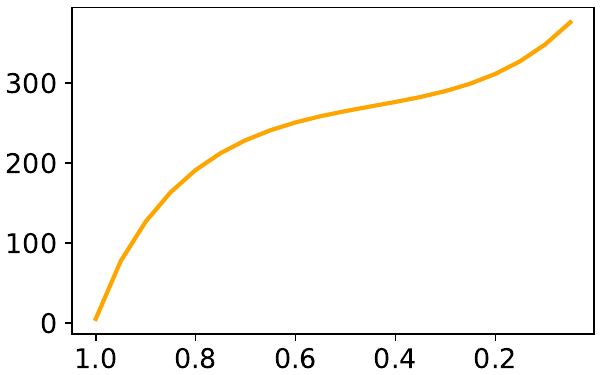} & \includegraphics[height=0.15\textwidth]{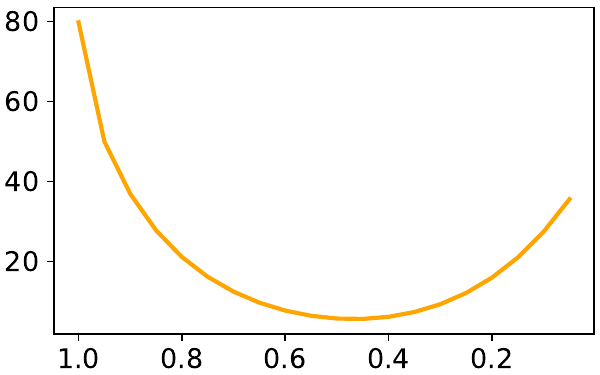} & \includegraphics[height=0.15\textwidth]{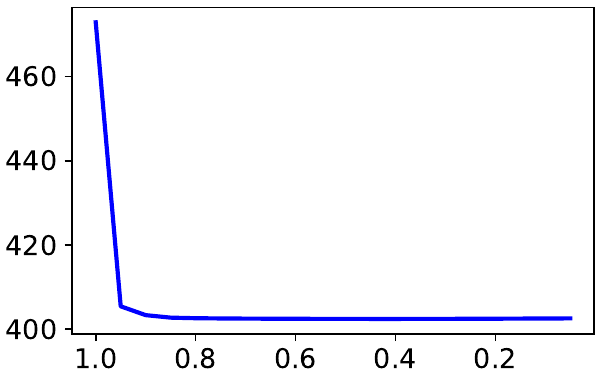} & \includegraphics[height=0.15\textwidth]{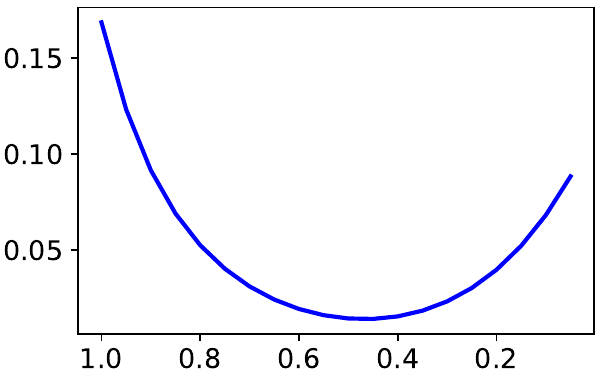}\tabularnewline
 & \multicolumn{1}{c|}{} & \multirow{1}{*}[0.075\textwidth]{1e-3} & \multirow{1}{*}[0.075\textwidth]{20} & \includegraphics[height=0.15\textwidth]{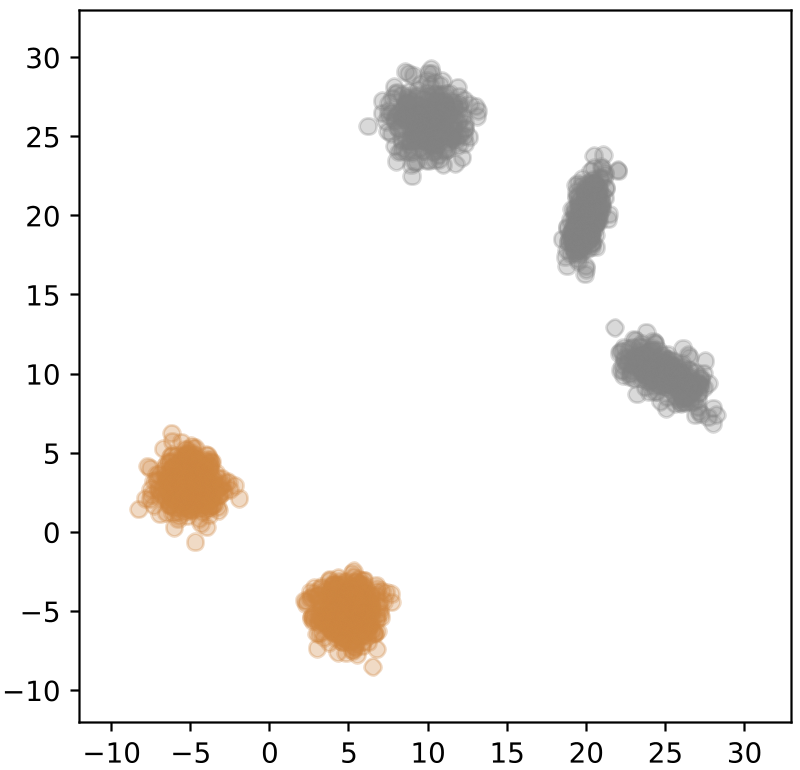} & \includegraphics[height=0.15\textwidth]{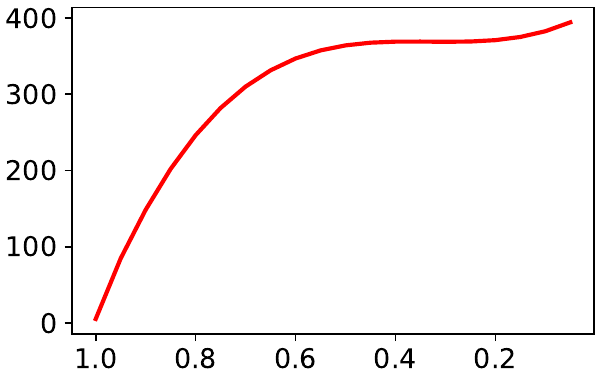} & \includegraphics[height=0.15\textwidth]{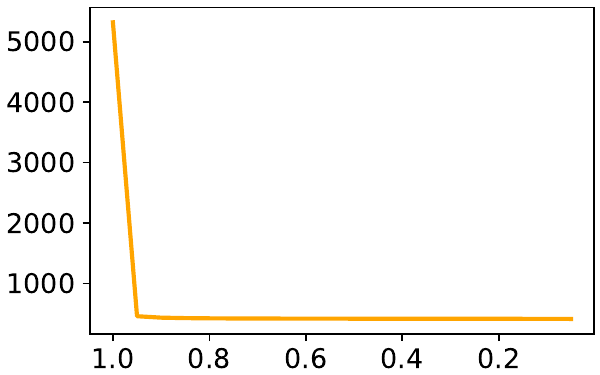} & \includegraphics[height=0.15\textwidth]{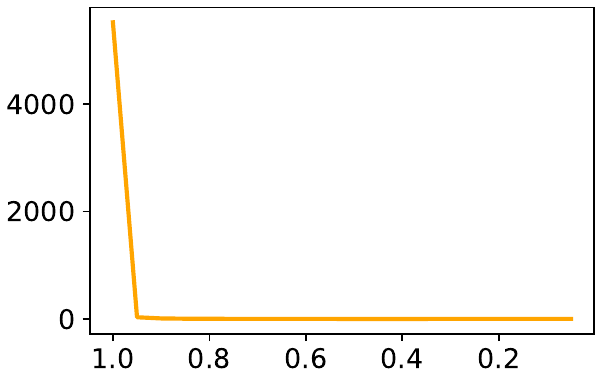} & \includegraphics[height=0.15\textwidth]{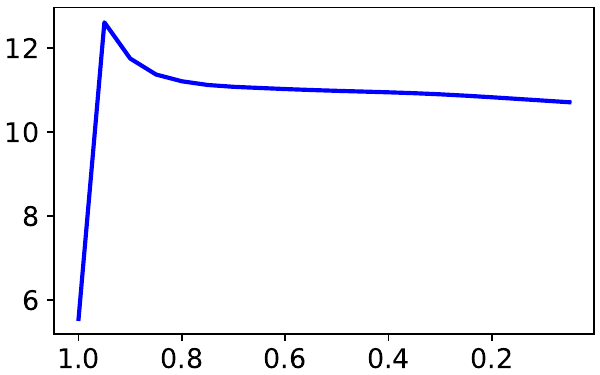} & \includegraphics[height=0.15\textwidth]{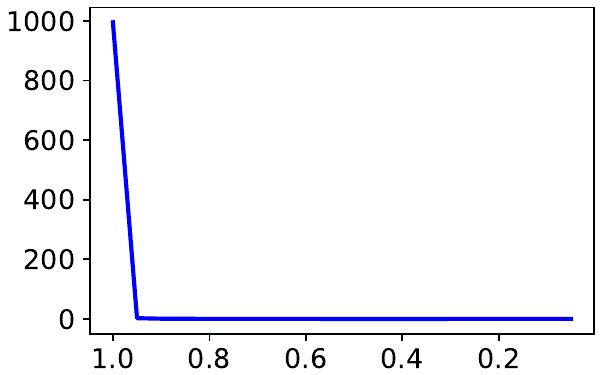}\tabularnewline
 & \multicolumn{1}{c|}{\multirow{1}{*}[0.075\textwidth]{Eq.~\ref{eq:straight_constant_speed_stochastic_2}}} & \multirow{1}{*}[0.075\textwidth]{1e-3} & \multirow{1}{*}[0.075\textwidth]{1000} & \includegraphics[height=0.15\textwidth]{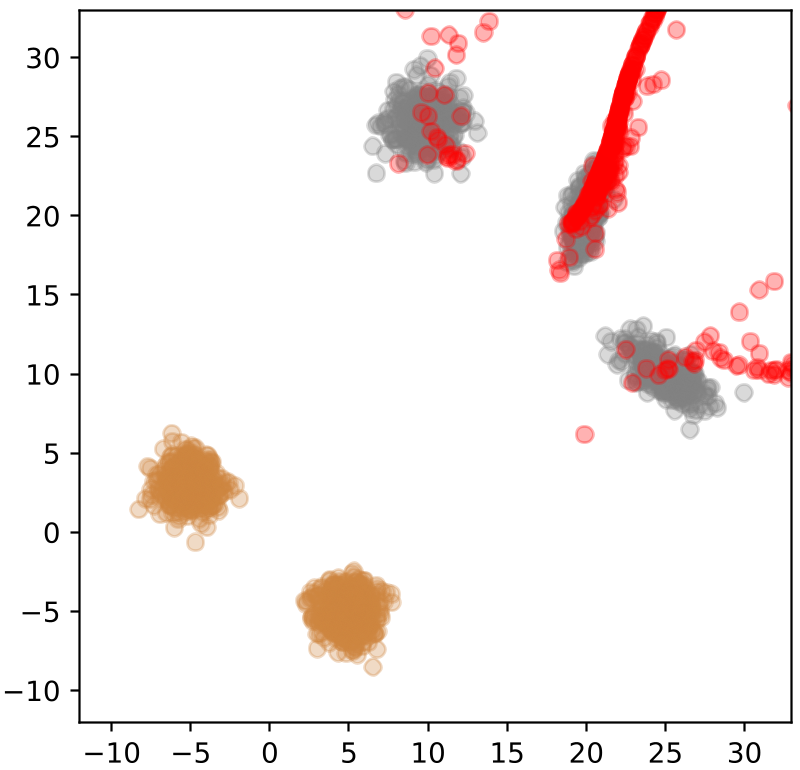} & \includegraphics[height=0.15\textwidth]{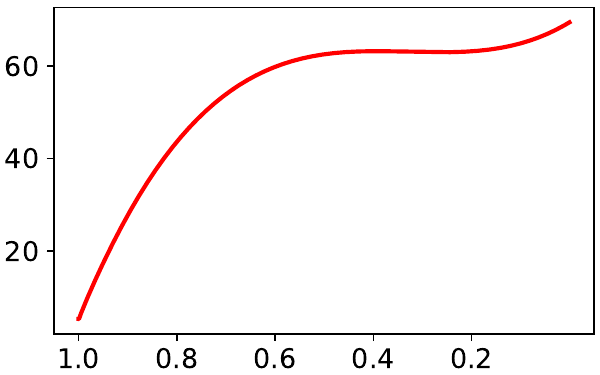} & \includegraphics[height=0.15\textwidth]{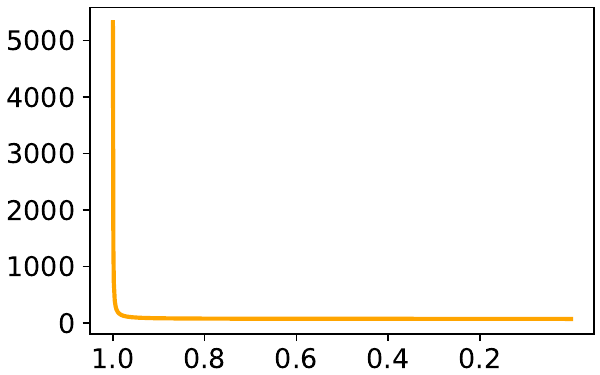} & \includegraphics[height=0.15\textwidth]{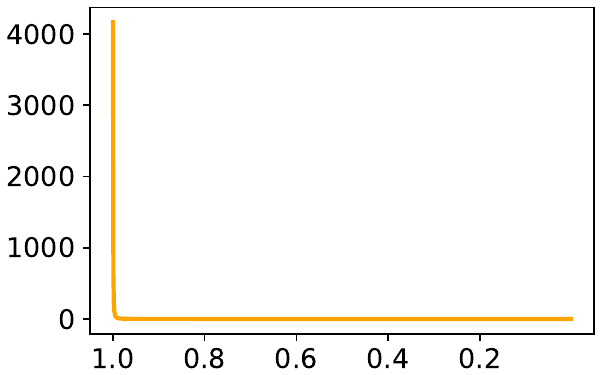} & \includegraphics[height=0.15\textwidth]{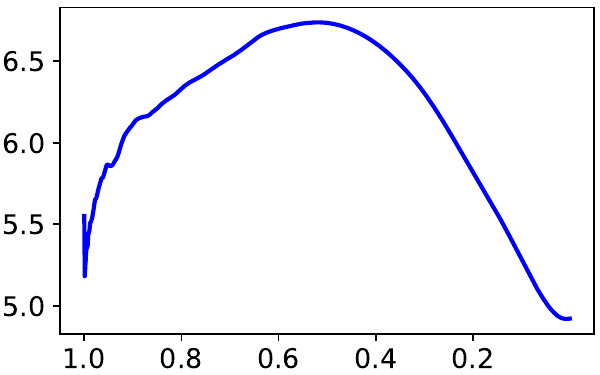} & \includegraphics[height=0.15\textwidth]{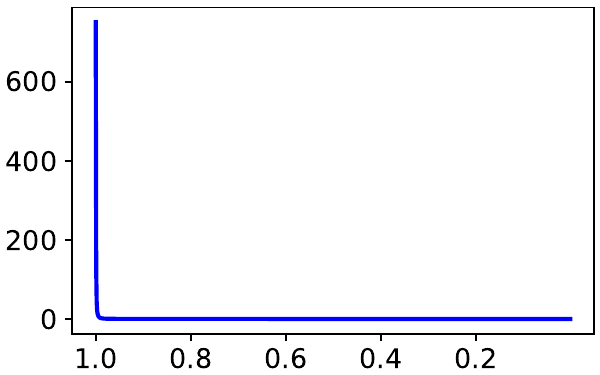}\tabularnewline
 & \multicolumn{1}{c|}{} & \multirow{1}{*}[0.075\textwidth]{1e-6} & \multirow{1}{*}[0.075\textwidth]{1000} & \includegraphics[height=0.15\textwidth]{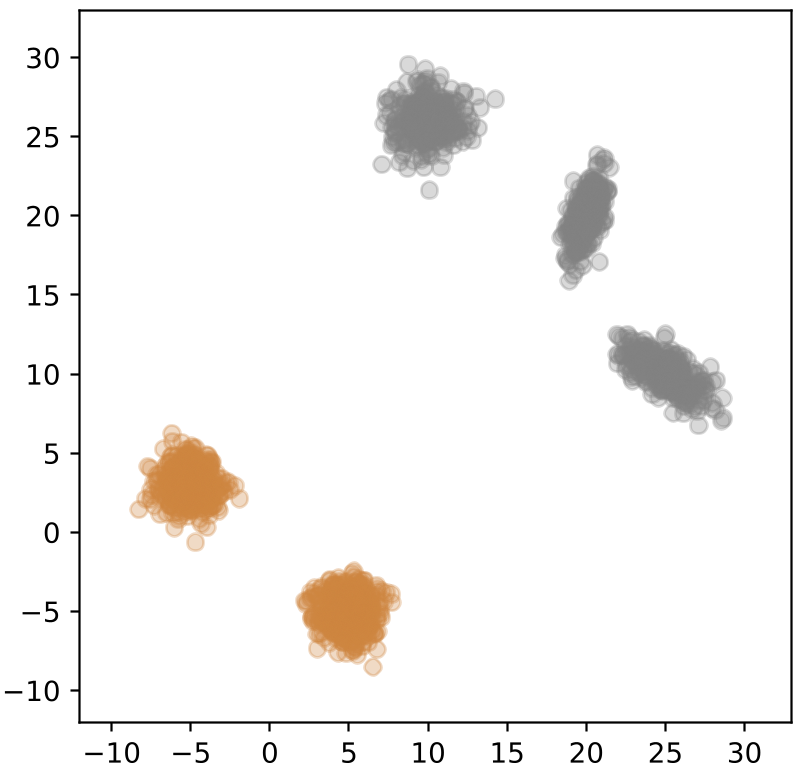} & \includegraphics[height=0.15\textwidth]{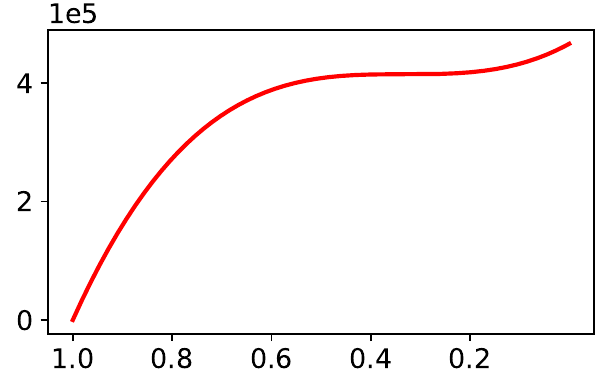} & \includegraphics[height=0.15\textwidth]{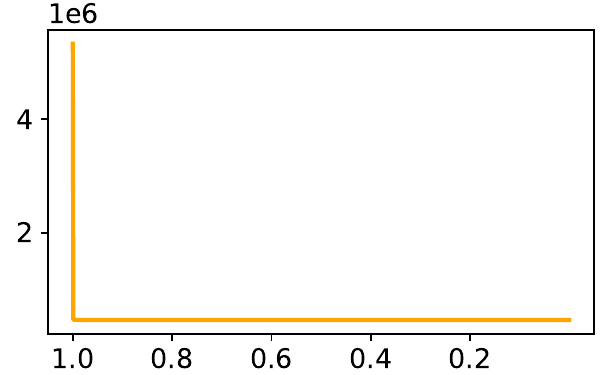} & \includegraphics[height=0.15\textwidth]{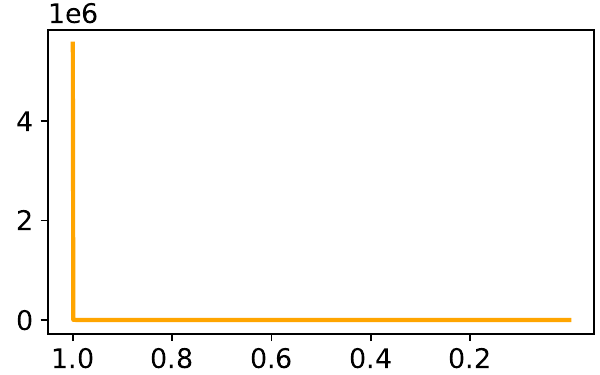} & \includegraphics[height=0.15\textwidth]{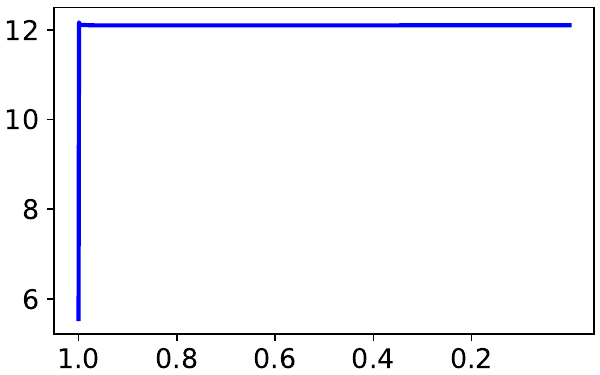} & \includegraphics[height=0.15\textwidth]{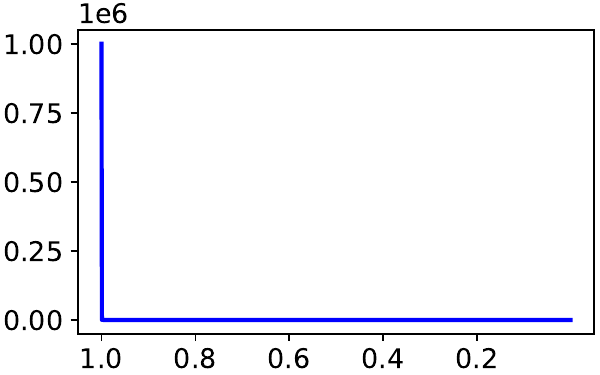}\tabularnewline
\multicolumn{1}{c}{} &  &  &  & (a) & (b) & (c) & (d) & (e) & (f)\tabularnewline
\end{tabular}}}
\par\end{centering}
\caption{Generated samples (a) and the evolution of the maximum absolute values
of different tracked variables over time (b-f) when sampling along
the SC flow of the third-degree process, with different types of models,
types of SC flows, values of clipping threshold $\epsilon$, and numbers
of steps $N$.\label{fig:Toy_NumericalRobust_ThirdDegree}}
\end{figure*}

\begin{figure*}
\begin{centering}
{\resizebox{\textwidth}{!}{%
\par\end{centering}
\begin{centering}
\begin{tabular}{c|ccccccccc}
\multicolumn{1}{c}{model} & SC type & $\epsilon$ & $N$ & samples & $x_{t}$ & $\bar{x}_{t}$ & $\left(\varphi_{t'}-\varphi_{t}\right)v_{\theta}\left(\bar{x}_{t},t\right)$ & $v_{\theta}\left(\bar{x}_{t},t\right)$ & $\varphi_{t'}-\varphi_{t}$\tabularnewline
\multirow{2}{*}{$v_{\theta}\left(x_{t},t\right)$} & \multirow{1}{*}[0.075\textwidth]{Eq.~\ref{eq:straight_constant_speed_stochastic_1}} & \multirow{1}{*}[0.075\textwidth]{1e-6} & \multirow{1}{*}[0.075\textwidth]{20} & \includegraphics[height=0.15\textwidth]{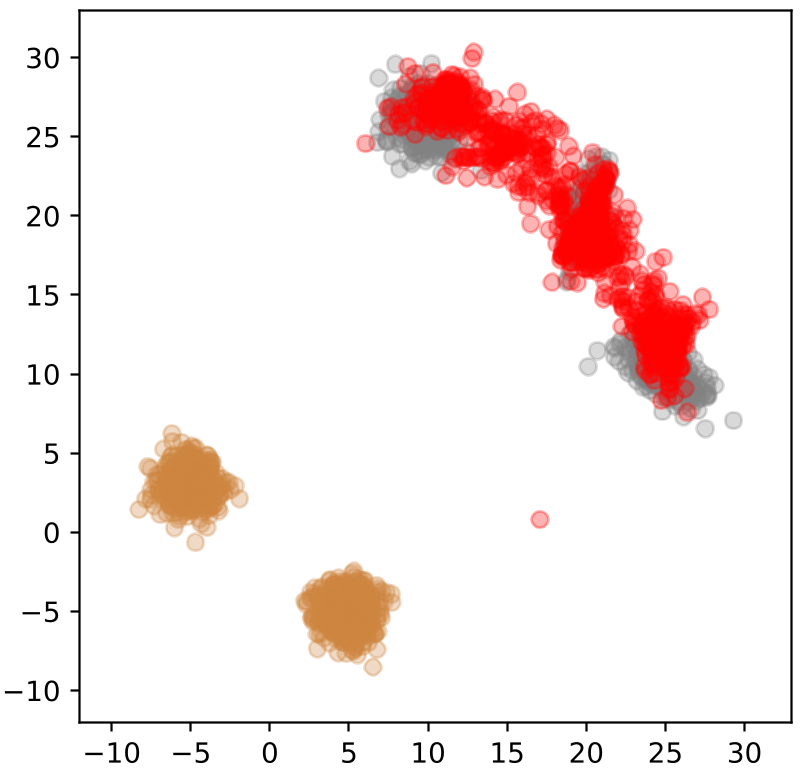} & \includegraphics[height=0.15\textwidth]{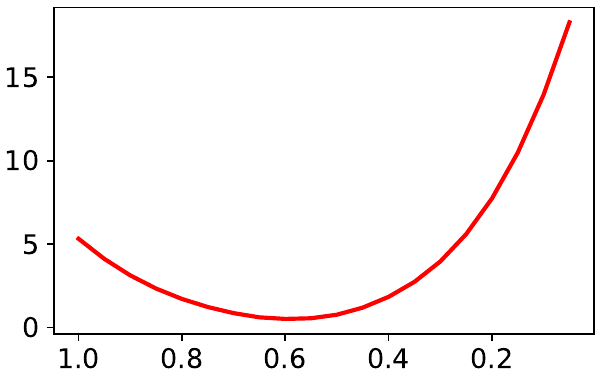} & \includegraphics[height=0.15\textwidth]{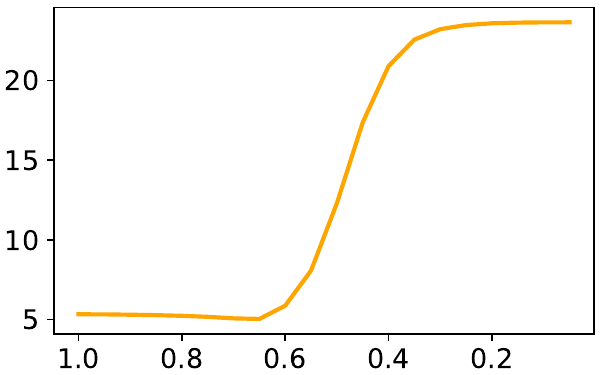} & \includegraphics[height=0.15\textwidth]{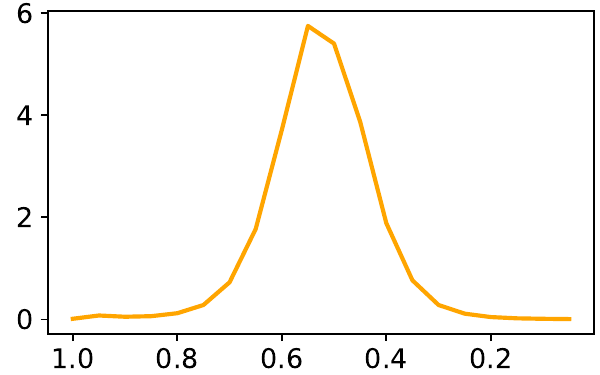} & \includegraphics[height=0.15\textwidth]{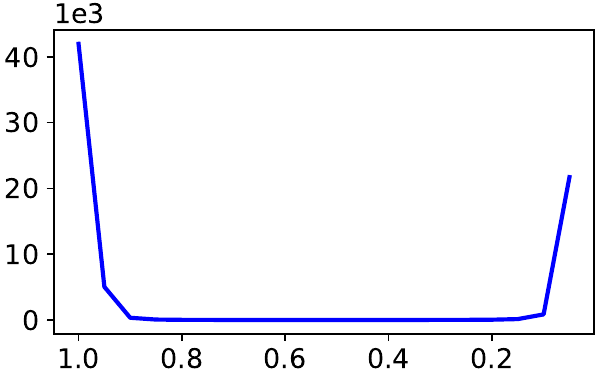} & \includegraphics[height=0.15\textwidth]{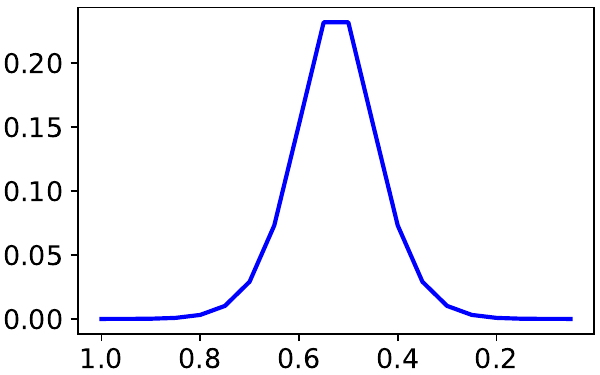}\tabularnewline
 & \multirow{1}{*}[0.075\textwidth]{Eq.~\ref{eq:straight_constant_speed_stochastic_2}} & \multirow{1}{*}[0.075\textwidth]{1e-6} & \multirow{1}{*}[0.075\textwidth]{20} & \includegraphics[height=0.15\textwidth]{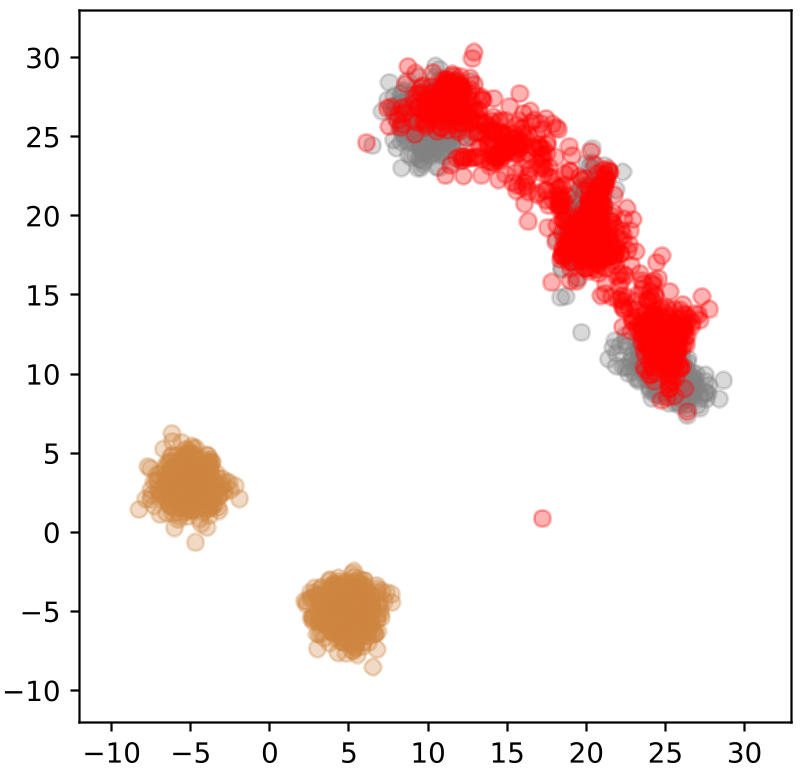} & \includegraphics[height=0.15\textwidth]{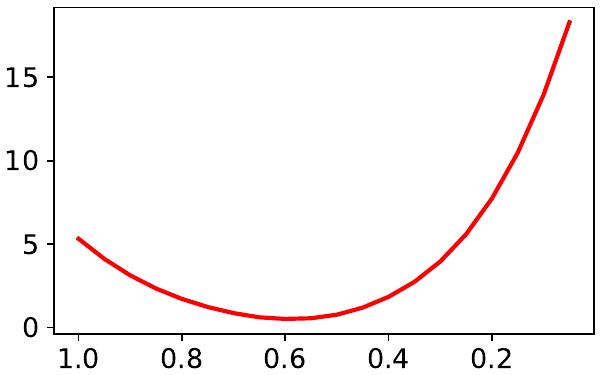} & \includegraphics[height=0.15\textwidth]{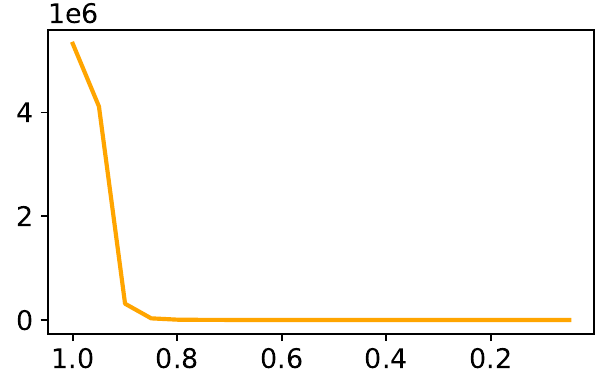} & \includegraphics[height=0.15\textwidth]{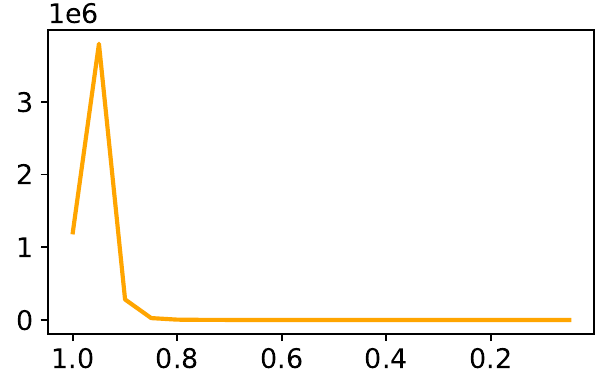} & \includegraphics[height=0.15\textwidth]{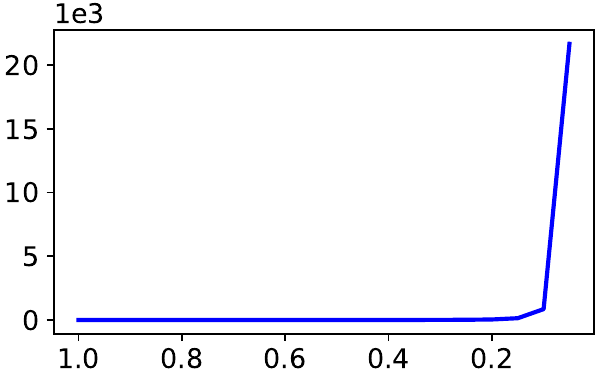} & \includegraphics[height=0.15\textwidth]{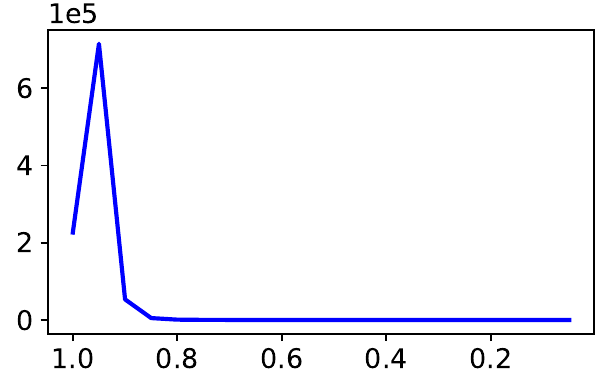}\tabularnewline
\multicolumn{1}{c}{} &  &  &  &  &  &  &  &  & \tabularnewline
\multirow{2}{*}{$x_{1,\theta}\left(x_{t},t\right)$} & \multirow{1}{*}[0.075\textwidth]{Eq.~\ref{eq:straight_constant_speed_stochastic_1}} & \multirow{1}{*}[0.075\textwidth]{1e-6} & \multirow{1}{*}[0.075\textwidth]{20} & \includegraphics[height=0.15\textwidth]{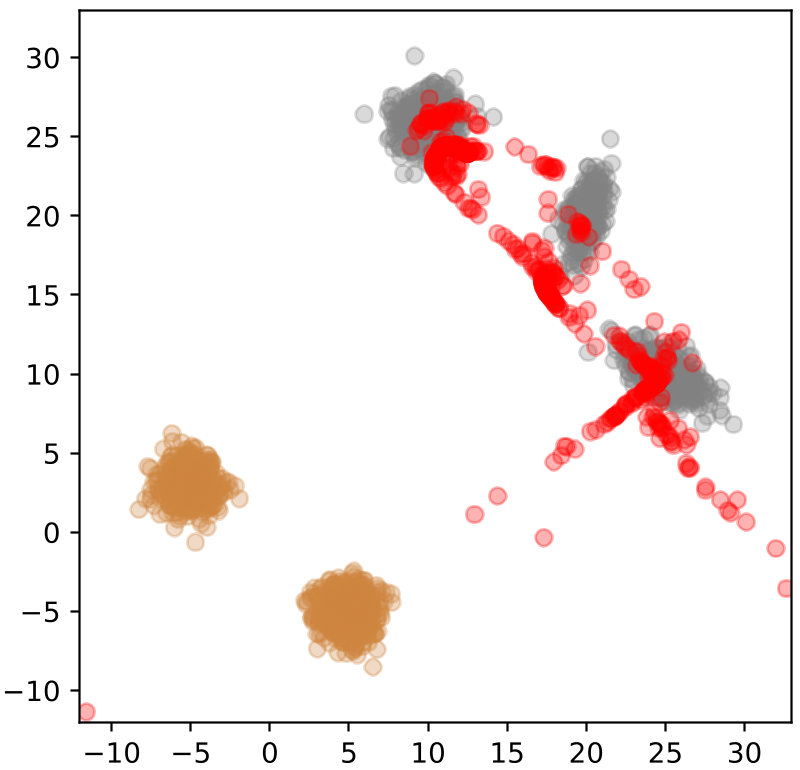} & \includegraphics[height=0.15\textwidth]{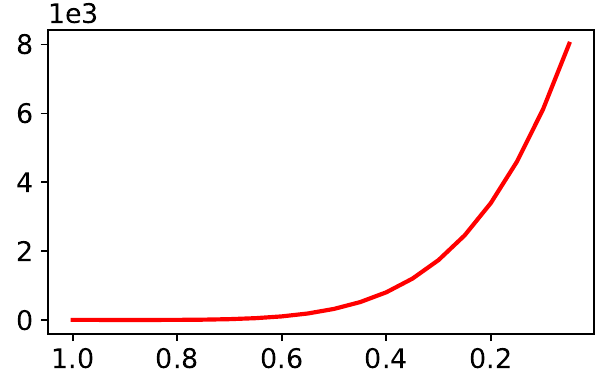} & \includegraphics[height=0.15\textwidth]{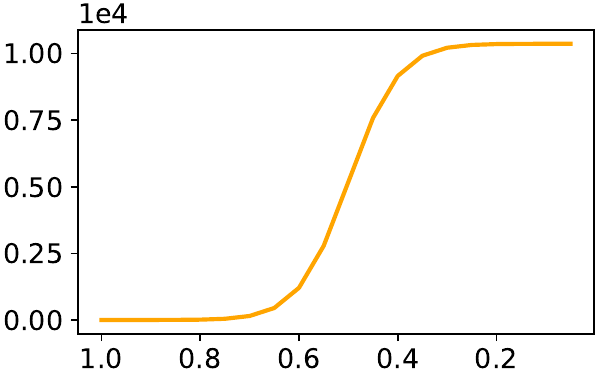} & \includegraphics[height=0.15\textwidth]{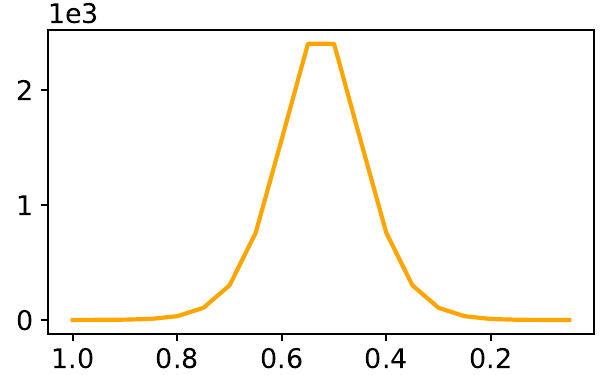} & \includegraphics[height=0.15\textwidth]{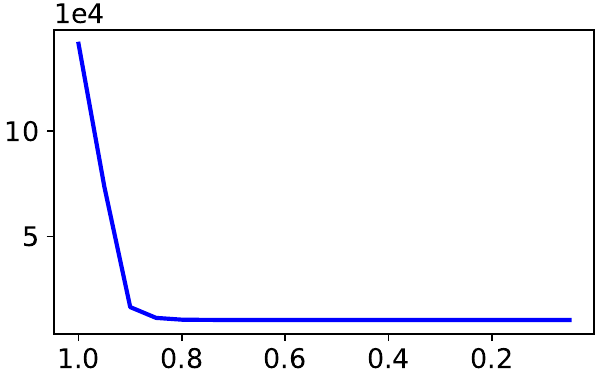} & \includegraphics[height=0.15\textwidth]{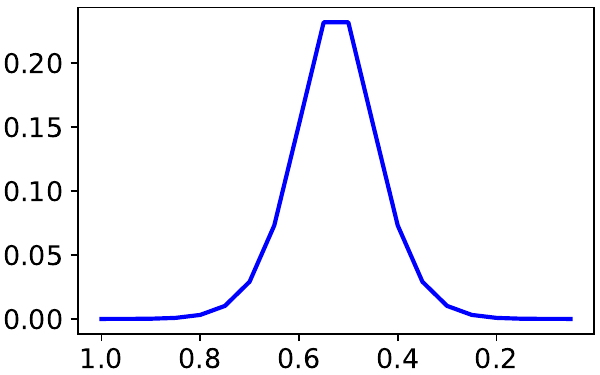}\tabularnewline
 & \multirow{1}{*}[0.075\textwidth]{Eq.~\ref{eq:straight_constant_speed_stochastic_2}} & \multirow{1}{*}[0.075\textwidth]{1e-6} & \multirow{1}{*}[0.075\textwidth]{20} & \includegraphics[height=0.15\textwidth]{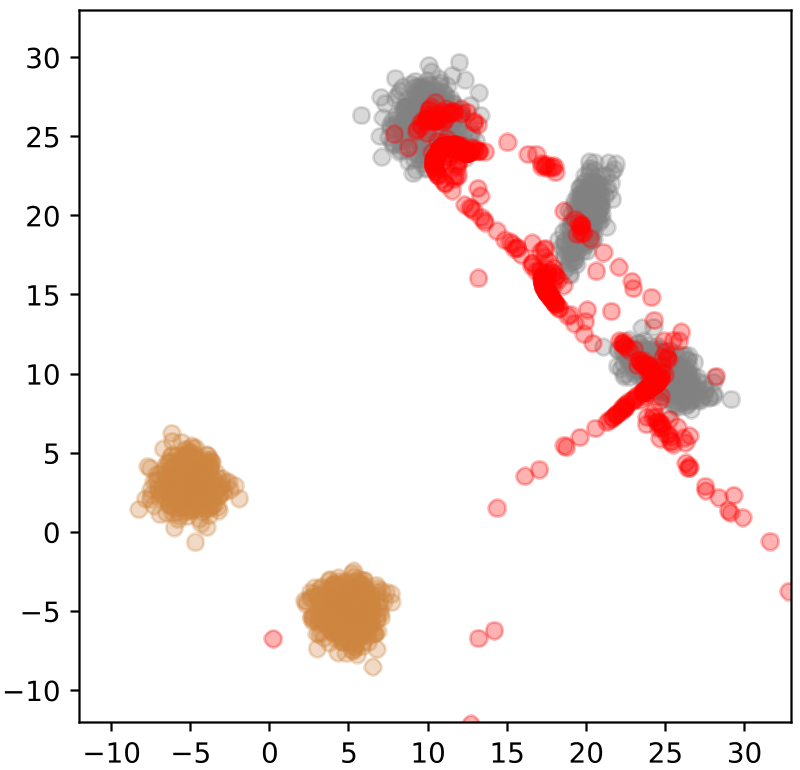} & \includegraphics[height=0.15\textwidth]{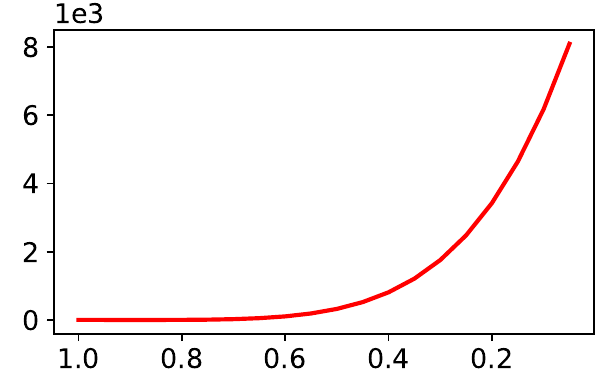} & \includegraphics[height=0.15\textwidth]{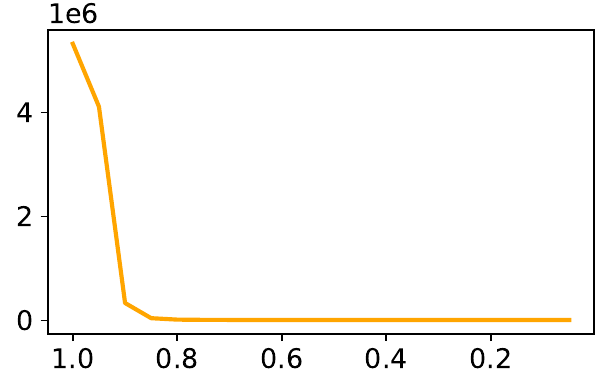} & \includegraphics[height=0.15\textwidth]{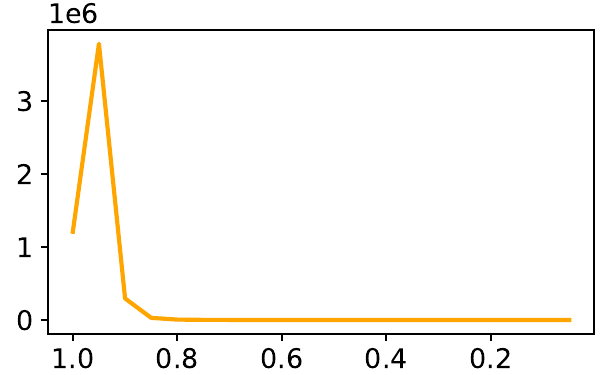} & \includegraphics[height=0.15\textwidth]{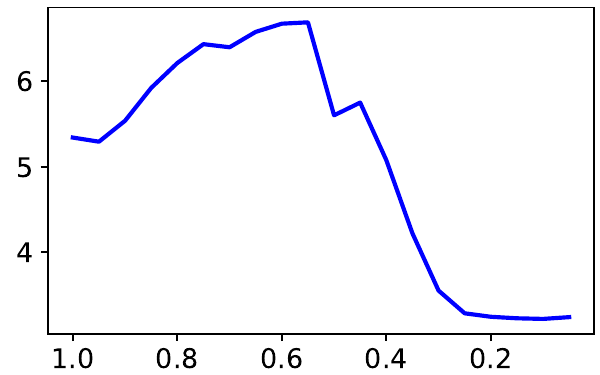} & \includegraphics[height=0.15\textwidth]{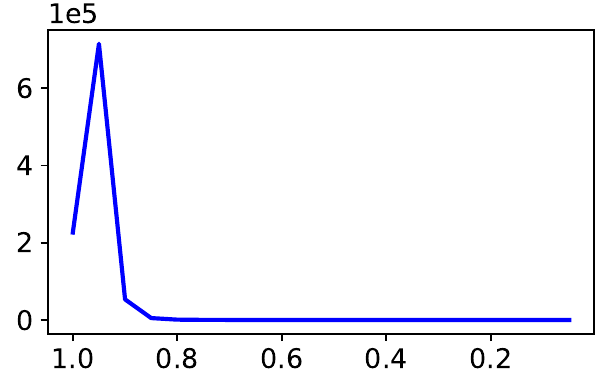}\tabularnewline
\multicolumn{1}{c}{} &  &  &  & (a) & (b) & (c) & (d) & (e) & (f)\tabularnewline
\end{tabular}}}
\par\end{centering}
\caption{Generated samples (a) and the evolution of the maximum absolute values
of different tracked variables over time (b-f) when sampling along
the SC flow of the fifth-degree process, with different types of models,
and types of SC flows.\label{fig:Toy_NumericalRobust_FifthDegree}}
\end{figure*}

\begin{figure*}
\begin{centering}
{\resizebox{\textwidth}{!}{%
\par\end{centering}
\begin{centering}
\begin{tabular}{c|ccccccccc}
\multicolumn{1}{c}{model} & SC type & $\epsilon$ & $N$ & samples & $x_{t}$ & $\bar{x}_{t}$ & $\left(\varphi_{t'}-\varphi_{t}\right)v_{\theta}\left(\bar{x}_{t},t\right)$ & $v_{\theta}\left(\bar{x}_{t},t\right)$ & $\varphi_{t'}-\varphi_{t}$\tabularnewline
\multirow{2}{*}{$v_{\theta}\left(x_{t},t\right)$} & \multirow{1}{*}[0.075\textwidth]{Eq.~\ref{eq:straight_constant_speed_stochastic_1}} & \multirow{1}{*}[0.075\textwidth]{1e-6} & \multirow{1}{*}[0.075\textwidth]{20} & \includegraphics[height=0.15\textwidth]{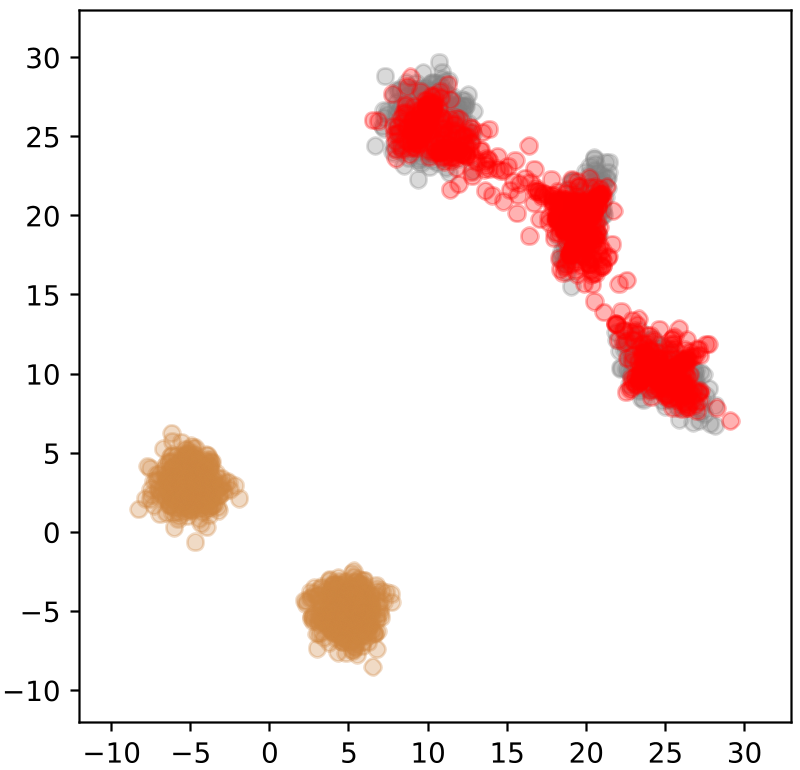} & \includegraphics[height=0.15\textwidth]{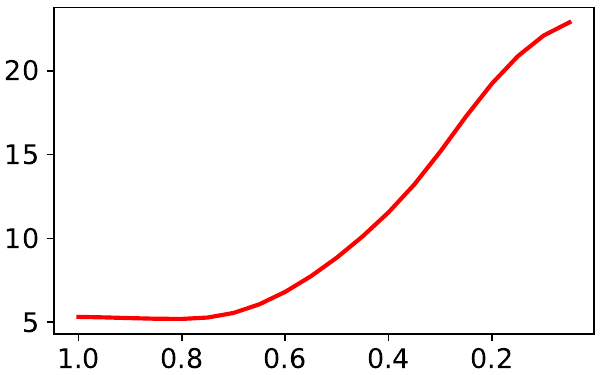} & \includegraphics[height=0.15\textwidth]{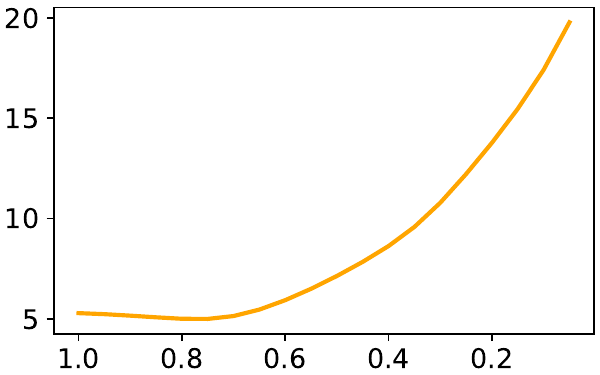} & \includegraphics[height=0.15\textwidth]{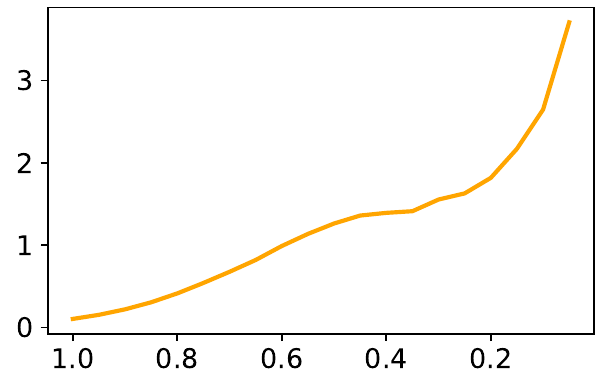} & \includegraphics[height=0.15\textwidth]{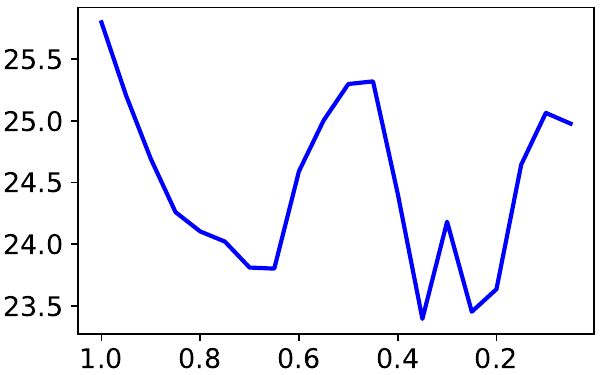} & \includegraphics[height=0.15\textwidth]{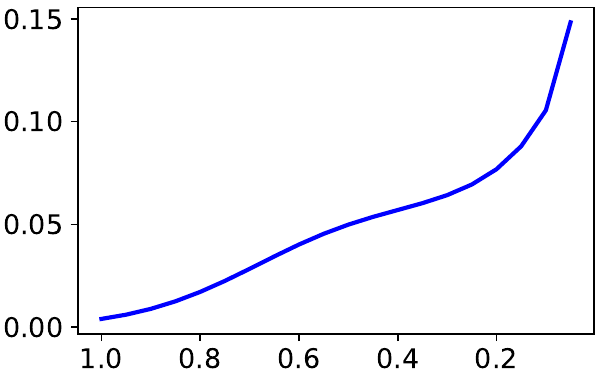}\tabularnewline
 & \multirow{1}{*}[0.075\textwidth]{Eq.~\ref{eq:straight_constant_speed_stochastic_2}} & \multirow{1}{*}[0.075\textwidth]{1e-6} & \multirow{1}{*}[0.075\textwidth]{20} & \includegraphics[height=0.15\textwidth]{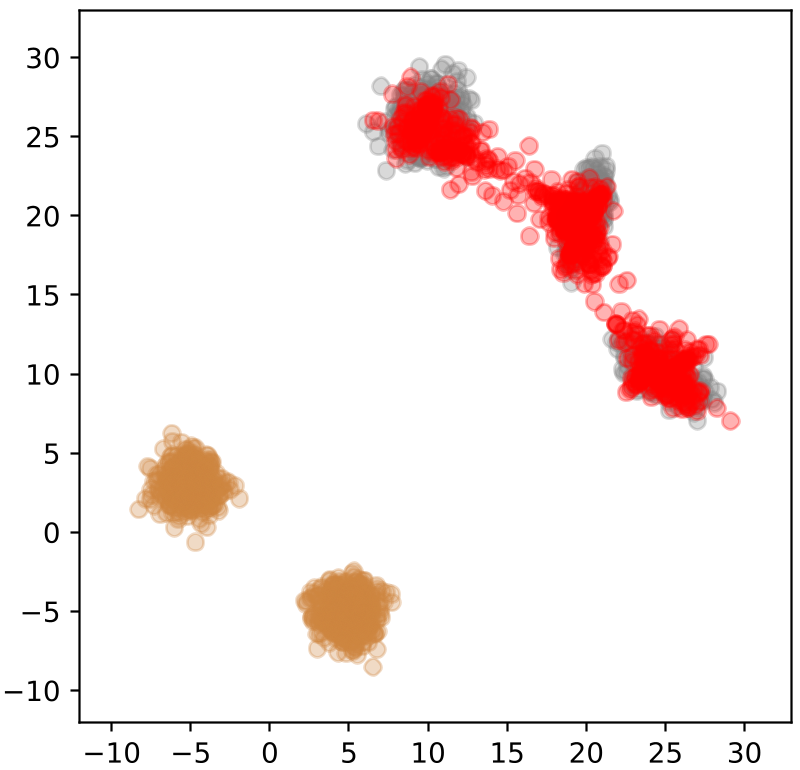} & \includegraphics[height=0.15\textwidth]{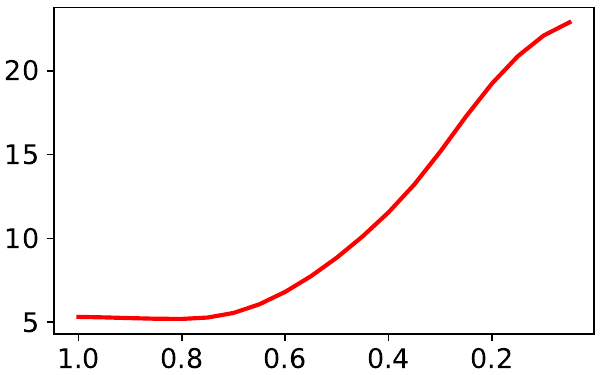} & \includegraphics[height=0.15\textwidth]{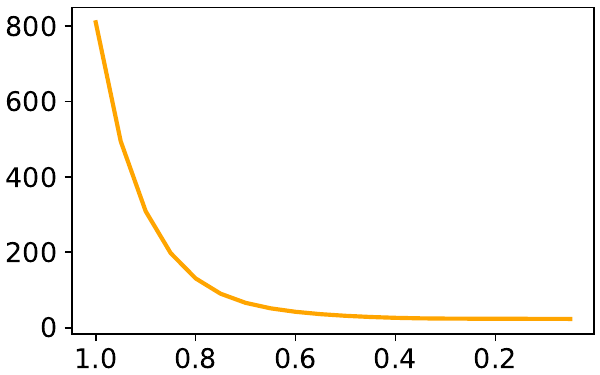} & \includegraphics[height=0.15\textwidth]{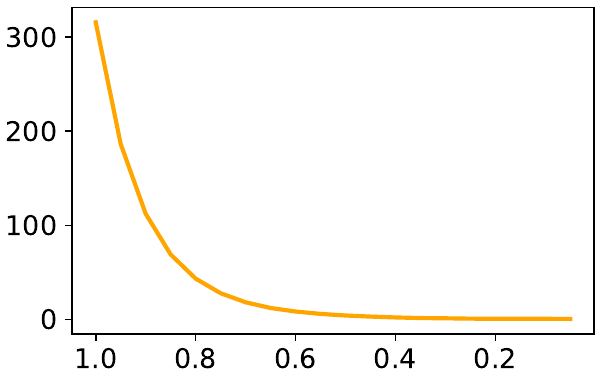} & \includegraphics[height=0.15\textwidth]{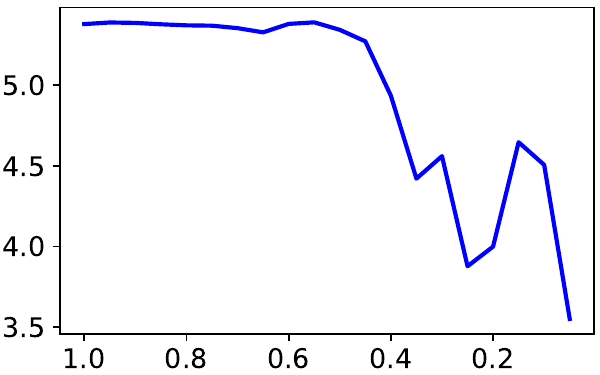} & \includegraphics[height=0.15\textwidth]{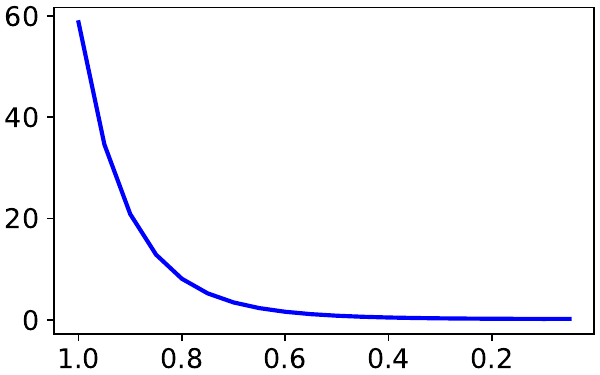}\tabularnewline
\multicolumn{1}{c}{} &  &  &  &  &  &  &  &  & \tabularnewline
\multirow{2}{*}{$x_{1,\theta}\left(x_{t},t\right)$} & \multirow{1}{*}[0.075\textwidth]{Eq.~\ref{eq:straight_constant_speed_stochastic_1}} & \multirow{1}{*}[0.075\textwidth]{1e-6} & \multirow{1}{*}[0.075\textwidth]{20} & \includegraphics[height=0.15\textwidth]{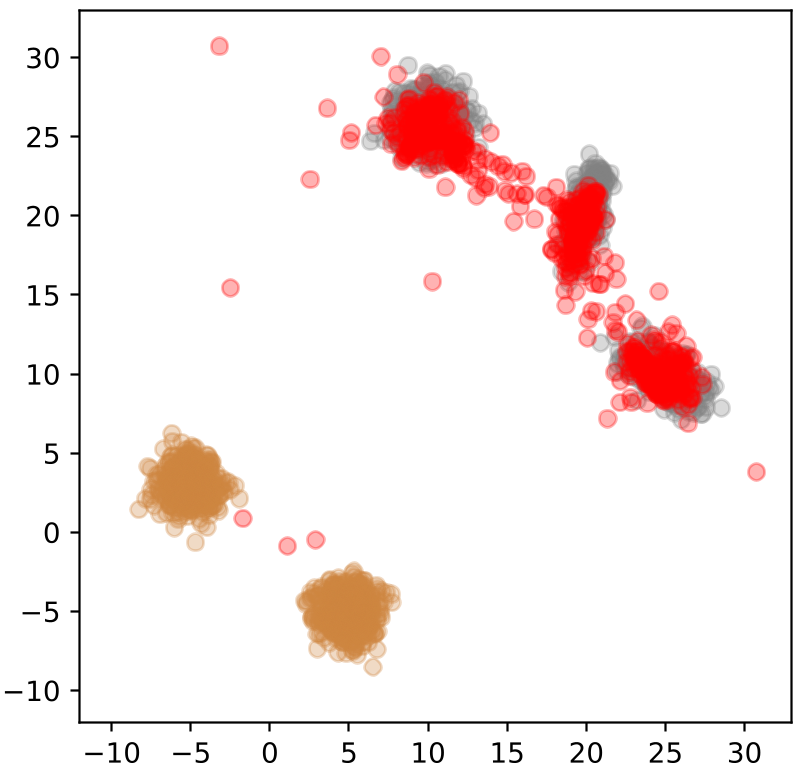} & \includegraphics[height=0.15\textwidth]{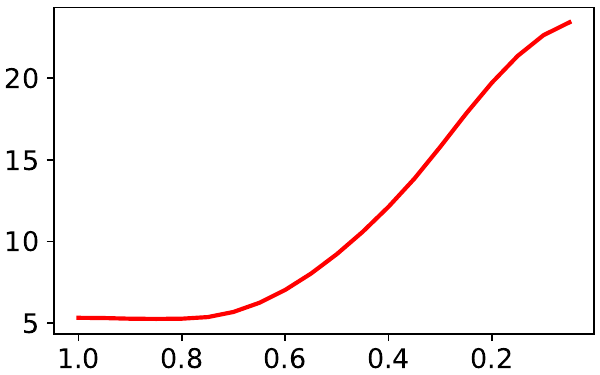} & \includegraphics[height=0.15\textwidth]{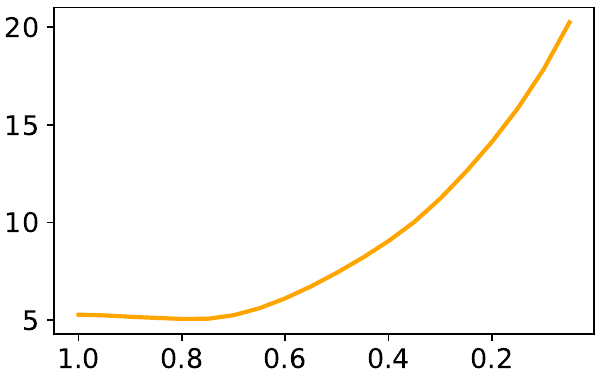} & \includegraphics[height=0.15\textwidth]{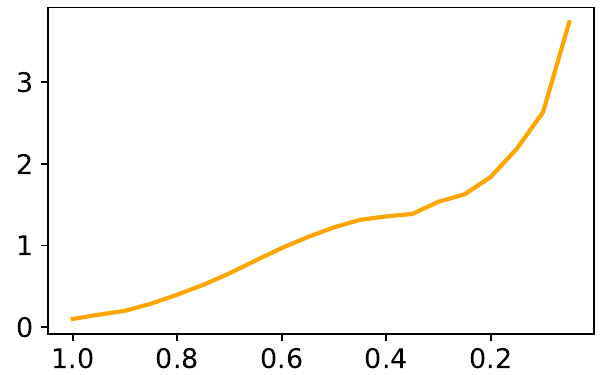} & \includegraphics[height=0.15\textwidth]{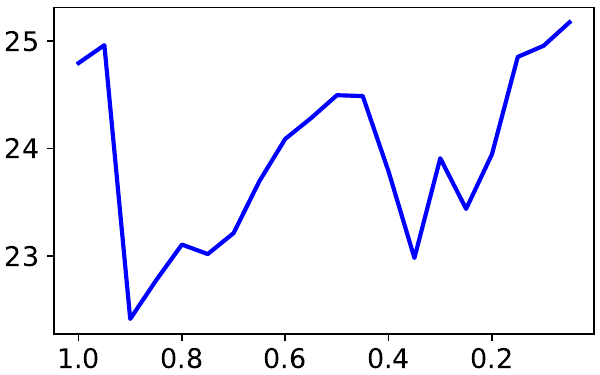} & \includegraphics[height=0.15\textwidth]{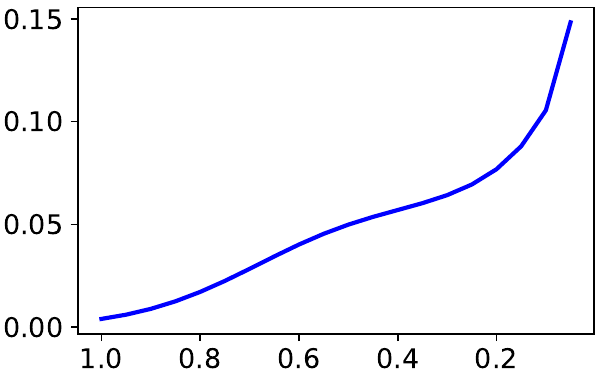}\tabularnewline
 & \multirow{1}{*}[0.075\textwidth]{Eq.~\ref{eq:straight_constant_speed_stochastic_2}} & \multirow{1}{*}[0.075\textwidth]{1e-6} & \multirow{1}{*}[0.075\textwidth]{20} & \includegraphics[height=0.15\textwidth]{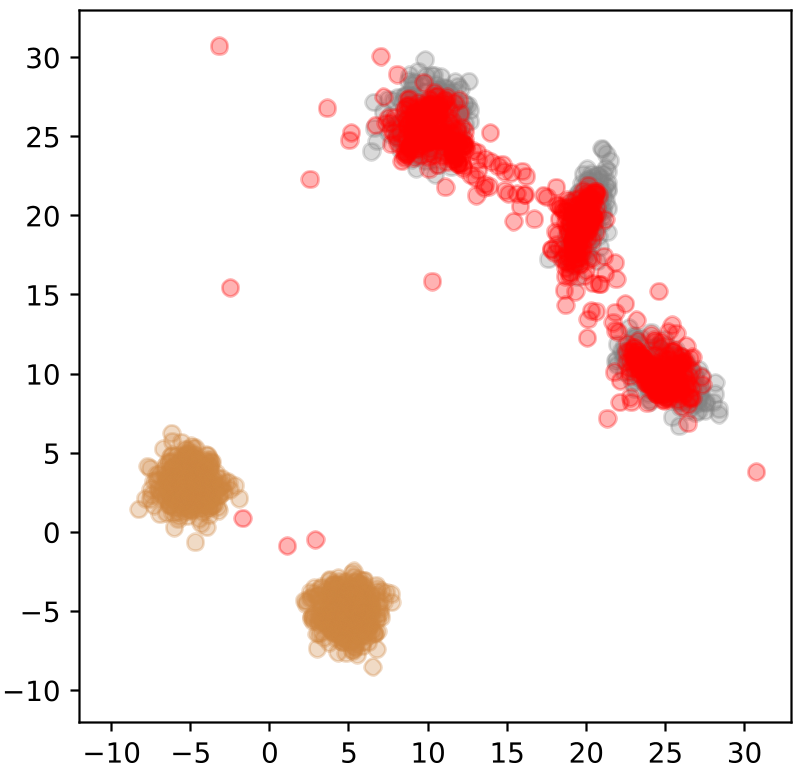} & \includegraphics[height=0.15\textwidth]{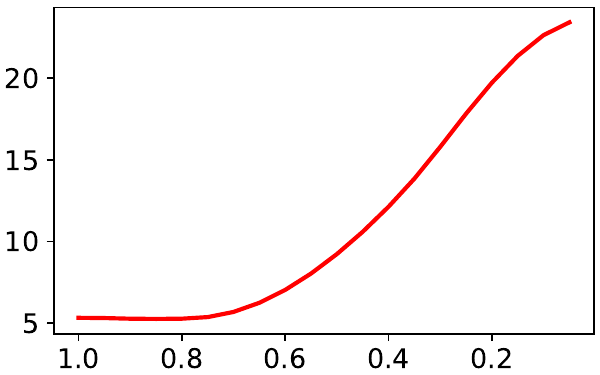} & \includegraphics[height=0.15\textwidth]{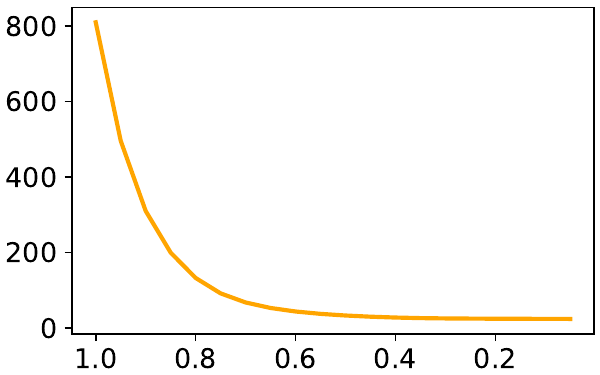} & \includegraphics[height=0.15\textwidth]{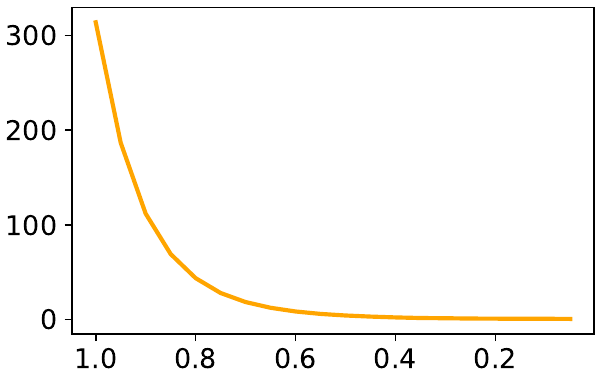} & \includegraphics[height=0.15\textwidth]{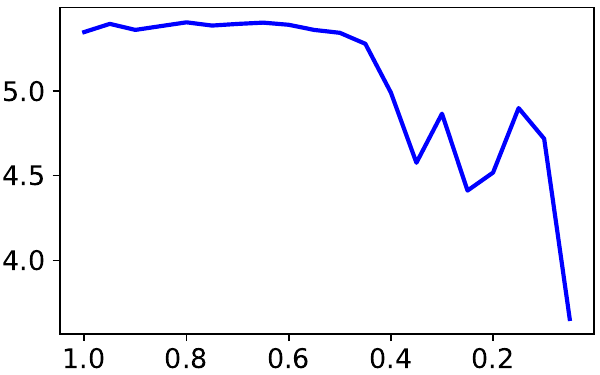} & \includegraphics[height=0.15\textwidth]{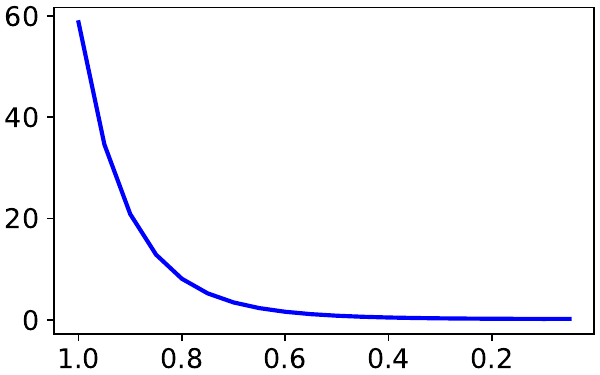}\tabularnewline
\multicolumn{1}{c}{} &  &  &  & (a) & (b) & (c) & (d) & (e) & (f)\tabularnewline
\end{tabular}}}
\par\end{centering}
\caption{Generated samples (a) and the evolution of the maximum absolute values
of different tracked variables over time (b-f) when sampling along
the SC flow of the VP-SDE-like process, with different types of models,
and types of SC flows.\label{fig:Toy_NumericalRobust_VPSDELike}}
\end{figure*}

During our experiments, we have observed that if the model is $v_{\theta}\left(x_{t},t\right)$,
using the simple clamping technique presented in Section~\ref{subsec:Numerical-robustness}
with $\epsilon$ as small as 1e-6 is enough to ensure the numerical
stability of sampling along the SC flow for all the three processes
(third-degree, fifth-degree and VP-SDE-like) as illustrated in Figs.~\ref{fig:Toy_NumericalRobust_ThirdDegree},
\ref{fig:Toy_NumericalRobust_FifthDegree}, \ref{fig:Toy_NumericalRobust_VPSDELike}.
On the other hand, if the model is $x_{1,\theta}\left(x_{t},t\right)$,
we achieve numerical stability \emph{only for the VP-SDE-like process}
when using this clamping technique with $\epsilon$ being 1e-6, which
can be seen from the average maximum absolute values of $x_{t}$ for
the VP-SDE-like process (the red curves in rows 3, 4 column (b) of
Fig.~\ref{fig:Toy_NumericalRobust_VPSDELike}) at $t$ = 0 being
normal (about 30) while the counterparts for the third-degree and
fifth-degree process (the red curves in rows 3, 4 column (b) of Figs.~\ref{fig:Toy_NumericalRobust_ThirdDegree},
\ref{fig:Toy_NumericalRobust_FifthDegree}) being much larger than
normal. The numerical instability for the fifth-degree process only
happens for a small fraction of generated samples as the majority
can still be visualized in rows 3, 4 column (a) of Fig.~\ref{fig:Toy_NumericalRobust_FifthDegree},
yet for the third-degree process, it happens for almost all samples.
This numerical problem can only be partially alleviated for the third-degree
process by increasing the clipping threshold $\epsilon$ and the number
of sampling steps $N$.

\subsubsection{Comparison between modeling $v^{*}\left(x_{t},t\right)$ and modeling
$x_{1|t}$}

\begin{figure*}
\begin{centering}
{\resizebox{0.92\textwidth}{!}{%
\par\end{centering}
\begin{centering}
\begin{tabular}{cccc|ccc}
$N$ & \multicolumn{6}{c}{\includegraphics[height=0.045\textwidth]{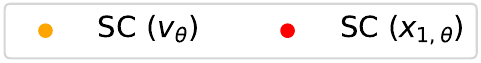}\hspace{0.02\textwidth}\includegraphics[height=0.045\textwidth]{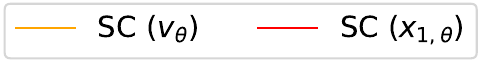}}\tabularnewline
\multirow{1}{*}[0.07\textwidth]{5} & \includegraphics[width=0.18\textwidth]{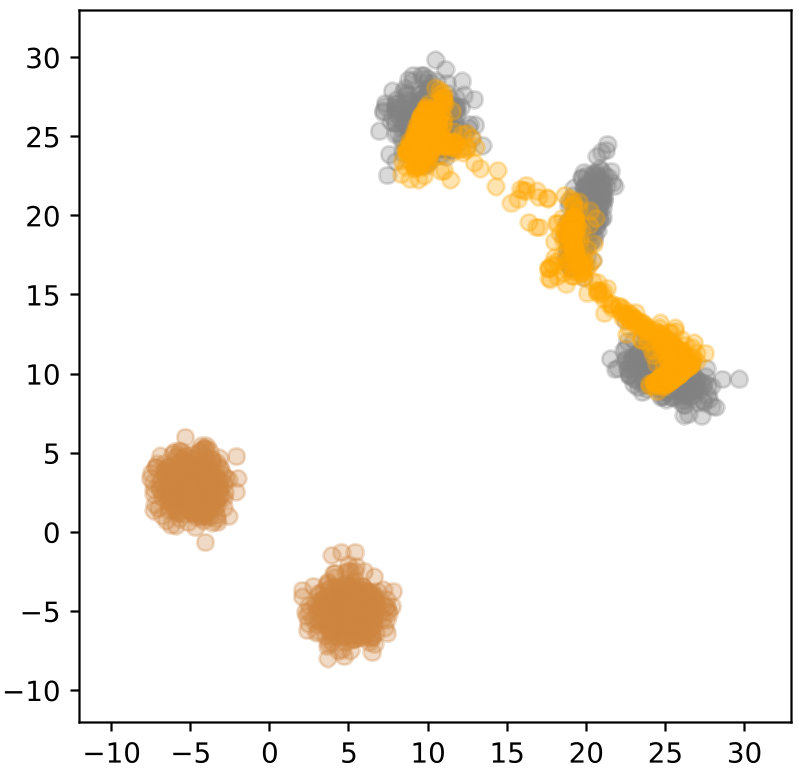} & \includegraphics[width=0.18\textwidth]{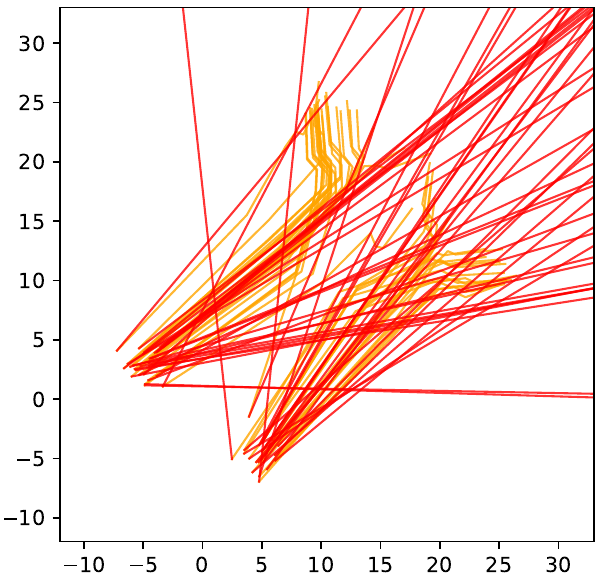} & \includegraphics[width=0.18\textwidth]{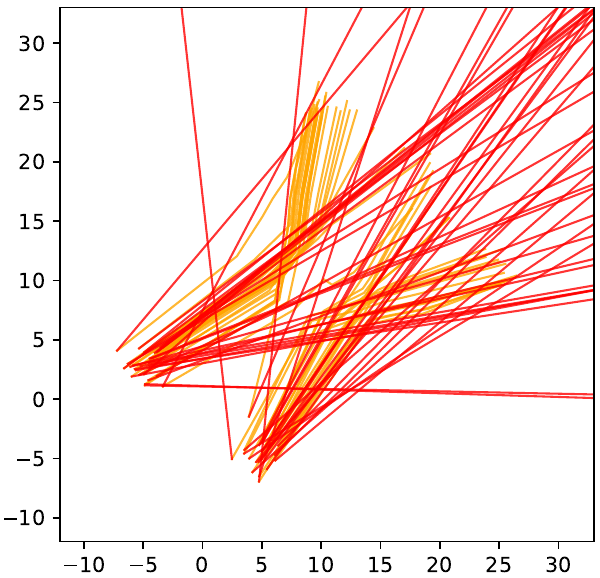} & \includegraphics[width=0.18\textwidth]{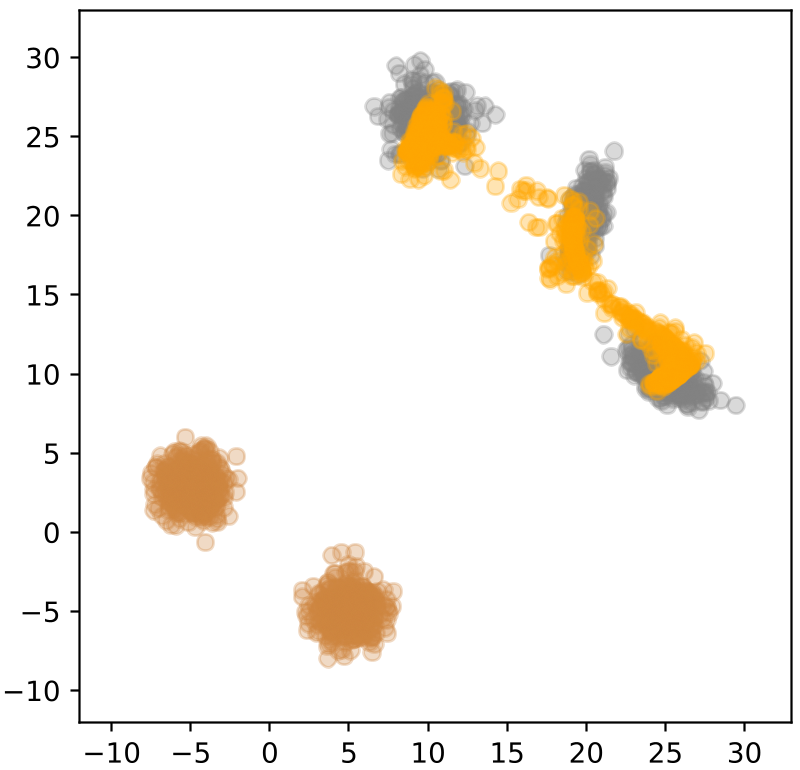} & \includegraphics[width=0.18\textwidth]{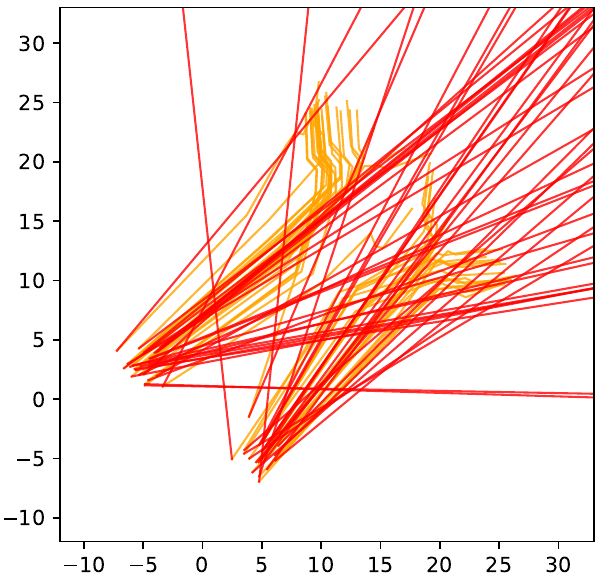} & \includegraphics[width=0.18\textwidth]{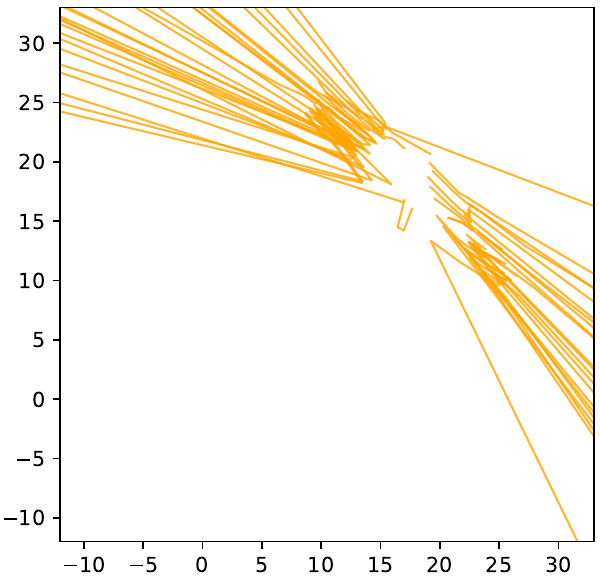}\tabularnewline
\multirow{1}{*}[0.07\textwidth]{50} & \includegraphics[width=0.18\textwidth]{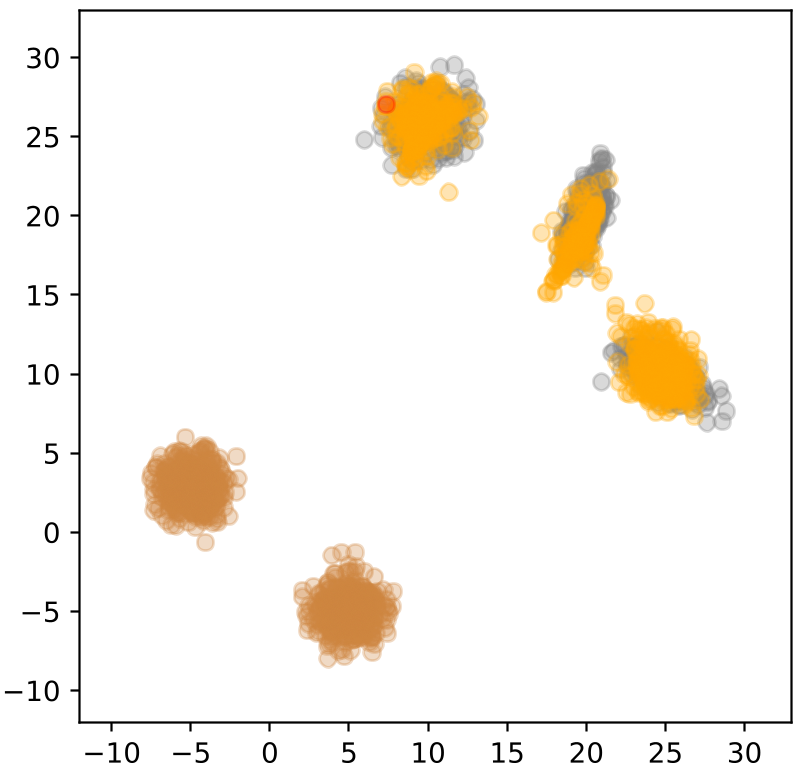} & \includegraphics[width=0.18\textwidth]{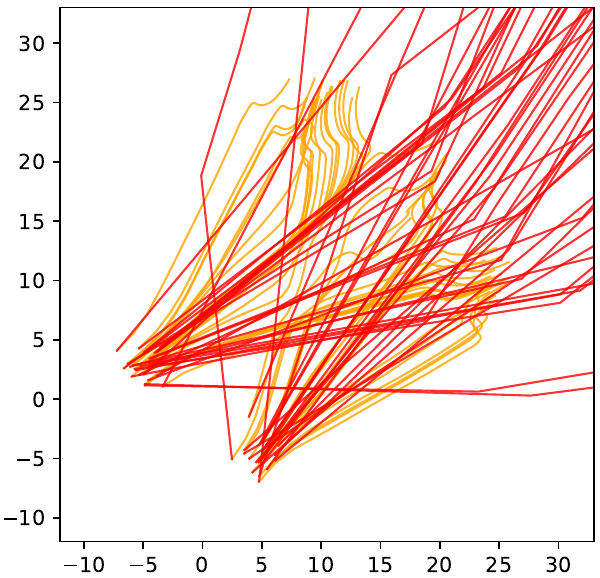} & \includegraphics[width=0.18\textwidth]{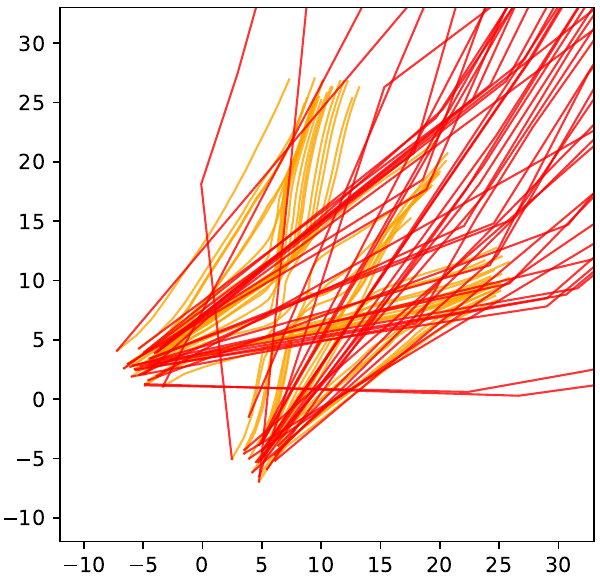} & \includegraphics[width=0.18\textwidth]{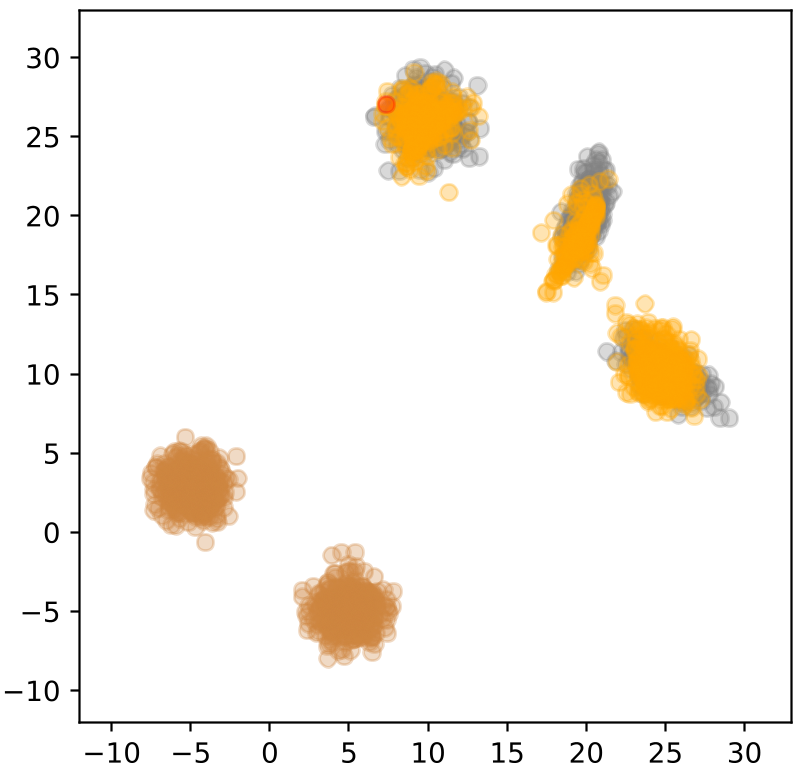} & \includegraphics[width=0.18\textwidth]{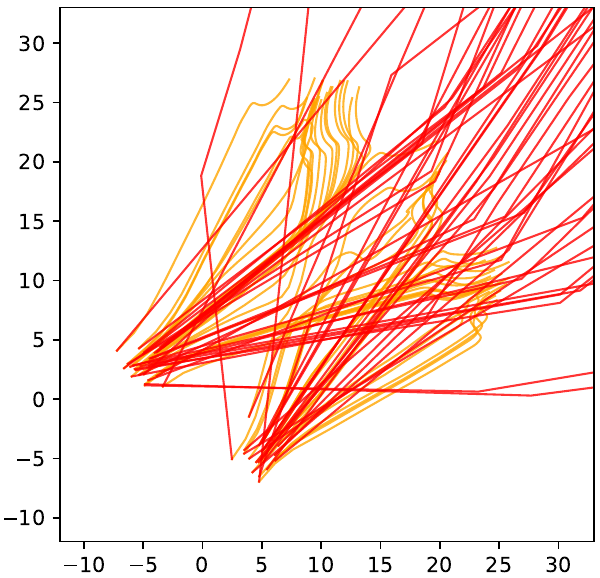} & \includegraphics[width=0.18\textwidth]{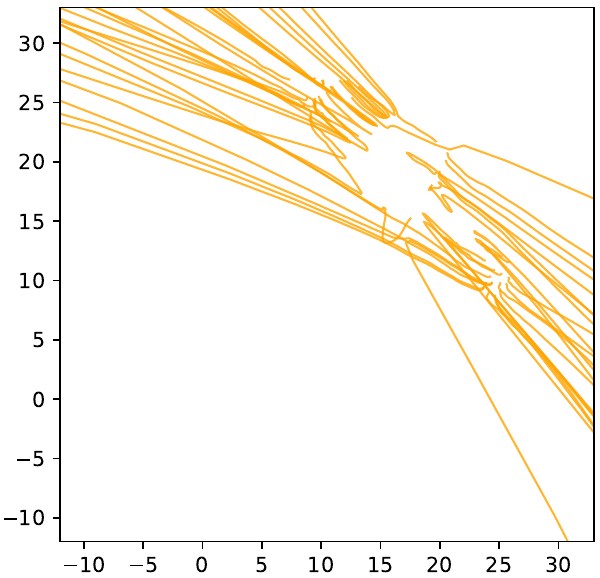}\tabularnewline
\multirow{1}{*}[0.07\textwidth]{1000} & \includegraphics[width=0.18\textwidth]{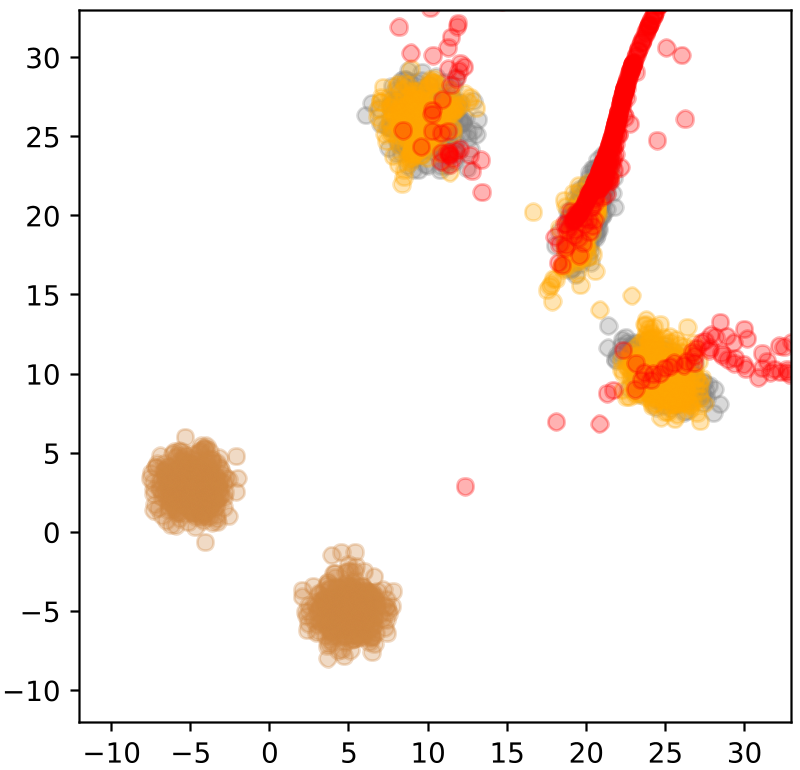} & \includegraphics[width=0.18\textwidth]{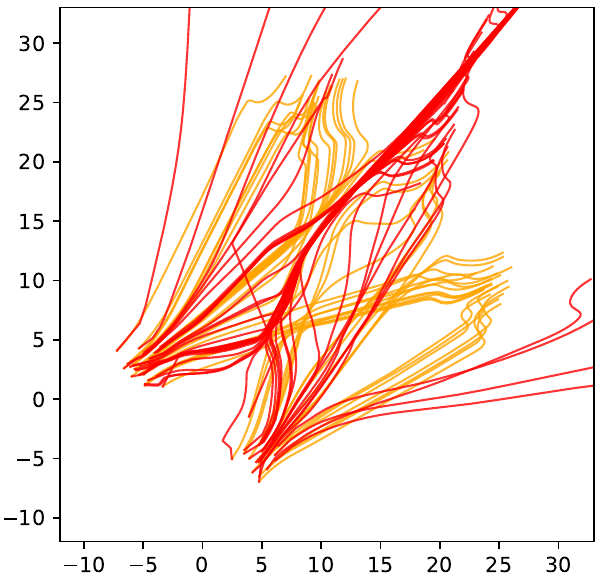} & \includegraphics[width=0.18\textwidth]{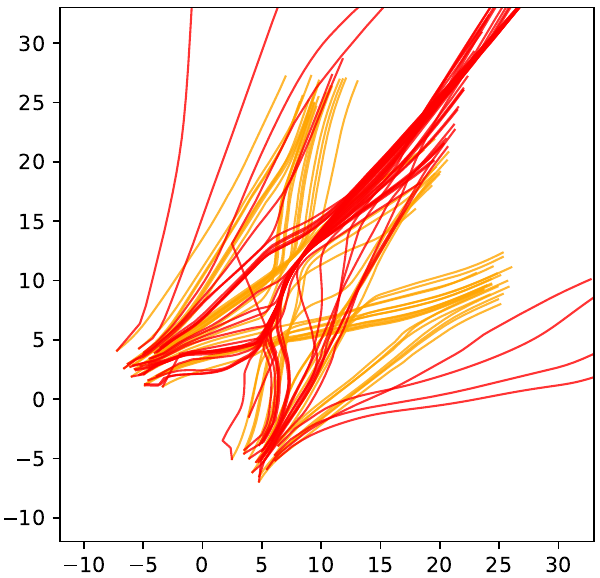} & \includegraphics[width=0.18\textwidth]{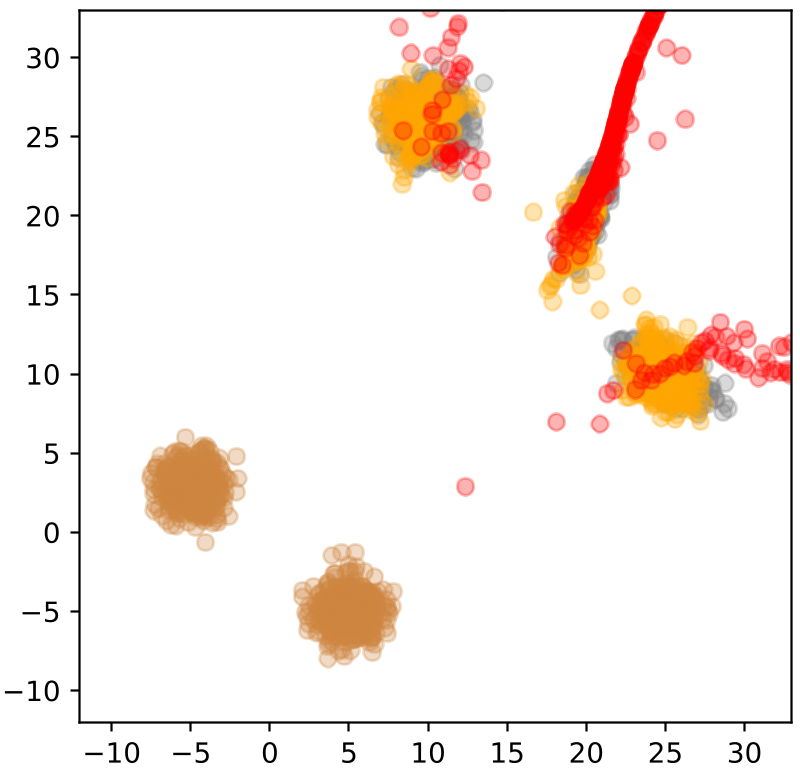} & \includegraphics[width=0.18\textwidth]{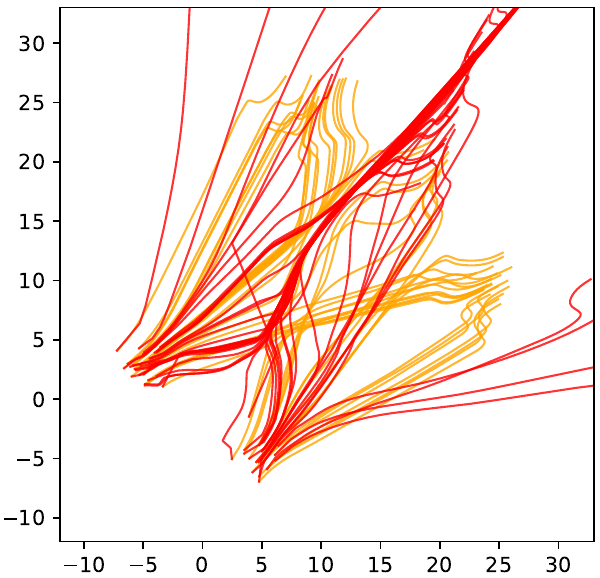} & \includegraphics[width=0.18\textwidth]{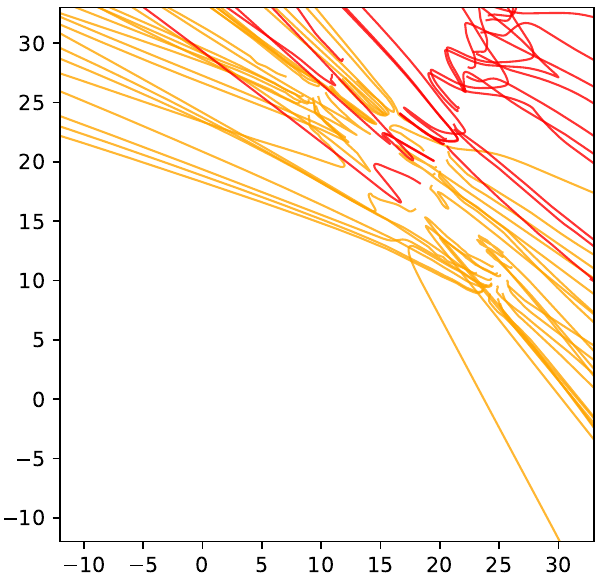}\tabularnewline
 & \multicolumn{3}{c}{$\bar{X}_{t}=\left(1-t\right)X_{0}+tX_{1}$} & \multicolumn{3}{c}{$\bar{X}_{t}=X_{0}+tX_{1}$}\tabularnewline
\end{tabular}}}
\par\end{centering}
\caption{Generated samples (left), $x_{t}$-trajectories (middle) and $\bar{x}_{t}$-trajectories
(right) of the SC flows of the \emph{third-degree} process, with different
number of updates $N$, different types of SC flows (two equations
in the bottom), and different types of parameterized models ($v_{\theta}$
vs $x_{1,\theta}$).\label{fig:Toy_VeloVSNoise_SC_ThirdDegree}}
\end{figure*}

\begin{figure*}
\begin{centering}
{\resizebox{0.92\textwidth}{!}{%
\par\end{centering}
\begin{centering}
\begin{tabular}{cccc|ccc}
$N$ & \multicolumn{6}{c}{\includegraphics[height=0.045\textwidth]{asset/ThreeGauss/legend_sample_velo_vs_noise}\hspace{0.02\textwidth}\includegraphics[height=0.045\textwidth]{asset/ThreeGauss/legend_traj_velo_vs_noise}}\tabularnewline
\multirow{1}{*}[0.07\textwidth]{5} & \includegraphics[width=0.18\textwidth]{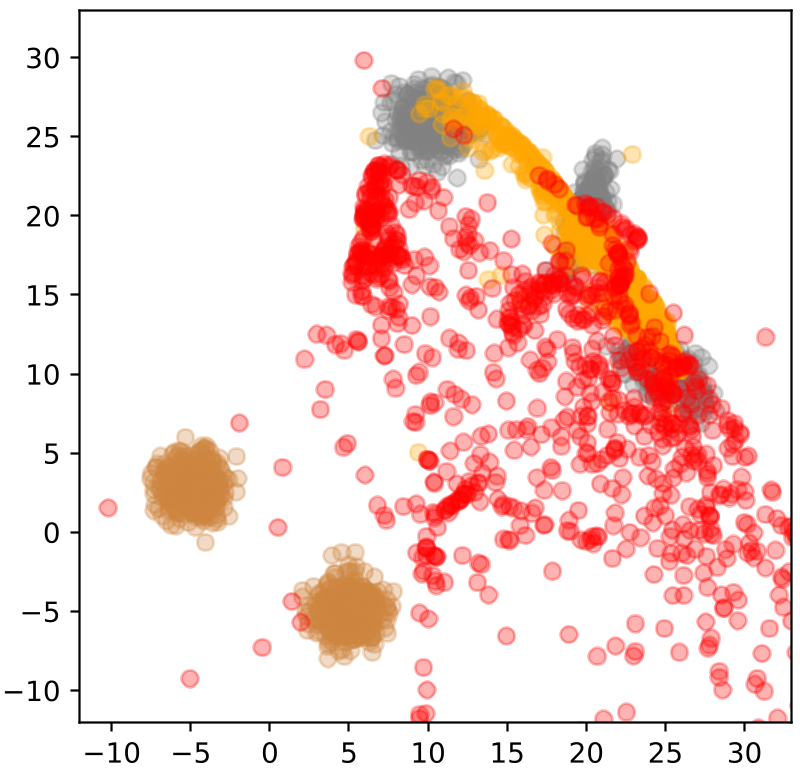} & \includegraphics[width=0.18\textwidth]{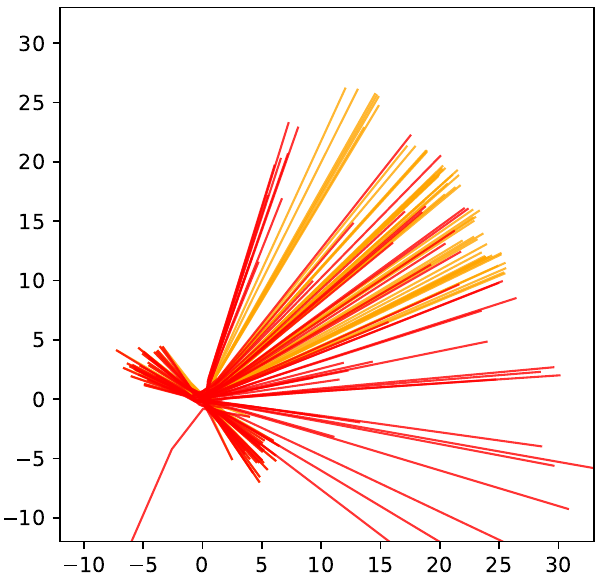} & \includegraphics[width=0.18\textwidth]{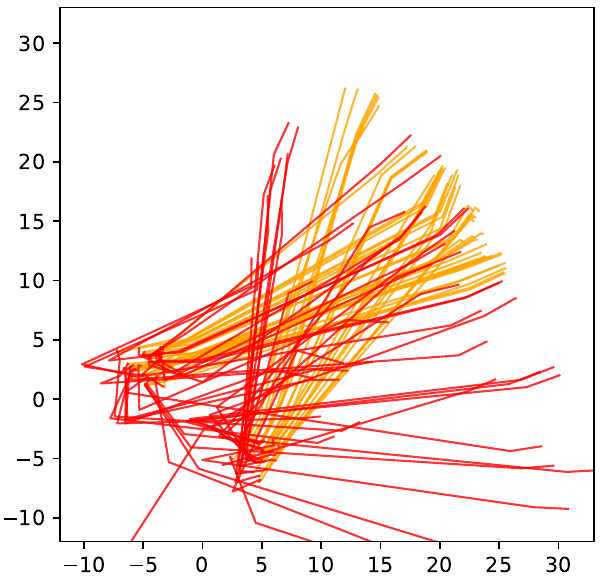} & \includegraphics[width=0.18\textwidth]{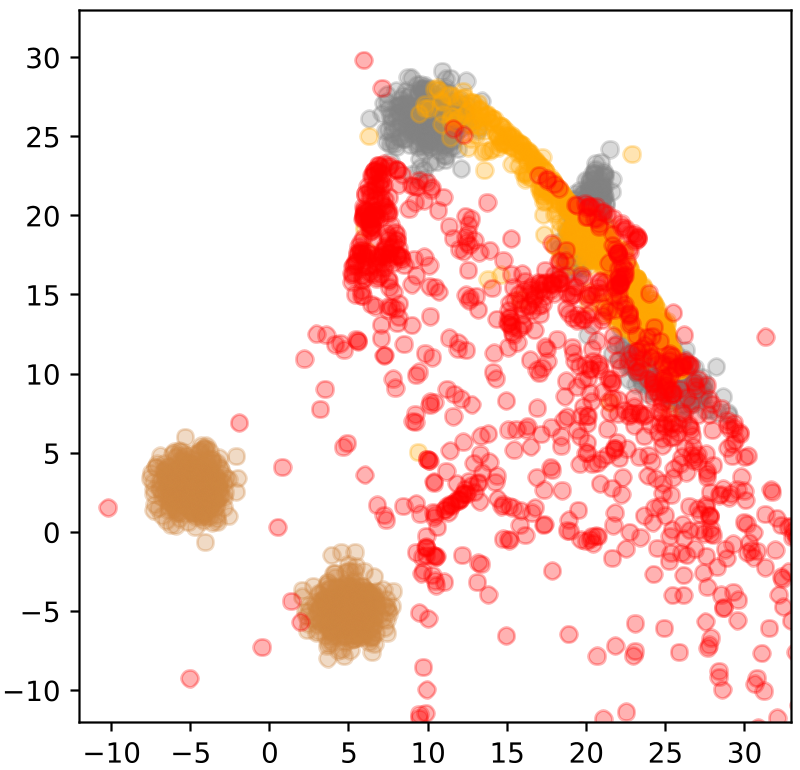} & \includegraphics[width=0.18\textwidth]{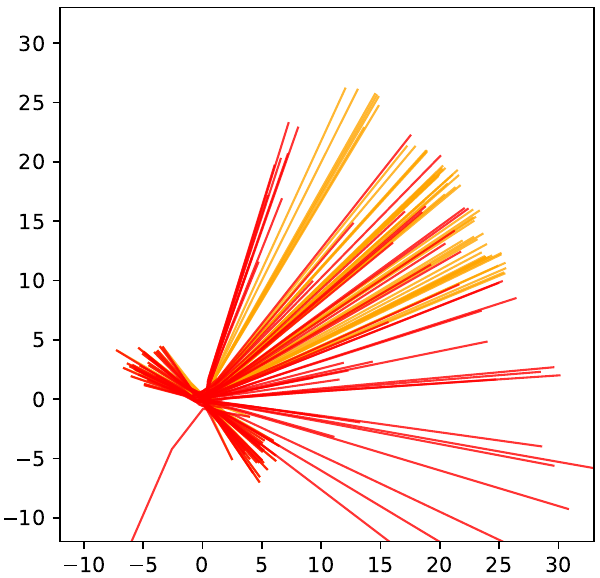} & \includegraphics[width=0.18\textwidth]{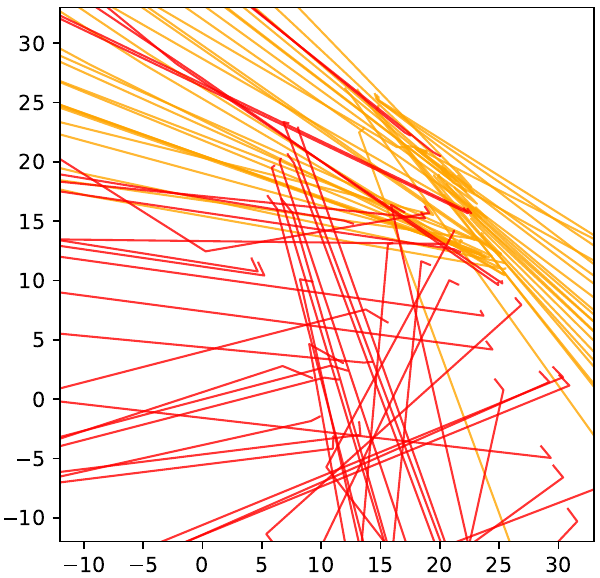}\tabularnewline
\multirow{1}{*}[0.07\textwidth]{50} & \includegraphics[width=0.18\textwidth]{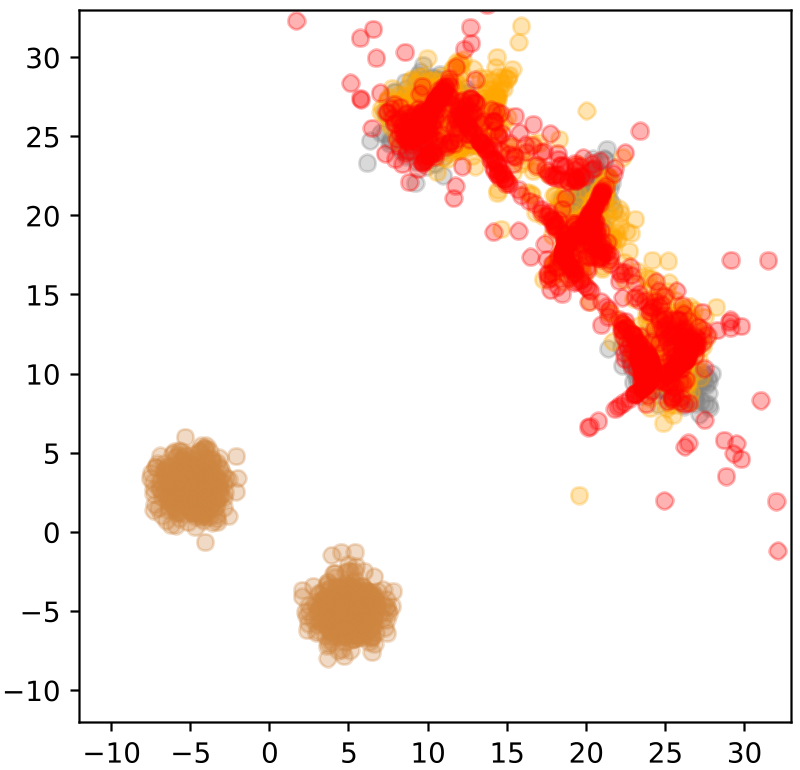} & \includegraphics[width=0.18\textwidth]{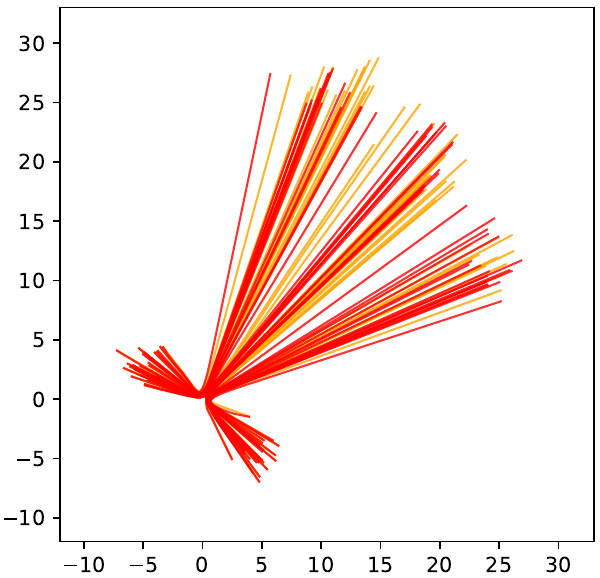} & \includegraphics[width=0.18\textwidth]{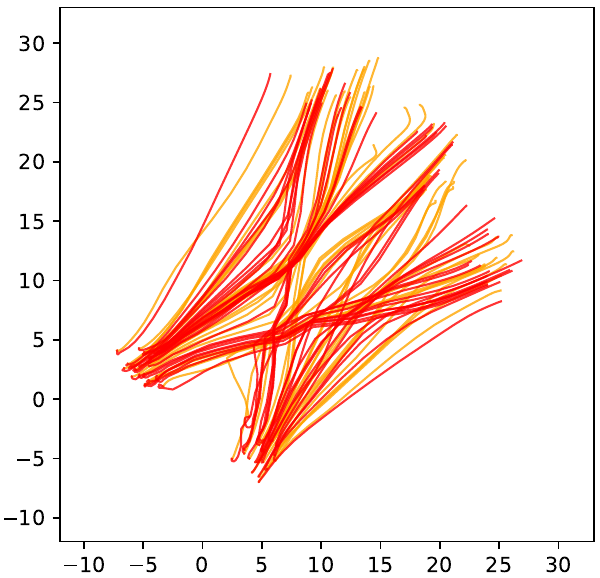} & \includegraphics[width=0.18\textwidth]{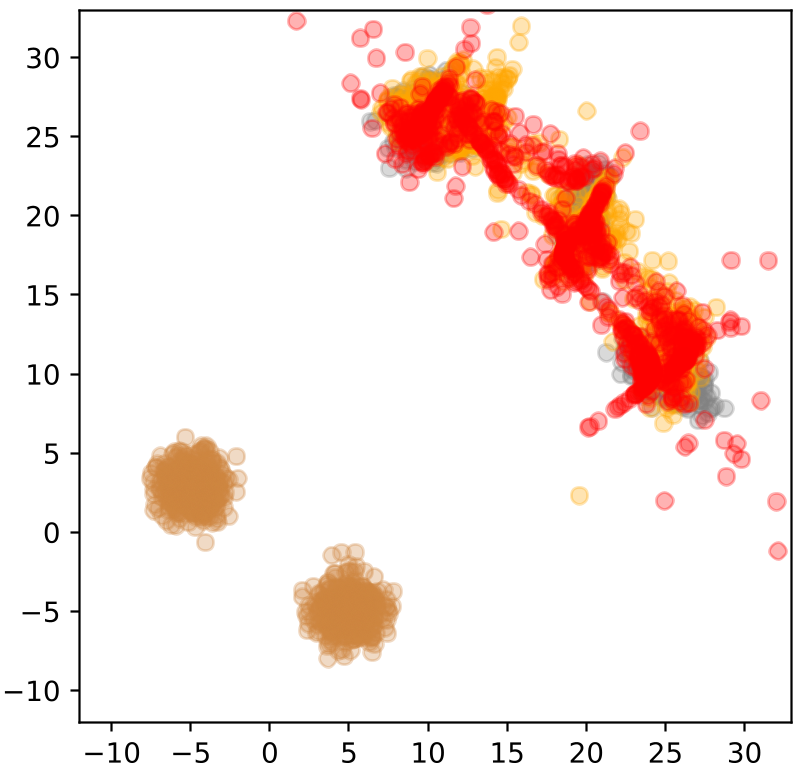} & \includegraphics[width=0.18\textwidth]{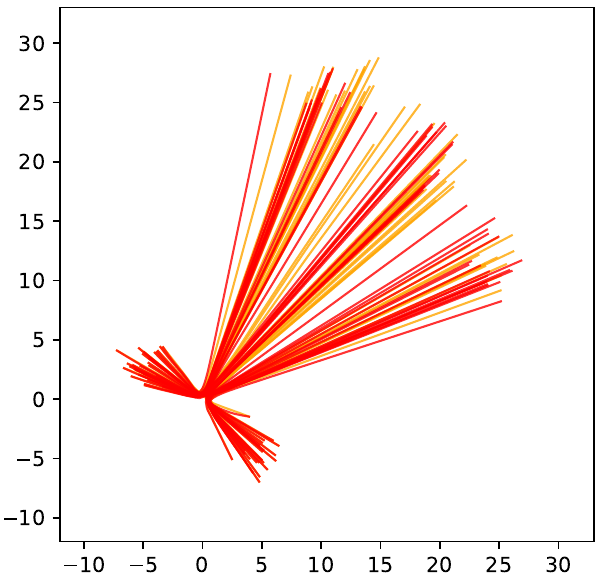} & \includegraphics[width=0.18\textwidth]{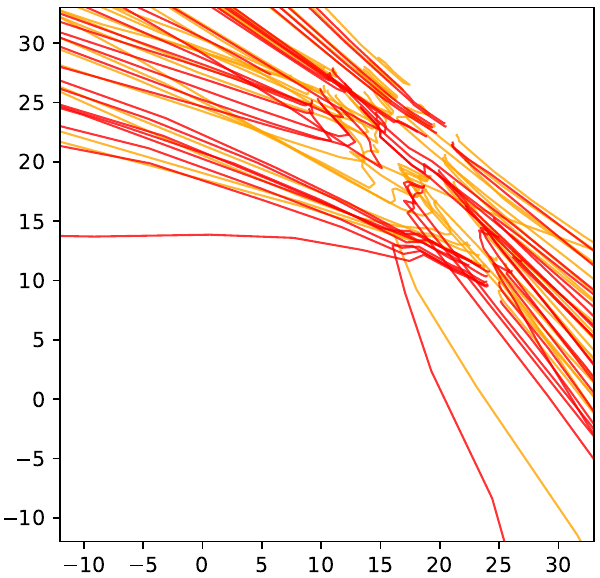}\tabularnewline
\multirow{1}{*}[0.07\textwidth]{1000} & \includegraphics[width=0.18\textwidth]{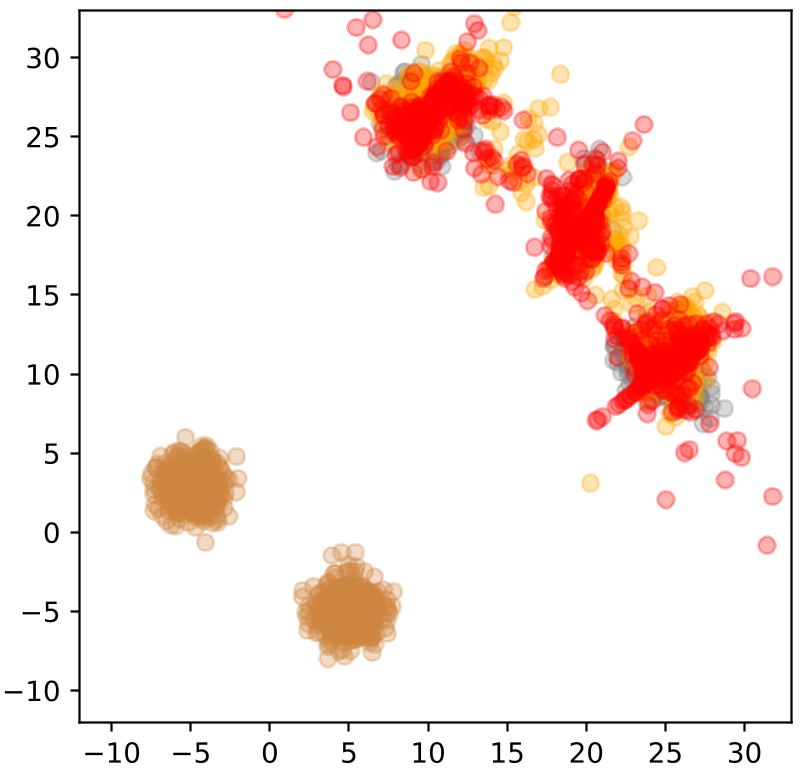} & \includegraphics[width=0.18\textwidth]{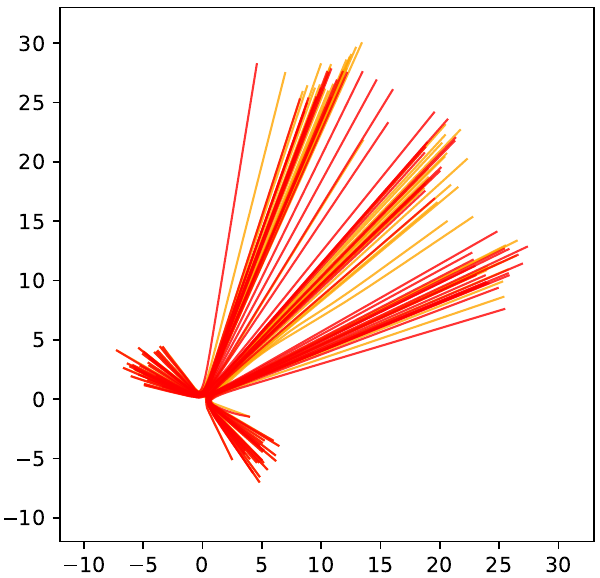} & \includegraphics[width=0.18\textwidth]{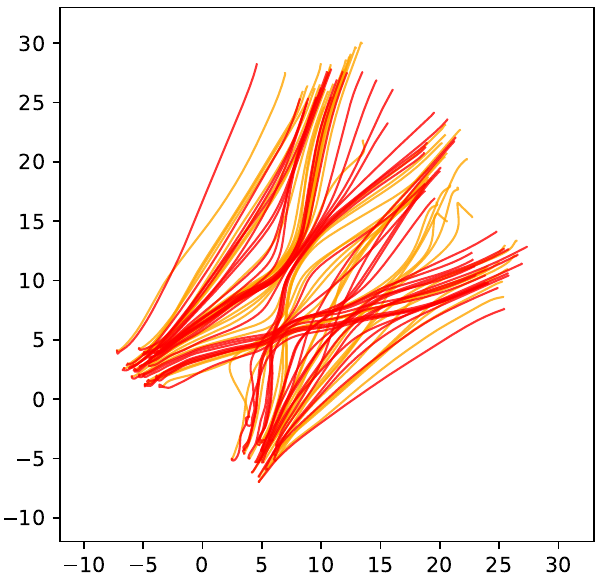} & \includegraphics[width=0.18\textwidth]{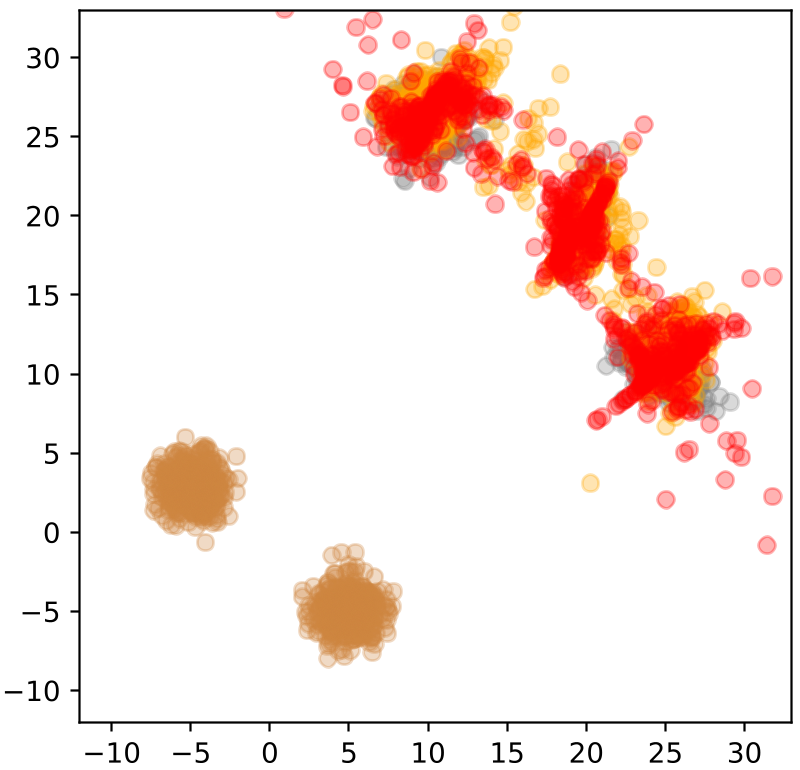} & \includegraphics[width=0.18\textwidth]{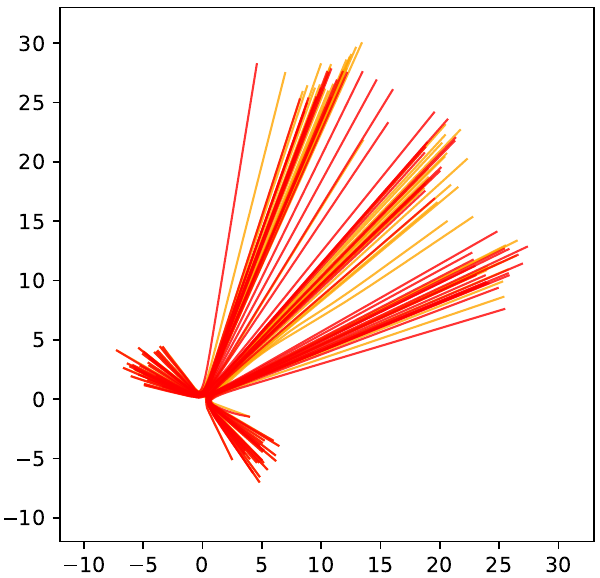} & \includegraphics[width=0.18\textwidth]{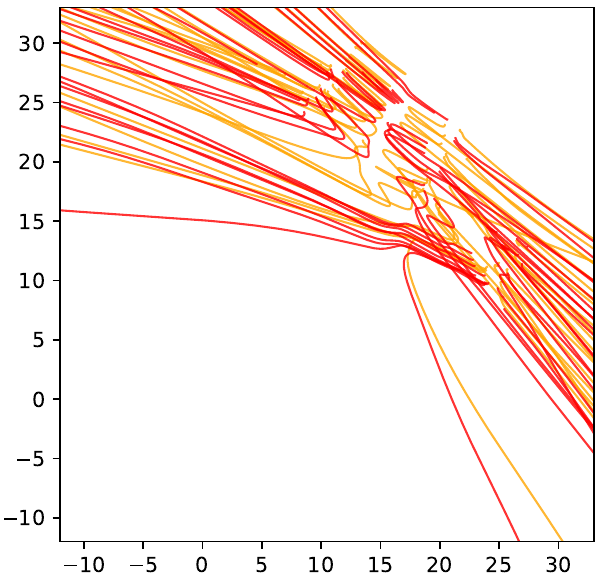}\tabularnewline
 & \multicolumn{3}{c}{$\bar{X}_{t}=\left(1-t\right)X_{0}+tX_{1}$} & \multicolumn{3}{c}{$\bar{X}_{t}=X_{0}+tX_{1}$}\tabularnewline
\end{tabular}}}
\par\end{centering}
\caption{Generated samples (left), $x_{t}$-trajectories (middle) and $\bar{x}_{t}$-trajectories
(right) of the SC flows of the \emph{fifth-degree} process, with different
numbers of update steps $N$, different SC types (two equations in
the bottom), and different types of models ($v_{\theta}$ vs $x_{1,\theta}$).\label{fig:Toy_VeloVSNoise_SC_FifthDegree}}
\end{figure*}

\begin{figure*}
\begin{centering}
{\resizebox{0.92\textwidth}{!}{%
\par\end{centering}
\begin{centering}
\begin{tabular}{cccc|ccc}
$N$ & \multicolumn{6}{c}{\includegraphics[height=0.045\textwidth]{asset/ThreeGauss/legend_sample_velo_vs_noise}\hspace{0.02\textwidth}\includegraphics[height=0.045\textwidth]{asset/ThreeGauss/legend_traj_velo_vs_noise}}\tabularnewline
\multirow{1}{*}[0.07\textwidth]{5} & \includegraphics[width=0.18\textwidth]{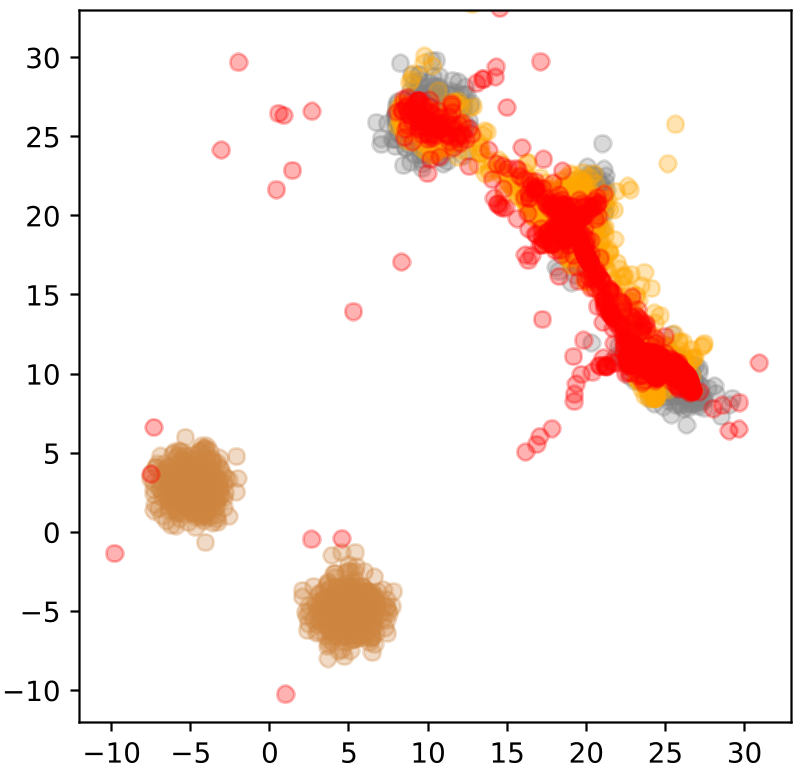} & \includegraphics[width=0.18\textwidth]{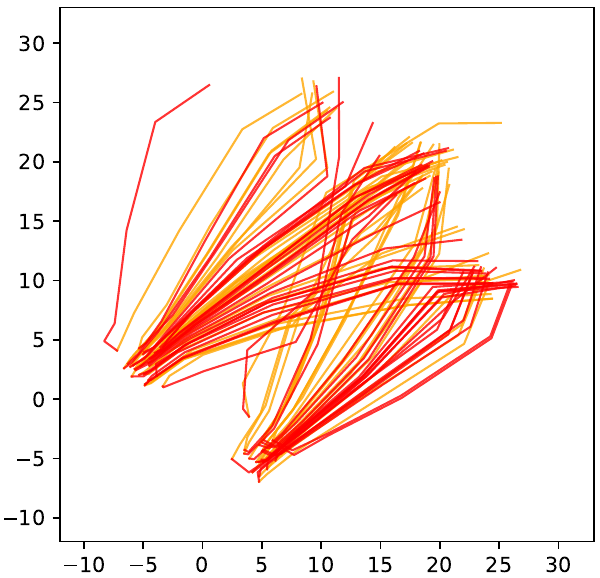} & \includegraphics[width=0.18\textwidth]{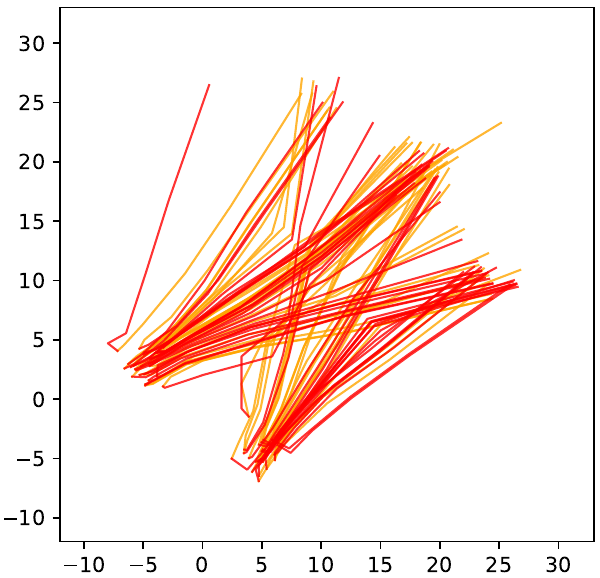} & \includegraphics[width=0.18\textwidth]{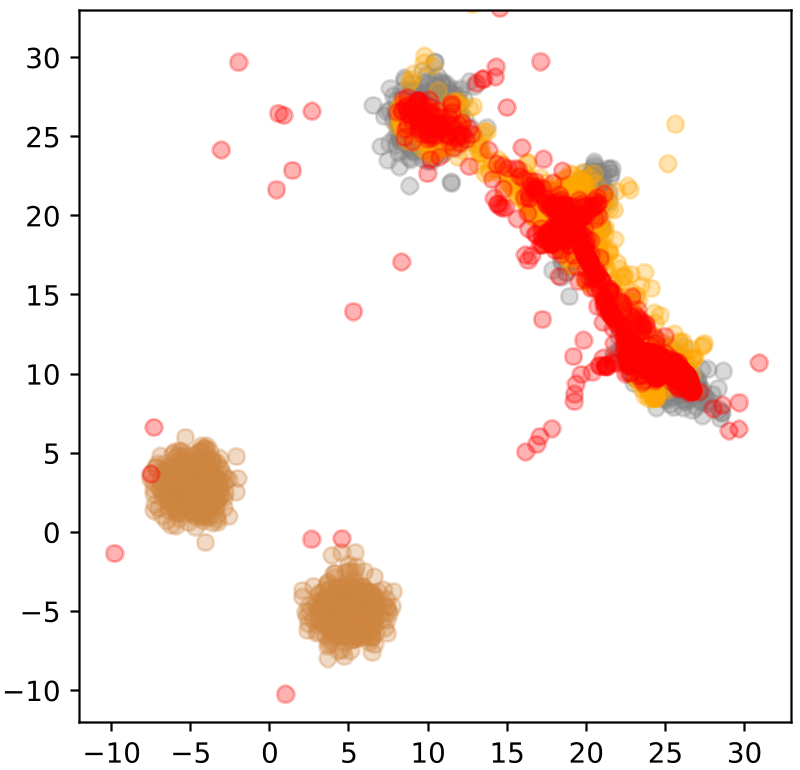} & \includegraphics[width=0.18\textwidth]{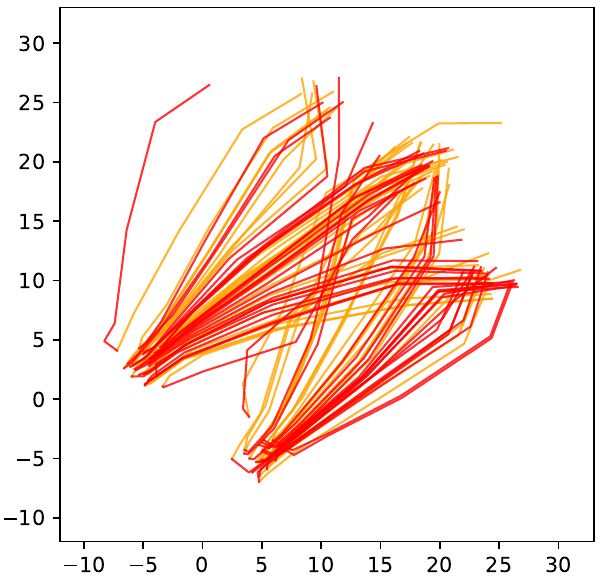} & \includegraphics[width=0.18\textwidth]{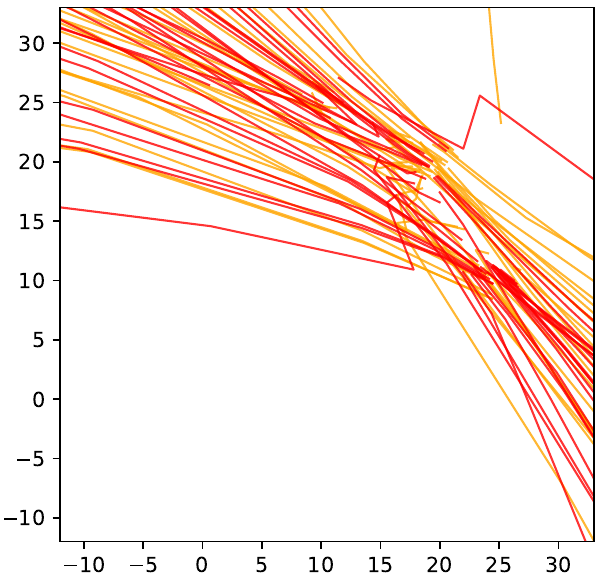}\tabularnewline
\multirow{1}{*}[0.07\textwidth]{50} & \includegraphics[width=0.18\textwidth]{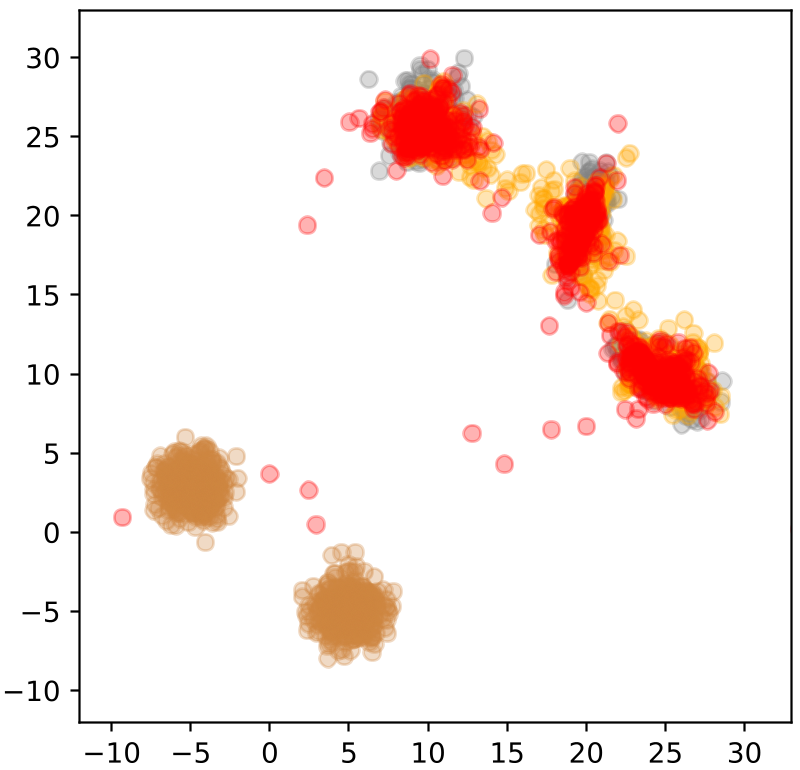} & \includegraphics[width=0.18\textwidth]{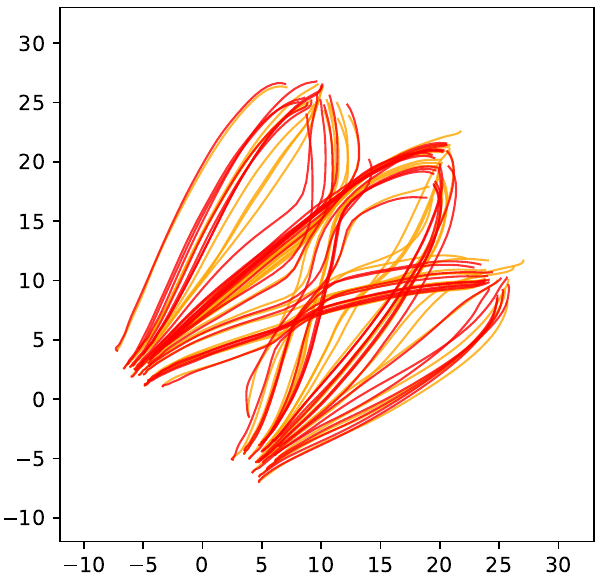} & \includegraphics[width=0.18\textwidth]{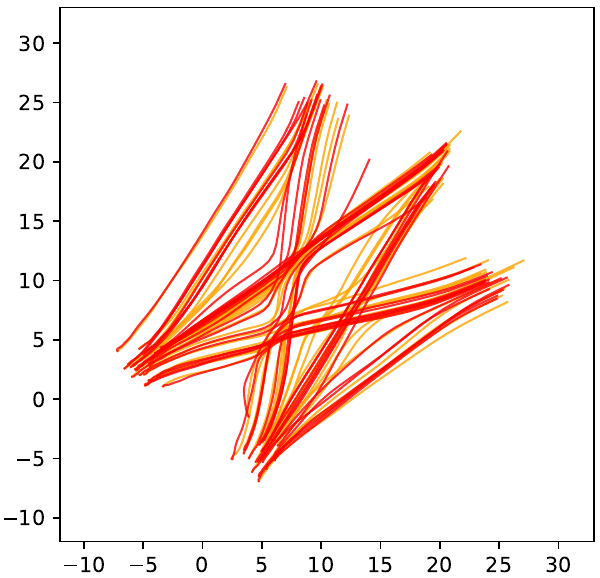} & \includegraphics[width=0.18\textwidth]{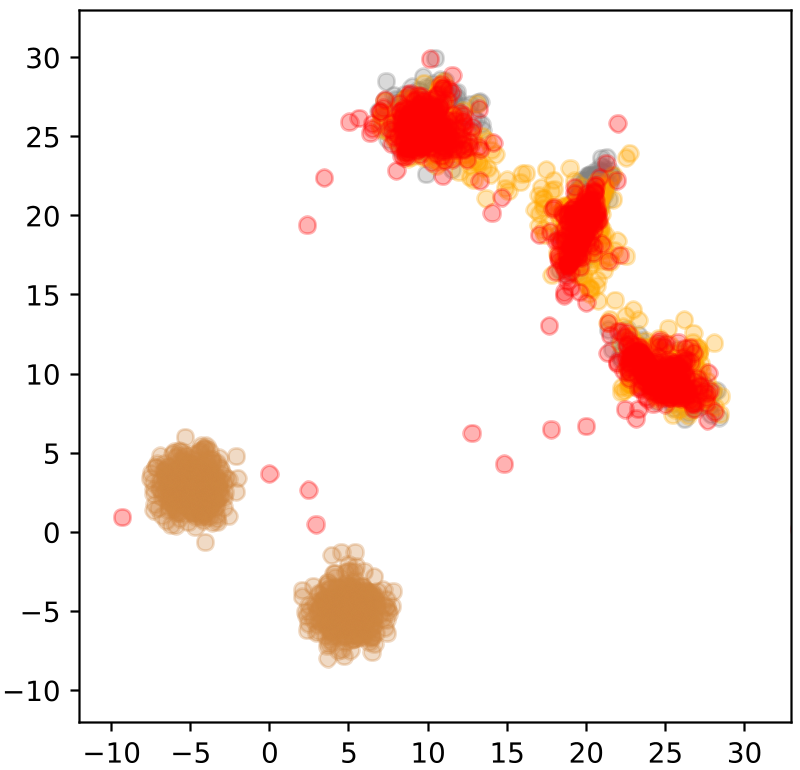} & \includegraphics[width=0.18\textwidth]{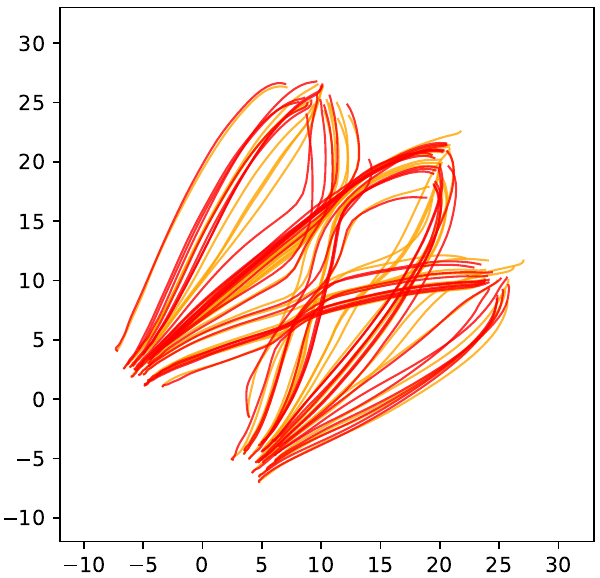} & \includegraphics[width=0.18\textwidth]{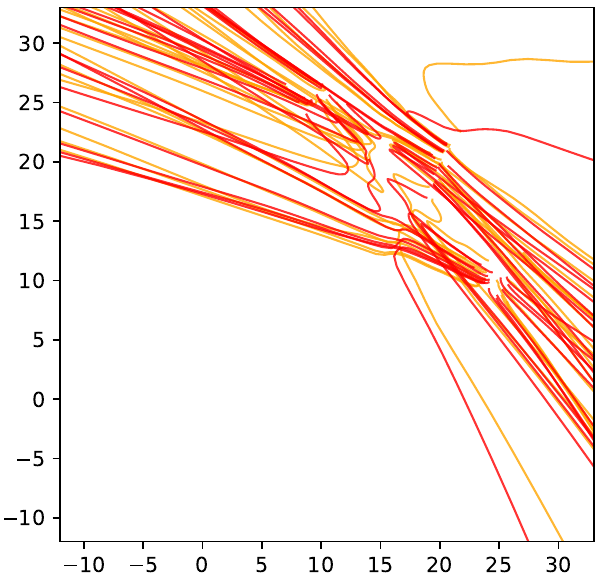}\tabularnewline
\multirow{1}{*}[0.07\textwidth]{1000} & \includegraphics[width=0.18\textwidth]{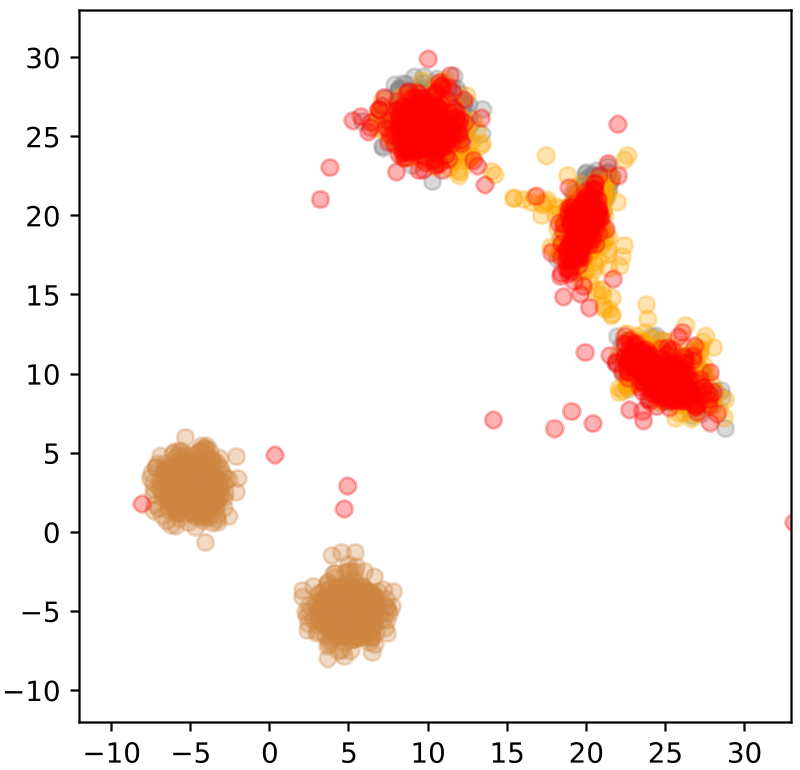} & \includegraphics[width=0.18\textwidth]{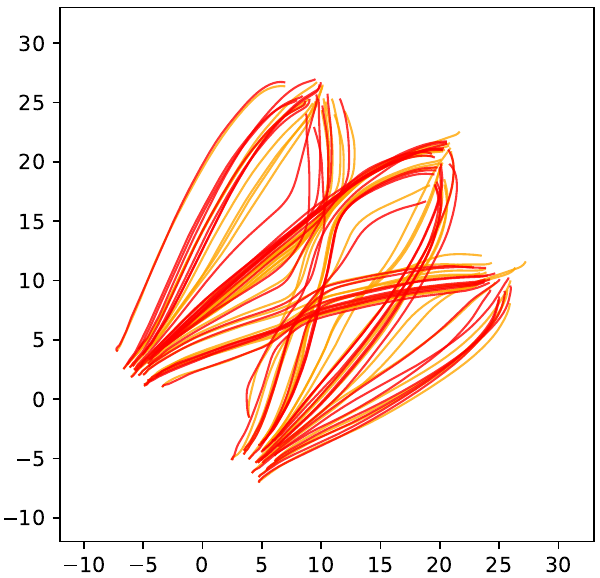} & \includegraphics[width=0.18\textwidth]{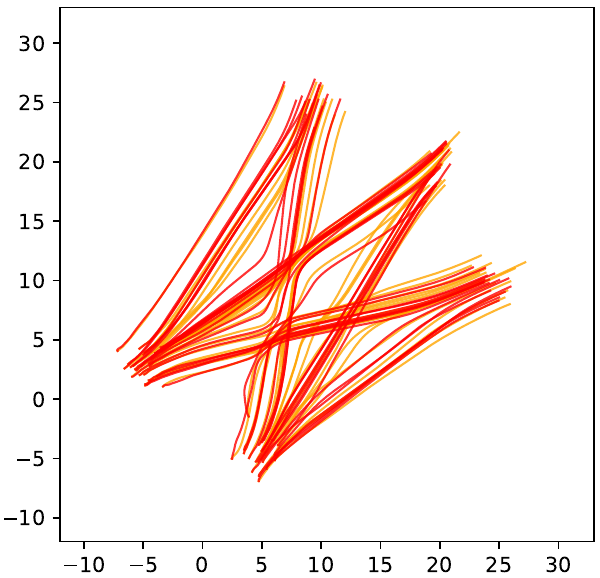} & \includegraphics[width=0.18\textwidth]{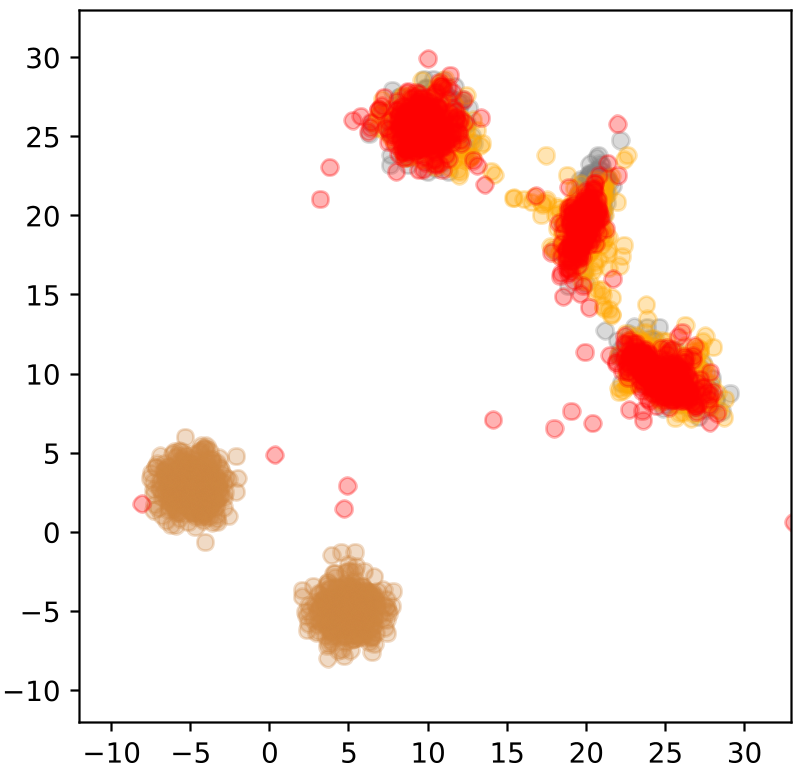} & \includegraphics[width=0.18\textwidth]{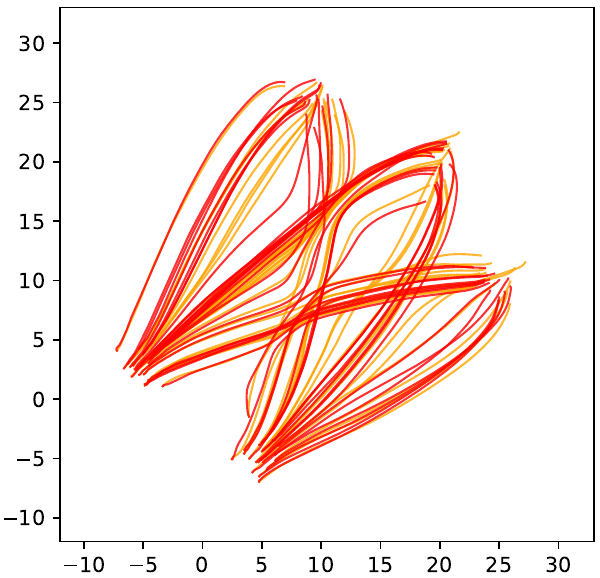} & \includegraphics[width=0.18\textwidth]{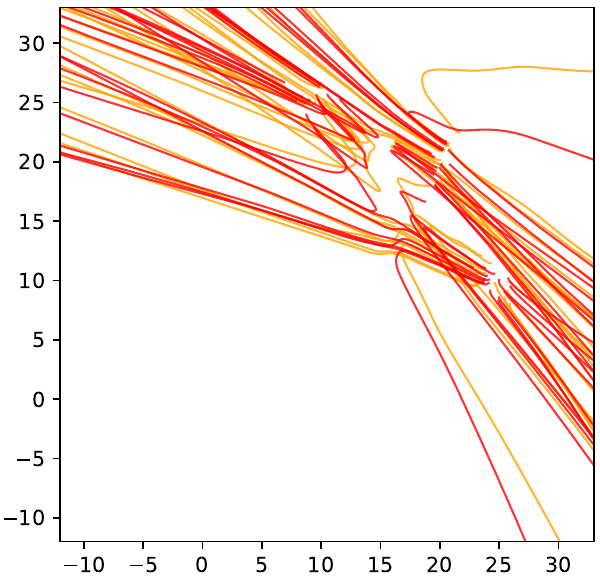}\tabularnewline
 & \multicolumn{3}{c}{$\bar{X}_{t}=\left(1-t\right)X_{0}+tX_{1}$} & \multicolumn{3}{c}{$\bar{X}_{t}=X_{0}+tX_{1}$}\tabularnewline
\end{tabular}}}
\par\end{centering}
\caption{Generated samples (left), $x_{t}$-trajectories (middle) and $\bar{x}_{t}$-trajectories
(right) of the SC flows of the \emph{VP-SDE-like} process, with different
numbers of update steps $N$, different SC types (two equations in
the bottom), and different types of models ($v_{\theta}$ vs $x_{1,\theta}$).\label{fig:Toy_VeloVSNoise_SC_VPSDELike}}
\end{figure*}

\begin{figure*}
\begin{centering}
{\resizebox{0.92\textwidth}{!}{%
\par\end{centering}
\begin{centering}
\begin{tabular}{cccc|ccc}
$N$ & \multicolumn{6}{c}{\includegraphics[height=0.045\textwidth]{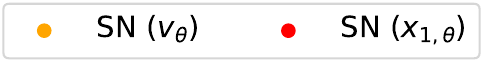}\hspace{0.02\textwidth}\includegraphics[height=0.045\textwidth]{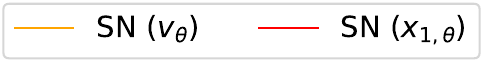}}\tabularnewline
\multirow{1}{*}[0.07\textwidth]{5} & \includegraphics[width=0.18\textwidth]{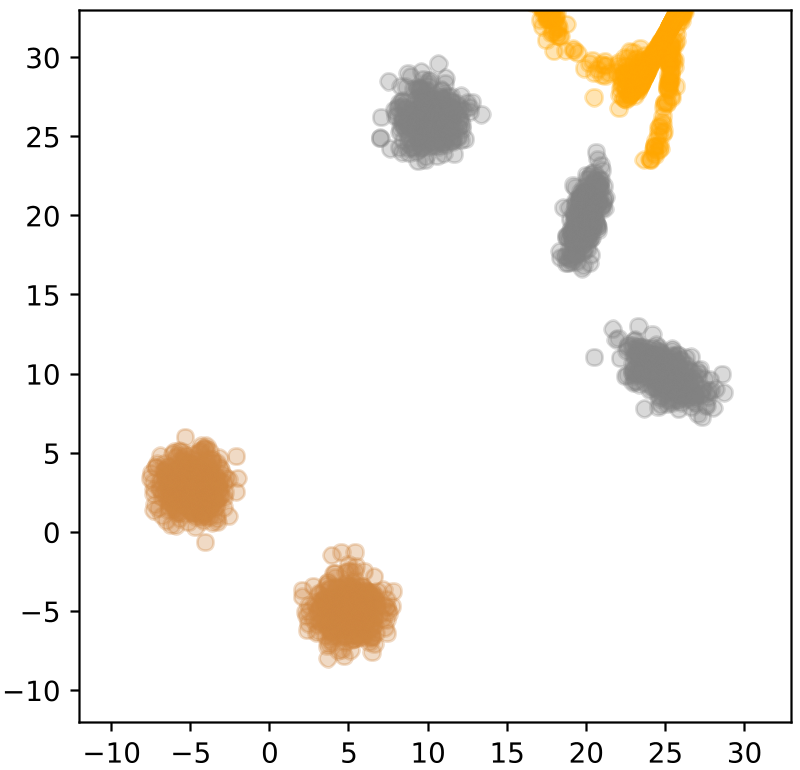} & \includegraphics[width=0.18\textwidth]{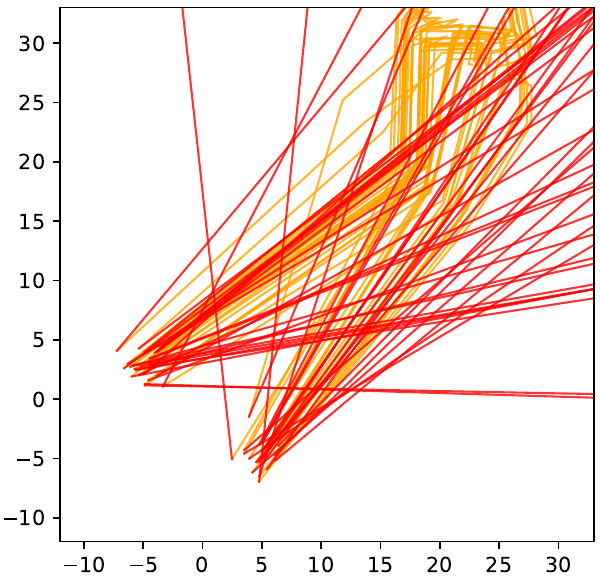} & \includegraphics[width=0.18\textwidth]{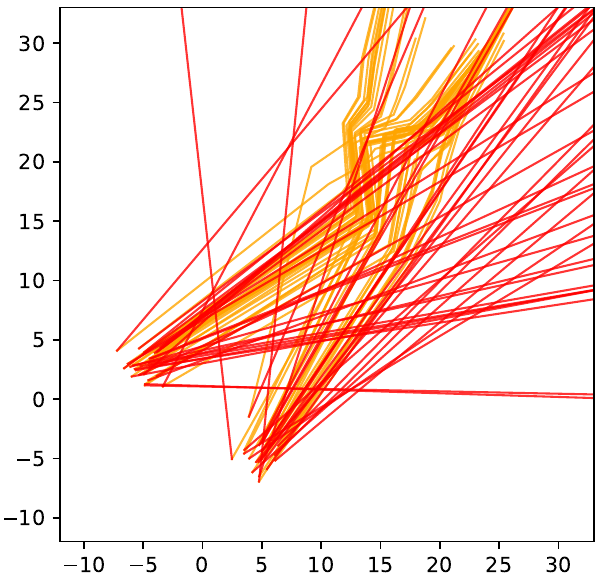} & \includegraphics[width=0.18\textwidth]{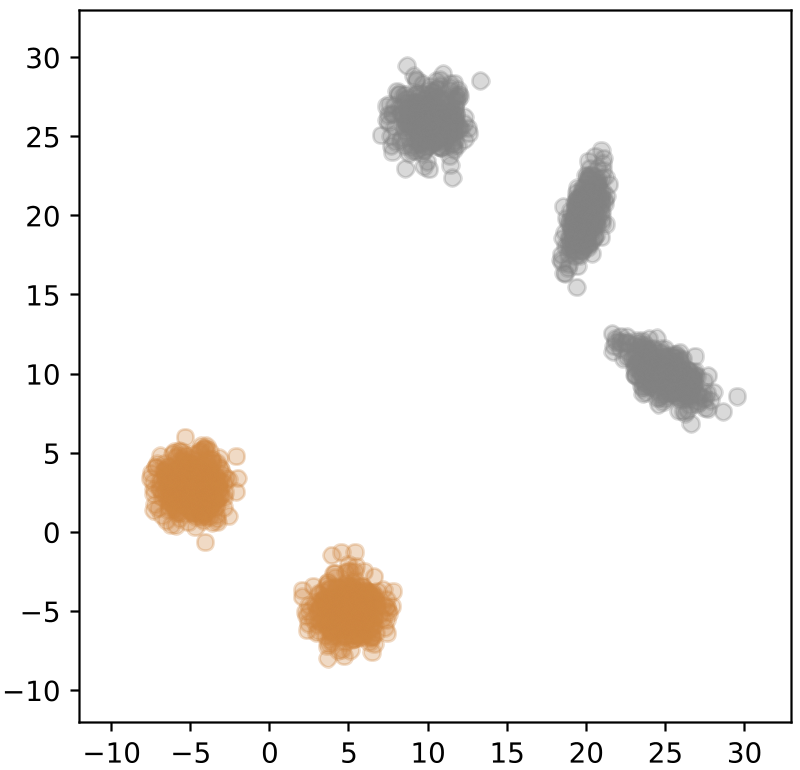} & \includegraphics[width=0.18\textwidth]{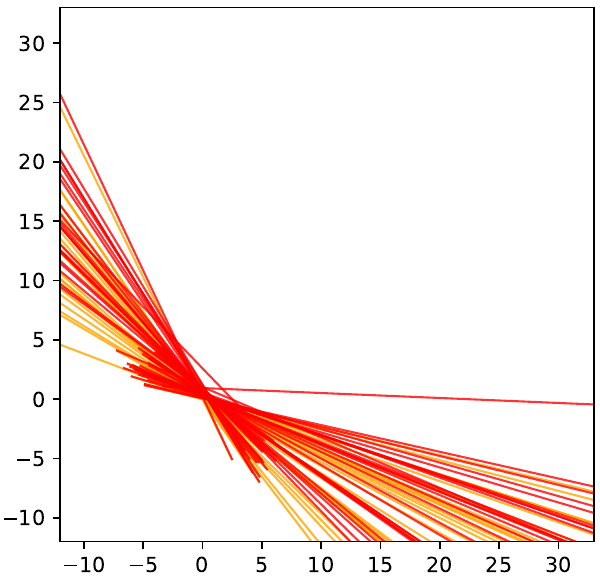} & \includegraphics[width=0.18\textwidth]{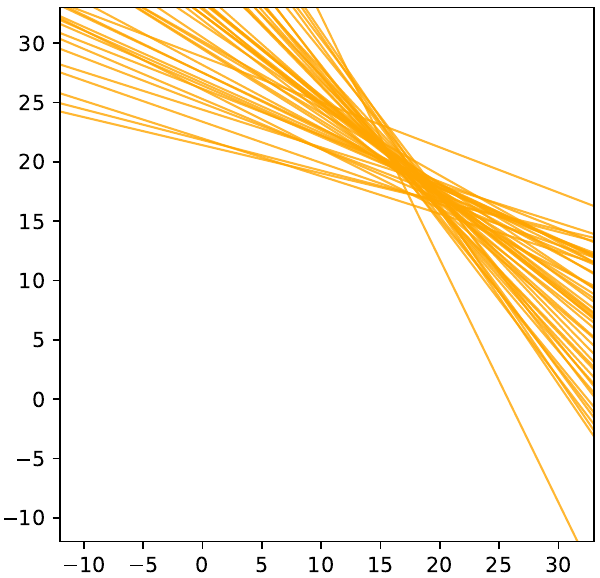}\tabularnewline
\multirow{1}{*}[0.07\textwidth]{50} & \includegraphics[width=0.18\textwidth]{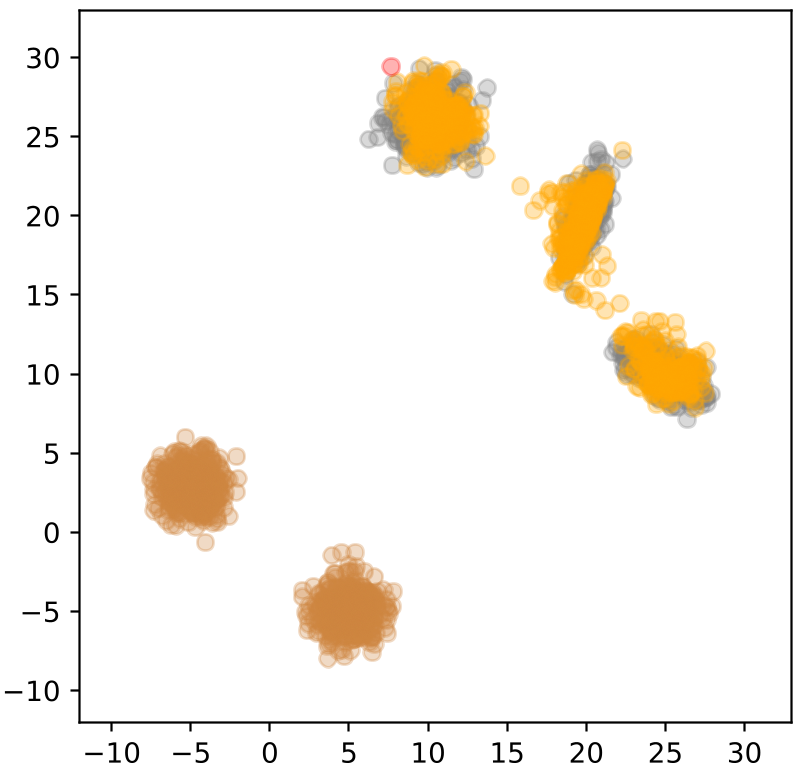} & \includegraphics[width=0.18\textwidth]{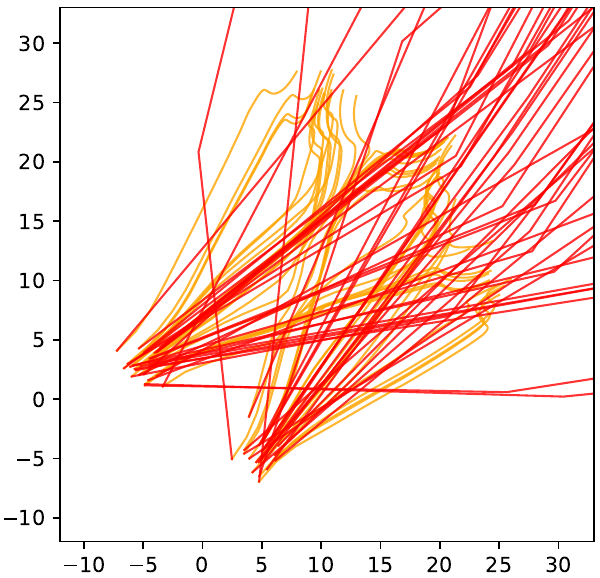} & \includegraphics[width=0.18\textwidth]{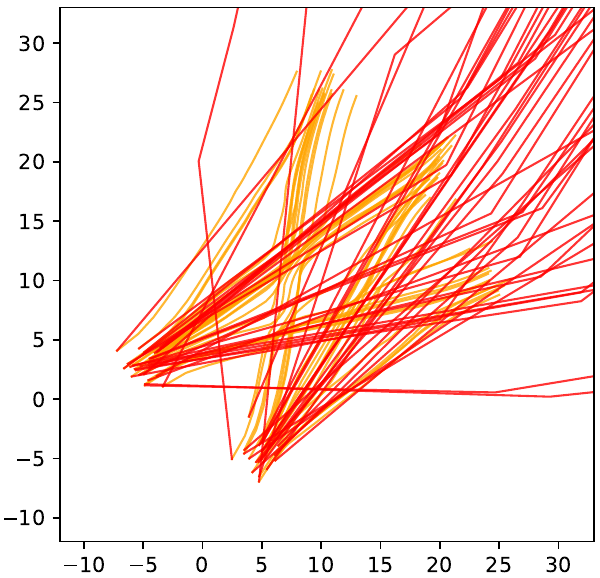} & \includegraphics[width=0.18\textwidth]{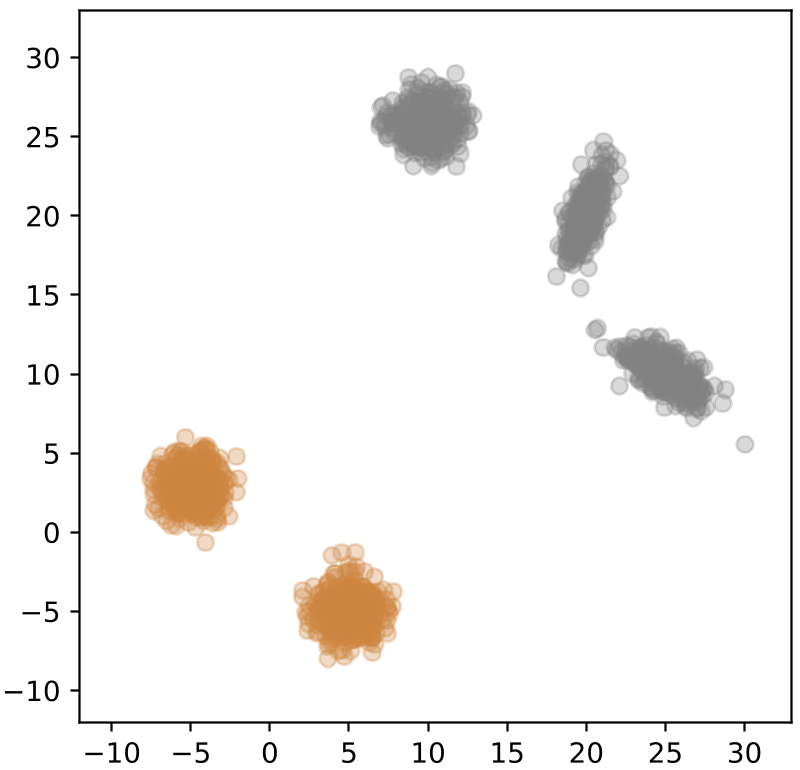} & \includegraphics[width=0.18\textwidth]{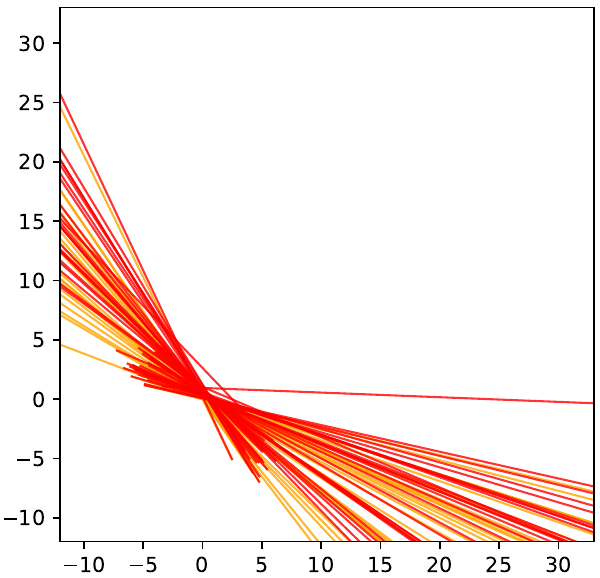} & \includegraphics[width=0.18\textwidth]{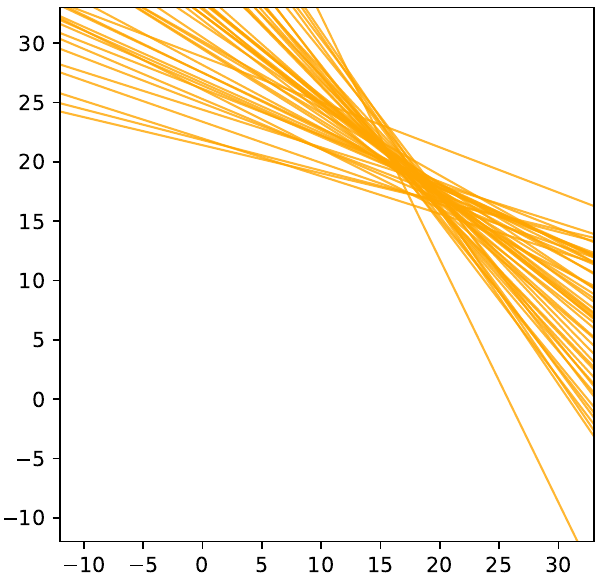}\tabularnewline
\multirow{1}{*}[0.07\textwidth]{1000} & \includegraphics[width=0.18\textwidth]{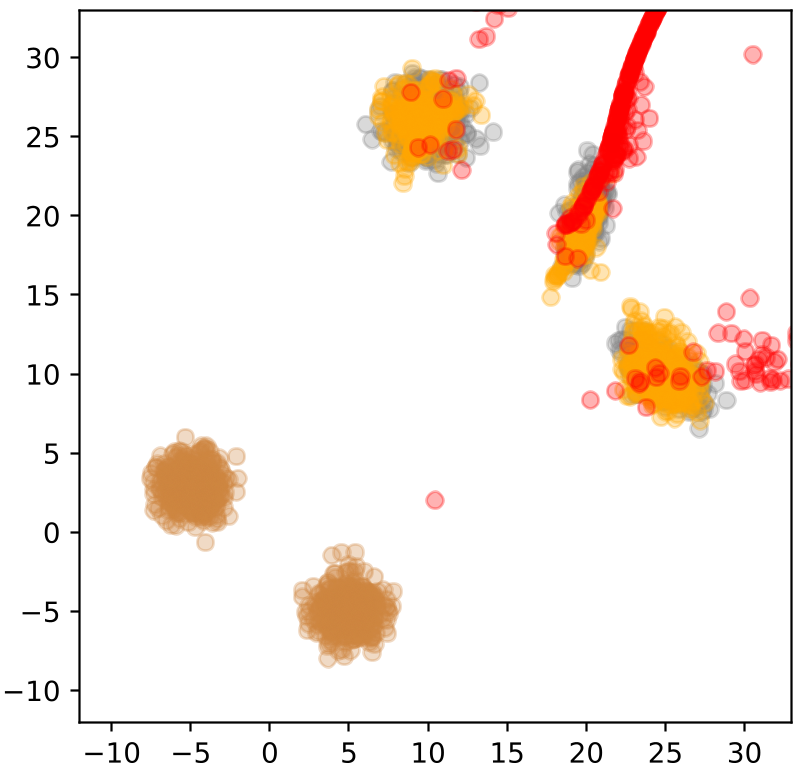} & \includegraphics[width=0.18\textwidth]{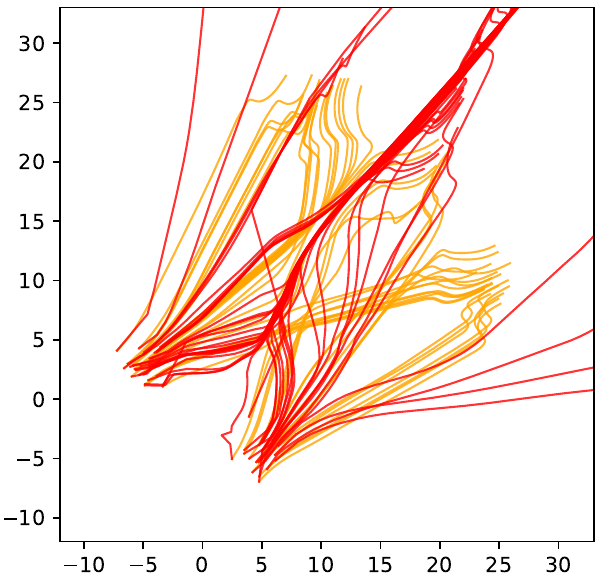} & \includegraphics[width=0.18\textwidth]{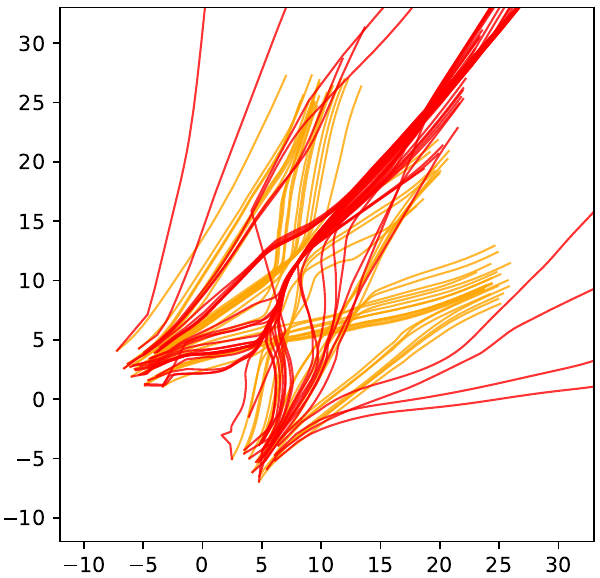} & \includegraphics[width=0.18\textwidth]{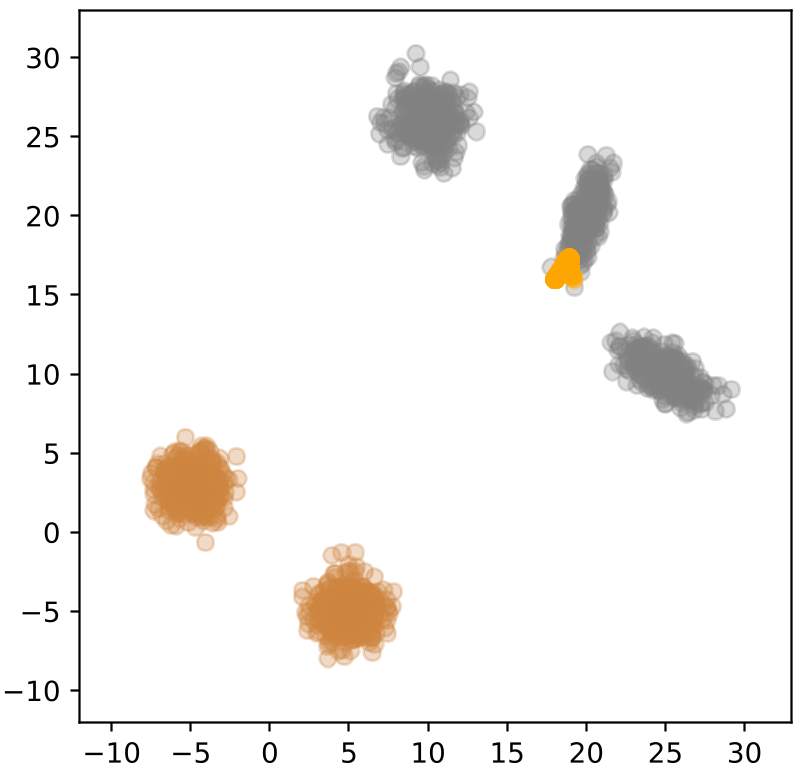} & \includegraphics[width=0.18\textwidth]{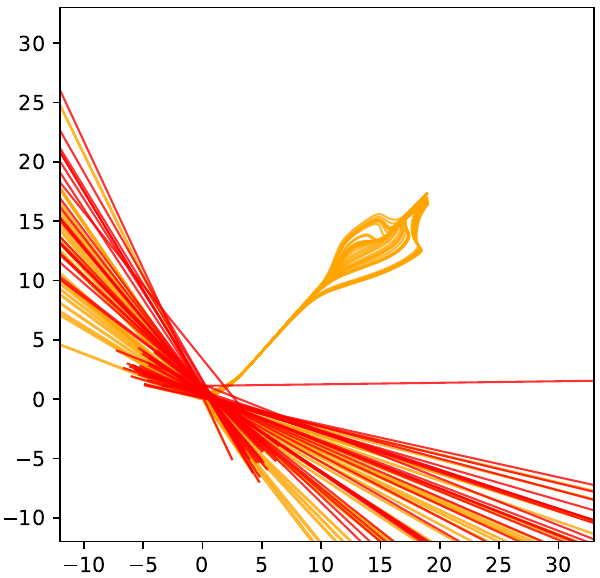} & \includegraphics[width=0.18\textwidth]{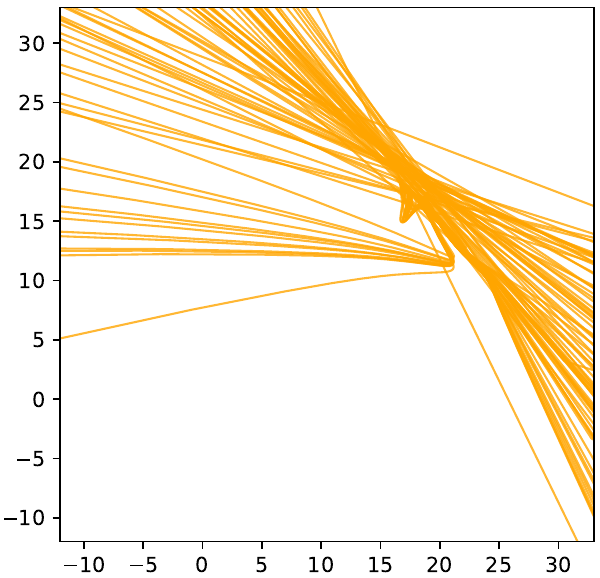}\tabularnewline
 & \multicolumn{3}{c}{$\bar{X}_{t}=\left(1-t\right)X_{0}+tX_{1}$} & \multicolumn{3}{c}{$\bar{X}_{t}=X_{0}+tX_{1}$}\tabularnewline
\end{tabular}}}
\par\end{centering}
\caption{Generated samples (left), $x_{t}$-trajectories (middle) and $\tilde{x}_{t}$-trajectories
(right) of the SN flows of the \emph{third-degree} process, with different
numbers of update steps $N$, different SC types (two equations in
the bottom), and different types of models ($v_{\theta}$ vs $x_{1,\theta}$).\label{fig:Toy_VeloVSNoise_SN_ThirdDegree}}
\end{figure*}

As discussed in Section~\ref{subsec:Transforming-a-flow}, we can
learn a posterior flow by modeling either $v^{*}\left(x_{t},t\right)$
or $x_{1|t}$. A model of $v^{*}\left(x_{t},t\right)$ is trained
via the velocity matching loss in Eq.~\ref{eq:velocity_matching_loss_reframed}
while a model of $x_{1|t}$ is trained via the noise matching loss
in Eq.~\ref{eq:noise_matching}. Fig.~\ref{fig:Toy_VeloVSNoise_SC_ThirdDegree}
shows the generation results w.r.t. two implementations of the SC
flow of the third-degree process, one using a model $v_{\theta}\left(x_{t},t\right)$
of $v^{*}\left(x_{t},t\right)$ (\textcolor{orange}{orange}) and
the other using a model $x_{1,\theta}\left(x_{t},t\right)$ of $x_{1|t}$
(\textcolor{red}{red}). It is evident that the SC flow using $v_{\theta}\left(x_{t},t\right)$
greatly outperforms the counterpart using $x_{1,\theta}\left(x_{t},t\right)$
for either of the two formulas $\bar{X}_{t}=\left(1-t\right)X_{0}+tX_{1}$
and $\bar{X}_{t}=X_{0}+tX_{1}$ and any number of update steps $N$.
Our experiments with the fifth-degree and the VP-SDE-like processes
also demonstrate that using $v_{\theta}\left(x_{t},t\right)$ leads
to better results than using $x_{1,\theta}\left(x_{t},t\right)$,
as depicted in Figs.~\ref{fig:Toy_VeloVSNoise_SC_FifthDegree}, \ref{fig:Toy_VeloVSNoise_SC_VPSDELike},
although the performance gaps in these cases are not as significant
as in the case of the third-degree process. These results suggest
that it is better to learn $v_{\theta}\left(x_{t},t\right)$ than
learn $x_{1,\theta}\left(x_{t},t\right)$. A possible reason is that
for different samples $x_{t}$ computed from the same $x_{1}$ and
different $x_{0}$ during training, $v_{\theta}\left(x_{t},t\right)$
will be matched to different target velocities $v_{\phi}\left(x_{0},x_{1},t\right)=\dot{a}_{t}x_{0}+\dot{\sigma}_{t}x_{1}$
while $x_{1,\theta}\left(x_{t},t\right)$ will be matched to the same
target sample $x_{1}$. This means $v_{\theta}\left(x_{t},t\right)$
tends to be more adaptive to the change of $x_{t}$ and produce more
accurate outputs than $x_{1,\theta}\left(x_{t},t\right)$ . We also
observed the same phenomenon for the SN flow, i.e., modeling the SN
flow using $v_{\theta}\left(x_{t},t\right)$ achieves better results
than using $x_{1,\theta}\left(x_{t},t\right)$, as shown in Fig.~\ref{fig:Toy_VeloVSNoise_SN_ThirdDegree}.

\subsection{Text-to-Image Generation with Stable Diffusion}

\begin{figure*}
\begin{centering}
{\resizebox{\textwidth}{!}{%
\par\end{centering}
\begin{centering}
\begin{tabular}{ccc}
\includegraphics[width=0.3\textwidth]{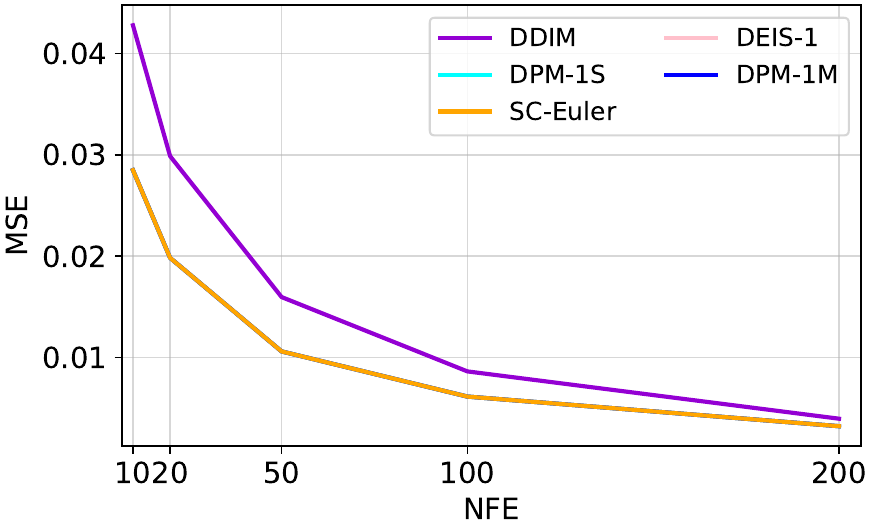} & \includegraphics[width=0.3\textwidth]{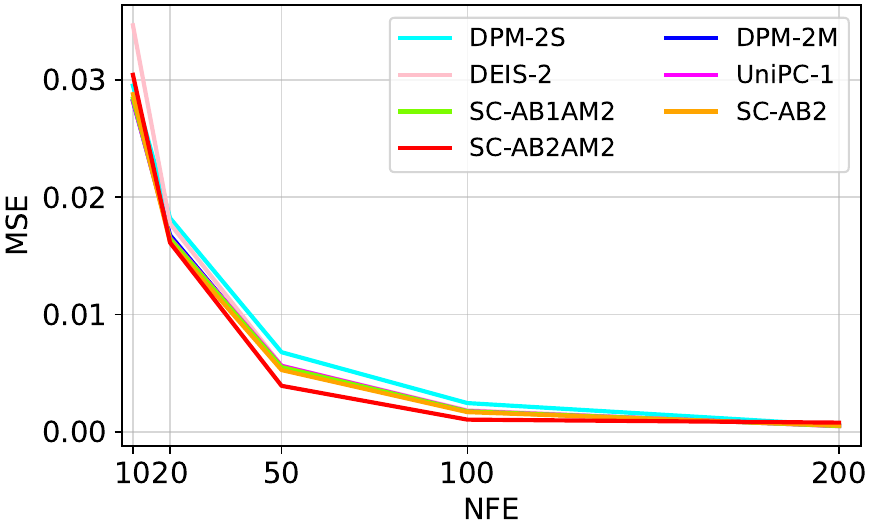} & \includegraphics[width=0.3\textwidth]{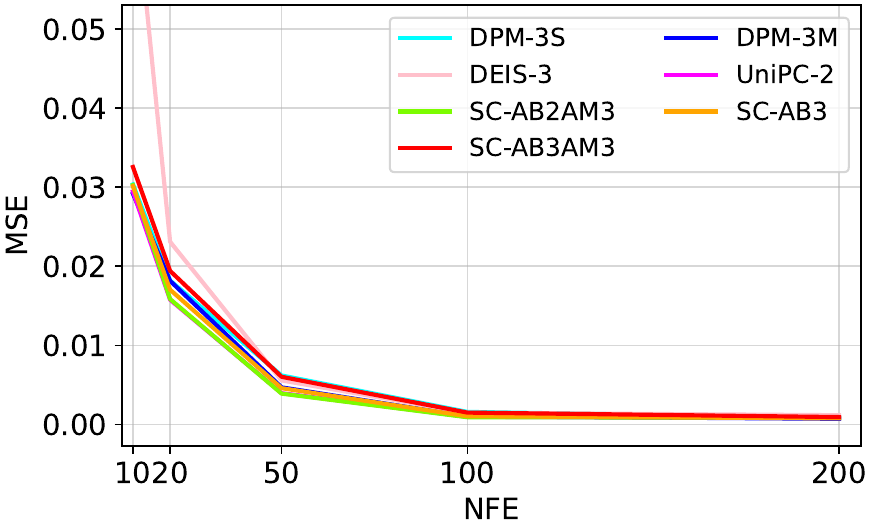}\tabularnewline
(a) First-order & (b) Second-order & (c) Third-order\tabularnewline
\end{tabular}}}
\par\end{centering}
\caption{MSEs between 1000 512$\times$512 images generated by a 1000-step
DDIM and those generated by our proposed solvers (SC-{*}) and baseline
solvers. The model is Stable Diffusion and the text prompts are COCO2014
captions. DPM stands for DPM-Solver++.\label{fig:MSE-StableDifussion-image}}
\end{figure*}

Fig.~\ref{fig:MSE-StableDifussion-image} shows the convergence errors
of our solvers and baselines when the number of function evaluations
(NFE) increases from 10 to 200. It is evident that all solvers converge
as the NFE increases and higher-order solvers tend to converge faster
than lower-order counterparts.

\begin{figure*}
\begin{centering}
{\resizebox{\textwidth}{!}{%
\par\end{centering}
\begin{centering}
\begin{tabular}{cccccc}
Real Image & DDIM{*} & DDIM & DEIS-1 & DPM-Solver++ (1M) & SC-Euler\tabularnewline
\includegraphics[width=0.2\textwidth]{asset/CocoCaption/gen_images/groundtruth/img36} & \includegraphics[width=0.2\textwidth]{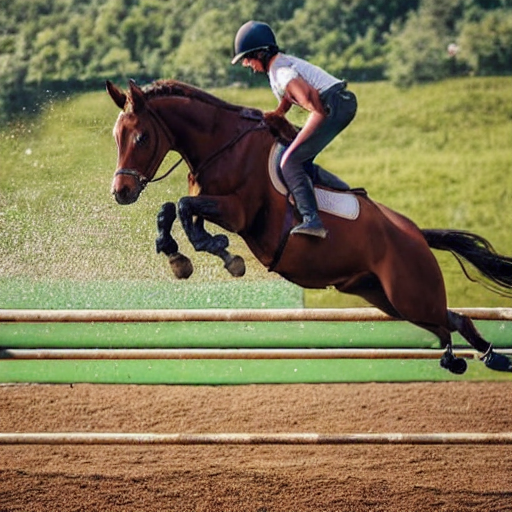} & \includegraphics[width=0.2\textwidth]{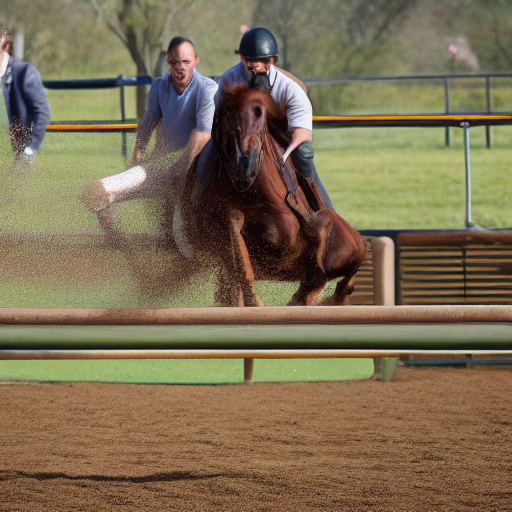} & \includegraphics[width=0.2\textwidth]{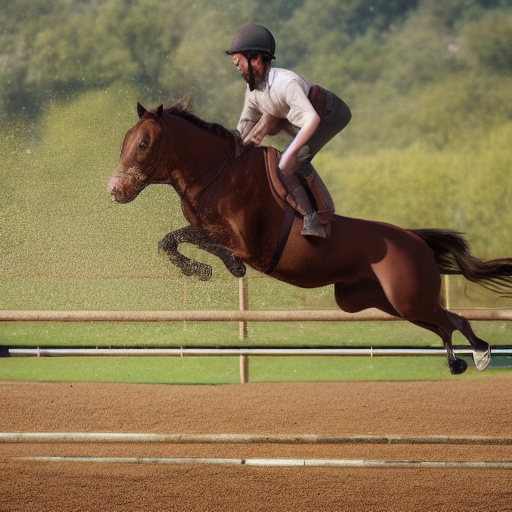} & \includegraphics[width=0.2\textwidth]{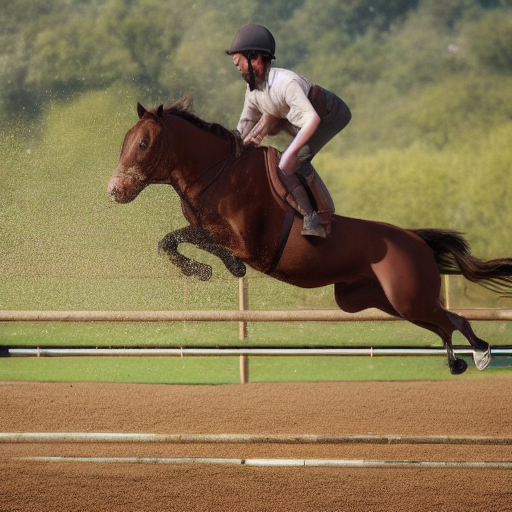} & \includegraphics[width=0.2\textwidth]{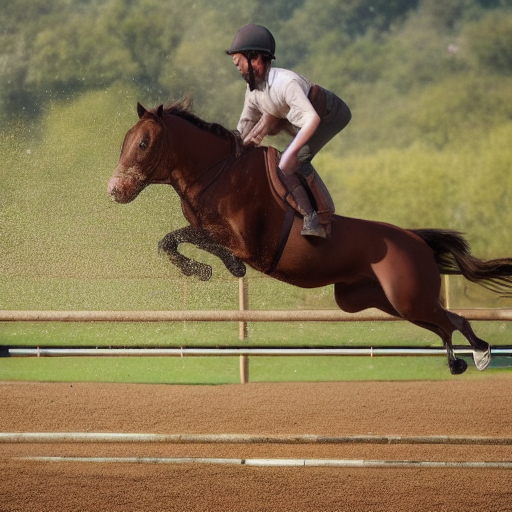}\tabularnewline
\multicolumn{5}{c}{``A man jumping a brown horse over an obstacle.''} & \tabularnewline
 &  &  &  &  & \tabularnewline
\includegraphics[width=0.2\textwidth]{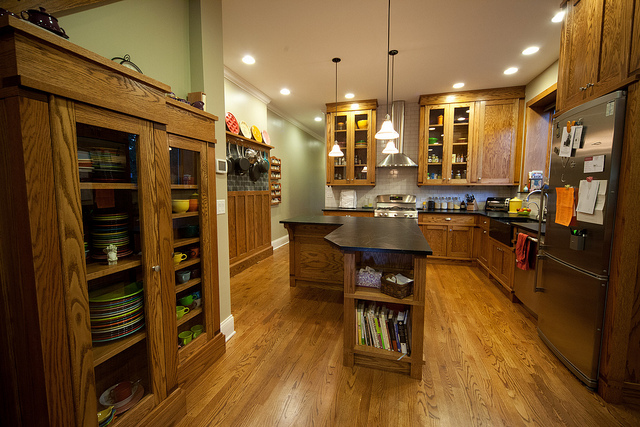} & \includegraphics[width=0.2\textwidth]{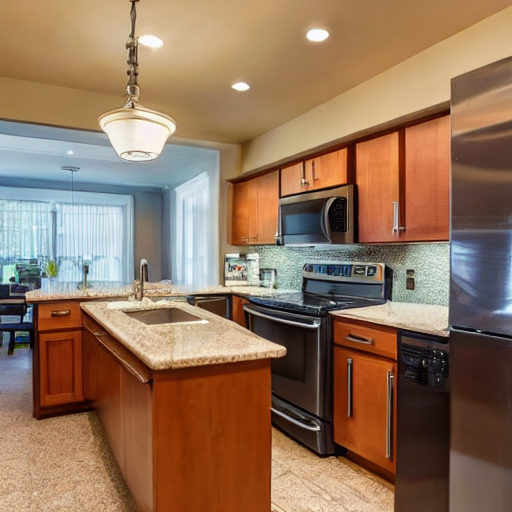} & \includegraphics[width=0.2\textwidth]{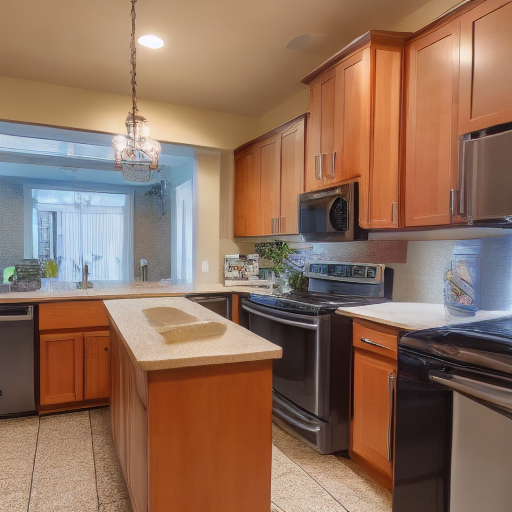} & \includegraphics[width=0.2\textwidth]{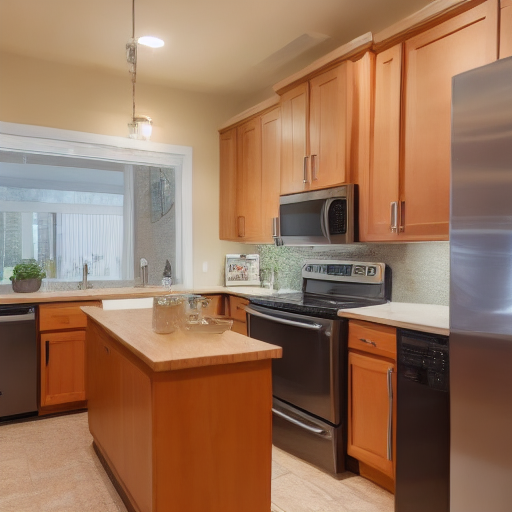} & \includegraphics[width=0.2\textwidth]{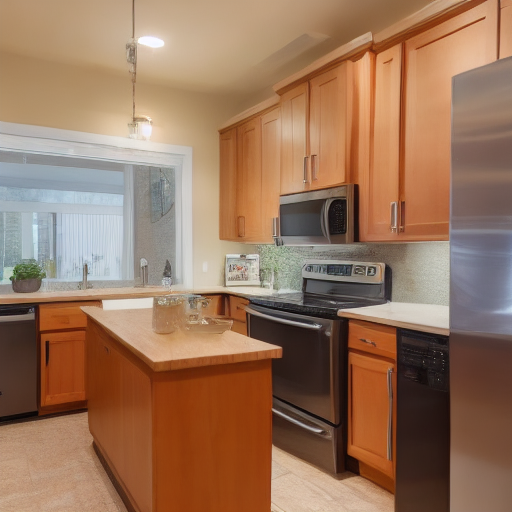} & \includegraphics[width=0.2\textwidth]{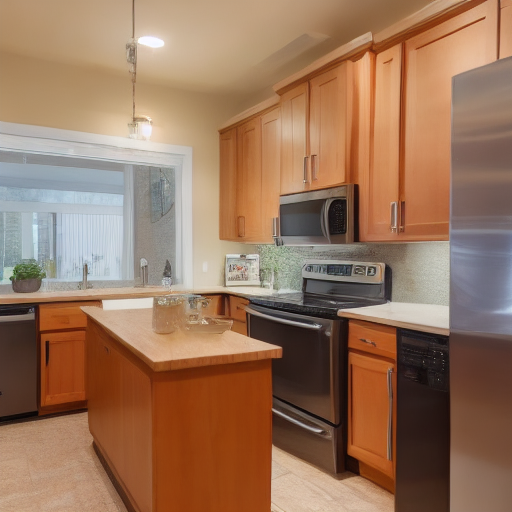}\tabularnewline
\multicolumn{6}{c}{``A kitchen with a center island next to a refrigerator freezer.''}\tabularnewline
\end{tabular}}}
\par\end{centering}
\caption{Real images (first column) and images generated by first-order numerical
solvers with NFE=10. The model is Stable Diffusion and the text prompts
are COCO2014 captions.\label{fig:GenImages_StableDiffusion_Order1}}
\end{figure*}

\begin{figure*}
\begin{centering}
{\resizebox{\textwidth}{!}{%
\par\end{centering}
\begin{centering}
\begin{tabular}{cccccc}
DEIS-2 & DPM-Solver++ (2M) & UniPC-1 & SC-AB2 & SC-AB1AM2 & SC-AB2AM2\tabularnewline
\includegraphics[width=0.2\textwidth]{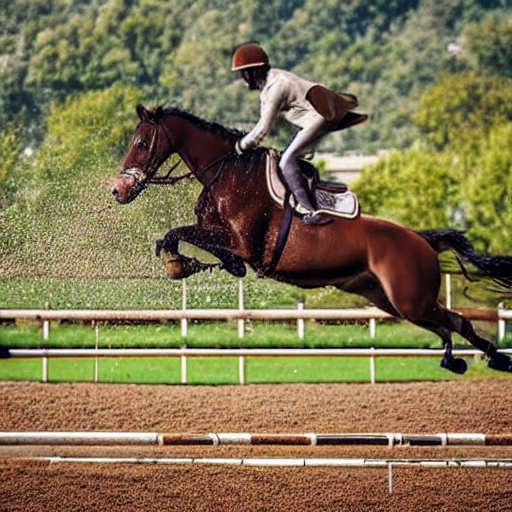} & \includegraphics[width=0.2\textwidth]{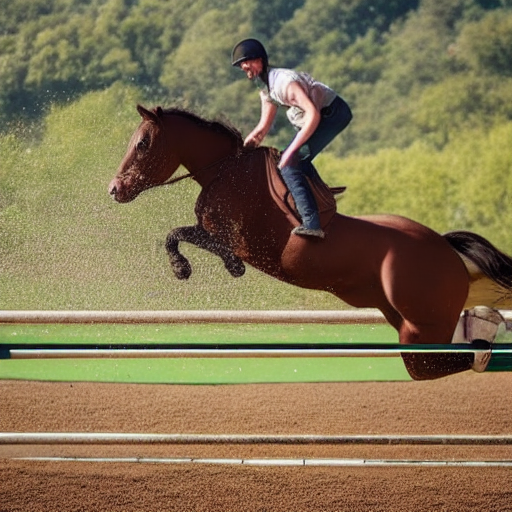} & \includegraphics[width=0.2\textwidth]{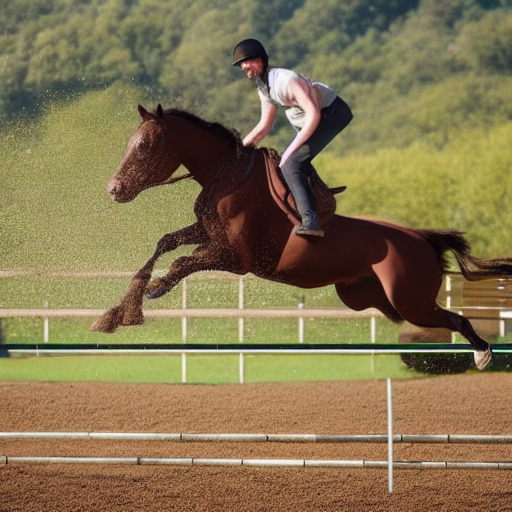} & \includegraphics[width=0.2\textwidth]{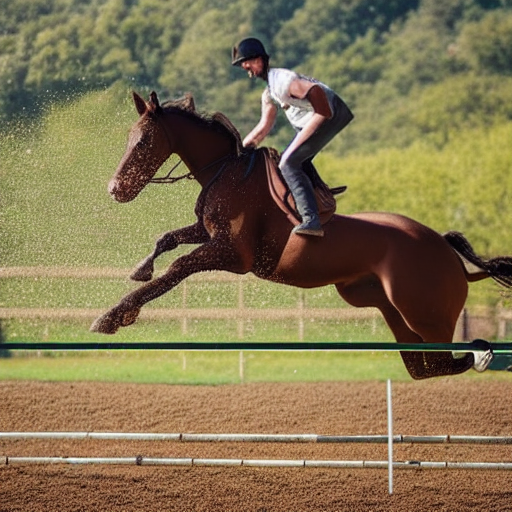} & \includegraphics[width=0.2\textwidth]{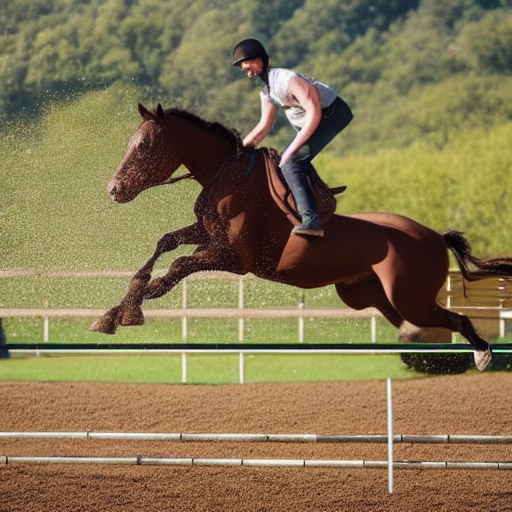} & \includegraphics[width=0.2\textwidth]{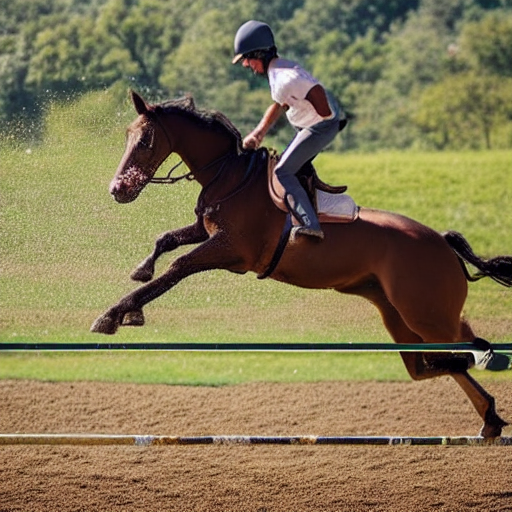}\tabularnewline
\multicolumn{6}{c}{``A man jumping a brown horse over an obstacle.''}\tabularnewline
 &  &  &  &  & \tabularnewline
\includegraphics[width=0.2\textwidth]{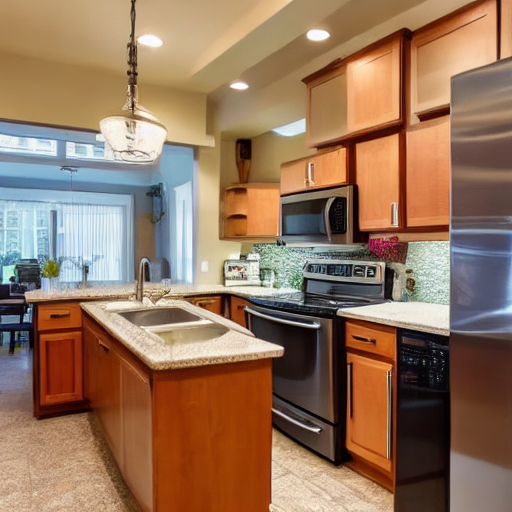} & \includegraphics[width=0.2\textwidth]{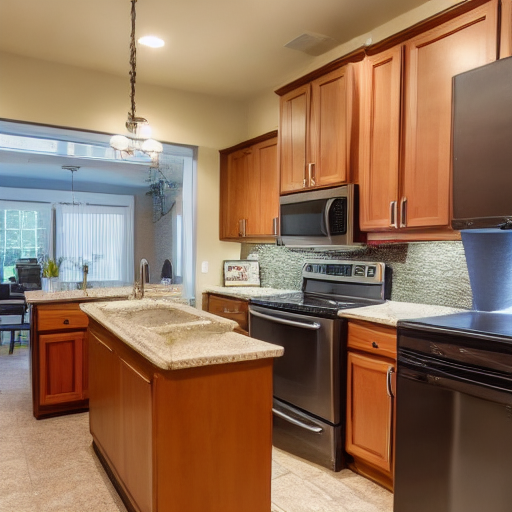} & \includegraphics[width=0.2\textwidth]{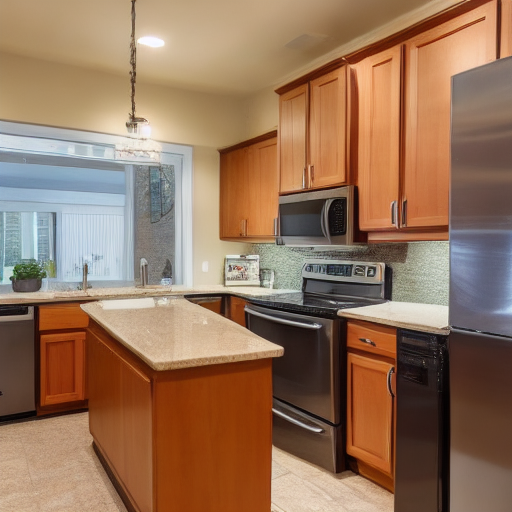} & \includegraphics[width=0.2\textwidth]{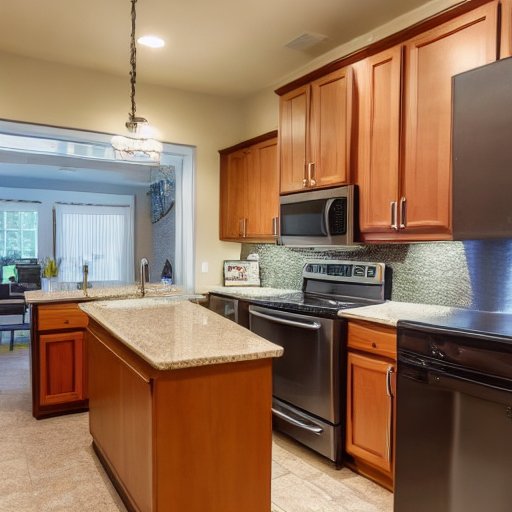} & \includegraphics[width=0.2\textwidth]{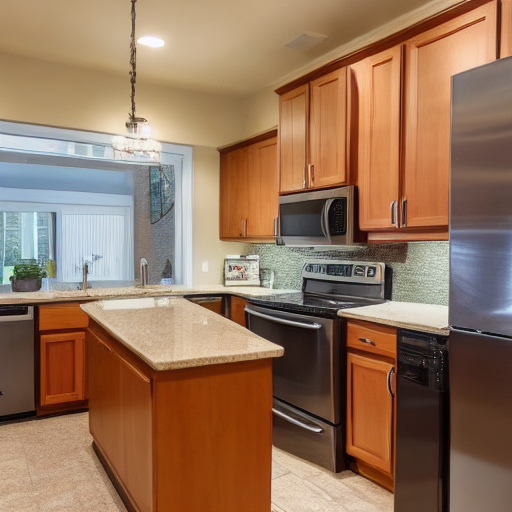} & \includegraphics[width=0.2\textwidth]{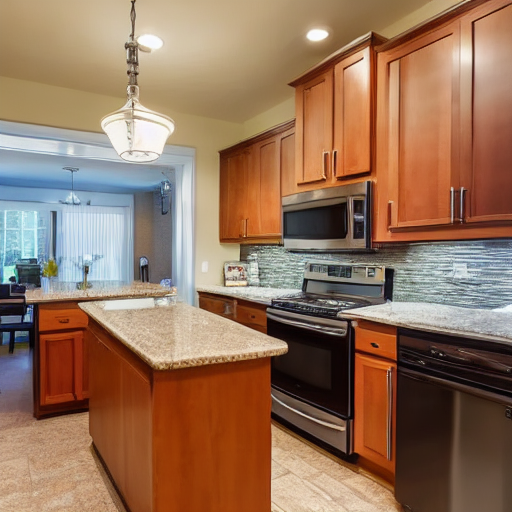}\tabularnewline
\multicolumn{6}{c}{``A kitchen with a center island next to a refrigerator freezer.''}\tabularnewline
\end{tabular}}}
\par\end{centering}
\caption{Images generated by second-order numerical solvers with NFE=10. The
model is Stable Diffusion and the text prompts are COCO2014 captions.\label{fig:GenImages_StableDiffusion_Order2}}
\end{figure*}

\begin{figure*}
\begin{centering}
{\resizebox{\textwidth}{!}{%
\par\end{centering}
\begin{centering}
\begin{tabular}{cccccc}
DEIS-3 & DPM-Solver++ (3M) & UniPC-2 & SC-AB3 & SC-AB2AM3 & SC-AB3AM3\tabularnewline
\includegraphics[width=0.2\textwidth]{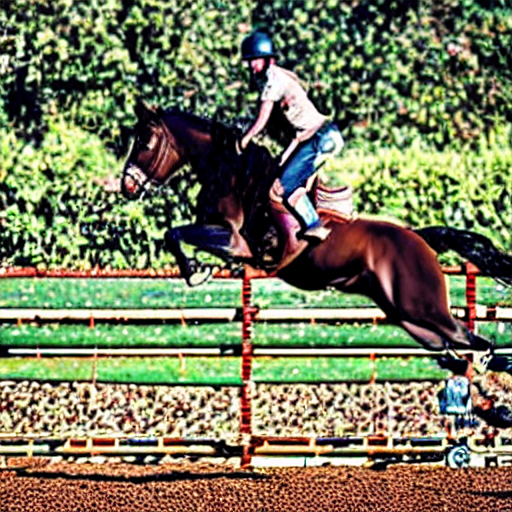} & \includegraphics[width=0.2\textwidth]{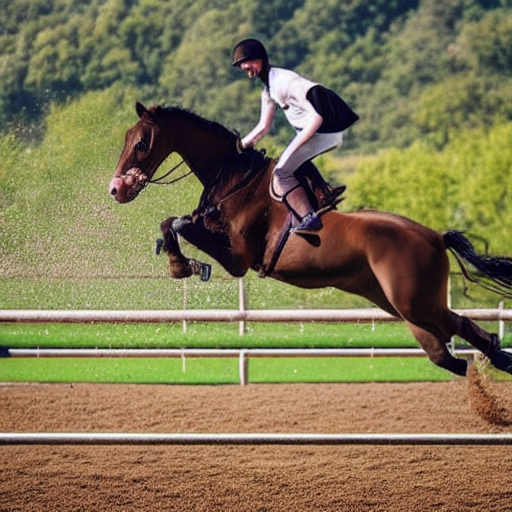} & \includegraphics[width=0.2\textwidth]{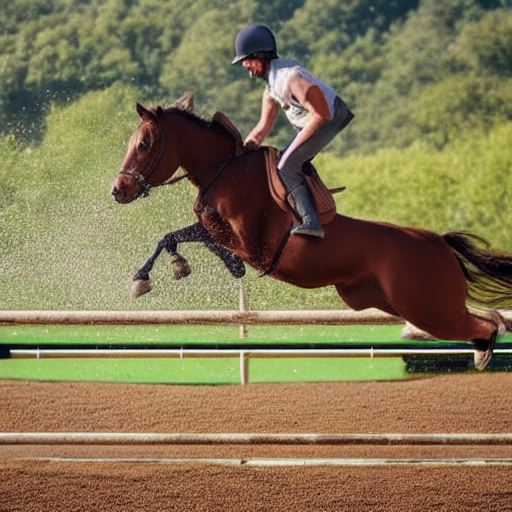} & \includegraphics[width=0.2\textwidth]{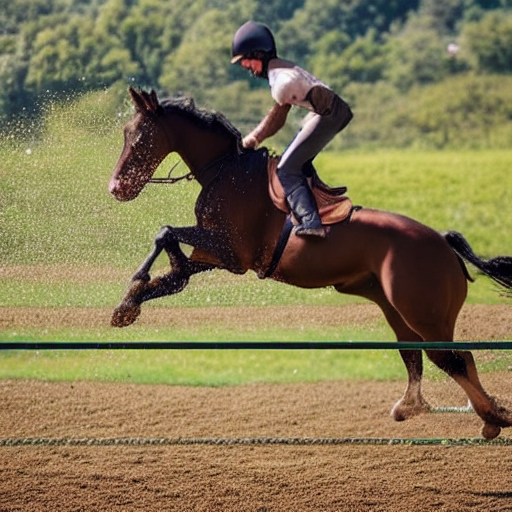} & \includegraphics[width=0.2\textwidth]{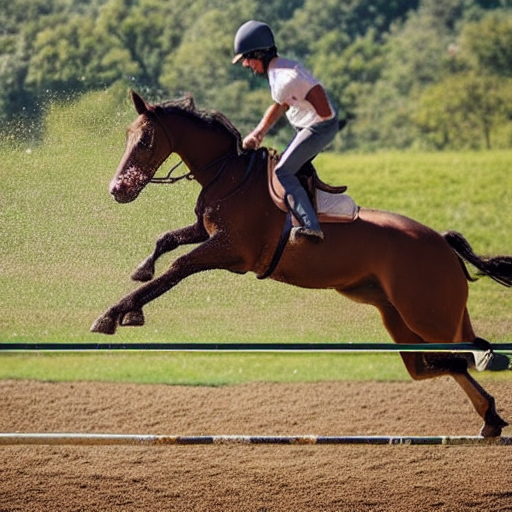} & \includegraphics[width=0.2\textwidth]{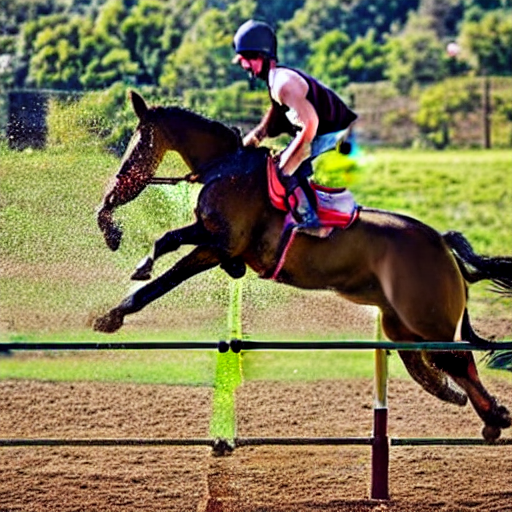}\tabularnewline
\multicolumn{6}{c}{``A man jumping a brown horse over an obstacle.''}\tabularnewline
 &  &  &  &  & \tabularnewline
\includegraphics[width=0.2\textwidth]{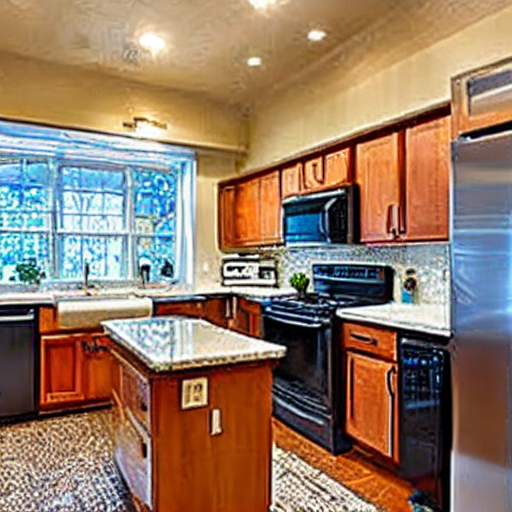} & \includegraphics[width=0.2\textwidth]{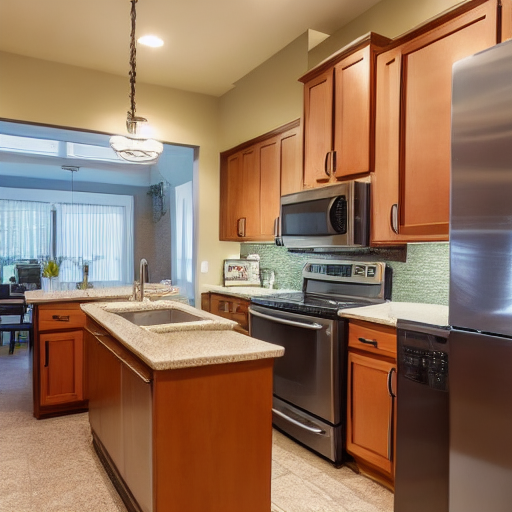} & \includegraphics[width=0.2\textwidth]{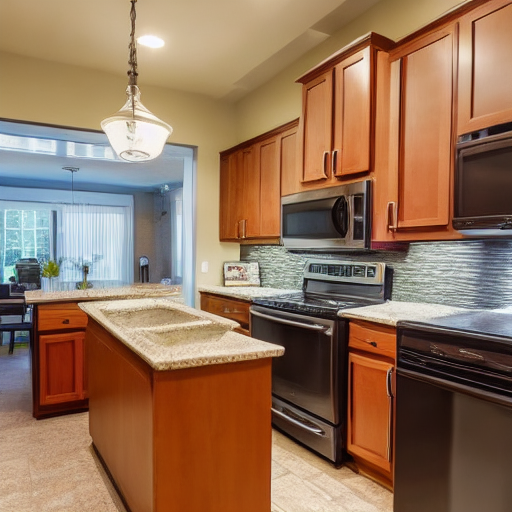} & \includegraphics[width=0.2\textwidth]{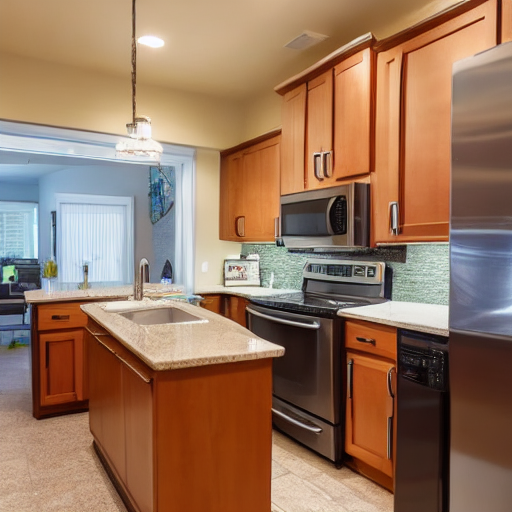} & \includegraphics[width=0.2\textwidth]{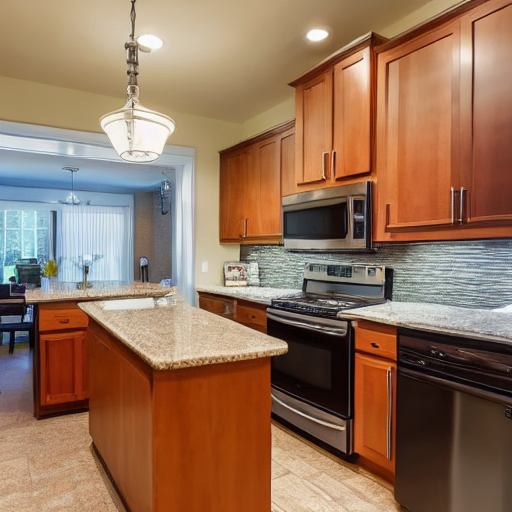} & \includegraphics[width=0.2\textwidth]{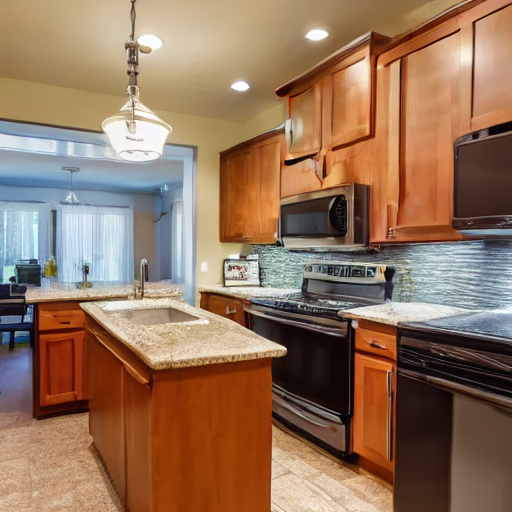}\tabularnewline
\multicolumn{6}{c}{``A kitchen with a center island next to a refrigerator freezer.''}\tabularnewline
\end{tabular}}}
\par\end{centering}
\caption{Images generated by third-order numerical solvers with NFE=10. The
model is Stable Diffusion and the text prompts are COCO2014 captions.\label{fig:GenImages_StableDiffusion_Order3}}
\end{figure*}

In Figs.~\ref{fig:GenImages_StableDiffusion_Order1}, \ref{fig:GenImages_StableDiffusion_Order2},
\ref{fig:GenImages_StableDiffusion_Order3}, we display images generated
by our solvers alongside those generated by the baseline solvers,
using COCO captions as input prompts. Notably, images generated by
first-order solvers, except DDIM, are identical. However, there are
slight content variations in images generated by our high-order solvers
compared to those from the baseline solvers, though their overall
quality remains quite similar.

\begin{thebibliography}{10}

\bibitem{albergo2023stochastic}
Michael~S Albergo, Nicholas~M Boffi, and Eric Vanden-Eijnden.
\newblock Stochastic interpolants: A unifying framework for flows and
  diffusions.
\newblock {\em arXiv preprint arXiv:2303.08797}, 2023.

\bibitem{albergo2022building}
Michael~Samuel Albergo and Eric Vanden-Eijnden.
\newblock Building normalizing flows with stochastic interpolants.
\newblock In {\em The Eleventh International Conference on Learning
  Representations}, 2022.

\bibitem{atkinson2009numerical}
Kendall Atkinson, Weimin Han, and David~E Stewart.
\newblock {\em Numerical solution of ordinary differential equations},
  volume~81.
\newblock John Wiley \& Sons, 2009.

\bibitem{bao2022analytic}
Fan Bao, Chongxuan Li, Jun Zhu, and Bo~Zhang.
\newblock Analytic-dpm: an analytic estimate of the optimal reverse variance in
  diffusion probabilistic models.
\newblock In {\em {ICLR}}, 2022.

\bibitem{bishop2006pattern}
C~Bishop.
\newblock Pattern recognition and machine learning.
\newblock {\em Springer google schola}, 2:531--537, 2006.

\bibitem{brooks2023instructpix2pix}
Tim Brooks, Aleksander Holynski, and Alexei~A Efros.
\newblock Instructpix2pix: Learning to follow image editing instructions.
\newblock In {\em Proceedings of the IEEE/CVF Conference on Computer Vision and
  Pattern Recognition}, pages 18392--18402, 2023.

\bibitem{chen2018neural}
Ricky~TQ Chen, Yulia Rubanova, Jesse Bettencourt, and David~K Duvenaud.
\newblock Neural ordinary differential equations.
\newblock {\em Advances in neural information processing systems}, 31, 2018.

\bibitem{dhariwal2021diffusion}
Prafulla Dhariwal and Alexander Nichol.
\newblock Diffusion models beat gans on image synthesis.
\newblock {\em Advances in neural information processing systems},
  34:8780--8794, 2021.

\bibitem{efron2011tweedie}
Bradley Efron.
\newblock Tweedie's formula and selection bias.
\newblock {\em Journal of the American Statistical Association},
  106(496):1602--1614, 2011.

\bibitem{finlay2020train}
Chris Finlay, J{\"o}rn-Henrik Jacobsen, Levon Nurbekyan, and Adam Oberman.
\newblock How to train your neural ode: the world of jacobian and kinetic
  regularization.
\newblock In {\em International conference on machine learning}, pages
  3154--3164. PMLR, 2020.

\bibitem{griffiths2010numerical}
David~Francis Griffiths and Desmond~J Higham.
\newblock {\em Numerical methods for ordinary differential equations: initial
  value problems}, volume~5.
\newblock Springer, 2010.

\bibitem{heusel2017gans}
Martin Heusel, Hubert Ramsauer, Thomas Unterthiner, Bernhard Nessler, and Sepp
  Hochreiter.
\newblock Gans trained by a two time-scale update rule converge to a local nash
  equilibrium.
\newblock {\em Advances in neural information processing systems}, 30, 2017.

\bibitem{ho2020denoising}
Jonathan Ho, Ajay Jain, and Pieter Abbeel.
\newblock Denoising diffusion probabilistic models.
\newblock {\em Advances in neural information processing systems},
  33:6840--6851, 2020.

\bibitem{ho2022video}
Jonathan Ho, Tim Salimans, Alexey Gritsenko, William Chan, Mohammad Norouzi,
  and David~J Fleet.
\newblock Video diffusion models.
\newblock {\em arXiv:2204.03458}, 2022.

\bibitem{hyvarinen2005estimation}
Aapo Hyv{\"a}rinen and Peter Dayan.
\newblock Estimation of non-normalized statistical models by score matching.
\newblock {\em Journal of Machine Learning Research}, 6(4), 2005.

\bibitem{karras2022elucidating}
Tero Karras, Miika Aittala, Timo Aila, and Samuli Laine.
\newblock Elucidating the design space of diffusion-based generative models.
\newblock {\em Advances in Neural Information Processing Systems},
  35:26565--26577, 2022.

\bibitem{kim2022maximum}
Dongjun Kim, Byeonghu Na, Se~Jung Kwon, Dongsoo Lee, Wanmo Kang, and Il-chul
  Moon.
\newblock Maximum likelihood training of implicit nonlinear diffusion model.
\newblock {\em Advances in Neural Information Processing Systems},
  35:32270--32284, 2022.

\bibitem{kingma2021variational}
Diederik Kingma, Tim Salimans, Ben Poole, and Jonathan Ho.
\newblock Variational diffusion models.
\newblock {\em Advances in neural information processing systems},
  34:21696--21707, 2021.

\bibitem{kong2021fast}
Zhifeng Kong and Wei Ping.
\newblock On fast sampling of diffusion probabilistic models.
\newblock In {\em ICML Workshop on Invertible Neural Networks, Normalizing
  Flows, and Explicit Likelihood Models}, 2021.

\bibitem{kong2020diffwave}
Zhifeng Kong, Wei Ping, Jiaji Huang, Kexin Zhao, and Bryan Catanzaro.
\newblock Diffwave: A versatile diffusion model for audio synthesis.
\newblock In {\em International Conference on Learning Representations}, 2021.

\bibitem{lin2014microsoft}
Tsung-Yi Lin, Michael Maire, Serge Belongie, James Hays, Pietro Perona, Deva
  Ramanan, Piotr Doll{\'a}r, and C~Lawrence Zitnick.
\newblock Microsoft coco: Common objects in context.
\newblock In {\em Computer Vision--ECCV 2014: 13th European Conference, Zurich,
  Switzerland, September 6-12, 2014, Proceedings, Part V 13}, pages 740--755.
  Springer, 2014.

\bibitem{lipman2022flow}
Yaron Lipman, Ricky~TQ Chen, Heli Ben-Hamu, Maximilian Nickel, and Matthew Le.
\newblock Flow matching for generative modeling.
\newblock In {\em The Eleventh International Conference on Learning
  Representations}, 2022.

\bibitem{liu2022pseudo}
Luping Liu, Yi~Ren, Zhijie Lin, and Zhou Zhao.
\newblock Pseudo numerical methods for diffusion models on manifolds.
\newblock In {\em {ICLR}}, 2022.

\bibitem{liu2022rectified}
Qiang Liu.
\newblock Rectified flow: A marginal preserving approach to optimal transport.
\newblock {\em arXiv preprint arXiv:2209.14577}, 2022.

\bibitem{liu2022flow}
Xingchao Liu, Chengyue Gong, et~al.
\newblock Flow straight and fast: Learning to generate and transfer data with
  rectified flow.
\newblock In {\em The Eleventh International Conference on Learning
  Representations}, 2022.

\bibitem{liu2023instaflow}
Xingchao Liu, Xiwen Zhang, Jianzhu Ma, Jian Peng, and Qiang Liu.
\newblock Instaflow: One step is enough for high-quality diffusion-based
  text-to-image generation.
\newblock {\em arXiv preprint arXiv:2309.06380}, 2023.

\bibitem{lu2022dpm}
Cheng Lu, Yuhao Zhou, Fan Bao, Jianfei Chen, Chongxuan Li, and Jun Zhu.
\newblock Dpm-solver: A fast ode solver for diffusion probabilistic model
  sampling in around 10 steps.
\newblock {\em Advances in Neural Information Processing Systems},
  35:5775--5787, 2022.

\bibitem{lu2022dpmpp}
Cheng Lu, Yuhao Zhou, Fan Bao, Jianfei Chen, Chongxuan Li, and Jun Zhu.
\newblock Dpm-solver++: Fast solver for guided sampling of diffusion
  probabilistic models.
\newblock {\em arXiv preprint arXiv:2211.01095}, 2022.

\bibitem{luhman2021knowledge}
Eric Luhman and Troy Luhman.
\newblock Knowledge distillation in iterative generative models for improved
  sampling speed.
\newblock {\em arXiv preprint arXiv:2101.02388}, 2021.

\bibitem{lyu2009interpretation}
Siwei Lyu.
\newblock Interpretation and generalization of score matching.
\newblock In {\em Proceedings of the Twenty-Fifth Conference on Uncertainty in
  Artificial Intelligence}, pages 359--366, 2009.

\bibitem{maaloe2019biva}
Lars Maal{\o}e, Marco Fraccaro, Valentin Li{\'e}vin, and Ole Winther.
\newblock Biva: A very deep hierarchy of latent variables for generative
  modeling.
\newblock {\em Advances in neural information processing systems}, 32, 2019.

\bibitem{maoutsa2020interacting}
Dimitra Maoutsa, Sebastian Reich, and Manfred Opper.
\newblock Interacting particle solutions of fokker--planck equations through
  gradient--log--density estimation.
\newblock {\em Entropy}, 22(8):802, 2020.

\bibitem{meng2021sdedit}
Chenlin Meng, Yutong He, Yang Song, Jiaming Song, Jiajun Wu, Jun-Yan Zhu, and
  Stefano Ermon.
\newblock Sdedit: Guided image synthesis and editing with stochastic
  differential equations.
\newblock In {\em International Conference on Learning Representations}, 2022.

\bibitem{moulton1926new}
Forest~Ray Moulton.
\newblock {\em New methods in exterior ballistics}.
\newblock University of Chicago Press, 1926.

\bibitem{nichol2021improved}
Alexander~Quinn Nichol and Prafulla Dhariwal.
\newblock Improved denoising diffusion probabilistic models.
\newblock In {\em {ICML}}, pages 8162--8171, 2021.

\bibitem{oksendal2013stochastic}
Bernt Oksendal.
\newblock {\em Stochastic differential equations: an introduction with
  applications}.
\newblock Springer Science \& Business Media, 2013.

\bibitem{onken2021ot}
Derek Onken, Samy~Wu Fung, Xingjian Li, and Lars Ruthotto.
\newblock Ot-flow: Fast and accurate continuous normalizing flows via optimal
  transport.
\newblock In {\em Proceedings of the AAAI Conference on Artificial
  Intelligence}, volume~35, pages 9223--9232, 2021.

\bibitem{pooladian2023multisample}
Aram-Alexandre Pooladian, Heli Ben-Hamu, Carles Domingo-Enrich, Brandon Amos,
  Yaron Lipman, and Ricky Chen.
\newblock Multisample flow matching: Straightening flows with minibatch
  couplings.
\newblock {\em arXiv preprint arXiv:2304.14772}, 2023.

\bibitem{rombach2022high}
Robin Rombach, Andreas Blattmann, Dominik Lorenz, Patrick Esser, and Bj{\"o}rn
  Ommer.
\newblock High-resolution image synthesis with latent diffusion models.
\newblock In {\em Proceedings of the IEEE/CVF conference on computer vision and
  pattern recognition}, pages 10684--10695, 2022.

\bibitem{saharia2022photorealistic}
Chitwan Saharia, William Chan, Saurabh Saxena, Lala Li, Jay Whang, Emily~L
  Denton, Kamyar Ghasemipour, Raphael Gontijo~Lopes, Burcu Karagol~Ayan, Tim
  Salimans, et~al.
\newblock Photorealistic text-to-image diffusion models with deep language
  understanding.
\newblock {\em Advances in Neural Information Processing Systems},
  35:36479--36494, 2022.

\bibitem{salimans2022progressive}
Tim Salimans and Jonathan Ho.
\newblock Progressive distillation for fast sampling of diffusion models.
\newblock {\em arXiv preprint arXiv:2202.00512}, 2022.

\bibitem{sohl2015deep}
Jascha Sohl-Dickstein, Eric Weiss, Niru Maheswaranathan, and Surya Ganguli.
\newblock Deep unsupervised learning using nonequilibrium thermodynamics.
\newblock In {\em International conference on machine learning}, pages
  2256--2265. PMLR, 2015.

\bibitem{song2020denoising}
Jiaming Song, Chenlin Meng, and Stefano Ermon.
\newblock Denoising diffusion implicit models.
\newblock In {\em International Conference on Learning Representations}, 2021.

\bibitem{song2023consistency}
Yang Song, Prafulla Dhariwal, Mark Chen, and Ilya Sutskever.
\newblock Consistency models.
\newblock {\em arXiv preprint arXiv:2303.01469}, 2023.

\bibitem{song2021maximum}
Yang Song, Conor Durkan, Iain Murray, and Stefano Ermon.
\newblock Maximum likelihood training of score-based diffusion models.
\newblock {\em Advances in Neural Information Processing Systems},
  34:1415--1428, 2021.

\bibitem{song2019generative}
Yang Song and Stefano Ermon.
\newblock Generative modeling by estimating gradients of the data distribution.
\newblock {\em Advances in neural information processing systems}, 32, 2019.

\bibitem{song2020improved}
Yang Song and Stefano Ermon.
\newblock Improved techniques for training score-based generative models.
\newblock {\em Advances in neural information processing systems},
  33:12438--12448, 2020.

\bibitem{song2020score}
Yang Song, Jascha Sohl-Dickstein, Diederik~P Kingma, Abhishek Kumar, Stefano
  Ermon, and Ben Poole.
\newblock Score-based generative modeling through stochastic differential
  equations.
\newblock In {\em International Conference on Learning Representations}, 2021.

\bibitem{vahdat2020nvae}
Arash Vahdat and Jan Kautz.
\newblock Nvae: A deep hierarchical variational autoencoder.
\newblock {\em Advances in neural information processing systems},
  33:19667--19679, 2020.

\bibitem{vincent2011connection}
Pascal Vincent.
\newblock A connection between score matching and denoising autoencoders.
\newblock {\em Neural computation}, 23(7):1661--1674, 2011.

\bibitem{vincent2010stacked}
Pascal Vincent, Hugo Larochelle, Isabelle Lajoie, Yoshua Bengio, Pierre-Antoine
  Manzagol, and L{\'e}on Bottou.
\newblock Stacked denoising autoencoders: Learning useful representations in a
  deep network with a local denoising criterion.
\newblock {\em Journal of machine learning research}, 11(12), 2010.

\bibitem{von-platen-etal-2022-diffusers}
Patrick von Platen, Suraj Patil, Anton Lozhkov, Pedro Cuenca, Nathan Lambert,
  Kashif Rasul, Mishig Davaadorj, and Thomas Wolf.
\newblock Diffusers: State-of-the-art diffusion models.
\newblock \url{https://github.com/huggingface/diffusers}, 2022.

\bibitem{watson2021learningfast}
Daniel Watson, William Chan, Jonathan Ho, and Mohammad Norouzi.
\newblock Learning fast samplers for diffusion models by differentiating
  through sample quality.
\newblock In {\em International Conference on Learning Representations}, 2022.

\bibitem{watson2021learning}
Daniel Watson, Jonathan Ho, Mohammad Norouzi, and William Chan.
\newblock Learning to efficiently sample from diffusion probabilistic models.
\newblock {\em arXiv preprint arXiv:2106.03802}, 2021.

\bibitem{Welling2011}
Max Welling and Yee~W Teh.
\newblock Bayesian learning via stochastic gradient langevin dynamics.
\newblock In {\em Proceedings of the 28th international conference on machine
  learning (ICML-11)}, pages 681--688, 2011.

\bibitem{zhang2023fast}
Qinsheng Zhang and Yongxin Chen.
\newblock Fast sampling of diffusion models with exponential integrator.
\newblock In {\em {ICLR}}, 2023.

\bibitem{zhao2023unipc}
Wenliang Zhao, Lujia Bai, Yongming Rao, Jie Zhou, and Jiwen Lu.
\newblock Unipc: A unified predictor-corrector framework for fast sampling of
  diffusion models.
\newblock {\em arXiv preprint arXiv:2302.04867}, 2023.

\end{thebibliography}
\end{document}